\documentclass[a4paper,fleqn]{cas-sc}

\usepackage{amsthm} 
\usepackage{booktabs}
\usepackage{siunitx}

\usepackage{subcaption}
\usepackage{bm}
\usepackage[numbers]{natbib}
\def\tsc#1{\csdef{#1}{\textsc{\lowercase{#1}}\xspace}}
\tsc{WGM}
\tsc{QE}

\newcommand{\USnum}[1]{%
  \num[
    locale=US,
    group-separator={,},
    group-minimum-digits=4,
    output-decimal-marker=.,
    scientific-notation=true,
    output-exponent-marker=\mathrm{e},
    tight-spacing=true
  ]{#1}%
}
\newtheorem{theorem}{Theorem} 
\newtheorem{lemma}{Lemma}  
\begin{document}
\let\WriteBookmarks\relax
\def\floatpagepagefraction{1}
\def\textpagefraction{.001}
\graphicspath{ {figs/} }
\shorttitle{LNN-PINN: Physics-Only Training}

\shortauthors{Z. Tao et al.}

\title [mode = title]{LNN-PINN: A Unified Physics-Only Training Framework with Liquid Residual Blocks}                      

%
\author[1]{Ze Tao}[orcid=0009-0004-0202-3641]
\credit{Calculation, data analyzing and manuscript writing}
\affiliation[1]{organization={Nanophotonics and Biophotonics Key Laboratory of Jilin Province, School of Physics, Changchun University of Science and Technology},
                city={Changchun},
                postcode={130022},
                country={P.R. China}}
\author[2]{Hanxuan Wang}[orcid=0000-0003-1830-5913]
\credit{Calculation and data analyzing}
\affiliation[2]{organization={Faculty of Chinese Medicine, Macau University of Science and 
Technology},
                city={Macau},
                postcode={999078},
                country={P.R. China}}
\author[1]{Fujun Liu}[orcid=0000-0002-8573-450X]
\credit{Review and Editing}
\cormark[1]
\ead{fjliu@cust.edu.cn}
\cortext[1]{Corresponding author}


\begin{abstract}
Physics-informed neural networks (PINNs) integrate physical priors with deep learning, but often struggle with predictive accuracy on complex problems. To address this, we propose Liquid Neural Networks-based PINN (LNN-PINN), a novel framework that enhances PINNs through a liquid residual gating architecture. This non-invasive upgrade introduces a lightweight gating mechanism solely within the hidden layers, preserving the original physics-based loss and training paradigm to ensure improvements arise purely from architectural refinement. Our method consistently reduces RMSE and MAE across diverse benchmark problems under identical training conditions, demonstrating robust adaptability to varying dimensions and boundary conditions. Consequently, LNN-PINN provides an effective, plug-and-play enhancement for improving the accuracy of physics-informed learning in challenging scientific and engineering applications.
\end{abstract}


\begin{highlights}
    \item A novel architecture that boosts the accuracy of physics-informed neural networks (PINNs) for complex problems.
    \item A non-invasive enhancement that requires no changes to the established physics-based loss or training setup.
    \item A versatile solution proven effective across a wide range of partial differential equations (PDEs).
    \item A practical and easy-to-implement upgrade for developing more reliable neural PDE solvers.
\end{highlights}


\maketitle

\section{Introduction}
Physics-informed neural networks (PINNs) embed prior knowledge of partial differential equations seamlessly into deep learning frameworks, offering a novel and efficient numerical tool for addressing physical modeling problems in science and engineering\cite{raissi2019physics}. This approach provides strong generalization capabilities and physical interpretability, and shows great potential for tackling high-dimensional, nonlinear, and multiphysics-coupled problems\cite{luo2025physics,hu2024physics,tao2025analytical,coutinho2023physics,zhang2022pinn,mcclenny2023self, Xing,kashefi2025physics,kashefi2022physics}. However, their potential has not yet been fully exploited, which limits their effectiveness in tasks demanding higher accuracy and complexity\cite{lee2025physics,li2024physics,toscano2025pinns,song2025vw,grossmann2024can}.

In recent years, numerous studies have sought to enhance the performance of PINNs through advances in network architecture design, sampling strategies, loss function weighting, and multi-scale feature modeling\cite{gladstone2025fo,wei2025ffv,ding20251d,yan2025dp,rezaei2022mixed}. These approaches have achieved notable improvements in convergence stability, training efficiency, and predictive accuracy, thereby expanding the applications of PINNs to diverse fields such as fluid mechanics\cite{lawal2025modeling,mochalin2025enhancement}, heat transfer\cite{shen2025multi,li2025improved,cai2025physics}, elasticity\cite{kianian2025pinn,limarchenko2025application}, and electromagnetics\cite{deng2025physics,wang2025dimensionless}. {Nevertheless, existing methods still face limitations when one seeks high-accuracy field reconstruction under purely physics-based supervision, especially in problems involving subtle couplings among multiple physical quantities and diverse boundary constraints. Recent critical analyses have further pointed out that idealized governing equations in PINN frameworks may overlook multiscale effects, boundary complexities, and nonlinear couplings, thereby creating the risk of physically plausible yet unreliable predictions in engineering settings\cite{naser2025fundamental}. However, most existing PINN enhancements act through sampling design\cite{hou2023enhancing}, loss reweighting\cite{chen2025self,gao2025physics}, multi-scale encoding\cite{bounnah2025physics,wang2026mfpinn}, or alternative problem formulations\cite{zhang2024weak,rojas2024robust}, whereas an architecture-only strategy that leaves the physics-only training pipeline unchanged while directly improving hidden-state propagation remains insufficiently explored. In the present work, the central question is therefore not whether multiple enhancement mechanisms can be combined, but whether a controlled modification of the internal feature-propagation law alone can improve physics-only neural PDE reconstruction.}

{Concurrently, liquid neural networks (LNNs), a novel architecture featuring dynamic state updates and gating mechanisms, have gained increasing attention because their gated residual modulation directly regulates how latent features are propagated and refined across depth. By leveraging learnable gating parameters to adaptively regulate information flow, LNNs enhance representational capacity and numerical stability while maintaining model compactness\cite{zhang2025dynamic,akpinar2025novel,shaheema2025explainable,narvare2025liqu,zhang2025invertible,aslan2025novel}. In the present context, this property is relevant not simply because some physical systems may be dynamic, but because the final PINN field approximation, its automatic-differentiation derivatives, and the resulting PDE residuals all depend on the internal feature-transport process. These characteristics therefore make LNNs a natural candidate for addressing the specific limitation studied here, namely whether improved hidden-state propagation can enhance physics-only PDE reconstruction under a matched physics-only training protocol.}

{Motivated by these developments, this work proposes LNN--PINN, a unified physics-constrained training framework that integrates a liquid residual gating architecture into physics-informed neural networks while keeping the physics modeling and optimization pipeline unchanged. The method replaces only the hidden-layer internal mapping, maintaining identical sampling strategies, loss compositions, and hyperparameter settings, so that the resulting performance difference is intended to isolate the architectural contribution. Specifically, this work makes three contributions. First, it formulates an architecture-level LNN--PINN framework that preserves the external physics-only training structure and modifies only the hidden-state propagation mechanism. Second, it establishes a controlled evaluation setting in which the contribution of liquid residual gating can be assessed without auxiliary changes in the training protocol. Third, it examines the proposed architecture on representative partial differential equation benchmarks and shows, through the reported fixed-grid error metrics, that LNN--PINN reduces reconstruction errors by benchmark-dependent margins relative to the plain PINN, ranging from substantial gains in the scalar benchmarks to more selective improvements in the strongly coupled multi-field setting.}

\section{LNN-PINN: Structural Mechanism and Modeling Framework}
\subsection{Overall Structure of PINNs and Physics-Only Loss}
{We introduce a unified formulation that covers both steady-state and time-dependent partial differential equations. Let $\xi$ denote the generic network input coordinate. For a steady-state problem, we set $\xi=x\in\Omega\subset\mathbb{R}^{d}$. For a time-dependent problem, we set $\xi=(x,t)\in\Omega\times(0,T]\subset\mathbb{R}^{d+1}$. Accordingly, the governing equation is written in the unified form
\begin{equation}
\mathcal{L}[u](\xi)=g(\xi), \qquad \xi\in\mathcal{Q},
\end{equation}
where $\mathcal{Q}=\Omega$ in the steady-state setting and $\mathcal{Q}=\Omega\times(0,T]$ in the time-dependent setting. The unknown field $u:\mathcal{Q}\to\mathbb{R}^{m}$ may represent either a single field or a coupled multi-field system. Boundary constraints are imposed on
\begin{equation}
\mathcal{B}[u](\xi)=0, \qquad \xi\in\Sigma_{\partial},
\end{equation}
where $\Sigma_{\partial}=\partial\Omega$ for steady-state problems and $\Sigma_{\partial}=\partial\Omega\times(0,T]$ for time-dependent problems. If the problem is time-dependent, we additionally impose the initial condition
\begin{equation}\label{3}
\mathcal{I}[u](x,0)=0, \qquad x\in\Omega.
\end{equation}
For steady-state problems, Eq.~\eqref{3} is absent and no initial-condition residual is introduced.

Before computation, we apply nondimensionalization and, when necessary, fixed residual-magnitude normalization in order to keep different physical channels on comparable numerical scales and to mitigate optimization ill-conditioning. We do not present this preprocessing as a novel ingredient of the present work. Rather, we use it as a standard conditioning step already discussed in the PINN literature and in related physics-informed architectures, including recent review and dimensionless/scaling-based discussions; for example, Refs.~\cite{luo2025physics,toscano2025pinns,wang2025dimensionless}. In the present manuscript, the scaling factors are fixed before training, they do not alter the governing operators, and they do not introduce any additional optimization component. The exact scale choices and loss-normalization rules adopted here are recorded in Appendix~\ref{A}.

The unknown field is approximated by a neural network
\begin{equation}
f_{\theta}:\mathcal{Q}\to\mathbb{R}^{m},
\end{equation}
parameterized by $\theta$. Automatic differentiation supplies all derivatives required by the operators in Eqs.~(1)–(3). We therefore define the corresponding strong-form residuals by
\begin{equation}
\begin{aligned}
r_{\mathcal{Q}}(\xi;\theta)&:=\mathcal{L}[f_{\theta}](\xi)-g(\xi), \qquad &&\xi\in\mathcal{Q},\\
r_{\partial}(\xi;\theta)&:=\mathcal{B}[f_{\theta}](\xi), \qquad &&\xi\in\Sigma_{\partial},\\
r_{0}(x;\theta)&:=\mathcal{I}[f_{\theta}](x,0), \qquad &&x\in\Omega \quad \text{(time-dependent problems only)}.
\end{aligned}
\end{equation}

At the continuous level, we define the unified physics-only objective by
\begin{equation}
\mathcal{J}(\theta)
=
\int_{\mathcal{Q}}\|r_{\mathcal{Q}}(\xi;\theta)\|^{2}\,\mathrm{d}\mu_{\mathcal{Q}}(\xi)
+
\lambda_{\partial}\int_{\Sigma_{\partial}}\|r_{\partial}(\xi;\theta)\|^{2}\,\mathrm{d}\mu_{\partial}(\xi)
+
\lambda_{0}\int_{\Omega}\|r_{0}(x;\theta)\|^{2}\,\mathrm{d}\mu_{0}(x),
\end{equation}
where $\mu_{\mathcal{Q}}$, $\mu_{\partial}$, and $\mu_{0}$ are the sampling measures for the interior, boundary, and initial manifolds, respectively, and $\lambda_{\partial}>0$, $\lambda_{0}>0$ are weighting coefficients. In the steady-state setting, the third term is omitted. The weighted continuous-risk extension and the corresponding empirical-risk interpretation are given in Appendix~\ref{B}.

For practical training, we draw finite sample sets
\begin{equation}
\mathcal{D}_{\mathcal{Q}}=\{\xi_{i}\}_{i=1}^{N_{\mathcal{Q}}}\sim\mu_{\mathcal{Q}},
\qquad
\mathcal{D}_{\partial}=\{\xi_{j}^{\partial}\}_{j=1}^{N_{\partial}}\sim\mu_{\partial},
\qquad
\mathcal{D}_{0}=\{x_{k}^{0}\}_{k=1}^{N_{0}}\sim\mu_{0},
\end{equation}
where $\mathcal{D}_{0}$ appears only in the time-dependent setting. The Monte Carlo empirical objective then reads
\begin{equation}
\widehat{\mathcal{J}}(\theta)
=
\frac{1}{N_{\mathcal{Q}}}\sum_{\xi_{i}\in\mathcal{D}_{\mathcal{Q}}}\|r_{\mathcal{Q}}(\xi_{i};\theta)\|^{2}
+
\lambda_{\partial}\frac{1}{N_{\partial}}\sum_{\xi_{j}^{\partial}\in\mathcal{D}_{\partial}}\|r_{\partial}(\xi_{j}^{\partial};\theta)\|^{2}
+
\lambda_{0}\frac{1}{N_{0}}\sum_{x_{k}^{0}\in\mathcal{D}_{0}}\|r_{0}(x_{k}^{0};\theta)\|^{2}.
\end{equation}
Again, the third term is omitted for steady-state problems. With independent and identically distributed draws, $\widehat{\mathcal{J}}$ is an unbiased estimator of $\mathcal{J}$ and converges almost surely to $\mathcal{J}$ as the corresponding sample numbers tend to infinity. The precise measure-theoretic statement is given in Appendix~\ref{B}.

To display the optimization structure, we define the network Jacobian
\begin{equation}
J_{\theta}(\xi):=\frac{\partial f_{\theta}(\xi)}{\partial\theta}\in\mathbb{R}^{m\times|\theta|},
\end{equation}
and denote the Fr\'echet derivatives of $\mathcal{L}$, $\mathcal{B}$, and, when applicable, $\mathcal{I}$ with respect to $u$, evaluated at $f_{\theta}$, by $D_{u}\mathcal{L}[f_{\theta}](\xi)$, $D_{u}\mathcal{B}[f_{\theta}](\xi)$, and $D_{u}\mathcal{I}[f_{\theta}](x,0)$. The gradient of the empirical objective takes the abstract form
\begin{equation}
\begin{split}
\nabla_{\theta}\widehat{\mathcal{J}}(\theta)
&=
\frac{2}{N_{\mathcal{Q}}}\sum_{\xi_{i}\in\mathcal{D}_{\mathcal{Q}}}
J_{\theta}(\xi_{i})^{\top}
D_{u}\mathcal{L}[f_{\theta}](\xi_{i})^{\top}
r_{\mathcal{Q}}(\xi_{i};\theta)\\
&\quad+
\lambda_{\partial}\frac{2}{N_{\partial}}\sum_{\xi_{j}^{\partial}\in\mathcal{D}_{\partial}}
J_{\theta}(\xi_{j}^{\partial})^{\top}
D_{u}\mathcal{B}[f_{\theta}](\xi_{j}^{\partial})^{\top}
r_{\partial}(\xi_{j}^{\partial};\theta)\\
&\quad+
\lambda_{0}\frac{2}{N_{0}}\sum_{x_{k}^{0}\in\mathcal{D}_{0}}
J_{\theta}(x_{k}^{0},0)^{\top}
D_{u}\mathcal{I}[f_{\theta}](x_{k}^{0},0)^{\top}
r_{0}(x_{k}^{0};\theta),
\end{split}
\end{equation}
where the last line is absent for steady-state problems. This expression follows the automatic-differentiation chain ``network $\to$ operator $\to$ residual''. Parameters are updated by a first-order optimizer such as Adam:
\begin{equation}
\theta^{(t+1)}=\theta^{(t)}-\eta_{t}\nabla_{\theta}\widehat{\mathcal{J}}\!\left(\theta^{(t)}\right),
\end{equation}
where $\eta_{t}$ denotes the learning rate.

For reporting purposes, we use the generic componentwise MSE diagnostics
\begin{equation}
J_{\mathrm{PDE}}(\theta)
=
\frac{1}{N_{\mathcal{Q}}}
\sum_{\xi_{i}\in\mathcal{D}_{\mathcal{Q}}}
\|r_{\mathcal{Q}}(\xi_{i};\theta)\|^{2},
\end{equation}
\begin{equation}
J_{\mathrm{BC}}(\theta)
=
\frac{1}{N_{\partial}}
\sum_{\xi_{j}^{\partial}\in\mathcal{D}_{\partial}}
\|r_{\partial}(\xi_{j}^{\partial};\theta)\|^{2},
\end{equation}
and, for time-dependent problems only,
\begin{equation}
J_{\mathrm{IC}}(\theta)
=
\frac{1}{N_{0}}
\sum_{x_{k}^{0}\in\mathcal{D}_{0}}
\|r_{0}(x_{k}^{0};\theta)\|^{2}.
\end{equation}
Accordingly, the empirical objective can be written compactly as
\begin{equation}\label{LOSS Total}
\widehat{\mathcal{J}}(\theta)
=
J_{\mathrm{PDE}}(\theta)
+
\lambda_{\partial}J_{\mathrm{BC}}(\theta)
+
\lambda_{0}J_{\mathrm{IC}}(\theta),
\end{equation}
where the initial-condition term is omitted for steady-state problems. When a particular benchmark contains several boundary segments or several boundary-condition types, we regard them as additive components of the boundary contribution inside Eq.~\eqref{LOSS Total}. Therefore, all benchmark-specific training objectives in Section~\ref{sec3} are treated as direct instantiations of Eq.~\eqref{LOSS Total}, and we do not restate separate case-level total-loss formulas there.

This completes the generic PINN specification for both steady-state and time-dependent problems under a unified physics-only objective. Fig.~\ref{LNN-PINN} then highlights how LNN--PINN preserves this external training structure while replacing only the internal hidden-state propagation law of the neural trial solution.}
\subsection{Integrating LNN into PINN: Structural Replacement and Implementation}

{To facilitate direct comparison, we keep the physics-only PINN specification established in Section 2.1 unchanged and modify only the parametric form of the neural trial solution. In this sense, the integration of LNN into PINN is realized at the level of hidden-state propagation, rather than through any additional loss term, altered sampling rule, or separate optimization procedure. Consequently, the governing equations, collocation sets, residual definitions, and MSE-based physics objective remain exactly the same as those introduced above, while the internal feature propagation law of the neural approximator is replaced by a liquid residual mechanism.

Inputs undergo a nondimensional affine transformation to yield $\hat{x}\in\hat{\Omega}$, and the network output is written as $f_{\theta}(\hat{x})\in\mathbb{R}^{m}$. Both the baseline PINN and LNN--PINN share the same input preprocessing and final linear readout. We write
\begin{equation}
z_{0}(\hat{x})=\phi_{\mathrm{in}}(\hat{x}), \qquad
f_{\theta}(\hat{x})=W_{\mathrm{out}}\,s_{L}(\hat{x})+b_{\mathrm{out}},
\end{equation}
where $w$ denotes the hidden width, $s_{L}(\hat{x})\in\mathbb{R}^{w}$ is the last hidden representation, and $W_{\mathrm{out}}\in\mathbb{R}^{m\times w}$.

For the baseline MLP--PINN, hidden features are propagated purely by feedforward nonlinear layers:
\begin{equation}
s_{l+1}=\tanh\!\left(W_{l}s_{l}+b_{l}\right), \qquad l=0,\dots,L-1.
\end{equation}
Therefore, the baseline trial map is generated solely by repeated compositions of affine transformations and pointwise $\tanh$ activations.

In the uploaded LNN--PINN implementations, the liquid module is a width-preserving residual operator acting on the current hidden feature. For any $h\in\mathbb{R}^{w}$, we write
\begin{equation}
\mathcal{L}_{l}(h)=h+\mathrm{Diag}(\alpha_{l})\,\tanh\!\left(A_{l}h+c_{l}\right),
\end{equation}
where $\alpha_{l}\in\mathbb{R}^{w}$ is a trainable channel-wise modulation vector, and $A_{l}\in\mathbb{R}^{w\times w}$, $c_{l}\in\mathbb{R}^{w}$ are the internal parameters of the liquid block. Since the correction term lies in $\mathbb{R}^{w}$, the liquid operator preserves the hidden width exactly. In the uploaded codes, the channel-wise modulation vectors are initialized componentwise by
\begin{equation}
\alpha_{l}^{(0)}=0.5\,\mathbf{1}.
\end{equation}

\begin{figure*}[t] 
  \centering
  \includegraphics[width=\textwidth]{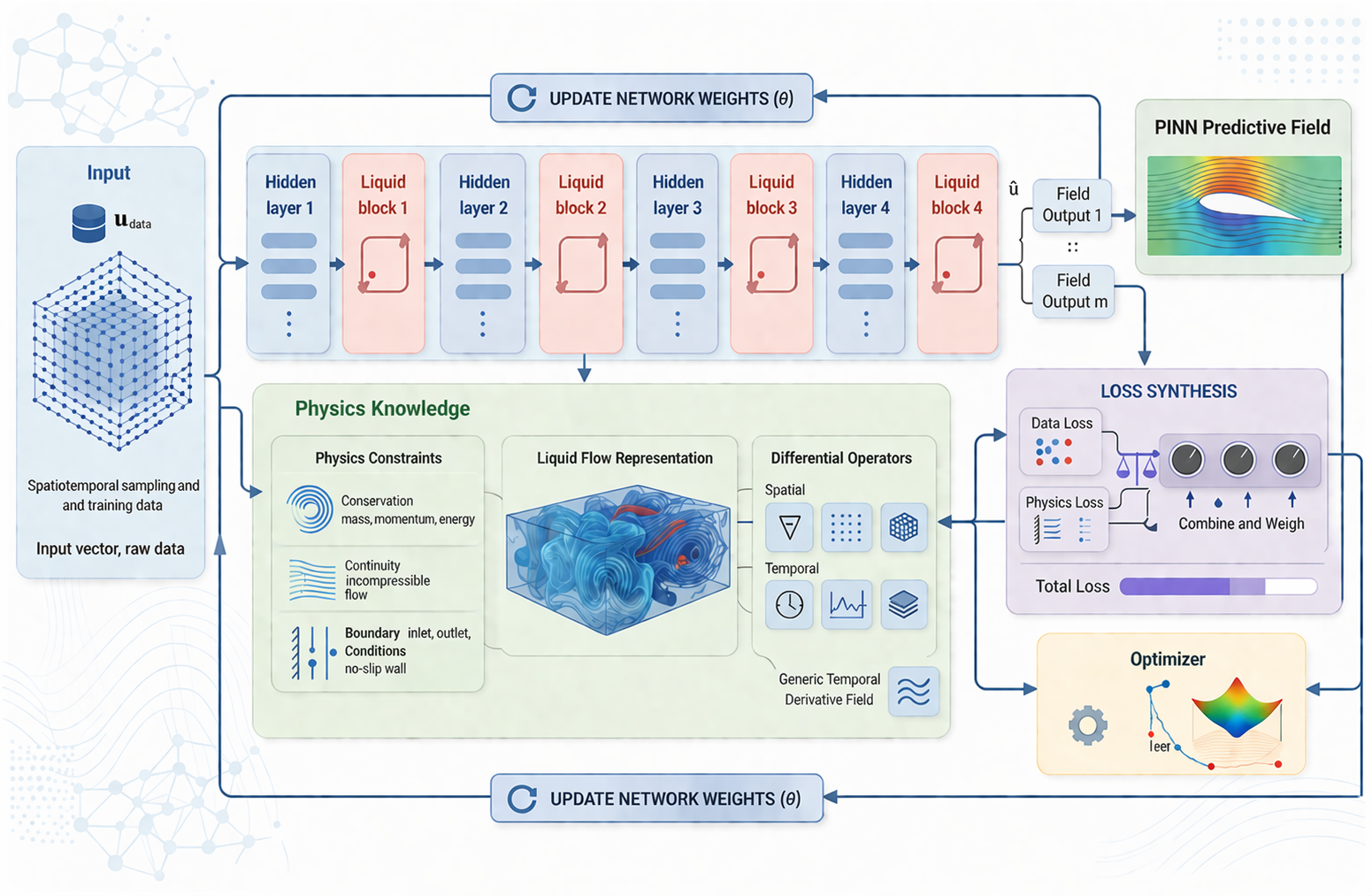} 
 \caption{{Architecture of the LNN--PINN model. The framework preserves the standard physics-only neural solver pipeline while introducing liquid residual blocks as an internal representation mechanism. The network consists of a strictly sequential composition of hidden layers and liquid blocks, where each liquid block performs a width-preserving residual refinement on the latent features generated by the preceding layer. This design isolates architectural enhancement from the governing equations and training formulation, so that the effect of improved hidden-state propagation can be assessed under the same governing equations and training formulation. The resulting model provides a unified and robust mapping from coordinate inputs to physical field predictions across both single-field and multi-field scenarios.}}
  \label{LNN-PINN}
\end{figure*}

Accordingly, LNN--PINN replaces the pure hidden propagation law by a two-step update consisting of a stage map followed by a liquid residual correction. We write
\begin{equation}
\tilde{s}_{l+1}=\varphi_{l}\!\left(W_{l}s_{l}+b_{l}\right), \qquad
s_{l+1}=\mathcal{L}_{l}\!\left(\tilde{s}_{l+1}\right), \qquad l=0,\dots,L-1.
\end{equation}
Here $\varphi_{l}$ denotes the stage activation adopted by the concrete implementation. In the dominant implementation pattern used in the present study, $\varphi_{l}=\tanh$, so that each liquid block is inserted after the corresponding hidden $\tanh$ layer. In the reduced-depth implementation, the same liquid residual operator is retained while the hidden-stage arrangement is shortened. In both cases, the integration principle is identical: the hidden width is preserved, and the liquid block acts as an additive residual refinement on the current latent representation.

The trainable parameter set of LNN--PINN therefore consists of the input map, the output readout, the affine hidden-layer parameters, and the liquid-block parameters. Writing
\begin{equation}
\theta=\Bigl\{\theta_{\mathrm{in}},\theta_{\mathrm{out}},\{W_{l},b_{l},A_{l},c_{l},\alpha_{l}\}_{l=0}^{L-1}\Bigr\},
\end{equation}
all components are optimized jointly through the same physics-only objective introduced in Eq.~\eqref{LOSS Total}, namely
\begin{equation}
\theta^{(t+1)}=\theta^{(t)}-\eta_{t}\nabla_{\theta}\hat{\mathcal{J}}\!\left(\theta^{(t)}\right).
\end{equation}
Thus, the liquid parameters are not fitted by any auxiliary supervision or separate criterion; they are calibrated solely through the back-propagated PDE, boundary, and initial-condition residuals already contained in the original PINN framework.

This formulation also clarifies the precise scope of the architectural replacement. The present implementation does not introduce a separate forgetting gate, an additional state variable with independent recurrence, a layer-wise scalar gate, or an auxiliary projection map. Instead, it replaces the hidden-layer propagation law itself by the width-preserving residual operator defined above. Therefore, the difference between PINN and LNN--PINN lies entirely in the internal parametric solution map, while the external physics-only training structure remains unchanged.

Accordingly, the contribution of LNN--PINN does not lie in constructing a new loss functional or a new optimizer, but in replacing the hidden-state propagation mechanism of the neural trial solution under a fixed physics-only objective. This replacement changes the trainable parametric solution family and the associated gradient transmission path under a controlled comparison setting, while preserving the same governing equations, collocation strategy, residual definitions, and loss construction.}

\section{Numerical Experiments: Method Comparison and Error Assessment}\label{sec3}
{
The loss histories reported in the upper panels of the main-text comparison figures are used here as optimization diagnostics rather than as a one-to-one surrogate for final reconstruction accuracy. This distinction is essential because the plotted training objective is evaluated on resampled stochastic collocation sets during Adam updates, whereas the reported RMSE, MSE, MAE, relative $L_2$ error, and MaxErr are computed afterward on fixed evaluation grids. Accordingly, visible short-scale oscillations in the loss curves do not by themselves imply inferior final field reconstruction. This point is demonstrated case by case in the appendix materials: for the drift--decay benchmark, Appendix~\ref{a} uses Eq.~\eqref{eq:appdd_train_eval_gap} together with Fig.~\ref{fig:appdd_lr}; for the mixed-boundary Laplace benchmark, Appendix~\ref{b} uses Eq.~\eqref{eq:lapapp_gap} together with Fig.~\ref{fig:lapapp_lr}; for the steady circular-heating benchmark, Appendix~\ref{c} uses Eq.~\eqref{eq:heat_app_gap} together with Fig.~\ref{Fapp:heat_lnn_lr}; for the anisotropic Poisson--beam benchmark, Appendix~\ref{d} uses Eq.~\eqref{eq:apbeapp_gap} together with Fig.~\ref{Fapp:apbe_lnn_lr}. For the two strongly coupled eight-field relativistic viscous-fluid benchmarks, Appendices~\ref{e} and~\ref{f} further show that the full six-rate loss histories expose optimization-path variability under the fixed LNN--PINN framework; see Fig.~\ref{Fapp:rel2_lnn_lr} together with Eq.~\eqref{eq:rel2app_samplingrange}, Eq.~\eqref{eq:rel2app_alpha_range_global}, and Eq.~\eqref{eq:rel2app_riemann}, and Fig.~\ref{Fapp:rel3_lnn_lr} together with Eq.~\eqref{eq:rel3app_sampling_ranges}, Eq.~\eqref{eq:rel3app_alpha_range_global}, and Eq.~\eqref{eq:rel3app_riemann}. Therefore, the comparative conclusions in this section are drawn primarily from the reported reconstruction errors and recorded computational cost, while the loss histories are interpreted as trajectory-level optimization information under the shared physics-only training setup\footnote{{Unless otherwise stated, all four compared methods are evaluated under the same benchmark-dependent collocation protocol, the same Adam training horizon, the same main-text learning-rate choice for that benchmark, and the same fixed evaluation grid used for post-training error reporting. The comparison is therefore protocol-matched rather than parameter-matched: the governing equations, residual definitions, sampling strategy, and optimizer remain aligned across methods, while LNN--PINN differs from the plain PINN only through the insertion of width-preserving liquid residual blocks into the hidden-state propagation law. Complete architecture specifications, trainable-parameter counts, learning-rate sweeps, and benchmark-specific implementation details are reported in Appendices~\ref{a}--\ref{f}. All recorded runtimes in the main-text tables are obtained under the same computational environment for the corresponding benchmark runs.}}
\subsection{1D Advection--Reaction (Drift--Decay)}

We first consider the one-dimensional advection--reaction problem on the space--time domain
\[
\Omega_{xt}=[0,2]\times[0,1].
\]
The governing initial--boundary value problem reads
\begin{subequations}\label{eq:dd_problem}
\begin{align}
\partial_t u-\frac{1}{2}\partial_x u+\frac{1}{2}u&=0,
\qquad (x,t)\in(0,2)\times(0,1), \label{eq:dd_pde}\\
u(x,0)&=6e^{-3x},
\qquad x\in[0,2], \label{eq:dd_ic}\\
u(2,t)&=6e^{-6-2t},
\qquad t\in[0,1]. \label{eq:dd_bc}
\end{align}
\end{subequations}
Since the advection velocity equals $-\tfrac{1}{2}<0$, the characteristic direction points from right to left, so the inflow boundary lies at $x=2$, which is exactly the boundary where Eq.~\eqref{eq:dd_bc} is prescribed.

The problem in Eq.~\eqref{eq:dd_problem} admits the analytical solution
\begin{equation}\label{eq:dd_exact}
u^{\ast}(x,t)=6e^{-3x-2t}.
\end{equation}
Indeed,
\begin{equation}\label{eq:dd_derivatives}
\partial_t u^{\ast}(x,t)=-2u^{\ast}(x,t),
\qquad
\partial_x u^{\ast}(x,t)=-3u^{\ast}(x,t).
\end{equation}
Substituting Eq.~\eqref{eq:dd_derivatives} into Eq.~\eqref{eq:dd_pde}, we obtain
\begin{equation}\label{eq:dd_verify_pde}
\partial_t u^{\ast}-\frac{1}{2}\partial_x u^{\ast}+\frac{1}{2}u^{\ast}
=
\left(-2+\frac{3}{2}+\frac{1}{2}\right)u^{\ast}
=0.
\end{equation}
Moreover, evaluating Eq.~\eqref{eq:dd_exact} at $t=0$ and $x=2$ yields
\begin{equation}\label{eq:dd_verify_data}
u^{\ast}(x,0)=6e^{-3x},
\qquad
u^{\ast}(2,t)=6e^{-6-2t},
\end{equation}
which agree with Eq.~\eqref{eq:dd_ic} and \eqref{eq:dd_bc}. Therefore, Eq.~\eqref{eq:dd_exact} solves the whole problem in Eq.~\eqref{eq:dd_problem} exactly.

Following the unified physics-only framework introduced in Section~2, we approximate the unknown field by a neural representation $u_{\theta}(x,t)$. For the present scalar problem, no additional residual rescaling enters the loss construction, so the case-specific objective directly specializes from Eq.~\eqref{LOSS Total}. We sample the interior, initial, and inflow-boundary collocation sets as
\begin{equation}\label{eq:dd_sets}
\mathcal{D}_{\Omega}=\{(x_i,t_i)\}_{i=1}^{N_{\Omega}},
\qquad
\mathcal{D}_{\mathrm{IC}}=\{x_j\}_{j=1}^{N_{\mathrm{IC}}},
\qquad
\mathcal{D}_{\mathrm{BD}}=\{t_k\}_{k=1}^{N_{\mathrm{BD}}},
\end{equation}
where $(x_i,t_i)\in\Omega_{xt}$, $(x_j,0)\in[0,2]\times\{0\}$, and $(2,t_k)\in\{2\}\times[0,1]$. In the present case we set
\[
N_{\Omega}=2000,
\qquad
N_{\mathrm{IC}}=1000,
\qquad
N_{\mathrm{BD}}=1000.
\]

The corresponding residuals read
\begin{subequations}\label{eq:dd_residuals}
\begin{align}
r_{\Omega}(x,t;\theta)
&=
\partial_t u_{\theta}(x,t)-\frac{1}{2}\partial_x u_{\theta}(x,t)+\frac{1}{2}u_{\theta}(x,t), \label{eq:dd_r_omega}\\
r_{\mathrm{IC}}(x;\theta)
&=
u_{\theta}(x,0)-6e^{-3x}, \label{eq:dd_r_ic}\\
r_{\mathrm{BD}}(t;\theta)
&=
u_{\theta}(2,t)-6e^{-6-2t}. \label{eq:dd_r_bd}
\end{align}
\end{subequations}
These residual components enter the unified physics-only objective in Eq.~\eqref{LOSS Total}; for the present time-dependent benchmark, Eq.~\eqref{LOSS Total} is instantiated through the interior, initial-condition, and inflow-boundary contributions evaluated on $\mathcal{D}_{\Omega}$, $\mathcal{D}_{\mathrm{IC}}$, and $\mathcal{D}_{\mathrm{BD}}$, respectively.

\begin{figure}[t]
\centering
\includegraphics[width=1\textwidth]{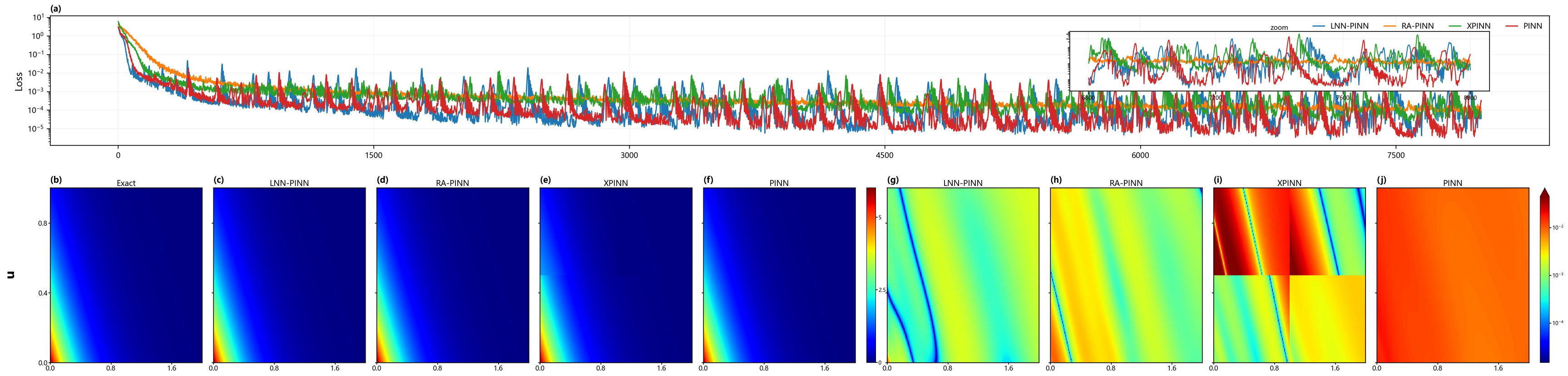}
\caption{{Integrated comparison for the 1D advection--reaction problem. The upper panel reports the loss histories, while the lower panel simultaneously presents the analytical field in Eq.~\eqref{eq:dd_exact}, the reconstructed fields, and the corresponding absolute-error distributions for LNN--PINN, RA--PINN, XPINN, and PINN under the same training budget and the same sampling protocol.}}
\label{F2}
\end{figure}

To complement the visual comparison in Fig.~\ref{F2}, Table~\ref{T2} collects the quantitative error metrics and computational cost for the representative main-text runs. All values are evaluated against the analytical solution in Eq.~\eqref{eq:dd_exact} on the same $200\times 200$ grid, namely $40000$ evaluation points in total, and all four methods use the same training budget of $8000$ Adam iterations with the selected main-text learning rate.

\begin{table}[t]
\centering
\setlength{\tabcolsep}{4.2pt}
\renewcommand{\arraystretch}{1.10}
\caption{{Quantitative comparison for the 1D advection--reaction problem corresponding to Fig.~\ref{F2}. All metrics are computed against the analytical solution in Eq.~\eqref{eq:dd_exact}. The training times are reported in seconds.}}
\label{T2}
\resizebox{\linewidth}{!}{
\begin{tabular}{lccccccc}
\toprule
Method & RMSE & MSE & MAE & rel$L_2$ & MaxErr & Train \\
\midrule
LNN--PINN & $\bm{1.156\times10^{-3}}$ & $\bm{1.337\times10^{-6}}$ & $\bm{9.982\times10^{-4}}$ & $\bm{1.327\times10^{-3}}$ & $1.975\times10^{-2}$ & $332.456$ \\
RA--PINN  & $2.154\times10^{-3}$ & $4.638\times10^{-6}$ & $1.775\times10^{-3}$ & $2.472\times10^{-3}$ & $\bm{1.665\times10^{-2}}$ & $\bm{189.041}$ \\
XPINN     & $2.187\times10^{-2}$ & $4.783\times10^{-4}$ & $8.562\times10^{-3}$ & $2.510\times10^{-2}$ & $4.816\times10^{-1}$ & $313.401$ \\
PINN      & $1.077\times10^{-2}$ & $1.160\times10^{-4}$ & $1.050\times10^{-2}$ & $1.236\times10^{-2}$ & $2.661\times10^{-2}$ & $756.460$ \\
\bottomrule
\end{tabular}
}
\end{table}

Fig.~\ref{F2} and Table~\ref{T2} together show the full picture of this benchmark. Fig.~\ref{F2} simultaneously reports the loss histories, the exact field, the reconstructed fields, and the corresponding absolute-error distributions, while Table~\ref{T2} quantifies the same comparison in terms of reconstruction accuracy and computational cost. Since the exact solution in Eq.~\eqref{eq:dd_exact} preserves a separable exponential drift--decay profile along the characteristic direction, this benchmark directly tests whether the competing models can accurately recover both transport and attenuation under identical physics-only training conditions. LNN--PINN attains the lowest error on RMSE, MSE, MAE, and relative $L_2$ error, with
\[
\mathrm{RMSE}=1.156\times10^{-3},
\qquad
\mathrm{MSE}=1.337\times10^{-6},
\qquad
\mathrm{MAE}=9.982\times10^{-4},
\qquad
\mathrm{rel}L_2=1.327\times10^{-3}.
\]
Relative to the plain PINN, LNN--PINN reduces RMSE, MSE, MAE, relative $L_2$ error, and MaxErr by 89.3\%, 98.8\%, 90.5\%, 89.3\%, and 25.8\%, respectively. Relative to RA--PINN, the corresponding reductions are 46.3\%, 71.2\%, 43.8\%, and 46.3\% for RMSE, MSE, MAE, and relative $L_2$, whereas MaxErr is 18.6\% larger for LNN--PINN than for RA--PINN. Relative to XPINN, the reductions reach 94.7\%, 99.7\%, 88.3\%, 94.7\%, and 95.9\% for RMSE, MSE, MAE, relative $L_2$, and MaxErr, respectively. Therefore, for the drift--decay benchmark, the improvement of LNN--PINN is quantitatively substantial in all global error metrics relative to the plain PINN and XPINN, while RA--PINN remains competitive only in training time and maximum absolute error. The learning-rate comparisons for this case, the mathematical parameter-sensitivity analysis, and the complete final hyperparameter configurations of the adopted models are provided in Appendix~\ref{a}.
\subsection{2D Laplace Equation with Mixed Dirichlet--Neumann Boundary Conditions}

We next consider the two-dimensional Laplace equation for the scalar field $\phi(x,y)$ on the unit square
\[
\Omega=[0,1]\times[0,1].
\]
The governing boundary-value problem reads
\begin{subequations}\label{eq:lap_problem}
\begin{align}
\partial_{xx}\phi+\partial_{yy}\phi&=0,
\qquad (x,y)\in(0,1)\times(0,1), \label{eq:lap_pde}\\
\phi(x,0)&=0,
\qquad x\in[0,1], \label{eq:lap_bc_bottom}\\
\phi(x,1)&=1,
\qquad x\in[0,1], \label{eq:lap_bc_top}\\
\partial_x\phi(0,y)&=0,
\qquad y\in[0,1], \label{eq:lap_bc_left}\\
\partial_x\phi(1,y)&=0,
\qquad y\in[0,1]. \label{eq:lap_bc_right}
\end{align}
\end{subequations}
The bottom and top boundaries impose Dirichlet data, while the left and right boundaries impose homogeneous Neumann data. The problem in Eq.~\eqref{eq:lap_problem} admits the analytical solution
\begin{equation}\label{eq:lap_exact}
\phi^{\ast}(x,y)=y.
\end{equation}
Indeed,
\begin{equation}\label{eq:lap_derivatives}
\partial_x\phi^{\ast}=0,
\qquad
\partial_{xx}\phi^{\ast}=0,
\qquad
\partial_y\phi^{\ast}=1,
\qquad
\partial_{yy}\phi^{\ast}=0.
\end{equation}
Substituting Eq.~\eqref{eq:lap_derivatives} into Eq.~\eqref{eq:lap_pde}, we obtain
\begin{equation}\label{eq:lap_verify_pde}
\partial_{xx}\phi^{\ast}+\partial_{yy}\phi^{\ast}=0.
\end{equation}
Moreover, evaluating Eq.~\eqref{eq:lap_exact} on the four boundaries yields
\begin{equation}\label{eq:lap_verify_bc}
\phi^{\ast}(x,0)=0,
\qquad
\phi^{\ast}(x,1)=1,
\qquad
\partial_x\phi^{\ast}(0,y)=0,
\qquad
\partial_x\phi^{\ast}(1,y)=0,
\end{equation}
which agree with Eqs.~\eqref{eq:lap_bc_bottom}--\eqref{eq:lap_bc_right}. Therefore, Eq.~\eqref{eq:lap_exact} solves the whole problem in Eq.~\eqref{eq:lap_problem} exactly.

Following the unified physics-only framework introduced in Section~2, we approximate the unknown field by a neural representation $\phi_{\theta}(x,y)$. For the present scalar elliptic problem, no additional residual rescaling enters the loss construction, so the case-specific objective directly specializes from Eq.~\eqref{LOSS Total}. We sample one interior collocation set and four boundary collocation sets as
\begin{equation}\label{eq:lap_sets}
\mathcal{D}_{\Omega}=\{(x_i,y_i)\}_{i=1}^{N_{\Omega}},
\qquad
\mathcal{D}_{\mathrm{bot}}=\{x_j\}_{j=1}^{N_{\mathrm{bot}}},
\qquad
\mathcal{D}_{\mathrm{top}}=\{x_k\}_{k=1}^{N_{\mathrm{top}}},
\qquad
\mathcal{D}_{\mathrm{L}}=\{y_{\ell}\}_{\ell=1}^{N_{\mathrm{L}}},
\qquad
\mathcal{D}_{\mathrm{R}}=\{y_m\}_{m=1}^{N_{\mathrm{R}}},
\end{equation}
where $(x_i,y_i)\in\Omega$, $(x_j,0)\in[0,1]\times\{0\}$, $(x_k,1)\in[0,1]\times\{1\}$, $(0,y_{\ell})\in\{0\}\times[0,1]$, and $(1,y_m)\in\{1\}\times[0,1]$. In the present case we set
\[
N_{\Omega}=1000,
\qquad
N_{\mathrm{bot}}=1000,
\qquad
N_{\mathrm{top}}=1000,
\qquad
N_{\mathrm{L}}=1000,
\qquad
N_{\mathrm{R}}=1000.
\]

The corresponding residuals read
\begin{subequations}\label{eq:lap_residuals}
\begin{align}
r_{\Omega}(x,y;\theta)
&=
\partial_{xx}\phi_{\theta}(x,y)+\partial_{yy}\phi_{\theta}(x,y), \label{eq:lap_r_omega}\\
r_{\mathrm{bot}}(x;\theta)
&=
\phi_{\theta}(x,0), \label{eq:lap_r_bot}\\
r_{\mathrm{top}}(x;\theta)
&=
\phi_{\theta}(x,1)-1, \label{eq:lap_r_top}\\
r_{\mathrm{L}}(y;\theta)
&=
\partial_x\phi_{\theta}(0,y), \label{eq:lap_r_left}\\
r_{\mathrm{R}}(y;\theta)
&=
\partial_x\phi_{\theta}(1,y). \label{eq:lap_r_right}
\end{align}
\end{subequations}
These residual components enter Eq.~\eqref{LOSS Total} through one interior contribution and four boundary contributions associated with the bottom, top, left, and right segments, respectively; no additional case-specific total-loss formula is introduced.

\begin{figure}[t]
\centering
\includegraphics[width=\linewidth]{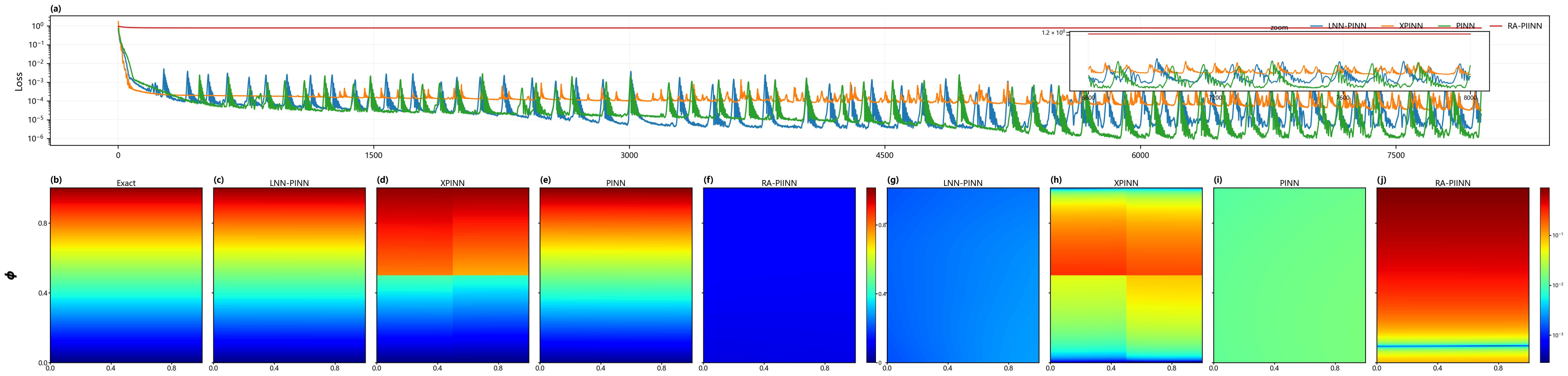}
\caption{{Integrated comparison for the 2D Laplace equation with mixed Dirichlet--Neumann boundary conditions. The upper panel reports the loss histories, while the lower panel simultaneously presents the analytical field in Eq.~\eqref{eq:lap_exact}, the reconstructed fields, and the corresponding absolute-error distributions for LNN--PINN, RA--PINN, XPINN, and PINN under the same training budget and the same sampling protocol.}}
\label{F3}
\end{figure}

To complement the visual comparison in Fig.~\ref{F3}, Table~\ref{T3} collects the quantitative error metrics and computational cost for the representative main-text runs. All values are evaluated against the analytical solution in Eq.~\eqref{eq:lap_exact} on the same $200\times 200$ grid, namely $40000$ evaluation points in total, and all four methods use the same training budget of $8000$ Adam iterations with the selected main-text learning rate.

\begin{table}[t]
\centering
\setlength{\tabcolsep}{4.2pt}
\renewcommand{\arraystretch}{1.10}
\caption{{Quantitative comparison for the 2D Laplace equation corresponding to Fig.~\ref{F3}. All metrics are computed against the analytical solution in Eq.~\eqref{eq:lap_exact}. The training times are reported in seconds.}}
\label{T3}
\resizebox{\linewidth}{!}{
\begin{tabular}{lccccccc}
\toprule
Method & RMSE & MSE & MAE & rel$L_2$ & MaxErr & Train \\
\midrule
LNN--PINN & $\bm{2.058\times10^{-3}}$ & $\bm{4.234\times10^{-6}}$ & $\bm{2.038\times10^{-3}}$ & $\bm{3.560\times10^{-3}}$ & $\bm{2.717\times10^{-3}}$ & $640.891$ \\
RA--PINN  & $4.820\times10^{-1}$ & $2.323\times10^{-1}$ & $4.003\times10^{-1}$ & $8.338\times10^{-1}$ & $8.811\times10^{-1}$ & $\bm{352.542}$ \\
XPINN     & $1.100\times10^{-1}$ & $1.210\times10^{-2}$ & $8.006\times10^{-2}$ & $1.903\times10^{-1}$ & $2.884\times10^{-1}$ & $566.724$ \\
PINN      & $1.367\times10^{-2}$ & $1.870\times10^{-4}$ & $1.364\times10^{-2}$ & $2.365\times10^{-2}$ & $1.553\times10^{-2}$ & $1374.119$ \\
\bottomrule
\end{tabular}
}
\end{table}

Fig.~\ref{F3} and Table~\ref{T3} together show the full picture of this benchmark. Fig.~\ref{F3} simultaneously reports the loss histories, the exact field, the reconstructed fields, and the corresponding absolute-error distributions, while Table~\ref{T3} quantifies the same comparison in terms of reconstruction accuracy and computational cost. Since the exact solution in Eq.~\eqref{eq:lap_exact} is linear in the vertical direction and uniform in the horizontal direction, this benchmark directly tests whether the competing models can faithfully recover a globally constrained harmonic field under mixed Dirichlet--Neumann conditions. LNN--PINN attains the lowest error on RMSE, MSE, MAE, relative $L_2$ error, and maximum absolute error, with
\[
\mathrm{RMSE}=2.058\times10^{-3},
\qquad
\mathrm{MSE}=4.234\times10^{-6},
\qquad
\mathrm{MAE}=2.038\times10^{-3},
\qquad
\mathrm{rel}L_2=3.560\times10^{-3}.
\]
Relative to the plain PINN, LNN--PINN reduces RMSE, MSE, MAE, relative $L_2$ error, and MaxErr by 84.9\%, 97.7\%, 85.1\%, 84.9\%, and 82.5\%, respectively. Relative to RA--PINN, the corresponding reductions are 99.6\%, 100.0\%, 99.5\%, 99.6\%, and 99.7\%, while relative to XPINN they are 98.1\%, 100.0\%, 97.5\%, 98.1\%, and 99.1\%, respectively. Hence the advantage of LNN--PINN in this mixed-boundary Laplace benchmark is not marginal but quantitatively dominant across all reported global error metrics, with the plain PINN remaining the second-most accurate method and RA--PINN and XPINN showing much larger reconstruction errors. The learning-rate comparisons for this case, the mathematical parameter-sensitivity analysis, and the complete final hyperparameter configurations of the adopted models are provided in Appendix~\ref{b}.
\subsection{Steady-State Heating of a Circular Silicon Plate with Convective Boundary}

We next consider steady heat conduction in a circular silicon plate occupying the disk
\[
\Omega_R=\left\{(x,y)\in\mathbb{R}^2:x^2+y^2\le R^2\right\}.
\]
Let $T(x,y)$ denote the temperature field measured relative to the ambient state. The governing boundary-value problem reads
\begin{subequations}\label{eq:heat_problem}
\begin{align}
k\left(\partial_{xx}T+\partial_{yy}T\right)+Q&=0,
\qquad (x,y)\in\Omega_R, \label{eq:heat_pde}\\
-k\,\partial_n T&=h\,T,
\qquad (x,y)\in\partial\Omega_R, \label{eq:heat_bc}
\end{align}
\end{subequations}
where $k$ denotes the thermal conductivity, $Q$ denotes the uniform volumetric heat source, $h$ denotes the convective heat-transfer coefficient, and $\partial_n$ denotes the outward normal derivative on the circular boundary. Since the domain and the loading are both radially symmetric, the solution depends only on
\[
r=\sqrt{x^2+y^2}.
\]
Hence we write
\[
T(x,y)=\mathcal{T}(r).
\]
Substituting this radial ansatz into Eq.~\eqref{eq:heat_pde} yields
\begin{equation}\label{eq:heat_radial}
k\left(\frac{d^2\mathcal{T}}{dr^2}+\frac{1}{r}\frac{d\mathcal{T}}{dr}\right)+Q=0,
\qquad 0<r<R.
\end{equation}
The boundary condition in Eq.~\eqref{eq:heat_bc} becomes
\begin{equation}\label{eq:heat_radial_bc}
-k\frac{d\mathcal{T}}{dr}\bigg|_{r=R}=h\,\mathcal{T}(R).
\end{equation}
Therefore, Eq.~\eqref{eq:heat_radial} and Eq.~\eqref{eq:heat_radial_bc} fully characterize the radial steady-heating problem on the disk.

For this benchmark, we take the high-resolution MATLAB result reported in Appendix~\ref{C} as the reference solution used for the true field in Fig.~\ref{F4} and for all error metrics in Table~\ref{T4}. Appendix~\ref{C} gives the complete numerical construction of that reference field on the circular domain together with the corresponding convergence and consistency checks. We denote that reference field by
\begin{equation}\label{eq:heat_ref}
T^{\mathrm{ref}}(x,y).
\end{equation}
The reference field in Eq.~\eqref{eq:heat_ref} satisfies the same governing problem in Eq.~\eqref{eq:heat_problem}, and we use it as the benchmark truth when evaluating all four learning-based solvers in the main text.

Following the unified physics-only framework introduced in Section~2, we approximate the unknown field by a neural representation $T_{\theta}(x,y)$. We sample one interior collocation set and one boundary collocation set,
\begin{equation}\label{eq:heat_sets}
\mathcal{D}_{\Omega}=\{(x_i,y_i)\}_{i=1}^{N_{\Omega}},
\qquad
\mathcal{D}_{\partial\Omega}=\{(x_j^{(b)},y_j^{(b)},n_{x,j},n_{y,j})\}_{j=1}^{N_{\partial\Omega}},
\end{equation}
where $(x_i,y_i)\in\Omega_R$, $(x_j^{(b)},y_j^{(b)})\in\partial\Omega_R$, and $(n_{x,j},n_{y,j})$ denotes the outward unit normal at the $j$th boundary point. In the present case we set
\[
N_{\Omega}=3000,
\qquad
N_{\partial\Omega}=500.
\]

The corresponding residuals read
\begin{subequations}\label{eq:heat_residuals}
\begin{align}
r_{\Omega}(x,y;\theta)
&=
k\left(\partial_{xx}T_{\theta}(x,y)+\partial_{yy}T_{\theta}(x,y)\right)+Q, \label{eq:heat_r_omega}\\
r_{\partial\Omega}(x,y;\theta)
&=
-k\left(n_x\partial_xT_{\theta}(x,y)+n_y\partial_yT_{\theta}(x,y)\right)-h\,T_{\theta}(x,y). \label{eq:heat_r_bc}
\end{align}
\end{subequations}
These residual components enter Eq.~\eqref{LOSS Total} through the interior PDE contribution and the Robin-boundary contribution on $\partial\Omega$; the fixed scale setting and residual normalization used in this benchmark are specified in Appendix~\ref{A}.

\begin{figure}[t]
\centering
\includegraphics[width=\linewidth]{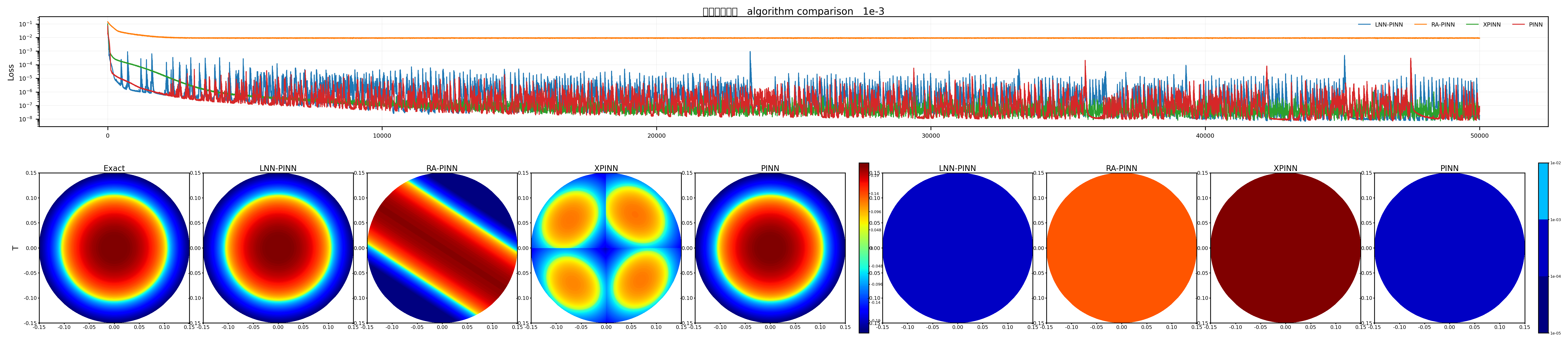}
\caption{{Integrated comparison for the steady-state heating of a circular silicon plate with convective boundary. The upper panel reports the loss histories, while the lower panel simultaneously presents the reference field from Appendix~\ref{C}, the reconstructed fields, and the corresponding absolute-error distributions for LNN--PINN, RA--PINN, XPINN, and PINN under the same training budget and the same sampling protocol.}}
\label{F4}
\end{figure}

To complement the visual comparison in Fig.~\ref{F4}, Table~\ref{T4} collects the quantitative error metrics and computational cost for the representative main-text runs. We compute all reported errors against the MATLAB reference field in Appendix~\ref{C}. All four methods use the same training budget of $50000$ Adam iterations with the selected main-text learning rate. The training times are reported in seconds.

\begin{table}[t]
\centering
\setlength{\tabcolsep}{4.2pt}
\renewcommand{\arraystretch}{1.10}
\caption{{Quantitative comparison for the steady-state heating of a circular silicon plate corresponding to Fig.~\ref{F4}. All metrics are computed against the MATLAB reference solution reported in Appendix~\ref{C}. The training times are reported in seconds.}}
\label{T4}
\resizebox{\linewidth}{!}{
\begin{tabular}{lcccccc}
\toprule
Method & RMSE & MSE & MAE & rel$L_2$ & MaxErr & Train \\
\midrule
LNN--PINN & $\bm{1.9785\times10^{-4}}$ & $\bm{3.9145\times10^{-8}}$ & $\bm{1.9406\times10^{-4}}$ & $\bm{2.4638\times10^{-7}}$ & $\bm{3.0518\times10^{-4}}$ & $1.3880\times10^{4}$ \\
RA--PINN  & $3.3818\times10^{-1}$ & $1.1437\times10^{-1}$ & $3.3717\times10^{-1}$ & $4.2113\times10^{-4}$ & $3.9807\times10^{-1}$ & $7.1440\times10^{2}$ \\
XPINN     & $1.6154\times10^{0}$ & $2.6094\times10^{0}$ & $1.6152\times10^{0}$ & $2.0116\times10^{-3}$ & $1.6778\times10^{0}$ & $1.9196\times10^{3}$ \\
PINN      & $7.0219\times10^{-4}$ & $4.9307\times10^{-7}$ & $6.9302\times10^{-4}$ & $8.7442\times10^{-7}$ & $9.1553\times10^{-4}$ & $\bm{3.3196\times10^{2}}$ \\
\bottomrule
\end{tabular}
}
\end{table}

Fig.~\ref{F4} and Table~\ref{T4} together show the full picture of this benchmark. Fig.~\ref{F4} simultaneously reports the loss histories, the reference field, the reconstructed fields, and the corresponding absolute-error distributions, while Table~\ref{T4} quantifies the same comparison in terms of reconstruction accuracy and computational cost. Since the reference solution in Appendix~\ref{C} exhibits a smooth radially symmetric temperature profile determined by the balance between uniform volumetric heating and convective boundary cooling, this benchmark directly tests whether the competing models can preserve both the interior curvature and the boundary heat-flux balance on a curved domain. LNN--PINN attains the lowest error on every reported accuracy metric, with
\[
\mathrm{RMSE}=1.9785\times10^{-4},
\qquad
\mathrm{MSE}=3.9145\times10^{-8},
\qquad
\mathrm{MAE}=1.9406\times10^{-4},
\qquad
\mathrm{rel}L_2=2.4638\times10^{-7}.
\]
Relative to the plain PINN, LNN--PINN reduces RMSE, MSE, MAE, relative $L_2$ error, and MaxErr by 71.8\%, 92.1\%, 72.0\%, 71.8\%, and 66.7\%, respectively. Relative to RA--PINN, the corresponding reductions are 99.9\%, 100.0\%, 99.9\%, 99.9\%, and 99.9\%, and relative to XPINN they are 100.0\%, 100.0\%, 100.0\%, 100.0\%, and 100.0\% up to the displayed precision. Therefore, for the steady heating problem on the circular domain, LNN--PINN provides a quantitatively strong improvement over the plain PINN and an overwhelming improvement over RA--PINN and XPINN across all reported accuracy metrics. The learning-rate comparisons for this case, the mathematical parameter-sensitivity analysis, and the complete final hyperparameter configurations of the adopted models are provided in Appendix~\ref{c}.
\subsection{Anisotropic Poisson--Beam Equation}

We next consider a two-dimensional anisotropic Poisson--beam equation for the scalar field $u(x,y)$ on the unit square
\[
\Omega=[0,1]\times[0,1].
\]
The governing boundary-value problem reads
\begin{subequations}\label{eq:apbe_problem}
\begin{align}
\partial_{xx}u-\partial_{yyyy}u&=(2-x^2)e^{-y},
\qquad (x,y)\in(0,1)\times(0,1), \label{eq:apbe_pde}\\
u(x,0)&=x^2,
\qquad x\in[0,1], \label{eq:apbe_bc_y0_u}\\
\partial_{yy}u(x,0)&=x^2,
\qquad x\in[0,1], \label{eq:apbe_bc_y0_yy}\\
u(x,1)&=x^2e^{-1},
\qquad x\in[0,1], \label{eq:apbe_bc_y1_u}\\
\partial_{yy}u(x,1)&=x^2e^{-1},
\qquad x\in[0,1], \label{eq:apbe_bc_y1_yy}\\
u(0,y)&=0,
\qquad y\in[0,1], \label{eq:apbe_bc_x0}\\
u(1,y)&=e^{-y},
\qquad y\in[0,1]. \label{eq:apbe_bc_x1}
\end{align}
\end{subequations}
The operator in Eq.~\eqref{eq:apbe_pde} combines a second-order term in the $x$ direction with a fourth-order term in the $y$ direction, so this benchmark tests whether the competing models can simultaneously resolve anisotropic differential orders and multiple boundary constraints.

The problem in Eq.~\eqref{eq:apbe_problem} admits the analytical solution
\begin{equation}\label{eq:apbe_exact}
u^{\ast}(x,y)=x^2e^{-y}.
\end{equation}
Indeed,
\begin{equation}\label{eq:apbe_derivatives}
\partial_{xx}u^{\ast}(x,y)=2e^{-y},
\qquad
\partial_{yy}u^{\ast}(x,y)=x^2e^{-y},
\qquad
\partial_{yyyy}u^{\ast}(x,y)=x^2e^{-y}.
\end{equation}
Substituting Eq.~\eqref{eq:apbe_derivatives} into Eq.~\eqref{eq:apbe_pde}, we obtain
\begin{equation}\label{eq:apbe_verify_pde}
\partial_{xx}u^{\ast}-\partial_{yyyy}u^{\ast}
=
2e^{-y}-x^2e^{-y}
=
(2-x^2)e^{-y},
\end{equation}
which coincides with the right-hand side of Eq.~\eqref{eq:apbe_pde}. Moreover, evaluating Eq.~\eqref{eq:apbe_exact} and Eq.~\eqref{eq:apbe_derivatives} on the boundaries gives
\begin{equation}\label{eq:apbe_verify_bc}
u^{\ast}(x,0)=x^2,
\qquad
\partial_{yy}u^{\ast}(x,0)=x^2,
\qquad
u^{\ast}(x,1)=x^2e^{-1},
\qquad
\partial_{yy}u^{\ast}(x,1)=x^2e^{-1},
\qquad
u^{\ast}(0,y)=0,
\qquad
u^{\ast}(1,y)=e^{-y},
\end{equation}
which agree with Eqs.~\eqref{eq:apbe_bc_y0_u}--\eqref{eq:apbe_bc_x1}. Therefore, Eq.~\eqref{eq:apbe_exact} solves the whole problem in Eq.~\eqref{eq:apbe_problem} exactly.

Following the unified physics-only framework introduced in Section~2, we approximate the unknown field by a neural representation $u_{\theta}(x,y)$. For the present scalar anisotropic problem, no additional residual rescaling enters the loss construction, so the case-specific objective directly specializes from Eq.~\eqref{LOSS Total}. We sample one interior collocation set together with six boundary collocation sets,
\begin{equation}\label{eq:apbe_sets}
\mathcal{D}_{\Omega}=\{(x_i,y_i)\}_{i=1}^{N_{\Omega}},
\qquad
\mathcal{D}_{y=0}^{(u)}=\{x_j\}_{j=1}^{N_{0,u}},
\qquad
\mathcal{D}_{y=0}^{(yy)}=\{x_k\}_{k=1}^{N_{0,yy}},
\qquad
\mathcal{D}_{y=1}^{(u)}=\{x_{\ell}\}_{\ell=1}^{N_{1,u}},
\end{equation}
\begin{equation}\label{eq:apbe_sets2}
\mathcal{D}_{y=1}^{(yy)}=\{x_m\}_{m=1}^{N_{1,yy}},
\qquad
\mathcal{D}_{x=0}=\{y_n\}_{n=1}^{N_{x0}},
\qquad
\mathcal{D}_{x=1}=\{y_p\}_{p=1}^{N_{x1}},
\end{equation}
where $(x_i,y_i)\in\Omega$, $(x_j,0)\in[0,1]\times\{0\}$, $(x_k,0)\in[0,1]\times\{0\}$, $(x_{\ell},1)\in[0,1]\times\{1\}$, $(x_m,1)\in[0,1]\times\{1\}$, $(0,y_n)\in\{0\}\times[0,1]$, and $(1,y_p)\in\{1\}\times[0,1]$. In the present case we set
\[
N_{\Omega}=1000,
\qquad
N_{0,u}=N_{0,yy}=N_{1,u}=N_{1,yy}=N_{x0}=N_{x1}=1000.
\]

The corresponding residuals read
\begin{subequations}\label{eq:apbe_residuals}
\begin{align}
r_{\Omega}(x,y;\theta)
&=
\partial_{xx}u_{\theta}(x,y)-\partial_{yyyy}u_{\theta}(x,y)-(2-x^2)e^{-y}, \label{eq:apbe_r_omega}\\
r_{y=0}^{(u)}(x;\theta)
&=
u_{\theta}(x,0)-x^2, \label{eq:apbe_r_y0_u}\\
r_{y=0}^{(yy)}(x;\theta)
&=
\partial_{yy}u_{\theta}(x,0)-x^2, \label{eq:apbe_r_y0_yy}\\
r_{y=1}^{(u)}(x;\theta)
&=
u_{\theta}(x,1)-x^2e^{-1}, \label{eq:apbe_r_y1_u}\\
r_{y=1}^{(yy)}(x;\theta)
&=
\partial_{yy}u_{\theta}(x,1)-x^2e^{-1}, \label{eq:apbe_r_y1_yy}\\
r_{x=0}(y;\theta)
&=
u_{\theta}(0,y), \label{eq:apbe_r_x0}\\
r_{x=1}(y;\theta)
&=
u_{\theta}(1,y)-e^{-y}. \label{eq:apbe_r_x1}
\end{align}
\end{subequations}
These residual components enter Eq.~\eqref{LOSS Total} through one interior contribution together with the boundary contributions associated with the present value and derivative constraints; we therefore do not restate a separate case-level total-loss formula.

\begin{figure}[t]
\centering
\includegraphics[width=\linewidth]{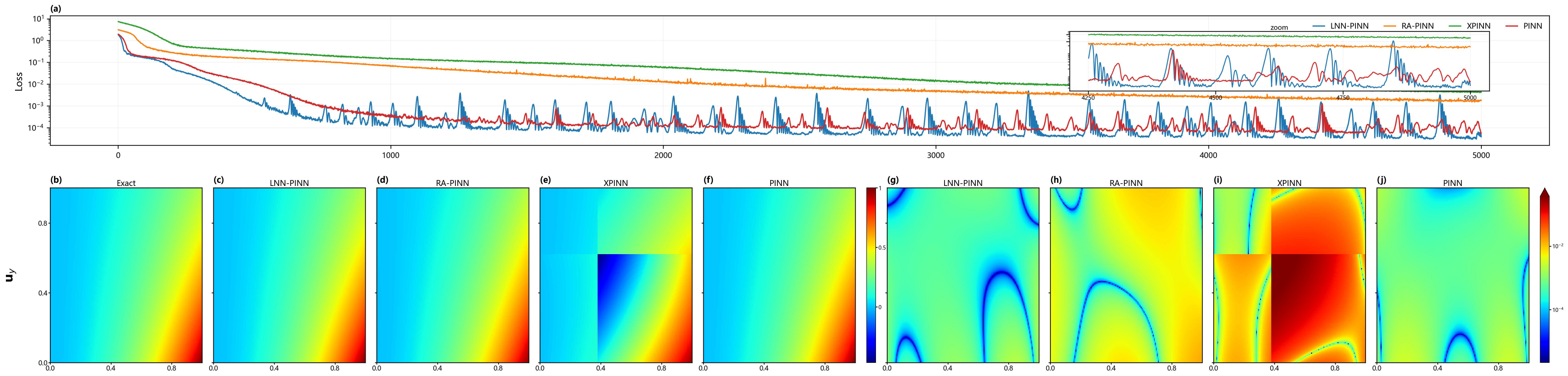}
\caption{{Integrated comparison for the anisotropic Poisson--beam equation. The upper panel reports the loss histories, while the lower panel simultaneously presents the analytical field in Eq.~\eqref{eq:apbe_exact}, the reconstructed fields, and the corresponding absolute-error distributions for LNN--PINN, RA--PINN, XPINN, and PINN under the same training budget and the same sampling protocol.}}
\label{F5}
\end{figure}

To complement the visual comparison in Fig.~\ref{F5}, Table~\ref{T5} collects the quantitative error metrics and computational cost for the representative main-text runs. All four methods use the same training budget of $5000$ Adam iterations with the selected main-text learning rate. 

\begin{table}[t]
\centering
\setlength{\tabcolsep}{4.2pt}
\renewcommand{\arraystretch}{1.10}
\caption{{Quantitative comparison for the anisotropic Poisson--beam equation corresponding to Fig.~\ref{F5}. All metrics are computed against the analytical solution in Eq.~\eqref{eq:apbe_exact}. The training times are reported in seconds.}}
\label{T5}
\resizebox{\linewidth}{!}{
\begin{tabular}{lccccccc}
\toprule
Method & RMSE & MSE & MAE & rel$L_2$ & MaxErr & Train \\
\midrule
LNN--PINN & $\bm{5.529\times10^{-4}}$ & $\bm{3.057\times10^{-7}}$ & $\bm{4.650\times10^{-4}}$ & $\bm{1.872\times10^{-3}}$ & $\bm{2.008\times10^{-3}}$ & $1799.516$ \\
RA--PINN  & $4.173\times10^{-3}$ & $1.741\times10^{-5}$ & $3.342\times10^{-3}$ & $1.410\times10^{-2}$ & $1.472\times10^{-2}$ & $1186.230$ \\
XPINN     & $1.118\times10^{-1}$ & $1.251\times10^{-2}$ & $5.892\times10^{-2}$ & $3.780\times10^{-1}$ & $5.478\times10^{-1}$ & $\bm{401.599}$ \\
PINN      & $9.643\times10^{-4}$ & $9.298\times10^{-7}$ & $8.045\times10^{-4}$ & $3.259\times10^{-3}$ & $3.067\times10^{-3}$ & $2585.032$ \\
\bottomrule
\end{tabular}
}
\end{table}

Fig.~\ref{F5} and Table~\ref{T5} together show the full picture of this benchmark. Fig.~\ref{F5} simultaneously reports the loss histories, the exact field, the reconstructed fields, and the corresponding absolute-error distributions, while Table~\ref{T5} quantifies the same comparison in terms of reconstruction accuracy and computational cost. Since the exact solution in Eq.~\eqref{eq:apbe_exact} couples a quadratic polynomial in $x$ with an exponential decay in $y$, and since Eq.~\eqref{eq:apbe_pde} combines second-order and fourth-order derivatives along different directions, this benchmark directly tests whether the competing models can maintain accuracy under strong operator anisotropy and mixed boundary constraints. LNN--PINN attains the lowest error on every reported accuracy metric, with
\[
\mathrm{RMSE}=5.529\times10^{-4},
\qquad
\mathrm{MSE}=3.057\times10^{-7},
\qquad
\mathrm{MAE}=4.650\times10^{-4},
\qquad
\mathrm{rel}L_2=1.872\times10^{-3}.
\]
Relative to the plain PINN, LNN--PINN reduces RMSE, MSE, MAE, relative $L_2$ error, and MaxErr by 42.7\%, 67.1\%, 42.2\%, 42.6\%, and 34.5\%, respectively. Relative to RA--PINN, the corresponding reductions are 86.8\%, 98.2\%, 86.1\%, 86.7\%, and 86.4\%, while relative to XPINN they are 99.5\%, 100.0\%, 99.2\%, 99.5\%, and 99.6\%, respectively. Hence the anisotropic Poisson--beam benchmark shows a moderate but still clear improvement relative to the plain PINN and a much stronger improvement relative to RA--PINN and XPINN, which is consistent across all reported scalar error indicators. The learning-rate comparisons for this case, the mathematical parameter-sensitivity analysis, and the complete final hyperparameter configurations of the adopted models are provided in Appendix~\ref{d}.
\subsection{2D Steady Relativistic Viscous Fluid Benchmark}

We next consider a two-dimensional steady relativistic viscous-fluid system on the unit square
\[
\Omega=[0,1]\times[0,1].
\]
Let the principal unknown fields be
\[
\mathbf{q}(x,y)=\big(v_x,v_y,\epsilon,n,\Pi,\pi_{xx},\pi_{xy},\pi_{yy}\big)^{\top}.
\]
The governing conservation--relaxation system reads
\begin{subequations}\label{eq:relB_system}
\begin{align}
\partial_x T^{x0}+\partial_y T^{y0}&=S_E(x,y), \label{eq:relB_energy}\\
\partial_x T^{xx}+\partial_y T^{yx}&=S_{Mx}(x,y), \label{eq:relB_mx}\\
\partial_x T^{xy}+\partial_y T^{yy}&=S_{My}(x,y), \label{eq:relB_my}\\
\partial_x N^{x}+\partial_y N^{y}&=S_n(x,y), \label{eq:relB_n}\\
\tau_{\Pi}D\Pi+\Pi+\zeta\theta&=S_{\Pi}(x,y), \label{eq:relB_Pi}\\
\tau_{\pi}D\pi_{xx}+\pi_{xx}-2\eta\sigma^{xx}&=S_{\pi xx}(x,y), \label{eq:relB_pixx}\\
\tau_{\pi}D\pi_{xy}+\pi_{xy}-2\eta\sigma^{xy}&=S_{\pi xy}(x,y), \label{eq:relB_pixy}\\
\tau_{\pi}D\pi_{yy}+\pi_{yy}-2\eta\sigma^{yy}&=S_{\pi yy}(x,y). \label{eq:relB_piyy}
\end{align}
\end{subequations}
Here
\begin{subequations}\label{eq:relB_constitutive}
\begin{align}
u^{\mu}&=\gamma(1,v_x,v_y,0), \qquad \gamma=\frac{1}{\sqrt{1-v_x^2-v_y^2}}, \label{eq:relB_u}\\
D&=u^x\partial_x+u^y\partial_y, \label{eq:relB_D}\\
N^{\mu}&=nu^{\mu}, \label{eq:relB_N}\\
T^{\mu\nu}&=(\epsilon+p+\Pi)u^{\mu}u^{\nu}+(p+\Pi)g^{\mu\nu}+\pi^{\mu\nu}, \label{eq:relB_T}\\
p&=(\Gamma_{\mathrm{ad}}-1)(\epsilon-mn), \label{eq:relB_p}\\
\Delta^{\mu\nu}&=g^{\mu\nu}+u^{\mu}u^{\nu}, \label{eq:relB_Delta}\\
\theta&=\partial_xu^x+\partial_yu^y, \label{eq:relB_theta}\\
\sigma^{\mu\nu}&=\Delta^{\mu\alpha}\Delta^{\nu\beta}\frac{\partial_{\alpha}u_{\beta}+\partial_{\beta}u_{\alpha}}{2}-\frac{1}{3}\Delta^{\mu\nu}\theta. \label{eq:relB_sigma}
\end{align}
\end{subequations}
We take
\begin{equation}\label{eq:relB_params}
\Gamma_{\mathrm{ad}}=1.4,
\qquad
m=0.8,
\qquad
\eta=0.06,
\qquad
\zeta=0.04,
\qquad
\tau_{\pi}=0.25,
\qquad
\tau_{\Pi}=0.22.
\end{equation}
On the whole boundary we prescribe the traces of the analytical state,
\begin{subequations}\label{eq:relB_bc}
\begin{align}
\mathbf{q}(x,0)&=\mathbf{q}^{\ast}(x,0),
\qquad
\mathbf{q}(x,1)=\mathbf{q}^{\ast}(x,1), \label{eq:relB_bc_tb}\\
\mathbf{q}(0,y)&=\mathbf{q}^{\ast}(0,y),
\qquad
\mathbf{q}(1,y)=\mathbf{q}^{\ast}(1,y). \label{eq:relB_bc_lr}
\end{align}
\end{subequations}

The system in Eq.~\eqref{eq:relB_system} admits the analytical state
\begin{subequations}\label{eq:relB_exact}
\begin{align}
v_x^{\ast}(x,y)&=A_{vx}\sin(\pi x)\sin(\pi y)\bigl(1+0.25\cos(\pi x)\bigr), \label{eq:relB_vx}\\
v_y^{\ast}(x,y)&=A_{vy}\sin(\pi x)\sin(\pi y)\bigl(1-0.25\cos(\pi y)\bigr), \label{eq:relB_vy}\\
\epsilon^{\ast}(x,y)&=E_0+A_{\epsilon 1}\cos(2\pi x)\sin(\pi y)+A_{\epsilon 2}\sin(\pi x)\cos(2\pi y), \label{eq:relB_eps}\\
n^{\ast}(x,y)&=N_0+A_{n1}\sin(2\pi x)\sin(\pi y)+A_{n2}\cos(\pi x)\sin(2\pi y), \label{eq:relB_n_exact}\\
\Pi^{\ast}(x,y)&=A_{\Pi}\Bigl[\sin(\pi x)\cos(2\pi y)+0.30\cos(2\pi x)\sin(\pi y)\Bigr], \label{eq:relB_Pi_exact}\\
\pi_{xx}^{\ast}(x,y)&=A_{xx}\Bigl[\cos(2\pi x)\sin(\pi y)+0.25\sin(\pi x)\sin(2\pi y)\Bigr], \label{eq:relB_pixx_exact}\\
\pi_{xy}^{\ast}(x,y)&=A_{xy}\Bigl[\sin(2\pi x)\sin(2\pi y)+0.20\cos(\pi x)\sin(\pi y)\Bigr], \label{eq:relB_pixy_exact}\\
\pi_{yy}^{\ast}(x,y)&=A_{yy}\Bigl[\sin(\pi x)\cos(2\pi y)-0.20\cos(2\pi x)\sin(\pi y)\Bigr]. \label{eq:relB_piyy_exact}
\end{align}
\end{subequations}
The coefficients are
\begin{equation}\label{eq:relB_coeffs1}
A_{vx}=0.14,\quad
A_{vy}=0.12,\quad
E_0=2.35,\quad
A_{\epsilon 1}=0.18,\quad
A_{\epsilon 2}=0.12,\quad
N_0=1.55,\quad
A_{n1}=0.10,\quad
A_{n2}=0.08,
\end{equation}
and
\begin{equation}\label{eq:relB_coeffs2}
A_{\Pi}=0.045,\qquad
A_{xx}=0.038,\qquad
A_{xy}=0.028,\qquad
A_{yy}=0.032.
\end{equation}
The analytical velocity remains subluminal throughout $\Omega$ because
\begin{equation}\label{eq:relB_velocity_bound}
|v_x^{\ast}(x,y)|\leq 1.25A_{vx}=0.175,
\qquad
|v_y^{\ast}(x,y)|\leq 1.25A_{vy}=0.150,
\end{equation}
which implies
\begin{equation}\label{eq:relB_subluminal}
(v_x^{\ast})^2+(v_y^{\ast})^2<1.
\end{equation}
Moreover,
\begin{equation}\label{eq:relB_positive_bounds}
\epsilon^{\ast}(x,y)\geq E_0-A_{\epsilon 1}-A_{\epsilon 2}=2.05,
\qquad
n^{\ast}(x,y)\geq N_0-A_{n1}-A_{n2}=1.37,
\end{equation}
so the energy density and particle density remain strictly positive.

Substituting Eq.~\eqref{eq:relB_exact} into Eq.~\eqref{eq:relB_constitutive} yields the analytical four-velocity $u^{\mu\ast}$, particle current $N^{\mu\ast}$, energy--momentum tensor $T^{\mu\nu\ast}$, expansion scalar $\theta^{\ast}$, and shear tensor $\sigma^{\mu\nu\ast}$. The source functions in Eq.~\eqref{eq:relB_system} are then exactly the smooth functions induced by that analytical state:
\begin{subequations}\label{eq:relB_sources}
\begin{align}
S_E(x,y)&=\partial_x T^{x0\ast}+\partial_y T^{y0\ast}, \label{eq:relB_SE}\\
S_{Mx}(x,y)&=\partial_x T^{xx\ast}+\partial_y T^{yx\ast}, \label{eq:relB_SMx}\\
S_{My}(x,y)&=\partial_x T^{xy\ast}+\partial_y T^{yy\ast}, \label{eq:relB_SMy}\\
S_n(x,y)&=\partial_x N^{x\ast}+\partial_y N^{y\ast}, \label{eq:relB_Sn}\\
S_{\Pi}(x,y)&=\tau_{\Pi}D^{\ast}\Pi^{\ast}+\Pi^{\ast}+\zeta\theta^{\ast}, \label{eq:relB_SPi}\\
S_{\pi xx}(x,y)&=\tau_{\pi}D^{\ast}\pi_{xx}^{\ast}+\pi_{xx}^{\ast}-2\eta\sigma^{xx\ast}, \label{eq:relB_Spixx}\\
S_{\pi xy}(x,y)&=\tau_{\pi}D^{\ast}\pi_{xy}^{\ast}+\pi_{xy}^{\ast}-2\eta\sigma^{xy\ast}, \label{eq:relB_Spixy}\\
S_{\pi yy}(x,y)&=\tau_{\pi}D^{\ast}\pi_{yy}^{\ast}+\pi_{yy}^{\ast}-2\eta\sigma^{yy\ast}. \label{eq:relB_Spiyy}
\end{align}
\end{subequations}
Therefore, once Eq.~\eqref{eq:relB_exact} and Eq.~\eqref{eq:relB_sources} are inserted into Eq.~\eqref{eq:relB_system}, each governing equation is satisfied identically on $\Omega$. The full source expansions, the learning-rate comparisons, the mathematical parameter-sensitivity analysis, and the final hyperparameter configurations are deferred to the Appendix~\ref{e}.

Following the unified physics-only framework introduced in Section~2, we approximate the state vector by
\[
\mathbf{q}_{\theta}(x,y)=\big(v_{x,\theta},v_{y,\theta},\epsilon_{\theta},n_{\theta},\Pi_{\theta},\pi_{xx,\theta},\pi_{xy,\theta},\pi_{yy,\theta}\big)^{\top}.
\]
We sample one interior collocation set and four edge collocation sets,
\begin{equation}\label{eq:relB_sets}
\mathcal{D}_{\Omega}=\{(x_i,y_i)\}_{i=1}^{N_{\Omega}},
\qquad
\mathcal{D}_{\mathrm{bot}}=\{x_j\}_{j=1}^{N_{\mathrm{bot}}},
\qquad
\mathcal{D}_{\mathrm{top}}=\{x_k\}_{k=1}^{N_{\mathrm{top}}},
\qquad
\mathcal{D}_{\mathrm{L}}=\{y_{\ell}\}_{\ell=1}^{N_{\mathrm{L}}},
\qquad
\mathcal{D}_{\mathrm{R}}=\{y_m\}_{m=1}^{N_{\mathrm{R}}},
\end{equation}
with
\[
N_{\Omega}=1000,
\qquad
N_{\mathrm{bot}}=N_{\mathrm{top}}=N_{\mathrm{L}}=N_{\mathrm{R}}=1000.
\]
For the present eight-field system, we write the interior residual channels compactly as
\begin{equation*}
\mathbf{r}_{\Omega}^{\mathrm{relB}}(x,y;\theta)
=
\begin{pmatrix}
\partial_x T_{\theta}^{x0}+\partial_y T_{\theta}^{y0}-S_E(x,y)\\
\partial_x T_{\theta}^{xx}+\partial_y T_{\theta}^{yx}-S_{Mx}(x,y)\\
\partial_x T_{\theta}^{xy}+\partial_y T_{\theta}^{yy}-S_{My}(x,y)\\
\partial_x N_{\theta}^{x}+\partial_y N_{\theta}^{y}-S_n(x,y)\\
\tau_{\Pi}D_{\theta}\Pi_{\theta}+\Pi_{\theta}+\zeta\theta_{\theta}-S_{\Pi}(x,y)\\
\tau_{\pi}D_{\theta}\pi_{xx,\theta}+\pi_{xx,\theta}-2\eta\sigma_{\theta}^{xx}-S_{\pi xx}(x,y)\\
\tau_{\pi}D_{\theta}\pi_{xy,\theta}+\pi_{xy,\theta}-2\eta\sigma_{\theta}^{xy}-S_{\pi xy}(x,y)\\
\tau_{\pi}D_{\theta}\pi_{yy,\theta}+\pi_{yy,\theta}-2\eta\sigma_{\theta}^{yy}-S_{\pi yy}(x,y)
\end{pmatrix},
\end{equation*}
where $T_{\theta}^{\mu\nu}$, $N_{\theta}^{\mu}$, $D_{\theta}$, $\theta_{\theta}$, and $\sigma_{\theta}^{\mu\nu}$ are obtained by inserting $\mathbf{q}_{\theta}$ into the constitutive relations in Eqs.~\eqref{eq:relB_constitutive}. The boundary residual is defined by the trace mismatch
\begin{equation*}
\mathbf{r}_{\partial}^{\mathrm{relB}}(x,y;\theta)
=
\mathbf{q}_{\theta}(x,y)-\mathbf{q}^{\ast}(x,y),
\qquad (x,y)\in \partial\Omega,
\end{equation*}
with the four edge sample sets in Eq.~\eqref{eq:relB_sets} providing the discrete boundary contributions on the bottom, top, left, and right edges, respectively. These residual channels enter the unified physics-only objective in Eq.~\eqref{LOSS Total}; we therefore specialize the benchmark at the residual level and do not restate a separate case-level total-loss formula.

\begin{figure}[t]
\centering
\includegraphics[width=\linewidth]{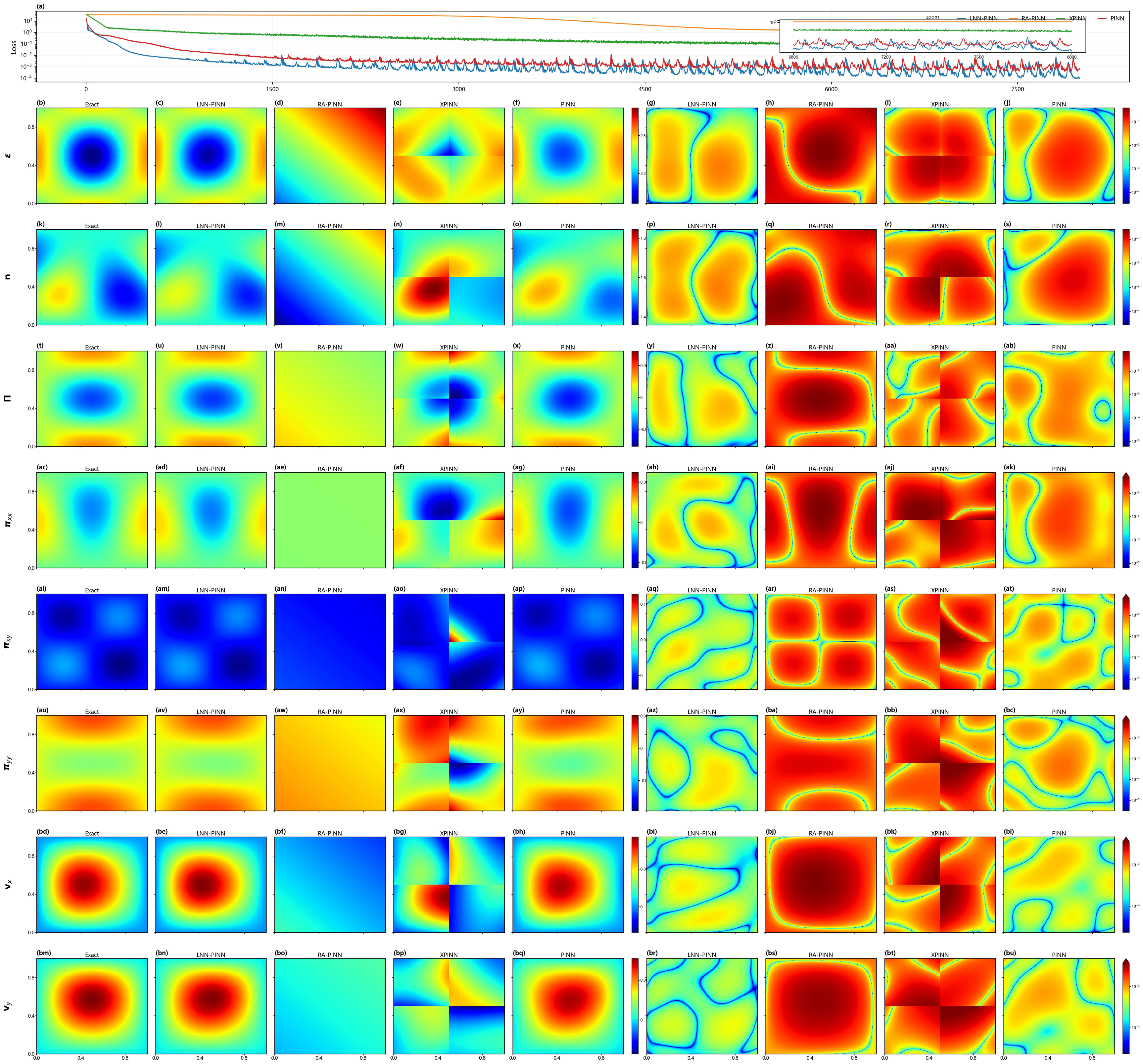}
\caption{{Integrated comparison for the 2D steady relativistic viscous-fluid benchmark. The upper panel reports the loss histories, while the lower panel simultaneously presents the analytical state in Eq.~\eqref{eq:relB_exact}, the reconstructed fields, and the corresponding absolute-error distributions for LNN--PINN, RA--PINN, XPINN, and PINN under the same training protocol.}}
\label{F6}
\end{figure}

To complement the visual comparison in Fig.~\ref{F6}, Table~\ref{T6} reports the field-wise error metrics and the computational cost for the representative main-text runs. For each method, the training time is repeated across all field rows only for compact presentation.

\begin{table*}[t]
\centering
\scriptsize
\setlength{\tabcolsep}{3.2pt}
\renewcommand{\arraystretch}{1.08}
\caption{{Field-wise quantitative comparison for the 2D steady relativistic viscous-fluid benchmark corresponding to Fig.~\ref{F6}. All metrics are computed field by field against the analytical state in Eq.~\eqref{eq:relB_exact}. The training times are reported in seconds.}}
\label{T6}
\resizebox{\textwidth}{!}{
\begin{tabular}{llccccccc}
\toprule
Field & Method & RMSE & MSE & MAE & rel$L_2$ & MaxErr & Train \\
\midrule
$v_x$ & LNN--PINN & $\bm{1.8907\times10^{-3}}$ & $\bm{3.5749\times10^{-6}}$ & $\bm{1.4565\times10^{-3}}$ & $\bm{2.6937\times10^{-2}}$ & $\bm{4.5969\times10^{-3}}$ & 1502.7868 \\
$v_x$ & RA--PINN  & $6.5583\times10^{-2}$ & $4.3012\times10^{-3}$ & $5.2630\times10^{-2}$ & $9.3434\times10^{-1}$ & $1.3659\times10^{-1}$ & 1967.8708 \\
$v_x$ & XPINN     & $4.6884\times10^{-2}$ & $2.1981\times10^{-3}$ & $3.1123\times10^{-2}$ & $6.6794\times10^{-1}$ & $1.7719\times10^{-1}$ & 6136.6556 \\
$v_x$ & PINN      & $2.9884\times10^{-3}$ & $8.9306\times10^{-6}$ & $2.4352\times10^{-3}$ & $4.2575\times10^{-2}$ & $8.3118\times10^{-3}$ & $\bm{1238.7301}$ \\
\midrule
$v_y$ & LNN--PINN & $\bm{1.3081\times10^{-3}}$ & $\bm{1.7112\times10^{-6}}$ & $\bm{1.0312\times10^{-3}}$ & $\bm{2.1742\times10^{-2}}$ & $\bm{3.7218\times10^{-3}}$ & 1502.7868 \\
$v_y$ & RA--PINN  & $5.6063\times10^{-2}$ & $3.1431\times10^{-3}$ & $4.4388\times10^{-2}$ & $9.3183\times10^{-1}$ & $1.1747\times10^{-1}$ & 1967.8708 \\
$v_y$ & XPINN     & $5.0605\times10^{-2}$ & $2.5609\times10^{-3}$ & $3.7017\times10^{-2}$ & $8.4111\times10^{-1}$ & $1.8085\times10^{-1}$ & 6136.6556 \\
$v_y$ & PINN      & $3.9372\times10^{-3}$ & $1.5501\times10^{-5}$ & $3.1721\times10^{-3}$ & $6.5440\times10^{-2}$ & $1.0089\times10^{-2}$ & $\bm{1238.7301}$ \\
\midrule
$\epsilon$ & LNN--PINN & $\bm{2.0589\times10^{-2}}$ & $\bm{4.2390\times10^{-4}}$ & $\bm{1.6145\times10^{-2}}$ & $\bm{8.7469\times10^{-3}}$ & $\bm{4.9864\times10^{-2}}$ & 1502.7868 \\
$\epsilon$ & RA--PINN  & $1.7889\times10^{-1}$ & $3.2002\times10^{-2}$ & $1.4839\times10^{-1}$ & $7.5999\times10^{-2}$ & $3.9827\times10^{-1}$ & 1967.8708 \\
$\epsilon$ & XPINN     & $1.1664\times10^{-1}$ & $1.3606\times10^{-2}$ & $8.9674\times10^{-2}$ & $4.9555\times10^{-2}$ & $2.8930\times10^{-1}$ & 6136.6556 \\
$\epsilon$ & PINN      & $5.4356\times10^{-2}$ & $2.9546\times10^{-3}$ & $3.7934\times10^{-2}$ & $2.3092\times10^{-2}$ & $1.3923\times10^{-1}$ & $\bm{1238.7301}$ \\
\midrule
$n$ & LNN--PINN & $\bm{1.5179\times10^{-2}}$ & $\bm{2.3041\times10^{-4}}$ & $\bm{1.2328\times10^{-2}}$ & $\bm{9.7847\times10^{-3}}$ & $\bm{3.3574\times10^{-2}}$ & 1502.7868 \\
$n$ & RA--PINN  & $1.1659\times10^{-1}$ & $1.3593\times10^{-2}$ & $9.7802\times10^{-2}$ & $7.5156\times10^{-2}$ & $2.4571\times10^{-1}$ & 1967.8708 \\
$n$ & XPINN     & $8.9108\times10^{-2}$ & $7.9402\times10^{-3}$ & $6.6530\times10^{-2}$ & $5.7440\times10^{-2}$ & $2.5880\times10^{-1}$ & 6136.6556 \\
$n$ & PINN      & $4.7567\times10^{-2}$ & $2.2626\times10^{-3}$ & $3.3379\times10^{-2}$ & $3.0662\times10^{-2}$ & $1.2149\times10^{-1}$ & $\bm{1238.7301}$ \\
\midrule
$\Pi$ & LNN--PINN & $\bm{2.1069\times10^{-3}}$ & $\bm{4.4392\times10^{-6}}$ & $\bm{1.6298\times10^{-3}}$ & $\bm{8.5590\times10^{-2}}$ & $\bm{5.2477\times10^{-3}}$ & 1502.7868 \\
$\Pi$ & RA--PINN  & $2.9413\times10^{-2}$ & $8.6514\times10^{-4}$ & $2.2606\times10^{-2}$ & $1.1949$ & $7.3473\times10^{-2}$ & 1967.8708 \\
$\Pi$ & XPINN     & $1.1828\times10^{-2}$ & $1.3990\times10^{-4}$ & $8.6206\times10^{-3}$ & $4.8049\times10^{-1}$ & $4.2418\times10^{-2}$ & 6136.6556 \\
$\Pi$ & PINN      & $4.9051\times10^{-3}$ & $2.4060\times10^{-5}$ & $4.1267\times10^{-3}$ & $1.9926\times10^{-1}$ & $9.7875\times10^{-3}$ & $\bm{1238.7301}$ \\
\midrule
$\pi_{xx}$ & LNN--PINN & $\bm{1.6356\times10^{-3}}$ & $\bm{2.6751\times10^{-6}}$ & $\bm{1.3007\times10^{-3}}$ & $\bm{8.3538\times10^{-2}}$ & $\bm{4.2931\times10^{-3}}$ & 1502.7868 \\
$\pi_{xx}$ & RA--PINN  & $2.1347\times10^{-2}$ & $4.5570\times10^{-4}$ & $1.7274\times10^{-2}$ & $1.0903$ & $5.0458\times10^{-2}$ & 1967.8708 \\
$\pi_{xx}$ & XPINN     & $2.1710\times10^{-2}$ & $4.7134\times10^{-4}$ & $1.6925\times10^{-2}$ & $1.1089$ & $6.1130\times10^{-2}$ & 6136.6556 \\
$\pi_{xx}$ & PINN      & $5.6295\times10^{-3}$ & $3.1691\times10^{-5}$ & $4.5736\times10^{-3}$ & $2.8753\times10^{-1}$ & $1.3096\times10^{-2}$ & $\bm{1238.7301}$ \\
\midrule
$\pi_{xy}$ & LNN--PINN & $\bm{9.7229\times10^{-4}}$ & $\bm{9.4535\times10^{-7}}$ & $\bm{7.8943\times10^{-4}}$ & $\bm{6.8430\times10^{-2}}$ & $\bm{3.8655\times10^{-3}}$ & 1502.7868 \\
$\pi_{xy}$ & RA--PINN  & $1.4453\times10^{-2}$ & $2.0890\times10^{-4}$ & $1.1812\times10^{-2}$ & $1.0172$ & $3.1923\times10^{-2}$ & 1967.8708 \\
$\pi_{xy}$ & XPINN     & $2.1870\times10^{-2}$ & $4.7832\times10^{-4}$ & $1.4371\times10^{-2}$ & $1.5392$ & $1.8263\times10^{-1}$ & 6136.6556 \\
$\pi_{xy}$ & PINN      & $2.3858\times10^{-3}$ & $5.6919\times10^{-6}$ & $1.9668\times10^{-3}$ & $1.6791\times10^{-1}$ & $7.6974\times10^{-3}$ & $\bm{1238.7301}$ \\
\midrule
$\pi_{yy}$ & LNN--PINN & $\bm{1.1941\times10^{-3}}$ & $\bm{1.4259\times10^{-6}}$ & $\bm{9.1848\times10^{-4}}$ & $\bm{7.5832\times10^{-2}}$ & $\bm{2.8278\times10^{-3}}$ & 1502.7868 \\
$\pi_{yy}$ & RA--PINN  & $1.7952\times10^{-2}$ & $3.2227\times10^{-4}$ & $1.5364\times10^{-2}$ & $1.1400$ & $3.3720\times10^{-2}$ & 1967.8708 \\
$\pi_{yy}$ & XPINN     & $2.6034\times10^{-2}$ & $6.7777\times10^{-4}$ & $1.8503\times10^{-2}$ & $1.6533$ & $9.1475\times10^{-2}$ & 6136.6556 \\
$\pi_{yy}$ & PINN      & $3.3167\times10^{-3}$ & $1.1000\times10^{-5}$ & $2.3582\times10^{-3}$ & $2.1062\times10^{-1}$ & $9.6874\times10^{-3}$ & $\bm{1238.7301}$ \\
\bottomrule
\end{tabular}
}
\end{table*}

Fig.~\ref{F6} and Table~\ref{T6} together show the full picture of this benchmark. Fig.~\ref{F6} simultaneously reports the loss histories, the analytical state, the reconstructed fields, and the corresponding absolute-error distributions, while Table~\ref{T6} lists the field-wise RMSE, MSE, MAE, rel$L_2$, and MaxErr for all eight principal fields together with the total training of each method. LNN--PINN attains the lowest value on every reported error metric for every field. In particular, for the dominant thermodynamic fields one obtains
\[
\mathrm{RMSE}_{\epsilon}=2.0589\times10^{-2},
\qquad
\mathrm{RMSE}_{n}=1.5179\times10^{-2},
\]
while for the dissipative sector one obtains
\[
\mathrm{RMSE}_{\Pi}=2.1069\times10^{-3},
\qquad
\mathrm{RMSE}_{\pi_{xx}}=1.6356\times10^{-3},
\qquad
\mathrm{RMSE}_{\pi_{xy}}=9.7229\times10^{-4},
\qquad
\mathrm{RMSE}_{\pi_{yy}}=1.1941\times10^{-3}.
\]
To quantify the advantage more precisely, we list the percentage reductions of LNN--PINN relative to the other methods for all eight principal fields, with each five-entry tuple ordered as (RMSE, MSE, MAE, relative $L_2$, MaxErr). Relative to the plain PINN, the reductions are
\[
v_x:(36.7,\,60.0,\,40.2,\,36.7,\,44.7)\%,
\quad
v_y:(66.8,\,89.0,\,67.5,\,66.8,\,63.1)\%,
\]
\[
\epsilon:(62.1,\,85.7,\,57.4,\,62.1,\,64.2)\%,
\quad
n:(68.1,\,89.8,\,63.1,\,68.1,\,72.4)\%,
\]
\[
\Pi:(57.0,\,81.5,\,60.5,\,57.0,\,46.4)\%,
\quad
\pi_{xx}:(70.9,\,91.6,\,71.6,\,70.9,\,67.2)\%,
\]
\[
\pi_{xy}:(59.2,\,83.4,\,59.9,\,59.2,\,49.8)\%,
\quad
\pi_{yy}:(64.0,\,87.0,\,61.1,\,64.0,\,70.8)\%.
\]
Relative to RA--PINN, the corresponding reductions are
\[
v_x:(97.1,\,99.9,\,97.2,\,97.1,\,96.6)\%,
\quad
v_y:(97.7,\,99.9,\,97.7,\,97.7,\,96.8)\%,
\]
\[
\epsilon:(88.5,\,98.7,\,89.1,\,88.5,\,87.5)\%,
\quad
n:(87.0,\,98.3,\,87.4,\,87.0,\,86.3)\%,
\]
\[
\Pi:(92.8,\,99.5,\,92.8,\,92.8,\,92.9)\%,
\quad
\pi_{xx}:(92.3,\,99.4,\,92.5,\,92.3,\,91.5)\%,
\]
\[
\pi_{xy}:(93.3,\,99.5,\,93.3,\,93.3,\,87.9)\%,
\quad
\pi_{yy}:(93.3,\,99.6,\,94.0,\,93.3,\,91.6)\%.
\]
Relative to XPINN, the reductions are
\[
v_x:(96.0,\,99.8,\,95.3,\,96.0,\,97.4)\%,
\quad
v_y:(97.4,\,99.9,\,97.2,\,97.4,\,97.9)\%,
\]
\[
\epsilon:(82.3,\,96.9,\,82.0,\,82.3,\,82.8)\%,
\quad
n:(83.0,\,97.1,\,81.5,\,83.0,\,87.0)\%,
\]
\[
\Pi:(82.2,\,96.8,\,81.1,\,82.2,\,87.6)\%,
\quad
\pi_{xx}:(92.5,\,99.4,\,92.3,\,92.5,\,93.0)\%,
\]
\[
\pi_{xy}:(95.6,\,99.8,\,94.5,\,95.6,\,97.9)\%,
\quad
\pi_{yy}:(95.4,\,99.8,\,95.0,\,95.4,\,96.9)\%.
\]
Therefore, in this first strongly coupled eight-field benchmark, the improvement of LNN--PINN is quantitatively consistent across all physical fields and all reported error metrics, while the plain PINN remains the second-most accurate model and the least expensive one in terms of runtime.The learning-rate comparisons, the mathematical parameter-sensitivity analysis, and the complete final hyperparameter configurations are provided in the Appendix~\ref{e}.
\subsection{2D Steady Relativistic Viscous Fluid Benchmark with an Alternative Analytical State}

We next consider a second steady relativistic viscous-fluid benchmark on the same unit square
\[
\Omega=[0,1]\times[0,1].
\]
The governing conservation--relaxation system, the constitutive closure, and the boundary prescription remain exactly those given in Eqs.~\eqref{eq:relB_system}--\eqref{eq:relB_bc}. Accordingly, the unknown state still takes the form
\[
\mathbf{q}(x,y)=\big(v_x,v_y,\epsilon,n,\Pi,\pi_{xx},\pi_{xy},\pi_{yy}\big)^{\top},
\]
and the physical parameters remain those in Eq.~\eqref{eq:relB_params}. The present benchmark differs from the preceding one through a different analytical state, namely
\begin{subequations}\label{eq:relC_exact}
\begin{align}
v_x^{\ast}(x,y)&=B_{vx1}\sin(2\pi x)\sin(\pi y)+B_{vx2}\sin(\pi x)\sin(2\pi y), \label{eq:relC_vx}\\
v_y^{\ast}(x,y)&=B_{vy1}\sin(\pi x)\sin(2\pi y)-B_{vy2}\sin(2\pi x)\sin(\pi y), \label{eq:relC_vy}\\
\epsilon^{\ast}(x,y)&=E_1+B_{\epsilon 1}\cos(2\pi x)\cos(\pi y)+B_{\epsilon 2}\sin(\pi x)\sin(2\pi y), \label{eq:relC_eps}\\
n^{\ast}(x,y)&=N_1+B_{n1}\sin(\pi x)\sin(\pi y)+B_{n2}\cos(2\pi x)\sin(2\pi y), \label{eq:relC_n}\\
\Pi^{\ast}(x,y)&=B_{\Pi 1}\cos(\pi x)\sin(2\pi y)+B_{\Pi 2}\sin(2\pi x)\cos(\pi y), \label{eq:relC_Pi}\\
\pi_{xx}^{\ast}(x,y)&=B_{xx1}\sin(2\pi x)\cos(\pi y)+B_{xx2}\cos(\pi x)\sin(2\pi y), \label{eq:relC_pixx}\\
\pi_{xy}^{\ast}(x,y)&=B_{xy1}\sin(\pi x)\sin(\pi y)+B_{xy2}\sin(2\pi x)\sin(2\pi y), \label{eq:relC_pixy}\\
\pi_{yy}^{\ast}(x,y)&=B_{yy1}\cos(2\pi x)\sin(\pi y)-B_{yy2}\sin(\pi x)\cos(2\pi y). \label{eq:relC_piyy}
\end{align}
\end{subequations}
The coefficients are
\begin{equation}\label{eq:relC_coeffs1}
B_{vx1}=0.085,\quad
B_{vx2}=0.045,\quad
B_{vy1}=0.075,\quad
B_{vy2}=0.040,\quad
E_1=2.42,\quad
B_{\epsilon 1}=0.17,\quad
B_{\epsilon 2}=0.11,
\end{equation}
\begin{equation}\label{eq:relC_coeffs2}
N_1=1.58,\quad
B_{n1}=0.095,\quad
B_{n2}=0.070,\quad
B_{\Pi 1}=0.032,\quad
B_{\Pi 2}=0.018,\quad
B_{xx1}=0.030,\quad
B_{xx2}=0.014,
\end{equation}
and
\begin{equation}\label{eq:relC_coeffs3}
B_{xy1}=0.026,\qquad
B_{xy2}=0.012,\qquad
B_{yy1}=0.028,\qquad
B_{yy2}=0.013.
\end{equation}

The analytical state in Eq.~\eqref{eq:relC_exact} remains physically admissible on $\Omega$. Indeed,
\begin{equation}\label{eq:relC_velocity_bounds}
|v_x^{\ast}(x,y)|\le |B_{vx1}|+|B_{vx2}|=0.130,
\qquad
|v_y^{\ast}(x,y)|\le |B_{vy1}|+|B_{vy2}|=0.115,
\end{equation}
so
\begin{equation}\label{eq:relC_subluminal}
(v_x^{\ast})^2+(v_y^{\ast})^2\le 0.130^2+0.115^2<1.
\end{equation}
Hence the Lorentz factor in Eq.~\eqref{eq:relB_u} is well defined throughout $\Omega$. Moreover,
\begin{equation}\label{eq:relC_positive_bounds}
\epsilon^{\ast}(x,y)\ge E_1-B_{\epsilon 1}-B_{\epsilon 2}=2.14,
\qquad
n^{\ast}(x,y)\ge N_1-B_{n1}-B_{n2}=1.415,
\end{equation}
so the energy density and particle density remain strictly positive on the whole computational domain.

Substituting Eq.~\eqref{eq:relC_exact} into the constitutive relations in Eq.~\eqref{eq:relB_constitutive} yields the analytical four-velocity $u^{\mu\ast}$, the particle current $N^{\mu\ast}$, the energy--momentum tensor $T^{\mu\nu\ast}$, the expansion scalar $\theta^{\ast}$, and the shear tensor $\sigma^{\mu\nu\ast}$. The source functions in Eq.~\eqref{eq:relB_system} are then precisely the smooth functions induced by that analytical state:
\begin{subequations}\label{eq:relC_sources}
\begin{align}
S_E(x,y)&=\partial_xT^{x0\ast}+\partial_yT^{y0\ast}, \label{eq:relC_SE}\\
S_{Mx}(x,y)&=\partial_xT^{xx\ast}+\partial_yT^{yx\ast}, \label{eq:relC_SMx}\\
S_{My}(x,y)&=\partial_xT^{xy\ast}+\partial_yT^{yy\ast}, \label{eq:relC_SMy}\\
S_n(x,y)&=\partial_xN^{x\ast}+\partial_yN^{y\ast}, \label{eq:relC_Sn}\\
S_{\Pi}(x,y)&=\tau_{\Pi}D^{\ast}\Pi^{\ast}+\Pi^{\ast}+\zeta\theta^{\ast}, \label{eq:relC_SPi}\\
S_{\pi xx}(x,y)&=\tau_{\pi}D^{\ast}\pi_{xx}^{\ast}+\pi_{xx}^{\ast}-2\eta\sigma^{xx\ast}, \label{eq:relC_Spixx}\\
S_{\pi xy}(x,y)&=\tau_{\pi}D^{\ast}\pi_{xy}^{\ast}+\pi_{xy}^{\ast}-2\eta\sigma^{xy\ast}, \label{eq:relC_Spixy}\\
S_{\pi yy}(x,y)&=\tau_{\pi}D^{\ast}\pi_{yy}^{\ast}+\pi_{yy}^{\ast}-2\eta\sigma^{yy\ast}. \label{eq:relC_Spiyy}
\end{align}
\end{subequations}
Therefore, once Eq.~\eqref{eq:relC_exact} and Eq.~\eqref{eq:relC_sources} are inserted into Eq.~\eqref{eq:relB_system}, every governing equation is satisfied identically on $\Omega$. The full expanded source expressions, the learning-rate comparisons, the mathematical parameter-sensitivity analysis, and the final hyperparameter configurations are provided in the Appendix~\ref{f}.

Following the same physics-only setup as in the preceding relativistic benchmark, we approximate the state by
\[
\mathbf{q}_{\theta}(x,y)=\big(v_{x,\theta},v_{y,\theta},\epsilon_{\theta},n_{\theta},\Pi_{\theta},\pi_{xx,\theta},\pi_{xy,\theta},\pi_{yy,\theta}\big)^{\top},
\]
and we retain the same sampling protocol as in Eq.~\eqref{eq:relB_sets}.For this alternative analytical state, the interior residual channels are written compactly as
\begin{equation*}
\mathbf{r}_{\Omega}^{\mathrm{relC}}(x,y;\theta)
=
\begin{pmatrix}
\partial_x T_{\theta}^{x0}+\partial_y T_{\theta}^{y0}-S_E(x,y)\\
\partial_x T_{\theta}^{xx}+\partial_y T_{\theta}^{yx}-S_{Mx}(x,y)\\
\partial_x T_{\theta}^{xy}+\partial_y T_{\theta}^{yy}-S_{My}(x,y)\\
\partial_x N_{\theta}^{x}+\partial_y N_{\theta}^{y}-S_n(x,y)\\
\tau_{\Pi}D_{\theta}\Pi_{\theta}+\Pi_{\theta}+\zeta\theta_{\theta}-S_{\Pi}(x,y)\\
\tau_{\pi}D_{\theta}\pi_{xx,\theta}+\pi_{xx,\theta}-2\eta\sigma_{\theta}^{xx}-S_{\pi xx}(x,y)\\
\tau_{\pi}D_{\theta}\pi_{xy,\theta}+\pi_{xy,\theta}-2\eta\sigma_{\theta}^{xy}-S_{\pi xy}(x,y)\\
\tau_{\pi}D_{\theta}\pi_{yy,\theta}+\pi_{yy,\theta}-2\eta\sigma_{\theta}^{yy}-S_{\pi yy}(x,y)
\end{pmatrix},
\end{equation*}
where the source functions are those defined in Eqs.~\eqref{eq:relC_sources}, while $T_{\theta}^{\mu\nu}$, $N_{\theta}^{\mu}$, $D_{\theta}$, $\theta_{\theta}$, and $\sigma_{\theta}^{\mu\nu}$ are still constructed through the constitutive closure in Eqs.~\eqref{eq:relB_constitutive}. The boundary residual is again the trace mismatch
\begin{equation*}
\mathbf{r}_{\partial}^{\mathrm{relC}}(x,y;\theta)
=
\mathbf{q}_{\theta}(x,y)-\mathbf{q}^{\ast}(x,y),
\qquad (x,y)\in \partial\Omega,
\end{equation*}
where $\mathbf{q}^{\ast}$ is now the analytical state in Eqs.~\eqref{eq:relC_exact}. These residual channels enter Eq.~\eqref{LOSS Total} directly, so the present benchmark differs from Section~3.5 only through the analytical state and source content, rather than through any change in the physics-only training structure.

\begin{figure}[t]
\centering
\includegraphics[width=\linewidth]{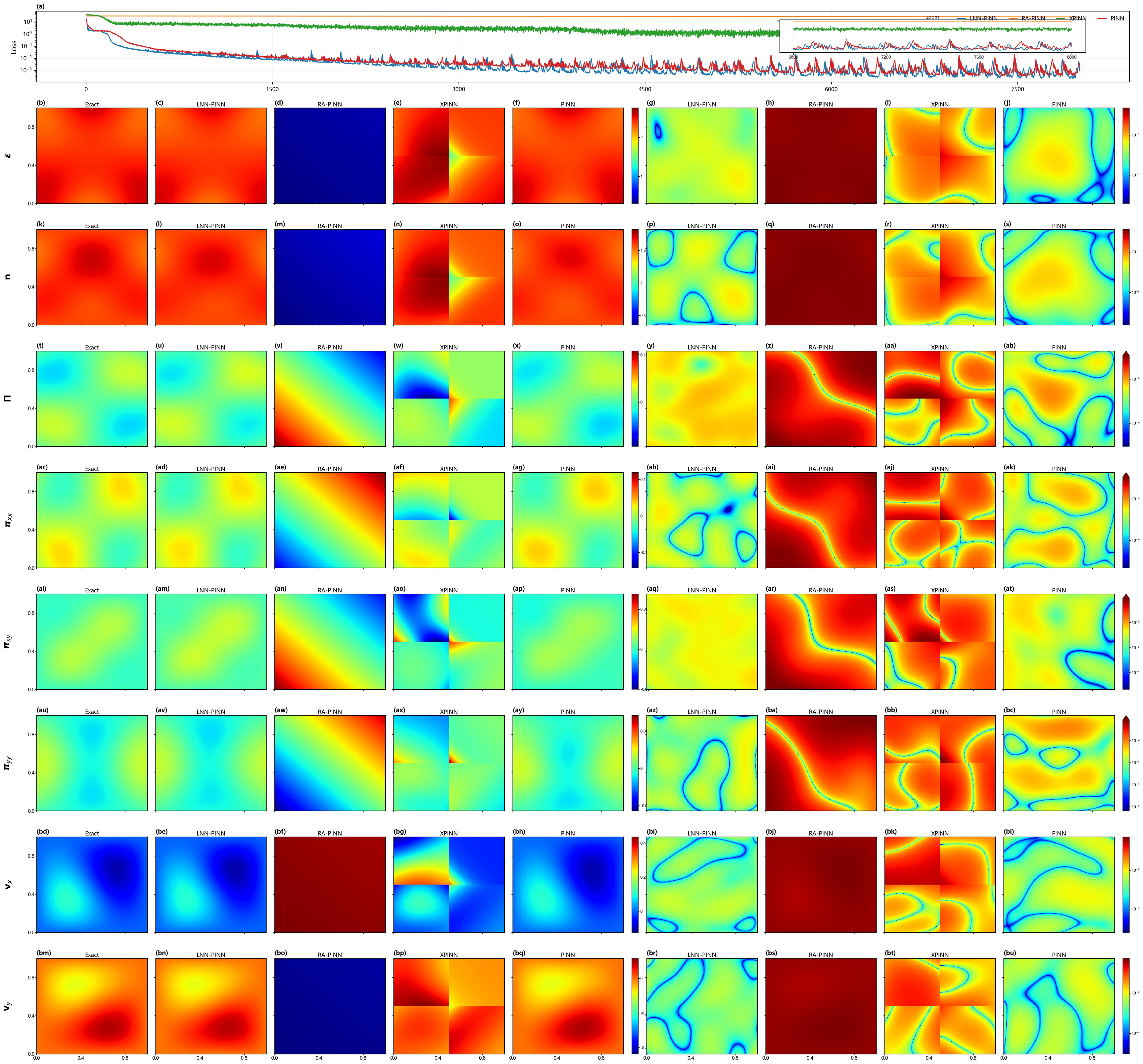}
\caption{{Integrated comparison for the second 2D steady relativistic viscous-fluid benchmark. The upper panel reports the loss histories, while the lower panel simultaneously presents the analytical state in Eq.~\eqref{eq:relC_exact}, the reconstructed fields, and the corresponding absolute-error distributions for LNN--PINN, RA--PINN, XPINN, and PINN under the same training protocol.}}
\label{F7}
\end{figure}

To complement the visual comparison in Fig.~\ref{F7}, Table~\ref{T7} reports the field-wise error metrics and the computational cost for the representative main-text runs. Each physical field has its own RMSE, MSE, MAE, rel$L_2$, and MaxErr. For compact presentation, the total training time of each method is repeated across the corresponding field rows.

\begin{table*}[t]
\centering
\scriptsize
\setlength{\tabcolsep}{3.2pt}
\renewcommand{\arraystretch}{1.08}
\caption{{Field-wise quantitative comparison for the second 2D steady relativistic viscous-fluid benchmark corresponding to Fig.~\ref{F7}. All metrics are computed field by field against the analytical state in Eq.~\eqref{eq:relC_exact}. The training times are reported in seconds.}}
\label{T7}
\resizebox{\textwidth}{!}{
\begin{tabular}{llcccccc}
\toprule
Field & Method & RMSE & MSE & MAE & rel$L_2$ & MaxErr & Train \\
\midrule
$v_x$ & LNN--PINN & $\bm{2.4618\times10^{-3}}$ & $\bm{6.0606\times10^{-6}}$ & $\bm{2.0669\times10^{-3}}$ & $\bm{5.1451\times10^{-2}}$ & $\bm{5.9224\times10^{-3}}$ & 1377.890 \\
$v_x$ & RA--PINN  & $4.3002\times10^{-1}$ & $1.8491\times10^{-1}$ & $4.2761\times10^{-1}$ & $8.9871$ & $5.2668\times10^{-1}$ & 1626.898 \\
$v_x$ & XPINN     & $9.4027\times10^{-2}$ & $8.8410\times10^{-3}$ & $6.0596\times10^{-2}$ & $1.9651$ & $3.2792\times10^{-1}$ & 4925.411 \\
$v_x$ & PINN      & $3.9688\times10^{-3}$ & $1.5751\times10^{-5}$ & $3.0554\times10^{-3}$ & $8.2946\times10^{-2}$ & $9.4572\times10^{-3}$ & $\bm{1125.587}$ \\
\midrule
$v_y$ & LNN--PINN & $\bm{2.2472\times10^{-3}}$ & $\bm{5.0501\times10^{-6}}$ & $\bm{1.9202\times10^{-3}}$ & $\bm{5.3142\times10^{-2}}$ & $\bm{7.8757\times10^{-3}}$ & 1377.890 \\
$v_y$ & RA--PINN  & $4.2981\times10^{-1}$ & $1.8473\times10^{-1}$ & $4.2763\times10^{-1}$ & $1.0164\times10^{1}$ & $5.1916\times10^{-1}$ & 1626.898 \\
$v_y$ & XPINN     & $5.4813\times10^{-2}$ & $3.0045\times10^{-3}$ & $3.9015\times10^{-2}$ & $1.2962$ & $1.5766\times10^{-1}$ & 4925.411 \\
$v_y$ & PINN      & $4.0453\times10^{-3}$ & $1.6364\times10^{-5}$ & $3.2306\times10^{-3}$ & $9.5661\times10^{-2}$ & $1.0142\times10^{-2}$ & $\bm{1125.587}$ \\
\midrule
$\epsilon$ & LNN--PINN & $\bm{2.4286\times10^{-2}}$ & $\bm{5.8981\times10^{-4}}$ & $\bm{2.2164\times10^{-2}}$ & $\bm{1.0030\times10^{-2}}$ & $\bm{5.0655\times10^{-2}}$ & 1377.890 \\
$\epsilon$ & RA--PINN  & $2.0626$ & $4.2545$ & $2.0605$ & $8.5183\times10^{-1}$ & $2.2839$ & 1626.898 \\
$\epsilon$ & XPINN     & $1.8851\times10^{-1}$ & $3.5535\times10^{-2}$ & $1.4102\times10^{-1}$ & $7.7850\times10^{-2}$ & $1.0986$ & 4925.411 \\
$\epsilon$ & PINN      & $2.4857\times10^{-2}$ & $6.1785\times10^{-4}$ & $1.7580\times10^{-2}$ & $1.0265\times10^{-2}$ & $6.4969\times10^{-2}$ & $\bm{1125.587}$ \\
\midrule
$n$ & LNN--PINN & $\bm{1.2341\times10^{-2}}$ & $\bm{1.5230\times10^{-4}}$ & $\bm{9.6952\times10^{-3}}$ & $\bm{7.6239\times10^{-3}}$ & $\bm{3.3557\times10^{-2}}$ & 1377.890 \\
$n$ & RA--PINN  & $1.2036$ & $1.4487$ & $1.2025$ & $7.4354\times10^{-1}$ & $1.2975$ & 1626.898 \\
$n$ & XPINN     & $1.2934\times10^{-1}$ & $1.6728\times10^{-2}$ & $8.8008\times10^{-2}$ & $7.9899\times10^{-2}$ & $7.0852\times10^{-1}$ & 4925.411 \\
$n$ & PINN      & $1.8411\times10^{-2}$ & $3.3895\times10^{-4}$ & $1.3645\times10^{-2}$ & $1.1373\times10^{-2}$ & $4.5067\times10^{-2}$ & $\bm{1125.587}$ \\
\midrule
$\Pi$ & LNN--PINN & $\bm{5.0470\times10^{-3}}$ & $\bm{2.5472\times10^{-5}}$ & $\bm{4.8040\times10^{-3}}$ & $2.1671\times10^{-1}$ & $\bm{8.0024\times10^{-3}}$ & 1377.890 \\
$\Pi$ & RA--PINN  & $6.3061\times10^{-2}$ & $3.9767\times10^{-3}$ & $5.5193\times10^{-2}$ & $2.7077$ & $1.6219\times10^{-1}$ & 1626.898 \\
$\Pi$ & XPINN     & $2.9676\times10^{-2}$ & $8.8069\times10^{-4}$ & $1.8766\times10^{-2}$ & $1.2742$ & $1.4059\times10^{-1}$ & 4925.411 \\
$\Pi$ & PINN      & $\bm{3.8789\times10^{-3}}$ & $\bm{1.5046\times10^{-5}}$ & $\bm{2.6903\times10^{-3}}$ & $\bm{1.6655\times10^{-1}}$ & $1.3186\times10^{-2}$ & $\bm{1125.587}$ \\
\midrule
$\pi_{xx}$ & LNN--PINN & $\bm{1.5071\times10^{-3}}$ & $\bm{2.2713\times10^{-6}}$ & $\bm{1.2294\times10^{-3}}$ & $\bm{7.3210\times10^{-2}}$ & $\bm{4.2305\times10^{-3}}$ & 1377.890 \\
$\pi_{xx}$ & RA--PINN  & $5.3893\times10^{-2}$ & $2.9044\times10^{-3}$ & $4.6022\times10^{-2}$ & $2.6180$ & $1.2159\times10^{-1}$ & 1626.898 \\
$\pi_{xx}$ & XPINN     & $2.3292\times10^{-2}$ & $5.4251\times10^{-4}$ & $1.7280\times10^{-2}$ & $1.1315$ & $1.4296\times10^{-1}$ & 4925.411 \\
$\pi_{xx}$ & PINN      & $2.8226\times10^{-3}$ & $7.9673\times10^{-6}$ & $2.2085\times10^{-3}$ & $1.3712\times10^{-1}$ & $7.7680\times10^{-3}$ & $\bm{1125.587}$ \\
\midrule
$\pi_{xy}$ & LNN--PINN & $4.3105\times10^{-3}$ & $1.8581\times10^{-5}$ & $4.2356\times10^{-3}$ & $3.0257\times10^{-1}$ & $\bm{6.2481\times10^{-3}}$ & 1377.890 \\
$\pi_{xy}$ & RA--PINN  & $3.7759\times10^{-2}$ & $1.4257\times10^{-3}$ & $3.1156\times10^{-2}$ & $2.6504$ & $1.1064\times10^{-1}$ & 1626.898 \\
$\pi_{xy}$ & XPINN     & $2.5333\times10^{-2}$ & $6.4177\times10^{-4}$ & $2.0143\times10^{-2}$ & $1.7782$ & $1.0528\times10^{-1}$ & 4925.411 \\
$\pi_{xy}$ & PINN      & $\bm{3.3916\times10^{-3}}$ & $\bm{1.1503\times10^{-5}}$ & $\bm{2.9805\times10^{-3}}$ & $\bm{2.3807\times10^{-1}}$ & $6.3427\times10^{-3}$ & $\bm{1125.587}$ \\
\midrule
$\pi_{yy}$ & LNN--PINN & $\bm{1.3110\times10^{-3}}$ & $\bm{1.7188\times10^{-6}}$ & $\bm{1.1051\times10^{-3}}$ & $\bm{9.1392\times10^{-2}}$ & $\bm{3.1393\times10^{-3}}$ & 1377.890 \\
$\pi_{yy}$ & RA--PINN  & $4.3647\times10^{-2}$ & $1.9051\times10^{-3}$ & $3.5629\times10^{-2}$ & $3.0427$ & $1.0167\times10^{-1}$ & 1626.898 \\
$\pi_{yy}$ & XPINN     & $1.7513\times10^{-2}$ & $3.0671\times10^{-4}$ & $1.3915\times10^{-2}$ & $1.2208$ & $1.2827\times10^{-1}$ & 4925.411 \\
$\pi_{yy}$ & PINN      & $3.4786\times10^{-3}$ & $1.2101\times10^{-5}$ & $2.6897\times10^{-3}$ & $2.4249\times10^{-1}$ & $9.8232\times10^{-3}$ & $\bm{1125.587}$ \\
\bottomrule
\end{tabular}
}
\end{table*}

Fig.~\ref{F7} and Table~\ref{T7} together show the full picture of this benchmark. Fig.~\ref{F7} simultaneously reports the loss histories, the analytical state, the reconstructed fields, and the corresponding absolute-error distributions, while Table~\ref{T7} lists the field-wise RMSE, MSE, MAE, rel$L_2$, and MaxErr for all eight principal fields together with the total training time of each method. LNN--PINN attains the lowest reported error metrics for six of the eight physical fields, whereas the plain PINN remains more accurate for $\Pi$ and $\pi_{xy}$. Quantitatively, the RMSE reduction relative to the plain PINN ranges from about 2.3\% to 62.3\% on the six improved fields.In particular, for the hydrodynamic variables one has
\[
\mathrm{RMSE}_{v_x}=2.4618\times10^{-3},
\qquad
\mathrm{RMSE}_{v_y}=2.2472\times10^{-3},
\qquad
\mathrm{RMSE}_{\epsilon}=2.4286\times10^{-2},
\qquad
\mathrm{RMSE}_{n}=1.2341\times10^{-2},
\]
while for the dissipative sector one has
\[
\mathrm{RMSE}_{\Pi}=5.0470\times10^{-3},
\qquad
\mathrm{RMSE}_{\pi_{xx}}=1.5071\times10^{-3},
\qquad
\mathrm{RMSE}_{\pi_{xy}}=4.3105\times10^{-3},
\qquad
\mathrm{RMSE}_{\pi_{yy}}=1.3110\times10^{-3}.
\]
For this second strongly coupled eight-field benchmark, the quantitative picture is more selective, so the percentage reductions must be reported field by field. With the same ordering (RMSE, MSE, MAE, relative $L_2$, MaxErr), the reductions of LNN--PINN relative to the plain PINN are
\[
v_x:(38.0,\,61.5,\,32.4,\,38.0,\,37.4)\%,
\quad
v_y:(44.4,\,69.1,\,40.6,\,44.4,\,22.3)\%,
\]
\[
\epsilon:(2.3,\,4.5,\,-26.1,\,2.3,\,22.0)\%,
\quad
n:(33.0,\,55.1,\,28.9,\,33.0,\,25.5)\%,
\]
\[
\Pi:(-30.1,\,-69.3,\,-78.6,\,-30.1,\,39.3)\%,
\quad
\pi_{xx}:(46.6,\,71.5,\,44.3,\,46.6,\,45.5)\%,
\]
\[
\pi_{xy}:(-27.1,\,-61.5,\,-42.1,\,-27.1,\,1.5)\%,
\quad
\pi_{yy}:(62.3,\,85.8,\,58.9,\,62.3,\,68.0)\%.
\]
Hence, relative to the plain PINN, LNN--PINN improves all five reported error metrics for $v_x$, $v_y$, $n$, $\pi_{xx}$, and $\pi_{yy}$; it improves RMSE, MSE, relative $L_2$, and MaxErr for $\epsilon$ but not MAE; it improves only MaxErr for $\Pi$; and it improves only MaxErr marginally for $\pi_{xy}$.
Relative to RA--PINN, the reductions are
\[
v_x:(99.4,\,100.0,\,99.5,\,99.4,\,98.9)\%,
\quad
v_y:(99.5,\,100.0,\,99.6,\,99.5,\,98.5)\%,
\]
\[
\epsilon:(98.8,\,100.0,\,98.9,\,98.8,\,97.8)\%,
\quad
n:(99.0,\,100.0,\,99.2,\,99.0,\,97.4)\%,
\]
\[
\Pi:(92.0,\,99.4,\,91.3,\,92.0,\,95.1)\%,
\quad
\pi_{xx}:(97.2,\,99.9,\,97.3,\,97.2,\,96.5)\%,
\]
\[
\pi_{xy}:(88.6,\,98.7,\,86.4,\,88.6,\,94.4)\%,
\quad
\pi_{yy}:(97.0,\,99.9,\,96.9,\,97.0,\,96.9)\%.
\]
Relative to XPINN, the reductions are
\[
v_x:(97.4,\,99.9,\,96.6,\,97.4,\,98.2)\%,
\quad
v_y:(95.9,\,99.8,\,95.1,\,95.9,\,95.0)\%,
\]
\[
\epsilon:(87.1,\,98.3,\,84.3,\,87.1,\,95.4)\%,
\quad
n:(90.5,\,99.1,\,89.0,\,90.5,\,95.3)\%,
\]
\[
\Pi:(83.0,\,97.1,\,74.4,\,83.0,\,94.3)\%,
\quad
\pi_{xx}:(93.5,\,99.6,\,92.9,\,93.5,\,97.0)\%,
\]
\[
\pi_{xy}:(83.0,\,97.1,\,79.0,\,83.0,\,94.1)\%,
\quad
\pi_{yy}:(92.5,\,99.4,\,92.1,\,92.5,\,97.6)\%.
\]
Therefore, the second relativistic benchmark shows that the advantage of LNN--PINN remains clear over RA--PINN and XPINN across all fields, whereas relative to the plain PINN the improvement is benchmark-dependent and field-dependent rather than uniform across every reported quantity. The learning-rate comparisons, the mathematical parameter-sensitivity analysis, and the complete final hyperparameter configurations are provided in the Appendix~\ref{f}.
}
\clearpage
\section{Conclusion}

{In summary, the numerical results across the six benchmark problems considered in this work show that, under an unchanged physics-only training setup, replacing the standard hidden-layer propagation in PINNs with liquid residual gating can improve solution reconstruction in a consistent overall sense. Across the reported time-dependent, mixed-boundary, curved-domain, higher-order, and strongly coupled multi-field examples, the advantage of LNN--PINN appears primarily in the fixed-grid error metrics, which is consistent with a more effective internal feature-propagation mechanism. At the same time, the improvement is reflected mainly in reconstruction quality rather than in a universal reduction of runtime or a uniformly smoother optimization path. Taken together, these results support LNN--PINN as a useful architecture-level refinement of physics-only neural PDE solvers for the benchmark classes studied here.}

\printcredits
\section*{Declaration of competing interest}
The authors declared that they have no conflicts of interest to this work. 
\section*{Acknowledgment}
This work is supported by the Developing Project of Science and Technology of Jilin Province (20240402042GH). 

\section*{Data availability}
{All the code for this article is available open access at a Github repository available at https://github.com/Uderwood-TZ/LNN-PINN-A-Unified-Physics-Only-Training-Framework-with-Liquid-Residual-Blocks.git.}

\appendix
\renewcommand{\theequation}{\thesection.\arabic{equation}}
\section{Scale Setting and Loss Normalization}\label{A}
To reduce ill-conditioning from disparate physical units and magnitudes, we introduce optional nondimensionalization and magnitude scaling without changing the purely physics-supervised MSE objectives or the residual definitions. During training and evaluation, we first normalize each residual by a fixed scale factor and then compute its mean square; this strategy preserves the physical meaning of the equations and boundary conditions and keeps the objective structure unchanged. Among the four examples, only the 2D steady heat-conduction case uses this scaling; the 1D advection–reaction case, the anisotropic Poisson–beam coupling, and the 2D Laplace case use no scaling (equivalently set all scale factors to 1).

Let $\|\cdot\|$ denote the Euclidean norm (absolute value for scalars). We choose a reference length $L_{\text{ref}}$ (and a reference time $T_{\text{ref}}$ when time appears) and a field scale $U_{\text{ref}}$, and we use the dimensionless variables:
$\hat{x}=\frac{x}{L_{\text{ref}}}$,
$\hat{t}=\frac{t}{T_{\text{ref}}}$,
$\hat{u}=\frac{u}{U_{\text{ref}}}$,
$\hat{\nabla}=L_{\text{ref}}\nabla$,
$\hat{\partial}_t=T_{\text{ref}}\partial_t.$
For a generic strong-form linear prototype $\mathcal{L}[u]=\sum_{|\alpha|\le p} c_\alpha(x)\, D^\alpha u$
we substitute the dimensionless variables and divide by $U_{\text{ref}}$ to obtain :
\begin{equation}
\hat{\mathcal{L}}[\hat{u}]
=\sum_{|\alpha|\le p}\Big(c_\alpha^* L_{\text{ref}}^{\alpha_x} T_{\text{ref}}^{\alpha_t}\Big)^{-1}
\Big(\frac{c_\alpha}{c_\alpha^*}\Big)\,
D^{\alpha}_{\hat{x}}\,\partial_{\hat{t}}^{\alpha_t}\hat{u},
\end{equation}
where $c_\alpha^*$ denotes a representative magnitude of $c_\alpha$ (e.g., mean/median/upper bound), and $\alpha=(\alpha_x,\alpha_t)$ records spatial and temporal orders. We use this rewrite only to determine residual scales and do not require code-level equation rewrites.

In the loss, we use MSEs of normalized residuals. Let $r_\Omega=\mathcal{L}[f_\theta]-g$, $r_D=f_\theta-u_D$, and $r_N=\mathbf{K}\nabla f_\theta\cdot\mathbf{n}-q_N$ denote the domain, Dirichlet, and Neumann residuals. We set the scale factors:
\begin{equation}
s_\Omega = U_{\text{ref}} \sum_{|\alpha|\le p} c_\alpha^* L_{\text{ref}}^{-\alpha_x} T_{\text{ref}}^{-\alpha_t},
\quad
s_D = U_{\text{ref}},
\quad
s_N = \frac{K_* U_{\text{ref}}}{L_{\text{ref}}},
\end{equation}
where $K_*$ denotes a representative magnitude of the coefficients in the principal part, independent of any specific physical name. For a principal part of the form:
\begin{equation}
    -\nabla\cdot(\mathbf{A}(x)\nabla u)\quad (\text{linear second-order elliptic term}),
\end{equation}
we define $K_* := \operatorname{stat}_{x\in\Omega}\,\|\mathbf{A}(x)\|_{\mathrm{op}},$
and evaluate $\|\cdot\|_{\mathrm{op}}$ as the operator norm (spectral norm / largest eigenvalue for matrices). We choose $\operatorname{stat}$ as a robust statistic such as the median, mean, or an $L^\infty$ upper bound and keep that choice fixed per task. For systems or anisotropy, $\mathbf{A}(x)$ may be block-structured; we then use the corresponding block-operator norm. From the dimensions of the flux-type boundary condition $\mathbf{q}=\mathbf{A}\nabla u\cdot\mathbf{n}$, we recover the Neumann scale $ s_N=\frac{K_* U_{\text{ref}}}{L_{\text{ref}}}.$
If the principal part takes a general $p$-th order form $\sum_{|\alpha|=p} a_\alpha(x) D^\alpha u$, we interpret $K_*$ as a representative magnitude of the same-order coefficient family, for example :$K_* := \operatorname{stat}_{x\in\Omega}\sum_{|\alpha|=p}\|a_\alpha(x)\|,$ 
and we pair it with the spatial factor $L_{\text{ref}}^{-p}$ in the scaling formula (this paper uses only second-order cases, hence $L_{\text{ref}}^{-2}$). We then write the empirical risk as:
\begin{equation}
\begin{aligned}
    \hat{\mathcal{J}}(\theta)
= \frac{1}{N_\Omega}\sum_{x\in\mathcal{D}_\Omega}\Big\|\tfrac{r_\Omega(x)}{s_\Omega}\Big\|^2
+ \lambda_D \frac{1}{N_D}\sum_{x\in\mathcal{D}_D}\Big\|\tfrac{r_D(x)}{s_D}\Big\|^2\\
+ \lambda_N \frac{1}{N_N}\sum_{x\in\mathcal{D}_N}\Big\|\tfrac{r_N(x)}{s_N}\Big\|^2,
\end{aligned}
\end{equation}
and we drop any nonexistent term in a specific case. We keep $\lambda_*>0$ as preference weights only.

In our experiments, the 2D steady heat-conduction case uses this scaling: we set $L_{\text{ref}}$ to a geometric length scale (e.g., side length or radius), $U_{\text{ref}}$ to the Dirichlet temperature amplitude, and $K_*$ to a representative thermal-conductivity magnitude (the constant value for homogeneous media; the domain median or mean for heterogeneous media). The other three cases apply no scaling, which equals setting $s_\Omega=s_D=s_N=1$. This design preserves the internal physical consistency of each example and avoids extra scale degrees of freedom when unnecessary.
\section{Continuous Risk and Monte Carlo Empirical Risk (Measure-Theoretic Setup and Convergence)}\label{B}
We concatenate the residuals by physical origin $r(x;\theta)=\big(r_\Omega(x;\theta),\ r_D(x;\theta),\ r_N(x;\theta),\ r_{\text{IC}}(x;\theta)\big),$
and we equip them with the product-sum measure $\mu=\mu_\Omega\oplus \mu_D\oplus \mu_N\oplus \mu_{\text{IC}}$
(when a residual type does not arise in a given problem, we drop its component and its measure).
To unify units and magnitudes, we evaluate residuals under fixed prior scales.
Specifically, we choose a parameter-independent diagonal positive-definite matrix:
\begin{equation}
    W=\operatorname{diag}\!\big(s_\Omega^{-1} I,\ \sqrt{\lambda_D}\,s_D^{-1} I,\ \sqrt{\lambda_N}\,s_N^{-1} I,\ \sqrt{\lambda_{\text{IC}}}\,s_{\text{IC}}^{-1} I\big),
\end{equation}
where $s_\bullet>0$ denote normalization scales for each residual block (we may also view $W$ as a weighting matrix), and $\lambda_\bullet>0$ follow the main text.
We then rewrite the continuous and empirical risks as:
\begin{equation}
    \mathcal{J}_W(\theta)=\int \big\| W\,r(x;\theta)\big\|^2\,\mathrm{d}\mu(x),
\quad
\widehat{\mathcal{J}}_W(\theta)=\frac{1}{N}\sum_{p=1}^N \big\| W\,r(\xi_p;\theta)\big\|^2 ,
\end{equation}
where the sequence $\{\xi_p\}_{p=1}^{N}$ consists of i.i.d. samples from $\mu$.
\begin{lemma}
Let $(\Omega,\mathcal{F})$ be a measurable space. Assume $r:\Omega\to\mathbb{R}^m$ is $\mathcal{F}/\mathcal{B}(\mathbb{R}^m)$–measurable and fix $W\in\mathbb{R}^{m\times m}$. Define:
\begin{equation}
    \phi(x):=\|Wr(x)\|^2.
\end{equation}
With this definition, we obtain an $\mathcal{F}/\mathcal{B}(\mathbb{R})$–measurable function $\phi:\Omega\to\mathbb{R}$.
\end{lemma}
\begin{proof}
Let $W=(w_{ij})$. For each $i=1,\ldots,m$, then we have:
\begin{equation}
    (W\boldsymbol{r})_i=\sum_{j=1}^m w_{ij}\, r_j .
\end{equation}
Multiplying a measurable function by a constant and forming a finite sum both preserve measurability, so we obtain measurability for every coordinate $(W\boldsymbol{r})_i$ and hence for the vector map $W\boldsymbol{r}:\Omega\to\mathbb{R}^m$. For the squared norm, write:
\begin{equation}
    \phi=\sum_{i=1}^m \big((W\boldsymbol{r})_i\big)^2 .
\end{equation}
Squaring real-valued measurable functions and taking finite sums again preserve measurability; therefore $\phi:\Omega\to\mathbb{R}$ defines an $\mathcal{F}/\mathcal{B}(\mathbb{R})$–measurable function.
\end{proof}
\begin{lemma}
For any $x\in\Omega$, set $\boldsymbol{v}=\boldsymbol{r}(x)$. By the definition of the operator norm:
\begin{equation}
    \|W\boldsymbol{r}(x)\|=\|W\boldsymbol{v}\|\le \|W\|_{\mathrm{op}}\;\|\boldsymbol{v}\|
= \|W\|_{\mathrm{op}}\;\|\boldsymbol{r}(x)\|.
\end{equation}
Square both sides and use the Rayleigh–quotient bound for the PSD matrix $W^\top W\succeq 0$ to obtain:
\begin{equation}
    \begin{aligned}
        \|W\boldsymbol{r}(x)\|^2
&= \boldsymbol{r}(x)^\top W^\top W\,\boldsymbol{r}(x)
\\&\le \lambda_{\max}(W^\top W)\,\|\boldsymbol{r}(x)\|^2
\\&= \|W\|_{\mathrm{op}}^{2}\,\|\boldsymbol{r}(x)\|^2.
    \end{aligned}
\end{equation}
\end{lemma}
\begin{proof}
We set $M:=W^\top W$ and obtain:
\begin{equation}
    \|Wr(x)\|^2 = r(x)^\top M\, r(x).
\end{equation}
Because $M=W^\top W$ inherits symmetry and positive semidefiniteness, we diagonalize $M=Q\Lambda Q^\top$ with $Q$ orthogonal and $\Lambda=\operatorname{diag}(\lambda_1,\ldots,\lambda_m)$, $\lambda_i\ge 0$. Write $y:=Q^\top r(x)$ (an orthogonal change of coordinates). Then we have:
\begin{equation}
\begin{aligned}
        r(x)^\top M\, r(x)
&= y^\top \Lambda y
= \sum_{i=1}^m \lambda_i y_i^2
\\&\le \Big(\max_i \lambda_i\Big)\sum_{i=1}^m y_i^2\\
&= \lambda_{\max}(M)\,\|y\|^2\\
&= \lambda_{\max}(W^\top W)\,\|r(x)\|^2.
\end{aligned}
\end{equation}
Since $\|W\|_{\mathrm{op}}=\sqrt{\lambda_{\max}(W^\top W)}$, we conclude:
\begin{equation}\label{点态不等式}
    \|Wr(x)\|^2 \;\le\; \|W\|_{\mathrm{op}}^{2}\,\|r(x)\|^2.
\end{equation}
\end{proof}
\begin{lemma}
Consider a probability space $(\Omega,\mathcal{F},\mu)$. Take a measurable map $r:\Omega\to\mathbb{R}^m$ with $r\in L^2(\mu;\mathbb{R}^m)$ so that
$\int_\Omega \|r(x)\|^2\,\mathrm{d}\mu(x)<\infty$.
Fix a matrix $W\in\mathbb{R}^{m\times m}$ independent of sampling and set:
\begin{equation}
    \phi(x)=\|W\,r(x)\|^2,\qquad
g(x)=\|W\|_{\mathrm{op}}^{2}\,\|r(x)\|^2.
\end{equation}
Consequently, $\phi\in L^1(\mu)$, and we have:
\begin{equation}
    \int_\Omega \phi(x)\,\mathrm{d}\mu(x)
\;\le\; \|W\|_{\mathrm{op}}^{2}\int_\Omega \|r(x)\|^{2}\,\mathrm{d}\mu(x)
\;<\;\infty .
\end{equation}
\end{lemma}
\begin{proof}
Fix $x\in\Omega$ and write $v=r(x)$. The inequality:
\begin{equation}
    \begin{aligned}
        \|W r(x)\|^2 &= v^\top W^\top W v\\ &\le \lambda_{\max}(W^\top W)\,\|v\|^2\\
&=\|W\|_{\mathrm{op}}^{\,2}\,\|r(x)\|^2
    \end{aligned}
\end{equation}
implies $0\le \phi(x)\le g(x)$ at every definition point of a representative of $r$, hence for $\mu$-a.e.\ $x$. Because $r\in L^2$, the function $\|r\|^2$ lies in $L^1$ and
$\int \|r\|^2\,\mathrm{d}\mu<\infty$; multiplying by $\|W\|_{\mathrm{op}}^{\,2}$ yields $g\in L^1$ and :
\begin{equation}
    \int_\Omega g\,\mathrm{d}\mu
\;=\; \|W\|_{\mathrm{op}}^{\,2}\int_\Omega \|r\|^2\,\mathrm{d}\mu
\;<\;\infty .
\end{equation}
Introduce the truncations $\phi_n:=\min\{\phi,n\}$; these functions are measurable, nonnegative, and increase to $\phi$, and they satisfy $\phi_n\le g$. Consequently we have:
\begin{equation}
    \int_\Omega \phi_n\,\mathrm{d}\mu \;\le\; \int_\Omega g\,\mathrm{d}\mu \;<\;\infty .
\end{equation}
By the Monotone Convergence Theorem, we have:
\begin{equation}
    \int_\Omega \phi\,\mathrm{d}\mu
= \lim_{n\to\infty}\int_\Omega \phi_n\,\mathrm{d}\mu
\le \int_\Omega g\,\mathrm{d}\mu
= \|W\|_{\mathrm{op}}^{\,2}\int_\Omega \|r\|^2\,\mathrm{d}\mu
< \infty .
\end{equation}
Thus $\phi\in L^1(\mu)$ and the stated integral bound follows.
\end{proof}
\begin{lemma}\label{l4}
Consider a probability space $(\mathcal{X},\Sigma,\mu)$. 
Take a measurable $r:\mathcal{X}\to\mathbb{R}^m$ with $r\in L^2(\mu;\mathbb{R}^m)$. 
Fix $W\in\mathbb{R}^{m\times m}$ (independent of the sampling) and set:
\begin{equation}
    \phi(x)=\|W\,r(x)\|^2,\quad
X_1,X_2,\ldots \stackrel{\mathrm{i.i.d.}}{\sim}\mu,\quad
Y_i=\phi(X_i)=\|W\,r(X_i)\|^2.
\end{equation}
We have $Y_1\in L^1$ and we also have:
\begin{equation}
    \mathbb{E}[Y_1]
=\int_{\mathcal{X}} \|W\,r(x)\|^2\,\mathrm{d}\mu(x)
\;\le\; \|W\|_{\mathrm{op}}^{2}\int_{\mathcal{X}} \|r(x)\|^2\,\mathrm{d}\mu(x)
\;<\;\infty.
\end{equation}
Hence Kolmogorov’s SLLN yields:
\begin{equation}
    \frac{1}{N}\sum_{i=1}^{N} Y_i
\;\stackrel{\text{a.s.}}{\underset{N\to\infty}{\longrightarrow}}\;
\mathbb{E}[Y_1]
=\int_{\mathcal{X}} \|W\,r(x)\|^2\,\mathrm{d}\mu(x).
\end{equation}
\end{lemma}
\begin{proof}
    Set $M:=W^\top W\succeq 0$. For any $x$, we have:
    \begin{equation}
        \|W r(x)\|^2
= r(x)^\top M\,r(x)
\;\le\; \lambda_{\max}(M)\,\|r(x)\|^2
= \|W\|_{\mathrm{op}}^{2}\,\|r(x)\|^2 .
    \end{equation}
Since $r\in L^2(\mu)$, the right-hand side lies in $L^1(\mu)$. Hence $\phi\in L^1(\mu)$ and:
\begin{equation}
    \int_{\mathcal{X}} \phi\,\mathrm{d}\mu
\;\le\; \|W\|_{\mathrm{op}}^{2}\int_{\mathcal{X}}\|r\|^2\,\mathrm{d}\mu
\;<\;\infty .
\end{equation}
Because $X_1$ follows $\mu$ (i.e., $X_1\#\mathbb{P}=\mu$), we apply the change-of-variables formula and compute:
\begin{equation}
    \mathbb{E}[Y_1]=\mathbb{E}[\phi(X_1)]
=\int_{\mathcal{X}} \phi(x)\,\mathrm{d}\mu(x)
=\int_{\mathcal{X}} \|W r(x)\|^2\,\mathrm{d}\mu(x).
\end{equation}
With $\{Y_i\}$ i.i.d. and $\mathbb{E}[|Y_1|]<\infty$ (as shown above), we invoke Kolmogorov’s SLLN and obtain:
\begin{equation}
    \frac{1}{N}\sum_{i=1}^N Y_i \;\xrightarrow[\;N\to\infty\;]{\text{a.s.}}\; \mathbb{E}[Y_1].
\end{equation}
\end{proof}
So we have:
\begin{theorem}\label{T1}
Consider a probability space $(\Omega,\mathcal{F},\mu)$ and a measurable map $r:\Omega\to\mathbb{R}^m$.
Fix $W\in\mathbb{R}^{m\times m}$ and write $\|W\|_{\mathrm{op}}=\sqrt{\lambda_{\max}(W^\top W)}$.
Set $\phi(x)=\|W\,r(x)\|^2.$ Assume $r\in L^2(\mu;\mathbb{R}^m)$.
Then $\phi\in L^1(\mu)$ and we have:
\begin{equation}
    \int_{\Omega}\phi(x)\,\mathrm{d}\mu(x)
\le \|W\|_{\mathrm{op}}^{2}\int_{\Omega}\|r(x)\|^{2}\,\mathrm{d}\mu(x)
< \infty.
\end{equation}
Thus, for $X\sim\mu$, the quantity $\mathbb{E}[\|W\,r(X)\|^2]$ exists and is finite.
With $X_i\stackrel{\mathrm{i.i.d.}}{\sim}\mu$ and $Y_i=\|W\,r(X_i)\|^2$, Kolmogorov’s SLLN guarantees:
\begin{equation}
    \frac{1}{N}\sum_{i=1}^{N} Y_i
\;\xrightarrow{\text{a.s.}}\;
\int_{\Omega}\|W\,r(x)\|^2\,\mathrm{d}\mu(x).
\end{equation}
\end{theorem}
\begin{theorem}
Work on a probability space $(\Omega,\mathcal{F},\mu)$. For a given parameter $\theta$, define:
\begin{equation}
    \phi_\theta(x):=\|W\,r(x;\theta)\|^2,
\end{equation}
where $W\succ 0$ stays fixed and independent of sampling, and assume $\phi_\theta\in L^1(\mu)$ (the Lemma \ref{l4} gives the sufficient condition $r\in L^2 \Rightarrow \phi_\theta\in L^1$).
Draw i.i.d.\ samples $\{\xi_p\}_{p=1}^N \stackrel{\mathrm{i.i.d.}}{\sim}\mu$ and set the empirical risk:
\begin{equation}
    \widehat{\mathcal{J}}_{W}(\theta)=\frac{1}{N}\sum_{p=1}^{N}\phi_\theta(\xi_p).
\end{equation}
Then we have:
\begin{equation}
    \mathbb{E}\!\big[\widehat{\mathcal{J}}_{W}(\theta)\big]
= \mathcal{J}_{W}(\theta)
:= \int_{\Omega} \phi_\theta(x)\,\mathrm{d}\mu(x).
\end{equation}
\end{theorem}
\begin{proof}
We verify finiteness and write:
\begin{equation}
    \mathbb{E}\!\big[\widehat{\mathcal{J}}_W(\theta)\big]
= \frac{1}{N}\sum_{p=1}^{N}\mathbb{E}\!\big[\phi_\theta(\xi_p)\big]
\quad \text{(use linearity).}
\end{equation}
Because $\xi_p$ are i.i.d.\ with law $\mu$, we obtain:
\begin{equation}
    \mathbb{E}\!\big[\phi_\theta(\xi_p)\big]
= \mathbb{E}\!\big[\phi_\theta(\xi_1)\big]
= \int_{\Omega}\phi_\theta(x)\,\mathrm{d}\mu(x)
\quad \text{(since } \xi_1\#\mathbb{P}=\mu\text{).}
\end{equation}
Combine the identities to get
\begin{equation}
    \mathbb{E}\!\big[\widehat{\mathcal{J}}_W(\theta)\big]
= \int_{\Omega}\phi_\theta\,\mathrm{d}\mu
= \mathcal{J}_W(\theta).
\end{equation}
\end{proof}
\begin{theorem}
Work with a compact parameter set $\Theta\subset\mathbb{R}^p$. For each $\theta\in\Theta$, define:
\begin{equation}
\begin{aligned}
\phi_\theta(x)&:=\|W\,r(x;\theta)\|^2\ge 0,\\
\mathcal{J}_W(\theta)&:=\mathbb{E}_\mu[\phi_\theta(X)],\\
\widehat{\mathcal{J}}_W(\theta)&:=\frac{1}{N}\sum_{p=1}^N \phi_\theta(\xi_p),
\end{aligned}
\end{equation}
where $\{\xi_p\}_{p=1}^N\stackrel{\mathrm{i.i.d.}}{\sim}\mu$ and $W\succ 0$ stays fixed and independent of sampling.

Assume:
\begin{enumerate}
\item[(A1)] (\textbf{Integrable envelope}) There exists $F\in L^1(\mu)$ such that
$\sup_{\theta\in\Theta}\phi_\theta(x)\le F(x)$ for $\mu$-a.e.\ $x$.
\item[(A2)] (\textbf{Lipschitz continuity}) There exists $L\in L^1(\mu)$ such that:
\begin{equation}
    |\phi_\theta(x)-\phi_{\theta'}(x)|\le L(x)\,\|\theta-\theta'\|
\quad\text{for $\mu$-a.e.\ }x,\ \forall\,\theta,\theta'\in\Theta.
\end{equation}
\item[(A3)] (\textbf{Measurability}) For each $\theta$, the map $\phi_\theta$ is measurable.
\end{enumerate}

\noindent\textbf{Remark.} If some $\theta_0$ satisfies $\phi_{\theta_0}\in L^1(\mu)$, then (A2) yields the envelope:
\begin{equation}
    F:=\phi_{\theta_0}+L\cdot \operatorname{diam}(\Theta)\in L^1(\mu),
\quad
\operatorname{diam}(\Theta):=\sup_{\theta,\theta'\in\Theta}\|\theta-\theta'\|,
\end{equation}
so (A1) holds automatically.

Under \emph{(A1)-(A3)}, the uniform law of large numbers holds:
\begin{equation}
    \sup_{\theta\in\Theta}\big|\widehat{\mathcal{J}}_W(\theta)-\mathcal{J}_W(\theta)\big|
\;\stackrel{\text{a.s.}}{\underset{N\to\infty}{\longrightarrow}}\;0.
\end{equation}
\end{theorem}
\begin{proof}
Take any sequence $\{\delta_m\}_{m\ge1}$ with $\delta_m \downarrow 0$. By the compactness of $\Theta$, choose a finite $\delta_m$–net $\{\theta^{(m,k)}\}_{k=1}^{K_m}\subset\Theta$ so that for every $\theta\in\Theta$ some $k$ satisfies $\|\theta-\theta^{(m,k)}\|\le\delta_m$.

For fixed $(m,k)$, the quantity $\widehat{\mathcal{J}}_W(\theta^{(m,k)})$ equals the sample mean of $\phi_{\theta^{(m,k)}}$, and (A1) gives $\phi_{\theta^{(m,k)}}\in L^1(\mu)$. Hence the strong law of large numbers yields:
\begin{equation}
    \widehat{\mathcal{J}}_W(\theta^{(m,k)}) \xrightarrow[\;N\to\infty\;]{\text{a.s.}} \mathcal{J}_W(\theta^{(m,k)}).
\end{equation}
Because the index set $\{(m,k)\}$ is countable, pick an event $\Omega_0$ with probability one on which these convergences hold for all $(m,k)$ simultaneously.

On $\Omega_0$, fix $\omega$ and $m$. For any $\theta\in\Theta$, choose its nearest net point $\theta^{(m,k)}$. Apply the triangle inequality and (A2) to obtain:
\begin{equation}
    \begin{aligned}
\big|\widehat{\mathcal{J}}_W(\theta)-\mathcal{J}_W(\theta)\big|
&\le \big|\widehat{\mathcal{J}}_W(\theta)-\widehat{\mathcal{J}}_W(\theta^{(m,k)})\big|
\\&+ \big|\widehat{\mathcal{J}}_W(\theta^{(m,k)})-\mathcal{J}_W(\theta^{(m,k)})\big|
\\&+ \big|\mathcal{J}_W(\theta^{(m,k)})-\mathcal{J}_W(\theta)\big| \\
&\le \frac{1}{N}\sum_{p=1}^N L(\xi_p)\,\|\theta-\theta^{(m,k)}\|
\\&+ \big|\widehat{\mathcal{J}}_W(\theta^{(m,k)})-\mathcal{J}_W(\theta^{(m,k)})\big|
\\&+ \mathbb{E}[L(X)]\,\|\theta-\theta^{(m,k)}\| \\
&\le \delta_m\!\left(\frac{1}{N}\sum_{p=1}^N L(\xi_p)+\mathbb{E}[L(X)]\right)
\\&+ \big|\widehat{\mathcal{J}}_W(\theta^{(m,k)})-\mathcal{J}_W(\theta^{(m,k)})\big|.
\end{aligned}
\end{equation}
Take $\sup_{\theta\in\Theta}$ and then $\max_{1\le k\le K_m}$ to get:
\begin{equation}
    \begin{aligned}
        \sup_{\theta\in\Theta}\big|\widehat{\mathcal{J}}_W(\theta)-\mathcal{J}_W(\theta)\big|
&\le \delta_m\!\left(\frac{1}{N}\sum_{p=1}^N L(\xi_p)+\mathbb{E}[L(X)]\right)
\\&+ \max_{1\le k\le K_m}\big|\widehat{\mathcal{J}}_W(\theta^{(m,k)})-\mathcal{J}_W(\theta^{(m,k)})\big|.
    \end{aligned}
\end{equation}
By the SLLN, $\frac{1}{N}\sum_{p=1}^N L(\xi_p)\to \mathbb{E}[L(X)]$ almost surely; and on $\Omega_0$ the maximum over the finitely many net points converges to $0$. Therefore, for fixed $m$, we have:
\begin{equation}
    \lim_{N\to\infty}\ \text{RHS}\;=\;2\,\delta_m\,\mathbb{E}[L(X)].
\end{equation}
Let $m\to\infty$ so that $\delta_m\downarrow 0$, and conclude:
\begin{equation}
    \limsup_{N\to\infty}\ \sup_{\theta\in\Theta}\big|\widehat{\mathcal{J}}_W(\theta)-\mathcal{J}_W(\theta)\big|=0
\quad\text{almost surely.}
\end{equation}
\end{proof}
\begin{theorem}
Consider $(\Omega,\mathcal{F},\mu)$ and a measurable residual $r:\Omega\to\mathbb{R}^m$ with $r(\cdot;\theta)\in L^4(\mu;\mathbb{R}^m)$ for a fixed $\theta$. Choose a fixed, sampling–independent $W\in\mathbb{R}^{m\times m}$ and write:
\begin{equation}
\phi_\theta(x)=\|W\,r(x;\theta)\|^2,\quad
\mathcal{J}_W(\theta)=\mathbb{E}_\mu[\phi_\theta(X)].
\end{equation}
With i.i.d.\ draws $\{\xi_i\}_{i=1}^N\sim\mu$, define:
\begin{equation}
\widehat{\mathcal{J}}_W(\theta)=\frac{1}{N}\sum_{i=1}^N \phi_\theta(\xi_i).
\end{equation}

\noindent\textbf{(i) Unbiasedness and variance.}
We have:
\begin{equation}
\mathbb{E}[\widehat{\mathcal{J}}_W(\theta)]=\mathcal{J}_W(\theta),
\end{equation}
and:
\begin{equation}\label{final var}
    \operatorname{Var}(\widehat{\mathcal{J}}_W(\theta))
=\frac{\operatorname{Var}(\phi_\theta(\xi_1))}{N}
\le \frac{\|W\|_{\mathrm{op}}^{4}}{N}\;\mathbb{E}\!\left[\|r(\xi_1;\theta)\|^{4}\right].
\end{equation}
\noindent\textbf{(ii) Gradient and Gauss–Newton with left preconditioning.}
Assume interchange conditions for expectation and differentiation and set:
\begin{equation}
    J_\theta(x)=\frac{\partial r(x;\theta)}{\partial\theta},\quad
L_u(x)=D_u r[f_\theta](x).
\end{equation}
Then we have:
\begin{equation}
    \begin{aligned}
        \nabla_\theta \mathcal{J}_W(\theta)&=2\,\mathbb{E}[J_\theta^\top L_u^\top W^\top W\,r],\\
H_{\mathrm{GN}}(\theta)&=\mathbb{E}\!\left[(W L_u J_\theta)^\top (W L_u J_\theta)\right]\succeq 0,
    \end{aligned}
\end{equation}
so $W$ acts as a fixed left preconditioner $v\mapsto Wv$.

\noindent\textbf{(iii) Invariance of minimizers when zero residual is achievable.}
If a parameter $\theta^*$ yields $r(\cdot;\theta^*)=0$ $\mu$-a.e., then for any $W\succ0$, we have:
\begin{equation}
    \mathcal{J}_W(\theta^*)=0=\inf_\theta \mathcal{J}_W(\theta),
\end{equation}
hence all weighted objectives share the same global minimizer. Adjusting $W$ influences variance scale and conditioning, not the optimizer’s location.
\end{theorem}
\begin{proof}
\noindent\textbf{(i) Unbiasedness and a variance bound.}
From the measurability and integrability established above, we have $\phi_\theta\in L^1(\mu)$. Hence:
\begin{equation}
    \mathbb{E}[\widehat{\mathcal{J}}_W]
= \frac{1}{N}\sum_{i=1}^N \mathbb{E}[\phi_\theta(\xi_i)]
= \mathbb{E}[\phi_\theta(\xi_1)]
= \int \phi_\theta \,\mathrm{d}\mu
= \mathcal{J}_W .
\end{equation}
Independence yields:
\begin{equation}\label{Var}
    \mathrm{Var}(\widehat{\mathcal{J}}_W)
= \frac{1}{N}\,\mathrm{Var}(\phi_\theta(\xi_1)).
\end{equation}
The Rayleigh–quotient inequality $\|W r\|^2 \le \|W\|_{\mathrm{op}}^{2}\,\|r\|^2$ implies:
\begin{equation}
    \phi_\theta(\xi_1)^2
= \|W r(\xi_1)\|^{4}
\le \|W\|_{\mathrm{op}}^{4}\,\|r(\xi_1)\|^{4}.
\end{equation}
Therefore $\phi_\theta(\xi_1)\in L^2$ and :
\begin{equation}\label{varp}
    \begin{aligned}
        \mathrm{Var}(\phi_\theta(\xi_1))
&= \mathbb{E}[\phi_\theta(\xi_1)^2]-\mathbb{E}[\phi_\theta(\xi_1)]^2\\
&\le \mathbb{E}[\phi_\theta(\xi_1)^2]
\le\|W\|_{\mathrm{op}}^{4}\,\mathbb{E}[\|r(\xi_1)\|^{4}] .
    \end{aligned}
\end{equation}
Combining Eq.~\eqref{Var} and \eqref{varp}, we derive Eq.~\eqref{final var}.\\
\noindent\textbf{(ii) Gradient and Gauss-Newton Forms}

\textbf{Assumptions:}
\begin{itemize}
\item[(B1)]\textbf{Measurable/integrable.} $r(\cdot;\theta)\in L^2(\mu;\mathbb{R}^m)$.
\item[(B2)] \textbf{Differentiability with an envelope.} For $\mu$-a.e.\ $x$, the map $\theta\mapsto r(x;\theta)$ is differentiable, and there exists $a\in L^2(\mu)$ such that, in a neighborhood of $\theta$, we have:
\begin{equation}
    \|L_u(x,\theta)\,J_\theta(x)\|\;\le\;a(x).
\end{equation}
\item[(B3)]\textbf{Second–order envelope (for Gauss–Newton).} \\There exists $b\in L^1(\mu)$ such that for any unit vectors $h,k$, we have:
\begin{equation}
    \|D_\theta^2 r(x;\theta)[h,k]\|\;\le\;b(x).
\end{equation}
\end{itemize}
Under (B1)–(B2), we have $\phi_\theta\in L^1(\mu)$ and:
\begin{equation}
    \bigl|\partial_\epsilon\,\phi_{\theta+\epsilon h}(x)\bigr|_{\epsilon=0}
\;\le\; 2\,\|W\|_{\mathrm{op}}^{2}\,\|r(x;\theta)\|\;\|L_u(x,\theta)\,J_\theta(x)\,h\|.
\end{equation}
By the Cauchy–Schwarz inequality together with (B1)–(B2), the expectation of the right-hand side is integrable; the dominated convergence theorem then justifies exchanging differentiation and expectation.

\noindent\textbf{Gâteaux derivative} (for any $h\in\mathbb{R}^p$):
\begin{equation}
    \begin{aligned}
D\mathcal{J}_W(\theta)[h]
&= \mathbb{E}\!\left[\partial_\epsilon \,\|W r(x;\theta+\epsilon h)\|^2\Big|_{\epsilon=0}\right] \\
&= \mathbb{E}\!\left[\,2\,\langle W r,\, W\,\partial_\epsilon r(x;\theta+\epsilon h)\rangle\Big|_{\epsilon=0}\right] \\
&= \mathbb{E}\!\left[\,2\,\langle W r,\, W\,D_\theta r(x;\theta)[h]\rangle\right].
\end{aligned}
\end{equation}
Invoke the Fréchet chain rule $D_\theta r=L_u J_\theta$ to obtain:
\begin{equation}
\begin{aligned}
        D\mathcal{J}_W(\theta)[h]
&= \mathbb{E}\!\left[\,2\,\langle W r,\, W L_u J_\theta h\rangle\right]\\
&= h^\top \underbrace{\,2\,\mathbb{E}\!\left[J_\theta^\top L_u^\top W^\top W r\right]}_{:=\,\nabla_\theta \mathcal{J}_W(\theta)}.
\end{aligned}
\end{equation}
Therefore, we have:
\begin{equation}
 \nabla_\theta \mathcal{J}_W(\theta)=2\,\mathbb{E}\!\left[J_\theta^\top L_u^\top W^\top W r\right]\in\mathbb{R}^p.
\end{equation}
\textbf{Second Gâteaux derivative} (for any unit $h,k\in\mathbb{R}^p$):
\begin{equation}
    \begin{aligned}
D^2 \mathcal{J}_W(\theta)[h,k]
&= \mathbb{E}\!\left[\,\partial_\epsilon \partial_\eta \,\|W r(x;\theta+\epsilon h+\eta k)\|^2 \Big|_{\epsilon=\eta=0}\right] \\
&= \mathbb{E}[\,2\,\langle W D_\theta r[h],\, W D_\theta r[k]\\
&+ 2\,\langle W r,\, W D_\theta^2 r[h,k] ].
\end{aligned}
\end{equation}
Here:
\begin{equation}
    \begin{aligned}
        &D_\theta r[h]=L_u J_\theta h,\\
&D_\theta^2 r[h,k]=(D_\theta L_u)[h]\,(J_\theta k)\;+\;L_u\,(D_\theta J_\theta)[h]\,k,
    \end{aligned}
\end{equation}
which applies the chain rule to $L_u\circ J_\theta$ at second order. Using (B2)–(B3) together with Cauchy–Schwarz, we verify the existence of both expectations. Assemble $h,k$ along coordinate directions to obtain the Hessian decomposition:
\begin{equation}
    \nabla_\theta^2 \mathcal{J}_W(\theta)
= 2\,\mathbb{E}\!\left[(L_u J_\theta)^\top W^\top W\,(L_u J_\theta)\right]
\;+\; 2\,\mathbb{E}\!\left[\mathcal{R}(x,\theta)\right],
\end{equation}
where the remainder satisfies:
\begin{equation}
    h^\top \mathcal{R}(x,\theta)\,k
= \langle W r(x;\theta),\, W D_\theta^2 r(x;\theta)[h,k]\rangle.
\end{equation}
Gauss-Newton approximation holds in two common settings:
(i) near a minimizer, $r$ stays small so $\mathbb{E}[\langle W r,\, W D_\theta^2 r\rangle]$ becomes negligible;
(ii) treat $r$ as linear in $u$ and approximate $J_\theta$ as constant, which yields $D_\theta^2 r\approx 0$.

Hence:
\begin{equation}
    H_{\mathrm{GN}}(\theta):=2\,\mathbb{E}\!\left[(W L_u J_\theta)^\top (W L_u J_\theta)\right]\succeq 0.
\end{equation}
By (B1)–(B2) and the Cauchy-Schwarz inequality,
\begin{equation}
    \begin{aligned}
        \mathbb{E}\!\left[\|W r\|\,\|W L_u J_\theta h\|\right]
&\le \|W\|_{\mathrm{op}}^{2}\,\big(\mathbb{E}\|r\|^{2}\big)^{1/2}\,\big(\mathbb{E}\|L_u J_\theta h\|^{2}\big)^{1/2}\\
&\le \|W\|_{\mathrm{op}}^{2}\,\|h\|\,\big(\mathbb{E}a^{2}\big)^{1/2}\,\big(\mathbb{E}\|r\|^{2}\big)^{1/2}<\infty,
    \end{aligned}
\end{equation}
so the expectation in the gradient exists. If we also assume (B3), then we have:
\begin{equation}
    \big|\langle W r,\, W D_\theta^2 r[h,k]\rangle\big|
\le \|W\|_{\mathrm{op}}^{2}\,\|r\|\, b(x)\,\|h\|\,\|k\|,
\end{equation}
and the conditions $\mathbb{E}\|r\|<\infty$ (from $r\in L^2$) and $b\in L^1$ ensure integrability of the second-order remainder.

Under the verifiable assumptions above, we obtain:
\begin{equation}
 \begin{aligned}
        \nabla_\theta \mathcal{J}_W(\theta)
&= 2\,\mathbb{E}\!\left[J_\theta^\top L_u^\top W^\top W\, r\right],
\\
H_{\mathrm{GN}}(\theta)&\approx 2\,\mathbb{E}\!\left[(W L_u J_\theta)^\top (W L_u J_\theta)\right].
\end{aligned}
\end{equation}
Thus the fixed matrix $W$ enters the first- and second-order geometry only as a \emph{left preconditioner}: it rescales the noise level of stochastic gradients and changes the conditioning of the Gauss–Newton curvature, while it leaves the risk expectation and the location of any “zero-residual” minimizer unchanged.\\
\noindent\textbf{(iii) Gradient and Gauss-Newton Forms}
Assume $r(\cdot;\theta)\in L^2(\mu)$, fix $W$, and define:
\begin{equation}
    \mathcal{J}_W(\theta)=\int_{\Omega}\|W r(x;\theta)\|^2\,\mathrm{d}\mu(x).
\end{equation}

\paragraph{Proof (sufficiency for the "Gradient and Gauss-Newton Forms").}
If $r(\cdot;\theta)=0$ holds $\mu$-a.e., then the integrand $\|W r(x;\theta)\|^2$ vanishes $\mu$-a.e., so $\mathcal{J}_W(\theta)=0$. For any parameter $\vartheta$, the inequality $\|W r(x;\vartheta)\|^2\ge 0$ yields $\mathcal{J}_W(\vartheta)\ge 0$. Hence $\mathcal{J}_W(\theta)=0$ equals the minimal value over all feasible parameters, and $\theta$ achieves the global minimum.

\paragraph{Proof (necessity for the "Gradient and Gauss-Newton Forms").}
Suppose there exists $\theta^*$ with $r(\cdot;\theta^*)=0$ $\mu$-a.e.\ (i.e., a zero-residual solution is attainable), and let $\hat\theta$ be a global minimizer of $\mathcal{J}_W$. From the previous part, $\mathcal{J}_W(\theta^*)=0$, and since $\mathcal{J}_W(\vartheta)\ge 0$ for all $\vartheta$, the minimum value equals $0$, so $\mathcal{J}_W(\hat\theta)=0$.

If, in addition, $W$ is positive definite (invertible), let $\lambda_{\min}>0$ denote the smallest eigenvalue of $W^\top W$. The Rayleigh bound gives:
\begin{equation}
    \|W r(x;\hat\theta)\|^2 \;\ge\; \lambda_{\min}\,\|r(x;\hat\theta)\|^2 \quad (\mu\text{-a.e.\ }x).
\end{equation}
Integrate both sides to obtain:
\begin{equation}
    \begin{aligned}
        0=\mathcal{J}_W(\hat\theta)&=\int_{\Omega}\|W r(x;\hat\theta)\|^2\,\mathrm{d}\mu(x)\\
&\ge \lambda_{\min}\int_{\Omega}\|r(x;\hat\theta)\|^2\,\mathrm{d}\mu(x).
    \end{aligned}
\end{equation}
The right-hand side is nonnegative and sits below $0$, so it must vanish:
$\int_{\Omega}\|r(x;\hat\theta)\|^2\,\mathrm{d}\mu(x)=0$. A nonnegative integrand with zero integral forces $\|r(x;\hat\theta)\|=0$ for $\mu$-a.e.\ $x$; thus $r(\cdot;\hat\theta)=0$ $\mu$-a.e. Therefore, changing the fixed scaling/weight matrix $W$ leaves the expected risk nonnegative and keeps any zero-residual solution—more generally, any solution with $r(\cdot;\theta)\in\ker W$—globally minimizing. This change only rescales the variance and the condition number of the Gauss-Newton curvature, so it affects sensible step sizes and numerical convergence without shifting the minimizers.
\end{proof}
\begin{theorem}\label{thm:is_unbiased_consistent}
Let $(\Omega,\mathcal{F})$ be a measurable space with probability measures $\mu$ and $\pi$ such that $\mu\ll\pi$. Write the Radon-Nikodym derivative as $w=\frac{\mathrm{d}\mu}{\mathrm{d}\pi}$, which is nonnegative and measurable.

For a given parameter $\theta$, take a measurable residual $r(\cdot;\theta):\Omega\to\mathbb{R}^m$. Fix $W\in\mathbb{R}^{m\times m}$ and define:
\begin{equation}
    \phi_\theta(x):=\|W r(x;\theta)\|^2.
\end{equation}
Assume $\phi_\theta\in L^1(\mu)$ (for example, this holds when $r(\cdot;\theta)\in L^2(\mu)$ since $\|W r\|^2\le \|W\|_{\mathrm{op}}^2\|r\|^2$). Draw i.i.d.\ samples $\{\xi_p\}_{p=1}^N\sim\pi$ and form the importance–sampling estimator:
\begin{equation}
    \widehat{\mathcal{J}}_{W}^{\mathrm{IS}}(\theta):=\frac{1}{N}\sum_{p=1}^{N} w(\xi_p)\,\phi_\theta(\xi_p).
\end{equation}
Denote the continuous risk by $\mathcal{J}_W(\theta):=\int_\Omega \phi_\theta(x)\,\mathrm{d}\mu(x)$.

\noindent Conclusion:

(a) Unbiasedness: $\mathbb{E}_{\pi}\!\left[\widehat{\mathcal{J}}_{W}^{\mathrm{IS}}(\theta)\right]=\mathcal{J}_W(\theta)$.

(b) Consistency: $\widehat{\mathcal{J}}_{W}^{\mathrm{IS}}(\theta)\xrightarrow[\;N\to\infty\;]{\text{a.s.}}\mathcal{J}_W(\theta)$.
\end{theorem}
\begin{proof}
We note that $r$ is measurable and the maps $v\mapsto Wv$ and $v\mapsto\|v\|^{2}$ are continuous, so $\phi_\theta$ is measurable; by assumption, $\phi_\theta\in L^{1}(\mu)$. With $\mu\ll\pi$ and $w=\frac{\mathrm{d}\mu}{\mathrm{d}\pi}$, the identity
$\int g\,\mathrm{d}\mu=\int g\,w\,\mathrm{d}\pi$ for nonnegative $g$ gives:
\begin{equation}
    \int_\Omega |\phi_\theta|\,\mathrm{d}\mu
= \int_\Omega w\,|\phi_\theta|\,\mathrm{d}\pi
<\infty,
\end{equation}
so $w\phi_\theta\in L^{1}(\pi)$. Define $Y_p=w(\xi_p)\phi_\theta(\xi_p)$. Then $Y_p$ are i.i.d. with $\mathbb{E}_\pi[|Y_1|]<\infty$, and:
\begin{equation}
    \mathbb{E}_\pi\!\big[\widehat{\mathcal{J}}_W^{\mathrm{IS}}(\theta)\big]
= \mathbb{E}_\pi[Y_1]
= \int_\Omega w\,\phi_\theta\,\mathrm{d}\pi
= \int_\Omega \phi_\theta\,\mathrm{d}\mu
= \mathcal{J}_W(\theta),
\end{equation}
which proves (a). Kolmogorov’s SLLN then ensures:
\begin{equation}
    \frac{1}{N}\sum_{p=1}^{N} Y_p
\stackrel{\text{a.s.}}{\longrightarrow}
\mathbb{E}_\pi[Y_1]
= \mathcal{J}_W(\theta),
\end{equation}
which establishes (b).
\end{proof}
By Theorems~\ref{T1}--\ref{thm:is_unbiased_consistent} and because $W$ stays fixed and independent of sampling, we obtain:
\begin{equation}
\mathbb{E}\!\big[\widehat{\mathcal{J}}_W(\theta)\big]=\mathcal{J}_W(\theta),
\quad
\widehat{\mathcal{J}}_W(\theta)\xrightarrow[\;N\to\infty\;]{\text{a.s.}}\mathcal{J}_W(\theta).
\end{equation}
Therefore, introducing a scale/weight does not change unbiasedness or consistency; it only multiplies the variance by a constant factor and guides sensible step-size choices in numerical optimization. We operationalize “comparable contributions” by evaluating at a reference parameter $\theta_\star$ (typically the initialization $\theta_0$ or a short warm start $\theta_{\mathrm{warm}}$). For each component, We define the normalized energy:
\begin{equation}
    C_i(\theta):=\mathbb{E}_{x\sim\mu}\!\left[\left\|\frac{r_i(x;\theta)}{s_i}\right\|^2\right],
\quad i\in\{\Omega,D,N,\mathrm{IC}\}.
\end{equation}
We then require a uniform balance: there exists $\kappa\ge1$ such that:
\begin{equation}
    \frac{1}{\kappa}\;\le\;\frac{C_i(\theta_\star)}{C_j(\theta_\star)}\;\le\;\kappa
\quad \text{for all } i,j.
\end{equation}
Equivalently, all components exhibit normalized energies of the same order of magnitude.

Let the four residual blocks, indexed by their physical origin, be $r_i(x;\theta)\in\mathbb{R}^{m_i}$ (e.g., $i\in\{\Omega,D,N,\mathrm{IC}\}$), with sampling measures $\mu_i$ and fixed weights/scales $\lambda_i>0$, $s_i>0$. We write both the empirical and continuous risks in the unified expectation form:
\begin{equation}
\begin{aligned}
        \mathcal{J}_W(\theta)
&= \sum_i \mathbb{E}_{x\sim\mu_i}\!\left[\left\|\sqrt{\lambda_i}\,\frac{r_i(x;\theta)}{s_i}\right\|^2\right]\\
&= \sum_i \frac{\lambda_i}{s_i^2}\,\mathbb{E}_{x\sim\mu_i}\!\left[\|r_i(x;\theta)\|^2\right].
\end{aligned} 
\end{equation}
Collect the left-multiplicative scalings into the fixed block-diagonal matrix:
\begin{equation}
    W=\operatorname{blkdiag}\!\left(\sqrt{\lambda_i}\,s_i^{-1} I_{m_i}\right)_i,
\end{equation}
so that $\mathcal{J}_W(\theta)=\mathbb{E}\!\left[\|W r(x;\theta)\|^2\right]$.Let $u_\theta(x)=f_\theta(x)\in\mathbb{R}^m$ denote the network output and $J_\theta(x)=\partial u_\theta(x)/\partial\theta\in\mathbb{R}^{m\times p}$ its Jacobian. For each block, assume the Fréchet derivative with respect to $u$ exists and write:
\begin{equation}
    L_i(x)=D_u r_i[u_\theta](x)\in\mathbb{R}^{m_i\times m}.
\end{equation}
Under the standard conditions that justify swapping expectation and differentiation (the previous section supplies an integrable envelope), for any direction $h\in\mathbb{R}^p$, we have:
\begin{equation}
    \begin{aligned}
D\mathcal{J}_W(\theta)[h]
&= \sum_i \mathbb{E}_{x\sim\mu_i}\!\left[\partial_\epsilon \left\|\sqrt{\lambda_i}\,\frac{r_i(x;\theta+\epsilon h)}{s_i}\right\|^2\Big|_{\epsilon=0}\right] \\
&= \sum_i \mathbb{E}\!\left[\frac{2\lambda_i}{s_i^2}\,\langle r_i(x;\theta),\,D_\theta r_i(x;\theta)[h]\rangle\right]\\
&= \sum_i \mathbb{E}\!\left[\frac{2\lambda_i}{s_i^2}\,\langle r_i(x;\theta),\,L_i(x) J_\theta(x) h\rangle\right] \\
&= h^\top\,\mathbb{E}\!\left[\sum_i \frac{2\lambda_i}{s_i^2}\,J_\theta(x)^\top L_i(x)^\top r_i(x;\theta)\right].
\end{aligned}
\end{equation}
Therefore the component-wise gradient reads:
\begin{equation}
    \;\nabla_\theta \mathcal{J}_W(\theta)
= 2 \sum_i \mathbb{E}_{x\sim\mu_i}\!\left[\frac{\lambda_i}{s_i^2}\,J_\theta(x)^\top L_i(x)^\top r_i(x;\theta)\right]\;\in\mathbb{R}^p.
\end{equation}
Fix any operator norm $\|\cdot\|$ consistent with the Euclidean vector norm. 
Assume constants $M$ and $L_i^*$ satisfy:
\begin{equation}
    \|J_\theta(x)\|\le M
\quad\text{and}\quad
\|L_i(x)\|\le L_i^*
\quad\text{for $\mu_i$-a.e. }x .
\end{equation}
Apply the triangle inequality $\|\mathbb{E}Z\|\le \mathbb{E}\|Z\|$ and the bound $\|A^\top b\|\le \|A\|\,\|b\|$ to obtain:
\begin{equation}
    \begin{aligned}
\|\nabla_\theta \mathcal{J}_W(\theta)\|
&= \left\|\,2\sum_i \mathbb{E}\!\left[\frac{\lambda_i}{s_i^2}\, J_\theta^\top L_i^\top r_i\right]\right\| \\
&\le 2\sum_i \frac{\lambda_i}{s_i^2}\,\left\|\mathbb{E}\!\left[J_\theta^\top L_i^\top r_i\right]\right\|\\
&\le 2\sum_i \frac{\lambda_i}{s_i^2}\,\mathbb{E}\!\left[\|J_\theta^\top L_i^\top r_i\|\right] \\
&\le 2\sum_i \frac{\lambda_i}{s_i^2}\,\mathbb{E}\!\left[\|J_\theta\|\,\|L_i\|\,\|r_i\|\right]\\
&\le 2M\sum_i \frac{\lambda_i L_i^*}{s_i^2}\,\mathbb{E}_{x\sim\mu_i}\!\left[\|r_i(x;\theta)\|\right].
\end{aligned}
\end{equation}
This bound makes the per-component “flux” into the total gradient explicit:$\frac{\lambda_i L_i^*}{s_i^2}\,\mathbb{E}\!\|r_i\|.$ Choose $s_i$ to match the order of $L_i^*$ and the typical magnitude of the $i$th residual (e.g., via the earlier physics-based scales or an early-stage RMS estimate). 
We then drive all component-wise gradient contributions to the same order and prevent any single physics term from dominating the optimization geometry.

In summary, when multi-physics coupling appears or units differ markedly, I introduce a fixed normalization scale or a weight matrix $W$ and thereby measure residuals under the inner product induced by $W^\top W$. This choice realizes a geometrically proportional balance of each residual component’s contributions to the gradient and the Gauss-Newton curvature, so all contributions operate on the same order in the optimization landscape. This modification keeps the empirical risk unbiased and consistent with the continuous risk and preserves the attainability of a zero-residual solution; it only affects numerical conditioning and step-size selection, and it expresses the trade-off criterion explicitly when constraints cannot be satisfied simultaneously.
\section{Steady Heat Conduction in a Circular Domain via MATLAB PDE Toolbox: Variational Principle, FEM Discretization, and Analytical Verification}\label{C}
\begin{figure*}[t] 
  \centering
  \includegraphics[width=\textwidth]{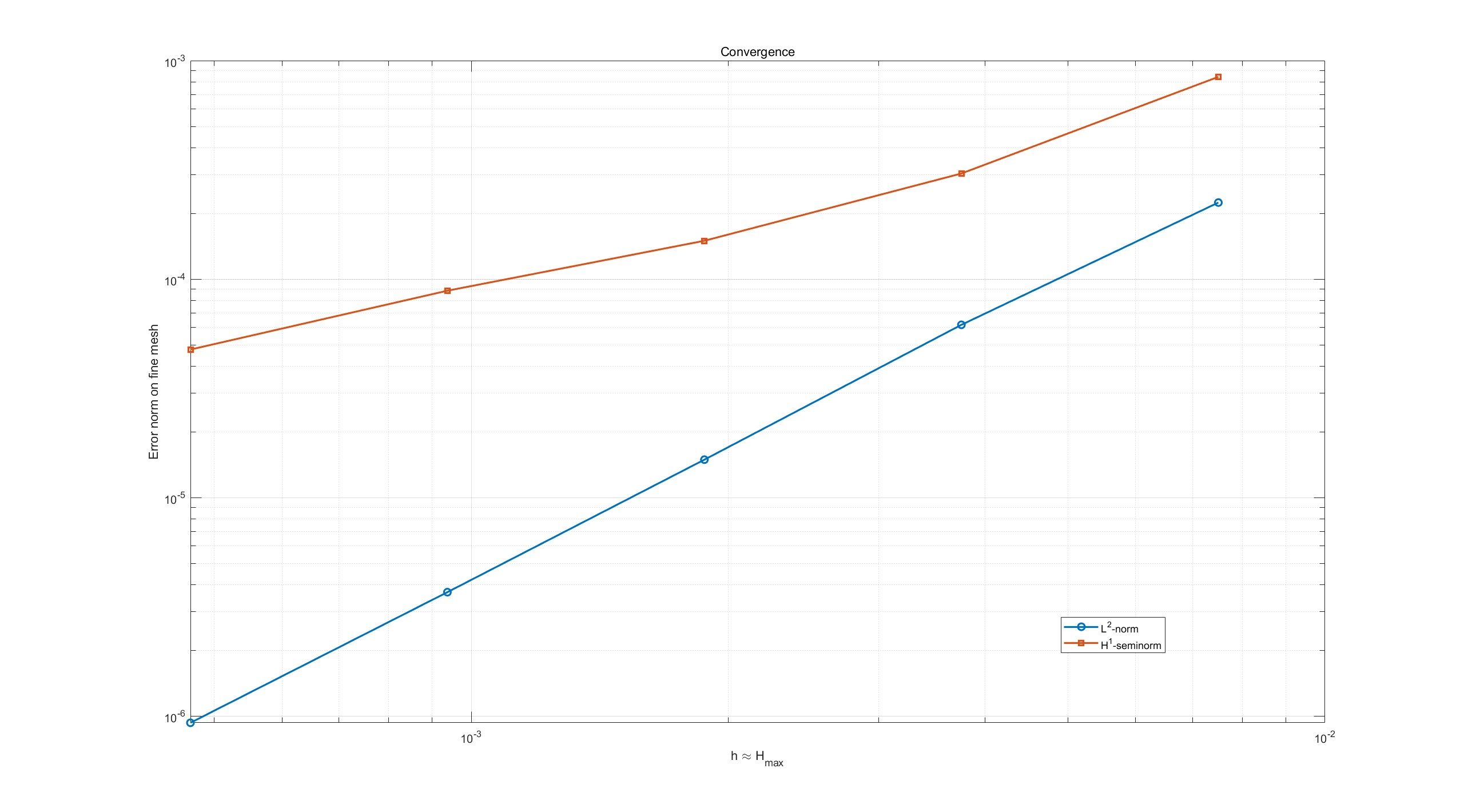} 
  \caption{Self-consistent convergence assessment without an analytic reference. On a mesh hierarchy refined with ratio $r = 2$, the coarse solution is projected onto the fine mesh $P_{h_c \to h_f} T_{h_c}$, and the inter-level difference $w = T_{h_f} - P_{h_c \to h_f} T_{h_c}$ is exactly integrated on the fine mesh using the P1 mass-matrix formula and constant per-element gradients. The horizontal axis is $h \approx H_{\text{max}}$ (log scale) and the vertical axis shows error norms (log scale). The blue curve (circles) reports $\| w \|_{L^2(\Omega)}$, while the red curve (squares) reports $\| \nabla w \|_{L^2(\Omega)}$ (the $H^1$ seminorm); the fitted lines indicate the expected optimal trends (approximately second order in $L^2$ and first order in $H^1$).  }
  \label{L2}
\end{figure*}
\begin{table*}[t]
  \centering
  \setlength{\tabcolsep}{6pt}
  \renewcommand{\arraystretch}{1.12}
  \caption{Convergence study without an analytic reference, reported in three blocks.
 (a) Mesh hierarchy: \emph{Level}—refinement index from coarse (1) to fine (6);
  $H_{\max}$—target maximum element size used in \texttt{generateMesh('Hmax',\,$H_{\max}$)};
  \emph{Nodes/Elems}—counts of mesh nodes and triangular elements in the disk.
  (b) Inter-level $L^2$ error on the fine mesh: \emph{Pair}—successive levels used to form the difference;
  $h_f$—$H_{\max}$ of the fine level; $\|\Delta\|_{L^2(\Omega)}$ with
  $\Delta=T_{h_f}-P_{h_c\to h_f}T_{h_c}$, obtained by exact element-wise integration on the fine mesh via the P1 mass-matrix formula.
 (c) Observed orders: \emph{Global $L^2$ slope}—slope of a linear fit of $\log\|\Delta\|_{L^2}$ vs.\ $\log h$;
  \emph{Local $i$–$i{+}1$–$i{+}2$}—three-level estimate $\log(D_i/D_{i+1})/\log(h_i/h_{i+1})$;
  \emph{Global $H^1$ slope}—slope of $\log\|\nabla\Delta\|_{L^2}$ vs.\ $\log h$, with $\|\nabla\Delta\|_{L^2}$ exactly integrated on the fine mesh.}
  \label{tab:conv_all_final}

  \begin{subtable}{\textwidth}
    \centering
    \caption{Mesh hierarchy}
    \begin{tabular*}{\textwidth}{@{\extracolsep{\fill}} r r r r @{}}
      \toprule
      Level & $H_{\max}$ & Nodes & Elems \\
      \midrule
      1 & \USnum{1.500e-2} & \USnum{1485}    & \USnum{710}    \\
      2 & \USnum{7.500e-3} & \USnum{6161}    & \USnum{3018}   \\
      3 & \USnum{3.750e-3} & \USnum{24901}   & \USnum{12324}  \\
      4 & \USnum{1.875e-3} & \USnum{92941}   & \USnum{46218}  \\
      5 & \USnum{9.370e-4} & \USnum{365565}  & \USnum{182280} \\
      6 & \USnum{4.690e-4} & \USnum{1377045} & \USnum{687516} \\
      \bottomrule
    \end{tabular*}
  \end{subtable}

  \vspace{4pt}

  \begin{subtable}{\textwidth}
    \centering
    \caption{Inter-level $L^2$ errors on the fine mesh}
    \begin{tabular*}{\textwidth}{@{\extracolsep{\fill}} r r r @{}}
      \toprule
      Pair & $h_f$ & $\|\Delta\|_{L^2(\Omega)}$ \\
      \midrule
      1 & \USnum{7.500e-3} & \USnum{2.2400e-4} \\
      2 & \USnum{3.750e-3} & \USnum{6.1744e-5} \\
      3 & \USnum{1.875e-3} & \USnum{1.4904e-5} \\
      4 & \USnum{9.370e-4} & \USnum{3.6877e-6} \\
      5 & \USnum{4.690e-4} & \USnum{9.3069e-7} \\
      \bottomrule
    \end{tabular*}
  \end{subtable}

  \vspace{4pt}

  \begin{subtable}{\textwidth}
    \centering
    \caption{Observed convergence orders}
    \begin{tabular*}{\textwidth}{@{\extracolsep{\fill}} l r @{}}
      \toprule
      Metric & Order \\
      \midrule
      Global $L^2$ slope  & 1.989 \\
      Local $1$–$2$–$3$   & 1.859 \\
      Local $2$–$3$–$4$   & 2.051 \\
      Local $3$–$4$–$5$   & 2.015 \\
      Local $4$–$5$–$6$   & 1.986 \\
      Global $H^1$ slope  & 1.008 \\
      \bottomrule
    \end{tabular*}
  \end{subtable}

\end{table*}
We consider the disk:
\begin{equation}
    \Omega:=\{(x,y)\in\mathbb{R}^2:\ x^2+y^2<R^2\},\quad
\Gamma:=\partial\Omega=\{r=R\}.
\end{equation}
Treat $k>0$ (thermal conductivity), $Q\in\mathbb{R}$ (volumetric heat source), $h\ge 0$ (convective heat–transfer coefficient), and $T_\infty\in\mathbb{R}$ (ambient temperature) as given constants. The steady heat equation reads:
\begin{equation}
    \begin{cases}
-\nabla\!\cdot\!\big(k\,\nabla T\big)=Q, & \text{in } \Omega,\\[6pt]
-\,k\,\dfrac{\partial T}{\partial n}=h\,\big(T-T_\infty\big), & \text{on } \Gamma,
\end{cases}
\end{equation}
where $n$ denotes the outward unit normal. The first line states a Poisson equation for isotropic constant conductivity; the second line imposes a convective boundary condition (Newton’s law of cooling). And we have the energy-balance check (any subdomain $\omega\subset\Omega$):
\begin{equation}
    \int_{\omega} Q\,\mathrm{d}x
= \int_{\partial\omega} \big(-k\,\nabla T\cdot n\big)\,\mathrm{d}s .
\end{equation}
We choose the trial/test space $V:=H^1(\Omega)$. For any $v\in V$, we multiply the strong form by $v$ and integrate by parts (Green’s identity) to obtain:
\begin{equation}
    \int_\Omega k\,\nabla T\cdot\nabla v\,\mathrm{d}x
\;+\;\int_\Gamma h\,T\,v\,\mathrm{d}s
\;=\;
\int_\Omega Q\,v\,\mathrm{d}x
\;+\;\int_\Gamma h\,T_\infty\,v\,\mathrm{d}s .
\end{equation}
We define the bilinear form and the linear functional:
\begin{equation}
\begin{aligned}
        a(u,v)&:=\int_\Omega k\,\nabla u\cdot\nabla v\,\mathrm{d}x
\;+\;\int_\Gamma h\,u\,v\,\mathrm{d}s,
\\
\ell(v)&:=\int_\Omega Q\,v\,\mathrm{d}x
\;+\;\int_\Gamma h\,T_\infty\,v\,\mathrm{d}s .
\end{aligned}
\end{equation}
We state the weak problem: find $T\in V$ such that:
\begin{equation}\label{c3}
    a(T,v)=\ell(v)\quad \forall\, v\in V.
\end{equation}
We also introduce the energy functional:
\begin{equation}
    \mathcal{E}(u)
=\frac12 \int_\Omega k\,|\nabla u|^2\,\mathrm{d}x
+\frac12 \int_\Gamma h\,(u-T_\infty)^2\,\mathrm{d}s
-\int_\Omega Q\,u\,\mathrm{d}x .
\end{equation}
We minimize $\mathcal{E}$ over $V$ and obtain the solution as $T=\arg\min_{u\in V}\mathcal{E}(u),$
and the Euler–Lagrange condition reads $a(T,v)=\ell(v)$ for all $v\in V$.
\begin{lemma}
We invoke the trace theorem and obtain $\|v\|_{L^2(\Gamma)} \le C_{\mathrm{tr}} \|v\|_{H^1(\Omega)}$ for every $v\in H^1(\Omega)$. Furthermore, we need (and we will prove) the following statement: if $\Gamma_0\subset \Gamma$ has positive arc measure, then there exists a constant $C_P>0$ such that:
\begin{equation}\label{Cl5}
    \|v\|_{L^2(\Omega)} \le C_P\!\left(\|\nabla v\|_{L^2(\Omega)}+\|v\|_{L^2(\Gamma_0)}\right),\quad \forall\, v\in H^1(\Omega). 
\end{equation}
\end{lemma}
\begin{proof}
We argue by contradiction. Otherwise, we can choose $v_n\in H^1(\Omega)$ with $\|v_n\|_{L^2(\Omega)}=1$ and $\|\nabla v_n\|_{L^2(\Omega)}+\|v_n\|_{L^2(\Gamma_0)}\to 0$. We invoke the Rellich-Kondrachov compact embedding and the trace theorem; after passing to a subsequence (and keeping the notation $v_n$), we obtain $v_n\to v$ in $L^2(\Omega)$, $v_n|_{\Gamma_0}\to v|_{\Gamma_0}$ in $L^2(\Gamma_0)$, and $\nabla v_n\to 0$ in $L^2(\Omega)$. These limits give $\nabla v=0$, so $v$ equals a constant; the boundary convergence yields $v|_{\Gamma_0}=0$, hence that constant equals $0$. This contradicts $\|v\|_{L^2(\Omega)}=\lim \|v_n\|_{L^2(\Omega)}=1$.
\end{proof}
\begin{theorem}
We work on a bounded $\Omega\subset\mathbb{R}^2$ with a piecewise $C^1$ boundary $\Gamma$. We impose $k\in L^\infty(\Omega)$ and $k(x)\ge k_0>0$ (a.e.); we impose $h\in L^\infty(\Gamma)$ with $h\ge 0$ (a.e.) and $h\not\equiv 0$ (so $\mathrm{meas}\,\Gamma_h>0$ with $\Gamma_h:=\{x\in\Gamma:\ h(x)>0\}$). We set $Q\in L^2(\Omega)$ and choose a constant $T_\infty\in\mathbb{R}$. We write:
\begin{equation}
    \begin{aligned}
        a(u,v)&:=\int_\Omega k\,\nabla u\cdot\nabla v\,\mathrm{d}x \;+\; \int_\Gamma h\,u\,v\,\mathrm{d}s,
\\
\ell(v)&:=\int_\Omega Q\,v\,\mathrm{d}x \;+\; \int_\Gamma h\,T_\infty\,v\,\mathrm{d}s.
    \end{aligned}
\end{equation}
We show $a(\cdot,\cdot)$ remains continuous and coercive on $H^1(\Omega)$ and $\ell(\cdot)$ remains continuous; hence the weak problem $\text{find } T\in H^1(\Omega)\ \text{such that } a(T,v)=\ell(v)\quad \forall\,v\in H^1(\Omega)$
has a unique solution.
\end{theorem}
\begin{proof}
    Since $h\not\equiv 0$ and $h\in L^\infty(\Gamma)$, we set $M:=\operatorname*{ess\,sup}_{\Gamma} h>0$ and choose $\eta:=M/2>0$. We define the set $\Gamma_\eta:=\{x\in\Gamma:\ h(x)\ge \eta\}$, which has positive measure (by the definition of the essential supremum). Hence:
\begin{equation}
    \int_{\Gamma} h\,v^{2}\,\mathrm{d}s
\;\ge\; \int_{\Gamma_\eta} h\,v^{2}\,\mathrm{d}s
\;\ge\; \eta\,\|v\|_{L^{2}(\Gamma_\eta)}^{2}.
\end{equation}
For any $v\in H^{1}(\Omega)$, we apply Eq.\eqref{Cl5} with $\Gamma_{0}=\Gamma_\eta$ and obtain:
\begin{equation}
 \begin{aligned}
        \|v\|_{L^{2}(\Omega)}^{2}
&\le C_{P}^{2}\!\left(\|\nabla v\|_{L^{2}(\Omega)}^{2}+\|v\|_{L^{2}(\Gamma_\eta)}^{2}\right)
\\&\le C_{P}^{2}\!\left(\|\nabla v\|_{L^{2}(\Omega)}^{2}+\frac{1}{\eta}\int_{\Gamma} h\,v^{2}\,\mathrm{d}s\right).
 \end{aligned}
\end{equation}
We combine this with $k(x)\ge k_{0}$ and obtain:
\begin{equation}
    \begin{aligned}
        a(v,v)
&=\int_{\Omega} k\,|\nabla v|^{2}\,\mathrm{d}x + \int_{\Gamma} h\,v^{2}\,\mathrm{d}s
\\&\ge k_{0}\,\|\nabla v\|_{L^{2}(\Omega)}^{2} + \int_{\Gamma} h\,v^{2}\,\mathrm{d}s
\\&\ge \alpha\!\left(\|\nabla v\|_{L^{2}(\Omega)}^{2}+\|v\|_{L^{2}(\Omega)}^{2}\right),
    \end{aligned}
\end{equation}
where:
\begin{equation}
    \alpha:=\min\!\left\{\,k_{0},\ \frac{\eta}{1+C_{P}^{2}}\,\right\}>0.
\end{equation}
Since the bracket on the right defines a norm equivalent to $\|v\|_{H^{1}(\Omega)}^{2}$, we write:
\begin{equation}
    a(v,v)\;\ge\; c\,\|v\|_{H^{1}(\Omega)}^{2}, 
\quad c:=\frac{\alpha}{\,1+C_{\mathrm{eq}}\,}>0,
\end{equation}
where $C_{\mathrm{eq}}$ denotes the equivalence constant between $\|\nabla v\|_{L^{2}}^{2}+\|v\|_{L^{2}}^{2}$ and $\|v\|_{H^{1}}^{2}$. Thus we establish the desired coercivity:
\begin{equation}
    a(v,v)\;\ge\; c\,\|v\|_{H^{1}(\Omega)}^{2},\quad \forall\, v\in H^{1}(\Omega).
\end{equation}
We apply Hölder and the trace estimate and bound:
\begin{equation}
    \begin{aligned}
|a(u,v)|
&\le \|k\|_{L^\infty(\Omega)}\,\|\nabla u\|_{L^2(\Omega)}\,\|\nabla v\|_{L^2(\Omega)}
\\&+ \|h\|_{L^\infty(\Gamma)}\,\|u\|_{L^2(\Gamma)}\,\|v\|_{L^2(\Gamma)} \\
&\le C_a\,\|u\|_{H^1(\Omega)}\,\|v\|_{H^1(\Omega)},
\end{aligned}
\end{equation}
with \(C_a := \|k\|_{L^\infty} + \|h\|_{L^\infty} C_{\mathrm{tr}}^{\,2}\). Similarly, we bound:
\begin{equation}
    \begin{aligned}
        |\ell(v)|
&\le \|Q\|_{L^2(\Omega)}\,\|v\|_{L^2(\Omega)}\\
&+ \|h\|_{L^\infty(\Gamma)}\,|T_\infty|\,|\Gamma|^{1/2}\,\|v\|_{L^2(\Gamma)}\\
&\le C_\ell\,\|v\|_{H^1(\Omega)},
    \end{aligned}
\end{equation}
for a constant $C_\ell$. We set the Hilbert space \(H:=H^1(\Omega)\) with norm \(\|v\|_H:=\|v\|_{H^1(\Omega)}\). We define the operator \(A:H\to H^*\) by:
\begin{equation}
    \langle A u, v\rangle_H \equiv a(u,v),\quad \forall\,u,v\in H.
\end{equation}
We take a coercivity constant \(c>0\) so that \(a(v,v)\ge c\,\|v\|_H^2\), and a continuity constant \(C_a>0\) so that \(|a(u,v)|\le C_a\,\|u\|_H\,\|v\|_H\). From (iv) we also have \(\ell\in H^*\) with \(|\ell(v)|\le C_\ell\,\|v\|_H\). Therefore \(A\) defines a bounded, coercive operator, and \(\ell\) defines a bounded linear functional. We apply Lax-Milgram and obtain a unique $T\in H$ that satisfies:
\begin{equation}
    a(T,v)=\ell(v)\quad \forall\, v\in H.
\end{equation}
We choose $v=T$ and get the a priori stability estimate (all constants explicit):
\begin{equation}
c\,\|T\|_{H}^{2}\le \ell(T)\le \|\ell\|_{H^*}\,\|T\|_{H},
\end{equation}
So we have:
\begin{equation}
    \|T\|_{H^1(\Omega)}\le \frac{\|\ell\|_{H^*}}{c}\le \frac{C_\ell}{c},
\end{equation}
where:
\begin{equation}
    C_\ell=\|Q\|_{L^2(\Omega)}+\|h\|_{L^\infty(\Gamma)}\,|T_\infty|\,C_{\mathrm{tr}}\,|\Gamma|^{1/2}.
\end{equation}
Given $(Q,T_\infty)$ and $(\tilde Q,\tilde T_\infty)$ with solutions $T$ and $\tilde T$, we obtain:
\begin{equation}
    \begin{aligned}
        \|T-\tilde T\|_{H}&\le \frac{1}{c}\,\|\ell-\tilde\ell\|_{H^*}
\\&\le \frac{1}{c}(\|Q-\tilde Q\|_{L^2(\Omega)}\\
&+\|h\|_{L^\infty(\Gamma)}\,|T_\infty-\tilde T_\infty|\,C_{\mathrm{tr}}\,|\Gamma|^{1/2}).
    \end{aligned}
\end{equation}
Therefore, we conclude that the solution is unique and stable.
\end{proof}
We take a triangulation $\mathcal{T}_h$ with node set $\{x_i\}_{i=1}^N$ and choose linear Lagrange shape functions $\{\varphi_i\}_{i=1}^N\subset H^1(\Omega)$. We write the numerical solution as:
\begin{equation}
    T_h(x)=\sum_{i=1}^N T_i\,\varphi_i(x).
\end{equation}
We substitute into (3) and set $v=\varphi_j$ to obtain the linear system:
\begin{equation}\label{ls}
    \underbrace{[K+C]}_{A}\,\mathbf{T}=\mathbf{f}+\mathbf{g},
\end{equation}
where:
\begin{equation}
    K_{ij}=\int_\Omega k\,\nabla\varphi_i\cdot\nabla\varphi_j\,\mathrm{d}x,\quad
C_{ij}=\int_\Gamma h\,\varphi_i\varphi_j\,\mathrm{d}s,
\end{equation}
and:
\begin{equation}
    f_j=\int_\Omega Q\,\varphi_j\,\mathrm{d}x,\quad
g_j=\int_\Gamma h\,T_\infty\,\varphi_j\,\mathrm{d}s.
\end{equation}
When $h>0$, the matrix $A$ is symmetric positive definite, and we solve it with methods such as conjugate gradients. MATLAB’s PDE Toolbox generates the mesh, assembles the elements, and solves the system. Consider a triangular element $e$ with vertices $(x_1,y_1),(x_2,y_2),(x_3,y_3)$ and area $A_e$. For P1 elements, the shape-function gradients are constant on $e$:
\begin{equation}
    \nabla N_i=\frac{1}{2A_e}\begin{bmatrix} b_i \\ c_i \end{bmatrix},\quad
\begin{aligned}
&b_1=y_2-y_3,\; b_2=y_3-y_1,\; b_3=y_1-y_2,\\
&c_1=x_3-x_2,\; c_2=x_1-x_3,\; c_3=x_2-x_1.
\end{aligned}
\end{equation}
Thus we have:
\begin{equation}
    (K_e)_{ij}=k\int_{\Omega_e}\nabla N_i\cdot\nabla N_j\,\mathrm{d}x
= k\,\frac{b_i b_j+c_i c_j}{4A_e}.
\end{equation}
For a constant source $Q$, we have:
\begin{equation}
    (f_e)_i=\int_{\Omega_e} Q\,N_i\,\mathrm{d}x=\frac{Q\,A_e}{3}.
\end{equation}
Along a convective edge $\gamma\subset\Gamma$ of length $L_\gamma$ (P1 on a segment), we obtain:
\begin{equation}
    (C_\gamma)=\frac{h L_\gamma}{6}\begin{bmatrix}2&1\\[2pt]1&2\end{bmatrix},\quad
(g_\gamma)=\frac{h T_\infty L_\gamma}{2}\begin{bmatrix}1\\[2pt]1\end{bmatrix}.
\end{equation}
We assemble all element and boundary contributions into global DOFs and recover Eq.~\eqref{ls}. Having formed the global linear system in Eq.\,\eqref{ls}, we solve it on a geometric mesh hierarchy $\{ h_\ell \}$ with refinement ratio $r = 2$, yielding solutions $T_{h_\ell}$. To quantify accuracy without an analytic reference, we compare successive levels by projecting the coarse solution onto the fine mesh, $P_{h_c \to h_f} T_{h_c}$, and exactly integrating on the fine mesh the inter-level difference $w := T_{h_f} - P_{h_c \to h_f} T_{h_c}$ to obtain $\| w \|_{L^2(\Omega)}$ and $\| \nabla w \|_{L^2(\Omega)}$. Linear fits on the $\log h$-$\log$ error plots give global rates $p_{L^2} = 1.989$ and $p_{H^1} = 1.008$, consistent with the optimal orders for P1 elements (second order in $L^2$, first order in $H^1$); see Fig.\,\ref{L2}, with level-wise data reported in Table.\,\ref{tab:conv_all_final}.

In conclusion, the self-consistent assessment—projecting the coarse solution onto the fine mesh and exactly integrating on the fine mesh—confirms optimal convergence consistent with linear (P1) theory, with assembly, boundary treatment, meshing, and post-processing being mutually consistent and numerically stable.
{
\section{Learning-rate comparison, complete implementation parameters, and strict framework-based parameter-sensitivity analysis for the LNN--PINN drift--decay benchmark}\label{a}

This appendix supplements the 1D advection--reaction benchmark in Section~3.1 and treats only the uploaded LNN--PINN implementation. The current main text does not assign local labels to the drift--decay equations in Section~3.1. Therefore, to avoid any false cross-reference, we rewrite the case-specific equations here in full and only retain the valid references to the framework objective, the framework figure in Fig.~\ref{LNN-PINN}, and the main-text training-history figure in Fig.~\ref{F2}. We do not introduce any subsection or paragraph split in order to keep the appendix as one continuous technical derivation.

The physical benchmark in Section~3.1 is the first-order linear advection--reaction problem
\begin{equation}
 u_t+a u_x+b u=0,
 \qquad
 a=-\frac12,
 \qquad
 b=\frac12,
 \qquad
 (x,t)\in[0,2]\times[0,1].
\label{eq:appdd_pde_raw}
\end{equation}
Multiplying Eq.~\eqref{eq:appdd_pde_raw} by $-2$ gives the exact implementation form used by the uploaded scripts,
\begin{equation}
 u_x-2u_t-u=0.
\label{eq:appdd_pde_impl}
\end{equation}
The initial and inflow constraints are
\begin{equation}
 u(x,0)=6e^{-3x},
 \qquad x\in[0,2],
\label{eq:appdd_ic}
\end{equation}
\begin{equation}
 u(2,t)=6e^{-6-2t},
 \qquad t\in[0,1],
\label{eq:appdd_bc}
\end{equation}
and the analytical reference field used for post-training evaluation is
\begin{equation}
 u^{\ast}(x,t)=6e^{-3x-2t}.
\label{eq:appdd_exact}
\end{equation}

The uploaded scripts use the interior sample set, the initial-line sample set, and the inflow-boundary sample set
\begin{equation}
 \mathcal S_{\Omega}=\{(x_i,t_i)\}_{i=1}^{N_{\Omega}},
 \qquad
 \mathcal S_{\mathrm{IC}}=\{(x_j,0)\}_{j=1}^{N_{\mathrm{IC}}},
 \qquad
 \mathcal S_{\mathrm{BD}}=\{(2,t_k)\}_{k=1}^{N_{\mathrm{BD}}},
\label{eq:appdd_sets}
\end{equation}
with the fixed numerical choices
\begin{equation}
 N_{\Omega}=2000,
 \qquad
 N_{\mathrm{IC}}=1000,
 \qquad
 N_{\mathrm{BD}}=1000.
\label{eq:appdd_counts}
\end{equation}
At every training iteration the scripts redraw these sets from the corresponding uniform laws. The residual vectors therefore are
\begin{equation}
 \mathbf r_{\mathrm{PDE}}(\theta)=\bigl[\partial_xu_{\theta}(x,t)-2\partial_tu_{\theta}(x,t)-u_{\theta}(x,t)\bigr]_{(x,t)\in\mathcal S_{\Omega}},
\label{eq:appdd_rpde}
\end{equation}
\begin{equation}
 \mathbf r_{\mathrm{IC}}(\theta)=\bigl[u_{\theta}(x,0)-6e^{-3x}\bigr]_{(x,0)\in\mathcal S_{\mathrm{IC}}},
\label{eq:appdd_ric}
\end{equation}
\begin{equation}
 \mathbf r_{\mathrm{BD}}(\theta)=\bigl[u_{\theta}(2,t)-6e^{-6-2t}\bigr]_{(2,t)\in\mathcal S_{\mathrm{BD}}}.
\label{eq:appdd_rbd}
\end{equation}
The corresponding mean-squared residual channels are
\begin{equation}
 L_{\mathrm{PDE}}(\theta)=\frac{1}{N_{\Omega}}\Vert\mathbf r_{\mathrm{PDE}}(\theta)\Vert_2^2,
 \qquad
 L_{\mathrm{IC}}(\theta)=\frac{1}{N_{\mathrm{IC}}}\Vert\mathbf r_{\mathrm{IC}}(\theta)\Vert_2^2,
 \qquad
 L_{\mathrm{BD}}(\theta)=\frac{1}{N_{\mathrm{BD}}}\Vert\mathbf r_{\mathrm{BD}}(\theta)\Vert_2^2,
\label{eq:appdd_losses}
\end{equation}
and the concrete case-level objective minimized by the uploaded LNN--PINN scripts is the direct sum
\begin{equation}
 \mathcal J_{\mathrm{DD}}(\theta)=L_{\mathrm{PDE}}(\theta)+L_{\mathrm{IC}}(\theta)+L_{\mathrm{BD}}(\theta).
\label{eq:appdd_J}
\end{equation}
Thus the implicit loss weights in the sense of Eq.~\eqref{LOSS Total} are exactly
\begin{equation}
 \lambda_{\mathrm{PDE}}=1,
 \qquad
 \lambda_{\mathrm{IC}}=1,
 \qquad
 \lambda_{\mathrm{BD}}=1.
\label{eq:appdd_lambdas}
\end{equation}
The training horizon is fixed to
\begin{equation}
 N_{\mathrm{train}}=8000,
\label{eq:appdd_ntrain}
\end{equation}
while the six uploaded learning-rate runs use
\begin{equation}
 \eta\in\left\{10^{-3},10^{-4},10^{-5},10^{-6},10^{-7},10^{-8}\right\}.
\label{eq:appdd_eta_set}
\end{equation}
The random seed fixed in the scripts is
\begin{equation}
 s=888888.
\label{eq:appdd_seed}
\end{equation}

The actual LNN--PINN forward map implemented in the scripts specializes the framework shown in Fig.~\ref{LNN-PINN}. Define the two-dimensional input vector
\begin{equation}
 \xi=\begin{bmatrix}x\\ t\end{bmatrix}\in\mathbb R^2.
\label{eq:appdd_input}
\end{equation}
The first hidden transformation reads
\begin{equation}
 z_1=\tanh(W_0\xi+b_0),
 \qquad
 h_1=z_1+\operatorname{Diag}(\alpha_1)\tanh(A_1 z_1+c_1).
\label{eq:appdd_h1}
\end{equation}
The next hidden stages are
\begin{equation}
 z_2=\tanh(W_1h_1+b_1),
 \qquad
 h_2=z_2+\operatorname{Diag}(\alpha_2)\tanh(A_2 z_2+c_2),
\label{eq:appdd_h2}
\end{equation}
\begin{equation}
 z_3=\tanh(W_2h_2+b_2),
 \qquad
 h_3=z_3+\operatorname{Diag}(\alpha_3)\tanh(A_3 z_3+c_3),
\label{eq:appdd_h3}
\end{equation}
\begin{equation}
 z_4=\tanh(W_3h_3+b_3),
 \qquad
 h_4=z_4+\operatorname{Diag}(\alpha_4)\tanh(A_4 z_4+c_4),
\label{eq:appdd_h4}
\end{equation}
and the scalar output is
\begin{equation}
 u_{\theta}(x,t)=w_{\mathrm{out}}^{\top}h_4+b_{\mathrm{out}}.
\label{eq:appdd_output}
\end{equation}
The explicit numerical architecture parameters are therefore
\begin{equation}
 d=2,
 \qquad
 o=1,
 \qquad
 w=64,
 \qquad
 H=4,
 \qquad
 q=4,
 \qquad
 \alpha_0=0.5,
 \qquad
 \alpha_{\ell}^{(0)}=\alpha_0\mathbf 1,
 \quad \ell=1,2,3,4.
\label{eq:appdd_archvals}
\end{equation}
Eq.~\eqref{eq:appdd_archvals} already lists each numerical architecture parameter one by one: input dimension $d$, output dimension $o$, hidden width $w$, hidden-stage number $H$, liquid-block number $q$, and the common gate initialization amplitude $\alpha_0$.

The exact trainable-parameter count follows from direct counting. The input affine map contributes
\begin{equation}
 P_{\mathrm{in}}=dw+w,
\label{eq:appdd_Pin}
\end{equation}
there are $(H-1)$ square hidden affine maps contributing
\begin{equation}
 P_{\mathrm{hid}}=(H-1)(w^2+w),
\label{eq:appdd_Phid}
\end{equation}
the scalar output map contributes
\begin{equation}
 P_{\mathrm{out}}=wo+o,
\label{eq:appdd_Pout}
\end{equation}
and each liquid block contributes
\begin{equation}
 P_{\mathrm{liq,one}}=w^2+w+w=w^2+2w.
\label{eq:appdd_Pliqone}
\end{equation}
Therefore the full trainable-parameter count is
\begin{equation}
 P_{\mathrm{LNN}}(d,o,w,H,q)=dw+w+(H-1)(w^2+w)+wo+o+q(w^2+2w).
\label{eq:appdd_PLNN_general}
\end{equation}
Substituting Eq.~\eqref{eq:appdd_archvals} gives
\begin{align}
 P_{\mathrm{LNN}}
 &=2\cdot64+64+3(64^2+64)+64\cdot1+1+4(64^2+2\cdot64)
 \nonumber\\
 &=192+12480+65+16896
 \nonumber\\
 &=29633.
\label{eq:appdd_PLNN_actual}
\end{align}

The uploaded scripts instantiate Adam as \texttt{torch.optim.Adam(model.parameters(), lr=LR)} and do not overwrite the other Adam arguments. Hence the effective numerical optimizer parameters are exactly
\begin{equation}
 \beta_1=0.9,
 \qquad
 \beta_2=0.999,
 \qquad
 \varepsilon_{\mathrm A}=10^{-8},
 \qquad
 \omega_{\mathrm d}=0,
 \qquad
 \chi_{\mathrm{AMS}}=0,
\label{eq:appdd_adamparams}
\end{equation}
where $\omega_{\mathrm d}$ denotes weight decay and $\chi_{\mathrm{AMS}}$ denotes the AMSGrad flag written numerically as $0$ for False. The optimizer recurrence is
\begin{equation}
 g_k=\nabla_{\theta}\mathcal J_{\mathrm{DD}}(\theta_k;\Xi_k),
\label{eq:appdd_gk}
\end{equation}
\begin{equation}
 m_k=\beta_1 m_{k-1}+(1-\beta_1)g_k,
\label{eq:appdd_mk}
\end{equation}
\begin{equation}
 v_k=\beta_2 v_{k-1}+(1-\beta_2)(g_k\odot g_k),
\label{eq:appdd_vk}
\end{equation}
\begin{equation}
 \widehat m_k=\frac{m_k}{1-\beta_1^k},
 \qquad
 \widehat v_k=\frac{v_k}{1-\beta_2^k},
\label{eq:appdd_hatmv}
\end{equation}
\begin{equation}
 d_k=\frac{\widehat m_k}{\sqrt{\widehat v_k}+\varepsilon_{\mathrm A}},
\label{eq:appdd_dk}
\end{equation}
\begin{equation}
 \theta_{k+1}=\theta_k-\eta d_k.
\label{eq:appdd_adam_update}
\end{equation}
We now carry out strict parameter sensitivity one numerical parameter at a time.

The learning-rate sensitivity follows directly from Eq.~\eqref{eq:appdd_adam_update}. Freeze $\theta_k$, $m_k$, $v_k$, and $\Xi_k$ and perturb $\eta\mapsto\eta+\delta\eta$. Then
\begin{equation}
 \theta_{k+1}(\eta+\delta\eta)-\theta_{k+1}(\eta)=-\delta\eta\,d_k,
\label{eq:appdd_eta_diff}
\end{equation}
so the exact local derivative is
\begin{equation}
 \left.\frac{\partial\theta_{k+1}}{\partial\eta}\right|_{(\theta_k,m_k,v_k,\Xi_k)}=-d_k,
 \qquad
 \|\delta\theta_{k+1}\|_2\le |\delta\eta|\,\|d_k\|_2.
\label{eq:appdd_eta_local}
\end{equation}
For objective sensitivity, Taylor expansion of $\mathcal J_{\mathrm{DD}}$ along the actual Adam direction yields
\begin{equation}
 \mathcal J_{\mathrm{DD}}(\theta_{k+1};\Xi_k)
 =
 \mathcal J_{\mathrm{DD}}(\theta_k;\Xi_k)
 -\eta g_k^{\top}d_k
 +\frac{\eta^2}{2}
 d_k^{\top}
 \nabla_\theta^2\mathcal J_{\mathrm{DD}}(\theta_k-\tau_k\eta d_k;\Xi_k)
 d_k
\label{eq:appdd_eta_taylor}
\end{equation}
for some $\tau_k\in(0,1)$. If
\begin{equation}
 g_k^{\top}d_k\ge c_k\|g_k\|_2\|d_k\|_2,
 \qquad
 \left\|\nabla_\theta^2\mathcal J_{\mathrm{DD}}(\theta_k-\tau_k\eta d_k;\Xi_k)\right\|_2\le M_k,
\label{eq:appdd_eta_cond}
\end{equation}
then a sufficient local descent interval is
\begin{equation}
 0<\eta<\frac{2c_k\|g_k\|_2}{M_k\|d_k\|_2}.
\label{eq:appdd_eta_range}
\end{equation}
Eq.~\eqref{eq:appdd_eta_range} is the rigorous meaning of a reasonable learning-rate range for the actual uploaded network and objective.

The epoch-count sensitivity follows from summing Eq.~\eqref{eq:appdd_adam_update}. After $K$ iterations,
\begin{equation}
 \theta_K=\theta_0-\eta\sum_{k=0}^{K-1}d_k.
\label{eq:appdd_thetaK}
\end{equation}
Thus changing the training horizon from $K$ to $K+\Delta K$ gives
\begin{equation}
 \theta_{K+\Delta K}-\theta_K=-\eta\sum_{k=K}^{K+\Delta K-1}d_k,
\label{eq:appdd_epochdiff}
\end{equation}
so that
\begin{equation}
 \|\theta_{K+\Delta K}-\theta_K\|_2\le \eta\sum_{k=K}^{K+\Delta K-1}\|d_k\|_2.
\label{eq:appdd_epochbound}
\end{equation}
Consequently the numerical parameter $N_{\mathrm{train}}=8000$ is reasonable only when the tail sum on the right-hand side is already sufficiently small. If one prescribes an admissible late-training displacement budget $\varepsilon_{\mathrm{ep}}$, then a sufficient stopping rule is
\begin{equation}
 \eta\sum_{k=K}^{K+\Delta K-1}\|d_k\|_2\le \varepsilon_{\mathrm{ep}}.
\label{eq:appdd_epochrange}
\end{equation}
Eq.~\eqref{eq:appdd_epochrange} gives the rigorous acceptable range criterion for the training-horizon parameter.

The sample-count sensitivities for $N_{\Omega}=2000$, $N_{\mathrm{IC}}=1000$, and $N_{\mathrm{BD}}=1000$ follow from the stochastic nature of Eq.~\eqref{eq:appdd_J}. Define the squared residual random variables
\begin{equation}
 X_{\Omega}=\bigl(\partial_xu_\theta(x,t)-2\partial_tu_\theta(x,t)-u_\theta(x,t)\bigr)^2,
 \qquad (x,t)\sim\mathrm{Unif}([0,2]\times[0,1]),
\label{eq:appdd_XOmega}
\end{equation}
\begin{equation}
 X_{\mathrm{IC}}=\bigl(u_\theta(x,0)-6e^{-3x}\bigr)^2,
 \qquad x\sim\mathrm{Unif}([0,2]),
\label{eq:appdd_XIC}
\end{equation}
\begin{equation}
 X_{\mathrm{BD}}=\bigl(u_\theta(2,t)-6e^{-6-2t}\bigr)^2,
 \qquad t\sim\mathrm{Unif}([0,1]).
\label{eq:appdd_XBD}
\end{equation}
Let
\begin{equation}
 \sigma_{\Omega}^2(\theta)=\operatorname{Var}(X_{\Omega}),
 \qquad
 \sigma_{\mathrm{IC}}^2(\theta)=\operatorname{Var}(X_{\mathrm{IC}}),
 \qquad
 \sigma_{\mathrm{BD}}^2(\theta)=\operatorname{Var}(X_{\mathrm{BD}}).
\label{eq:appdd_sigmas}
\end{equation}
Then
\begin{equation}
 \operatorname{Var}(L_{\mathrm{PDE}})=\frac{\sigma_{\Omega}^2(\theta)}{N_{\Omega}},
 \qquad
 \operatorname{Var}(L_{\mathrm{IC}})=\frac{\sigma_{\mathrm{IC}}^2(\theta)}{N_{\mathrm{IC}}},
 \qquad
 \operatorname{Var}(L_{\mathrm{BD}})=\frac{\sigma_{\mathrm{BD}}^2(\theta)}{N_{\mathrm{BD}}},
\label{eq:appdd_varchannels}
\end{equation}
and therefore
\begin{equation}
 \operatorname{Var}(\mathcal J_{\mathrm{DD}})=
 \frac{\sigma_{\Omega}^2(\theta)}{N_{\Omega}}+
 \frac{\sigma_{\mathrm{IC}}^2(\theta)}{N_{\mathrm{IC}}}+
 \frac{\sigma_{\mathrm{BD}}^2(\theta)}{N_{\mathrm{BD}}}.
\label{eq:appdd_vartotal}
\end{equation}
Differentiating Eq.~\eqref{eq:appdd_vartotal} gives the exact sensitivities
\begin{equation}
 \frac{\partial}{\partial N_{\Omega}}\operatorname{Var}(\mathcal J_{\mathrm{DD}})=-\frac{\sigma_{\Omega}^2(\theta)}{N_{\Omega}^2},
\label{eq:appdd_dvarNO}
\end{equation}
\begin{equation}
 \frac{\partial}{\partial N_{\mathrm{IC}}}\operatorname{Var}(\mathcal J_{\mathrm{DD}})=-\frac{\sigma_{\mathrm{IC}}^2(\theta)}{N_{\mathrm{IC}}^2},
\label{eq:appdd_dvarNIC}
\end{equation}
\begin{equation}
 \frac{\partial}{\partial N_{\mathrm{BD}}}\operatorname{Var}(\mathcal J_{\mathrm{DD}})=-\frac{\sigma_{\mathrm{BD}}^2(\theta)}{N_{\mathrm{BD}}^2}.
\label{eq:appdd_dvarNBD}
\end{equation}
Hence each sample count admits an exact reasonable-range criterion. If the acceptable variance budgets of the three channels are $\delta_{\Omega}^2$, $\delta_{\mathrm{IC}}^2$, and $\delta_{\mathrm{BD}}^2$, then sufficient lower bounds are
\begin{equation}
 N_{\Omega}\ge \frac{\sigma_{\Omega}^2(\theta)}{\delta_{\Omega}^2},
 \qquad
 N_{\mathrm{IC}}\ge \frac{\sigma_{\mathrm{IC}}^2(\theta)}{\delta_{\mathrm{IC}}^2},
 \qquad
 N_{\mathrm{BD}}\ge \frac{\sigma_{\mathrm{BD}}^2(\theta)}{\delta_{\mathrm{BD}}^2}.
\label{eq:appdd_Nranges}
\end{equation}
These are not heuristic statements; they are direct consequences of Eqs.~\eqref{eq:appdd_varchannels}--\eqref{eq:appdd_dvarNBD}.

The seed parameter $s=888888$ is discrete rather than differentiable. Its strict mathematical sensitivity is measured by seed-to-seed realization gaps. If $\Xi_k^{(s_1)}$ and $\Xi_k^{(s_2)}$ denote the collocation streams generated by two seeds, then the exact seed sensitivity at iteration $k$ is
\begin{equation}
 \Delta_{s_1,s_2}\mathcal J_{\mathrm{DD}}(\theta_k)=\mathcal J_{\mathrm{DD}}(\theta_k;\Xi_k^{(s_1)})-\mathcal J_{\mathrm{DD}}(\theta_k;\Xi_k^{(s_2)}).
\label{eq:appdd_seedgap}
\end{equation}
A reasonable seed policy for a fair learning-rate comparison therefore requires the same $s$ across all six runs, which the uploaded scripts indeed enforce.

We now treat the optimizer parameters $\beta_1$, $\beta_2$, and $\varepsilon_{\mathrm A}$ one by one. From Eq.~\eqref{eq:appdd_mk}, one-step differentiation with respect to $\beta_1$ gives
\begin{equation}
 \frac{\partial m_k}{\partial\beta_1}=m_{k-1}-g_k+\beta_1\frac{\partial m_{k-1}}{\partial\beta_1},
\label{eq:appdd_dm_db1}
\end{equation}
which unrolls to
\begin{equation}
 \frac{\partial m_k}{\partial\beta_1}=\sum_{j=1}^{k}\beta_1^{k-j}(m_{j-1}-g_j).
\label{eq:appdd_dm_db1_unroll}
\end{equation}
Since
\begin{equation}
 \widehat m_k=\frac{m_k}{1-\beta_1^k},
\label{eq:appdd_hatm_repeat}
\end{equation}
its derivative is
\begin{equation}
 \frac{\partial \widehat m_k}{\partial\beta_1}
 =
 \frac{(1-\beta_1^k)\frac{\partial m_k}{\partial\beta_1}+k\beta_1^{k-1}m_k}{(1-\beta_1^k)^2}.
\label{eq:appdd_dhatm_db1}
\end{equation}
Using Eq.~\eqref{eq:appdd_dk}, the exact sensitivity of the Adam direction with respect to $\beta_1$ is
\begin{equation}
 \frac{\partial d_k}{\partial\beta_1}
 =
 \frac{1}{\sqrt{\widehat v_k}+\varepsilon_{\mathrm A}}
 \frac{\partial\widehat m_k}{\partial\beta_1}.
\label{eq:appdd_dd_db1}
\end{equation}
Hence
\begin{equation}
 \frac{\partial\theta_{k+1}}{\partial\beta_1}=-\eta\frac{\partial d_k}{\partial\beta_1}.
\label{eq:appdd_dtheta_db1}
\end{equation}
Eq.~\eqref{eq:appdd_dtheta_db1} is the strict stepwise sensitivity law for the numerical parameter $\beta_1=0.9$. A reasonable range for $\beta_1$ requires the denominator $(1-\beta_1^k)^2$ in Eq.~\eqref{eq:appdd_dhatm_db1} to stay safely away from $0$ while still allowing enough first-moment smoothing; that range is the open interval
\begin{equation}
 0<\beta_1<1.
\label{eq:appdd_beta1_range_basic}
\end{equation}
Values too close to $1$ produce excessive memory and slow adaptation, while values too close to $0$ collapse the momentum effect.

For the numerical parameter $\beta_2=0.999$, Eq.~\eqref{eq:appdd_vk} gives
\begin{equation}
 \frac{\partial v_k}{\partial\beta_2}=v_{k-1}-(g_k\odot g_k)+\beta_2\frac{\partial v_{k-1}}{\partial\beta_2},
\label{eq:appdd_dv_db2}
\end{equation}
which unrolls to
\begin{equation}
 \frac{\partial v_k}{\partial\beta_2}=\sum_{j=1}^{k}\beta_2^{k-j}(v_{j-1}-g_j\odot g_j).
\label{eq:appdd_dv_db2_unroll}
\end{equation}
The bias-corrected second moment satisfies
\begin{equation}
 \frac{\partial\widehat v_k}{\partial\beta_2}
 =
 \frac{(1-\beta_2^k)\frac{\partial v_k}{\partial\beta_2}+k\beta_2^{k-1}v_k}{(1-\beta_2^k)^2}.
\label{eq:appdd_dhatv_db2}
\end{equation}
Differentiating Eq.~\eqref{eq:appdd_dk} gives
\begin{equation}
 \frac{\partial d_k}{\partial\beta_2}
 =
 -\frac{\widehat m_k}{2(\sqrt{\widehat v_k}+\varepsilon_{\mathrm A})^2\sqrt{\widehat v_k}}
 \frac{\partial\widehat v_k}{\partial\beta_2},
\label{eq:appdd_dd_db2}
\end{equation}
so that
\begin{equation}
 \frac{\partial\theta_{k+1}}{\partial\beta_2}=-\eta\frac{\partial d_k}{\partial\beta_2}.
\label{eq:appdd_dtheta_db2}
\end{equation}
The rigorous admissible interval again is
\begin{equation}
 0<\beta_2<1,
\label{eq:appdd_beta2_range_basic}
\end{equation}
because outside this interval the exponential moving-average structure in Eq.~\eqref{eq:appdd_vk} ceases to be a stable convex recursion. Values too near $1$ over-smooth the second moment, whereas smaller values raise variance in the denominator of Eq.~\eqref{eq:appdd_dk}.

For the numerical parameter $\varepsilon_{\mathrm A}=10^{-8}$, Eq.~\eqref{eq:appdd_dk} yields the exact derivative
\begin{equation}
 \frac{\partial d_k}{\partial\varepsilon_{\mathrm A}}=-\frac{\widehat m_k}{(\sqrt{\widehat v_k}+\varepsilon_{\mathrm A})^2},
\label{eq:appdd_dd_deps}
\end{equation}
so that
\begin{equation}
 \frac{\partial\theta_{k+1}}{\partial\varepsilon_{\mathrm A}}=\eta\frac{\widehat m_k}{(\sqrt{\widehat v_k}+\varepsilon_{\mathrm A})^2}.
\label{eq:appdd_dtheta_deps}
\end{equation}
This derivative is exact and shows that increasing $\varepsilon_{\mathrm A}$ weakens the step magnitude. The strict admissible range is simply
\begin{equation}
 \varepsilon_{\mathrm A}>0,
\label{eq:appdd_eps_range_basic}
\end{equation}
because the denominator in Eq.~\eqref{eq:appdd_dk} must remain strictly positive. Very large $\varepsilon_{\mathrm A}$ collapses the adaptive normalization; extremely tiny $\varepsilon_{\mathrm A}$ can make the step excessively sensitive where $\widehat v_k$ becomes small.

The loss-weight sensitivities for the numerical values in Eq.~\eqref{eq:appdd_lambdas} follow from the generalized weighted objective
\begin{equation}
 \mathcal J_{\mathrm{DD}}^{\mathrm{gen}}(\theta)=\lambda_{\mathrm{PDE}}L_{\mathrm{PDE}}(\theta)+\lambda_{\mathrm{IC}}L_{\mathrm{IC}}(\theta)+\lambda_{\mathrm{BD}}L_{\mathrm{BD}}(\theta).
\label{eq:appdd_Jgen}
\end{equation}
Its first variation is
\begin{equation}
 \delta\mathcal J_{\mathrm{DD}}^{\mathrm{gen}}=L_{\mathrm{PDE}}\,\delta\lambda_{\mathrm{PDE}}+L_{\mathrm{IC}}\,\delta\lambda_{\mathrm{IC}}+L_{\mathrm{BD}}\,\delta\lambda_{\mathrm{BD}},
\label{eq:appdd_dJlambda}
\end{equation}
so the exact partial derivatives are
\begin{equation}
 \frac{\partial\mathcal J_{\mathrm{DD}}^{\mathrm{gen}}}{\partial\lambda_{\mathrm{PDE}}}=L_{\mathrm{PDE}},
 \qquad
 \frac{\partial\mathcal J_{\mathrm{DD}}^{\mathrm{gen}}}{\partial\lambda_{\mathrm{IC}}}=L_{\mathrm{IC}},
 \qquad
 \frac{\partial\mathcal J_{\mathrm{DD}}^{\mathrm{gen}}}{\partial\lambda_{\mathrm{BD}}}=L_{\mathrm{BD}}.
\label{eq:appdd_dJdlambdas}
\end{equation}
Since the uploaded scripts fix all three weights at $1$, they compare learning rates under a pure equal-weight regime.

We now turn to the architecture parameter sensitivities. For the numerical input dimension $d=2$ and output dimension $o=1$, Eq.~\eqref{eq:appdd_PLNN_general} gives the exact partial derivatives of the parameter-count functional with respect to these two problem-interface dimensions,
\begin{equation}
 \frac{\partial P_{\mathrm{LNN}}}{\partial d}=w,
 \qquad
 \frac{\partial P_{\mathrm{LNN}}}{\partial o}=w+1.
\label{eq:appdd_dP_ddo}
\end{equation}
Since the benchmark is scalar and two-dimensional in $(x,t)$, these two numbers are fixed by the task and do not vary in the present case; nevertheless Eq.~\eqref{eq:appdd_dP_ddo} shows their exact algebraic role.

For the hidden width parameter $w=64$, the exact discrete increment at fixed $d$, $o$, $H$, and $q$ is
\begin{align}
 &P_{\mathrm{LNN}}(d,o,w+1,H,q)-P_{\mathrm{LNN}}(d,o,w,H,q)
 \nonumber\\
 &=(H-1+q)(2w+1)+(d+H+o+2q).
\label{eq:appdd_dP_dw_general}
\end{align}
Substituting Eq.~\eqref{eq:appdd_archvals} yields
\begin{equation}
 P_{\mathrm{LNN}}(65)-P_{\mathrm{LNN}}(64)=7\cdot129+15=918.
\label{eq:appdd_dP_dw_actual}
\end{equation}
This exact count increase of $918$ parameters per additional hidden channel shows the strict sensitivity of the model class to $w$. The width also enters the output sensitivity with respect to the gate parameter through the factor $\sqrt w$, as shown below in Eq.~\eqref{eq:appdd_du_dalpha0_bound}. Therefore the hidden-width parameter influences both capacity and response amplitude.

For the hidden-stage number $H=4$, the exact discrete increment at fixed $d$, $o$, $w$, and $q$ is
\begin{equation}
 P_{\mathrm{LNN}}(d,o,w,H+1,q)-P_{\mathrm{LNN}}(d,o,w,H,q)=w^2+w.
\label{eq:appdd_dP_dH_general}
\end{equation}
At $w=64$ this becomes
\begin{equation}
 P_{\mathrm{LNN}}(H+1)-P_{\mathrm{LNN}}(H)=64^2+64=4160.
\label{eq:appdd_dP_dH_actual}
\end{equation}
Thus one extra hidden stage changes the parameter count by $4160$. More importantly, depth also controls transport amplification. Writing each stage as
\begin{equation}
 \mathcal F_{\ell}(h)=\mathcal B_{\ell}(\tanh(W_{\ell-1}h+b_{\ell-1})),
\label{eq:appdd_Fl}
\end{equation}
with
\begin{equation}
 \mathcal B_{\ell}(z)=z+\operatorname{Diag}(\alpha_{\ell})\tanh(A_{\ell}z+c_{\ell}),
\label{eq:appdd_Bl}
\end{equation}
we obtain
\begin{equation}
 D\mathcal B_{\ell}(z)=I+\operatorname{Diag}(\alpha_{\ell})\operatorname{Diag}(\operatorname{sech}^2(A_{\ell}z+c_{\ell}))A_{\ell}.
\label{eq:appdd_DB}
\end{equation}
Hence
\begin{equation}
 \|D\mathcal B_{\ell}(z)\|_2\le 1+\|\operatorname{Diag}(\alpha_{\ell})\|_2\,\|A_{\ell}\|_2.
\label{eq:appdd_DBbound}
\end{equation}
If
\begin{equation}
 \|W_{\ell-1}\|_2\le \rho,
 \qquad
 \|A_{\ell}\|_2\le \gamma,
 \qquad
 \|\operatorname{Diag}(\alpha_{\ell})\|_2\le \alpha_0,
\label{eq:appdd_rhogamma}
\end{equation}
then
\begin{equation}
 \|D\mathcal F_{\ell}\|_2\le \rho(1+\alpha_0\gamma).
\label{eq:appdd_DFbound}
\end{equation}
Therefore an $H$-stage hidden transport obeys
\begin{equation}
 \operatorname{Lip}_H\le \bigl(\rho(1+\alpha_0\gamma)\bigr)^H.
\label{eq:appdd_LipH}
\end{equation}
If one prescribes an admissible amplification budget $\Lambda_{\star}>1$, then a strict depth admissibility condition is
\begin{equation}
 H\le \frac{\log\Lambda_{\star}}{\log(\rho(1+\alpha_0\gamma))}
 \qquad\text{whenever }\rho(1+\alpha_0\gamma)>1.
\label{eq:appdd_H_range}
\end{equation}
This formula gives the rigorous reasonable-range condition for the numerical depth parameter.

For the liquid-block count parameter $q=4$, the exact discrete parameter-count increment at fixed $d$, $o$, $w$, and $H$ is
\begin{equation}
 P_{\mathrm{LNN}}(d,o,w,H,q+1)-P_{\mathrm{LNN}}(d,o,w,H,q)=w^2+2w.
\label{eq:appdd_dP_dq_general}
\end{equation}
At $w=64$,
\begin{equation}
 P_{\mathrm{LNN}}(q+1)-P_{\mathrm{LNN}}(q)=64^2+2\cdot64=4224.
\label{eq:appdd_dP_dq_actual}
\end{equation}
Hence one extra liquid block changes the model size by $4224$ parameters. Since each liquid block contributes a factor bounded by $1+\alpha_0\gamma$ in Eq.~\eqref{eq:appdd_DFbound}, the exact transport bound for $q$ matched liquid blocks is
\begin{equation}
 \operatorname{Lip}_{q}\le \rho^H(1+\alpha_0\gamma)^q.
\label{eq:appdd_Lipq}
\end{equation}
If one prescribes the same amplification budget $\Lambda_{\star}$, then a strict admissible liquid-block range is
\begin{equation}
 q\le \frac{\log\Lambda_{\star}-H\log\rho}{\log(1+\alpha_0\gamma)}
 \qquad\text{whenever }1+\alpha_0\gamma>1.
\label{eq:appdd_q_range}
\end{equation}
Thus the numerical block count $q$ admits its own exact reasonable-range condition.

For the gate initialization parameter $\alpha_0=0.5$, strict sensitivity follows from differentiating the liquid block. Since
\begin{equation}
 \alpha_{\ell}^{(0)}=\alpha_0\mathbf 1,
\label{eq:appdd_alpha_repeat}
\end{equation}
we have
\begin{equation}
 \left.\frac{\partial \mathcal B_{\ell}(z)}{\partial\alpha_0}\right|_{\alpha_{\ell}=\alpha_0\mathbf 1}=\tanh(A_{\ell}z+c_{\ell}),
\label{eq:appdd_dB_dalpha0}
\end{equation}
which implies
\begin{equation}
 \left\|\left.\frac{\partial \mathcal B_{\ell}(z)}{\partial\alpha_0}\right|_{\alpha_{\ell}=\alpha_0\mathbf 1}\right\|_2\le \sqrt w.
\label{eq:appdd_dB_dalpha0_bound}
\end{equation}
Now
\begin{equation}
 u_{\theta}=w_{\mathrm{out}}^{\top}(\mathcal F_4\circ\mathcal F_3\circ\mathcal F_2\circ\mathcal F_1)(\xi)+b_{\mathrm{out}},
\label{eq:appdd_ucomp}
\end{equation}
so repeated application of the chain rule gives
\begin{align}
 \frac{\partial u_{\theta}}{\partial\alpha_0}
 &=w_{\mathrm{out}}^{\top}
 \sum_{j=1}^{4}
 \left(\prod_{\ell=j+1}^{4}D\mathcal F_{\ell}\right)
 \left.\frac{\partial \mathcal B_j}{\partial\alpha_0}\right|_{\alpha_j=\alpha_0\mathbf 1}.
\label{eq:appdd_du_dalpha0}
\end{align}
Using Eqs.~\eqref{eq:appdd_DFbound} and \eqref{eq:appdd_dB_dalpha0_bound}, we obtain the exact sensitivity bound
\begin{equation}
 \left\|\frac{\partial u_{\theta}}{\partial\alpha_0}\right\|_2
 \le
 \|w_{\mathrm{out}}\|_2
 \sum_{j=1}^{4}
 \sqrt w
 \prod_{\ell=j+1}^{4}\rho(1+\alpha_0\gamma).
\label{eq:appdd_du_dalpha0_bound}
\end{equation}
A strict reasonable range for $\alpha_0$ follows from the per-block amplification requirement
\begin{equation}
 1+\alpha_0\gamma\le \kappa,
\label{eq:appdd_alpha_constraint1}
\end{equation}
where $\kappa>1$ is the admissible per-block amplification ceiling. Eq.~\eqref{eq:appdd_alpha_constraint1} is equivalent to
\begin{equation}
 0\le \alpha_0\le \frac{\kappa-1}{\gamma}.
\label{eq:appdd_alpha_range}
\end{equation}
This formula provides the rigorous acceptable interval for the numerical gate-initialization parameter.

The evaluation-grid parameter in the uploaded scripts is
\begin{equation}
 h=200,
 \qquad
 M_h=h^2=40000.
\label{eq:appdd_h}
\end{equation}
On the rectangular domain $[0,2]\times[0,1]$, the corresponding mesh widths are
\begin{equation}
 \Delta x=\frac{2}{h-1},
 \qquad
 \Delta t=\frac{1}{h-1}.
\label{eq:appdd_dxdt}
\end{equation}
The discrete evaluation errors are
\begin{equation}
 e_m(\theta)=u_{\theta}(x_m,t_m)-u^{\ast}(x_m,t_m),
 \qquad m=1,\dots,M_h,
\label{eq:appdd_em}
\end{equation}
with reported metrics
\begin{equation}
 \mathrm{MSE}_h(\theta)=\frac{1}{M_h}\sum_{m=1}^{M_h}e_m(\theta)^2,
 \qquad
 \mathrm{RMSE}_h(\theta)=\mathrm{MSE}_h(\theta)^{1/2},
\label{eq:appdd_metrics1}
\end{equation}
\begin{equation}
 \mathrm{MAE}_h(\theta)=\frac{1}{M_h}\sum_{m=1}^{M_h}|e_m(\theta)|,
 \qquad
 \mathrm{MaxErr}_h(\theta)=\max_{1\le m\le M_h}|e_m(\theta)|,
\label{eq:appdd_metrics2}
\end{equation}
\begin{equation}
 \mathrm{rel}L_{2,h}(\theta)=\left(\frac{\sum_{m=1}^{M_h}e_m(\theta)^2}{\sum_{m=1}^{M_h}(u^{\ast}(x_m,t_m))^2}\right)^{1/2}.
\label{eq:appdd_metrics3}
\end{equation}
Define the continuum squared-error functional
\begin{equation}
 \mathcal E_{\mathrm{cont}}(\theta)=\frac{1}{2}\int_{0}^{2}\int_{0}^{1}|u_{\theta}(x,t)-u^{\ast}(x,t)|^2\,dt\,dx.
\label{eq:appdd_Econt}
\end{equation}
Then the uniform-grid quadrature obeys
\begin{equation}
 \mathrm{MSE}_h(\theta)=\mathcal E_{\mathrm{cont}}(\theta)+\mathcal O(\Delta x+\Delta t)=\mathcal E_{\mathrm{cont}}(\theta)+\mathcal O(h^{-1}).
\label{eq:appdd_eval_error}
\end{equation}
Hence the strict reasonable range for $h$ is determined by the requested quadrature bias. If one requires
\begin{equation}
 |\mathrm{MSE}_h(\theta)-\mathcal E_{\mathrm{cont}}(\theta)|\le \varepsilon_h,
\label{eq:appdd_h_constraint}
\end{equation}
then a sufficient grid condition is
\begin{equation}
 h\ge 1+\frac{C_h}{\varepsilon_h}
\label{eq:appdd_h_range}
\end{equation}
for a problem-dependent constant $C_h$.

The booleans $\chi_{\mathrm{AMS}}=0$ and the weight-decay value $\omega_{\mathrm d}=0$ can also be treated strictly. Since the actual optimizer call sets zero weight decay, the update contains no explicit regularization term $\omega_{\mathrm d}\theta_k$. Therefore
\begin{equation}
 \left.\frac{\partial\theta_{k+1}}{\partial\omega_{\mathrm d}}\right|_{\omega_{\mathrm d}=0}=0
\label{eq:appdd_weightdecay_sens}
\end{equation}
for the actual uploaded scripts. Likewise, because the AMSGrad branch is not activated, the current recurrence does not depend on the branch variable locally, so
\begin{equation}
 \left.\frac{\partial\theta_{k+1}}{\partial\chi_{\mathrm{AMS}}}\right|_{\chi_{\mathrm{AMS}}=0}=0.
\label{eq:appdd_amsgrad_sens}
\end{equation}
These two formulas complete the one-by-one sensitivity list for all explicit numerical optimizer parameters appearing in the uploaded scripts.

The unique appendix comparison figure for the six learning-rate runs is shown in Fig.~\ref{fig:appdd_lr}. It reports the six total-loss trajectories and the corresponding reconstructed fields and absolute-error distributions under the learning-rate family in Eq.~\eqref{eq:appdd_eta_set}.

\begin{figure*}[t]
\centering
\includegraphics[width=\textwidth]{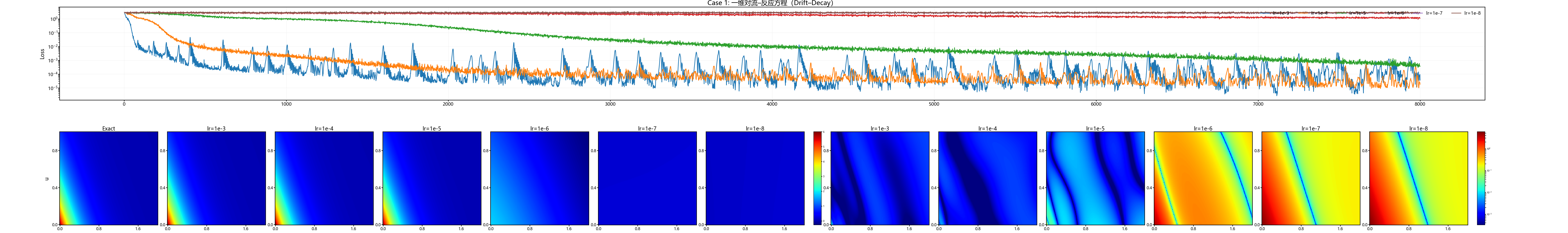}
\caption{{Learning-rate comparison for the LNN--PINN implementation of the 1D advection--reaction benchmark. The upper panel reports the total training-loss histories for the six learning rates in Eq.~\eqref{eq:appdd_eta_set}. The lower panel reports the exact field in Eq.~\eqref{eq:appdd_exact}, the six corresponding LNN--PINN reconstructions, and the associated absolute-error distributions. All runs keep the benchmark in Eqs.~\eqref{eq:appdd_pde_impl}--\eqref{eq:appdd_bc}, the architecture in Eqs.~\eqref{eq:appdd_h1}--\eqref{eq:appdd_output}, the collocation counts in Eq.~\eqref{eq:appdd_counts}, the optimizer parameters in Eq.~\eqref{eq:appdd_adamparams}, the training horizon in Eq.~\eqref{eq:appdd_ntrain}, and the evaluation grid in Eq.~\eqref{eq:appdd_h} fixed.}}
\label{fig:appdd_lr}
\end{figure*}

The terminal loss after 8000 iterations is
\begin{equation}
 \mathcal J_{\mathrm{DD}}^{(8000)}=\mathcal J_{\mathrm{DD}}(\theta_{8000}).
\label{eq:appdd_J8000}
\end{equation}
The numerical results extracted from the uploaded scripts and output files are summarized in Table~\ref{tab:appdd_lr}. The best global reconstruction over the six uploaded runs occurs at $\eta=10^{-3}$, while the smallest terminal training loss occurs at $\eta=10^{-4}$. This is mathematically consistent because for any two trained parameter vectors $\theta_a$ and $\theta_b$,
\begin{align}
 \mathrm{MSE}_h(\theta_a)-\mathrm{MSE}_h(\theta_b)
 &=\bigl(\mathrm{MSE}_h(\theta_a)-\mathcal J_{\mathrm{DD}}(\theta_a)\bigr)
 +\bigl(\mathcal J_{\mathrm{DD}}(\theta_a)-\mathcal J_{\mathrm{DD}}(\theta_b)\bigr)
 \nonumber\\
 &\quad+\bigl(\mathcal J_{\mathrm{DD}}(\theta_b)-\mathrm{MSE}_h(\theta_b)\bigr).
\label{eq:appdd_train_eval_gap}
\end{align}
The middle term is the ordering induced by the training objective; the two outer terms are the mismatch between stochastic collocation loss and uniform-grid reporting error.

\begin{table*}[t]
\centering
\scriptsize
\setlength{\tabcolsep}{4.5pt}
\renewcommand{\arraystretch}{1.10}
\caption{{Six-group learning-rate comparison for the uploaded LNN--PINN drift--decay scripts. All runs use the same architecture, collocation counts, optimizer defaults, seed, training horizon, and evaluation grid.}}
\label{tab:appdd_lr}
\resizebox{\textwidth}{!}{
\begin{tabular}{cccccccc}
\toprule
Learning rate & $\mathcal J_{\mathrm{DD}}^{(8000)}$ & RMSE & MSE & MAE & rel$L_2$ & MaxErr & Training time (s) \\
\midrule
$10^{-3}$ & $3.59356891\times 10^{-5}$ & $\mathbf{1.15614599\times 10^{-3}}$ & $\mathbf{1.33667356\times 10^{-6}}$ & $\mathbf{9.98238097\times 10^{-4}}$ & $\mathbf{1.32709096\times 10^{-3}}$ & $1.97520300\times 10^{-2}$ & $\mathbf{332.455518}$ \\
$10^{-4}$ & $\mathbf{1.81689593\times 10^{-5}}$ & $1.36438228\times 10^{-3}$ & $1.86154101\times 10^{-6}$ & $1.18533718\times 10^{-3}$ & $1.56611761\times 10^{-3}$ & $\mathbf{1.48377000\times 10^{-2}}$ & $710.268917$ \\
$10^{-5}$ & $3.65526328\times 10^{-4}$ & $4.50358354\times 10^{-3}$ & $2.02822650\times 10^{-5}$ & $3.23861500\times 10^{-3}$ & $5.16947379\times 10^{-3}$ & $1.13740000\times 10^{-1}$ & $709.547432$ \\
$10^{-6}$ & $1.11032331\times 10^{0}$ & $5.00800939\times 10^{-1}$ & $2.50801581\times 10^{-1}$ & $3.56996000\times 10^{-1}$ & $5.74848195\times 10^{-1}$ & $4.17252400\times 10^{0}$ & $710.328561$ \\
$10^{-7}$ & $2.57550192\times 10^{0}$ & $7.89371265\times 10^{-1}$ & $6.23107139\times 10^{-1}$ & $4.13184000\times 10^{-1}$ & $9.06085836\times 10^{-1}$ & $5.79175200\times 10^{0}$ & $713.114972$ \\
$10^{-8}$ & $2.70013857\times 10^{0}$ & $8.17163657\times 10^{-1}$ & $6.67756385\times 10^{-1}$ & $4.21431000\times 10^{-1}$ & $9.37987520\times 10^{-1}$ & $5.88180000\times 10^{0}$ & $715.001124$ \\
\bottomrule
\end{tabular}}
\end{table*}

Finally, Table~\ref{tab:appdd_params} lists the complete numerical parameter set appearing in the uploaded LNN--PINN implementation, with no omitted numeric control parameter.

\begin{table*}[t]
\centering
\scriptsize
\setlength{\tabcolsep}{4.5pt}
\renewcommand{\arraystretch}{1.10}
\caption{Complete numerical parameter list for the uploaded LNN--PINN drift--decay scripts.}
\label{tab:appdd_params}
\resizebox{\textwidth}{!}{
\begin{tabular}{llll}
\toprule
Category & Symbol / item & Value & Role \\
\midrule
Learning-rate family & $\eta$ & $\{10^{-3},10^{-4},10^{-5},10^{-6},10^{-7},10^{-8}\}$ & optimization step size \\
Training horizon & $N_{\mathrm{train}}$ & $8000$ & number of Adam iterations \\
Interior collocation count & $N_{\Omega}$ & $2000$ & PDE residual sampling size \\
Initial-line collocation count & $N_{\mathrm{IC}}$ & $1000$ & initial residual sampling size \\
Boundary collocation count & $N_{\mathrm{BD}}$ & $1000$ & inflow residual sampling size \\
Seed & $s$ & $888888$ & stochastic realization selector \\
Input dimension & $d$ & $2$ & input-space dimension \\
Output dimension & $o$ & $1$ & scalar output dimension \\
Hidden width & $w$ & $64$ & channel count per hidden stage \\
Hidden-stage number & $H$ & $4$ & number of hidden affine stages \\
Liquid-block number & $q$ & $4$ & number of residual liquid blocks \\
Gate initialization & $\alpha_0$ & $0.5$ & common gate amplitude initialization \\
Adam first-moment factor & $\beta_1$ & $0.9$ & first-moment smoothing \\
Adam second-moment factor & $\beta_2$ & $0.999$ & second-moment smoothing \\
Adam epsilon & $\varepsilon_{\mathrm A}$ & $10^{-8}$ & denominator stabilization \\
Weight decay & $\omega_{\mathrm d}$ & $0$ & explicit regularization coefficient \\
AMSGrad flag & $\chi_{\mathrm{AMS}}$ & $0$ & optimizer branch selector \\
Evaluation grid size & $h$ & $200$ & report-grid resolution \\
Evaluation-point count & $M_h$ & $40000$ & report-grid size \\
Trainable-parameter count & $P_{\mathrm{LNN}}$ & $29633$ & derived model size \\
\bottomrule
\end{tabular}}
\end{table*}

Eqs.~\eqref{eq:appdd_eta_range}, \eqref{eq:appdd_epochrange}, \eqref{eq:appdd_Nranges}, \eqref{eq:appdd_beta1_range_basic}, \eqref{eq:appdd_beta2_range_basic}, \eqref{eq:appdd_eps_range_basic}, \eqref{eq:appdd_H_range}, \eqref{eq:appdd_q_range}, \eqref{eq:appdd_alpha_range}, and \eqref{eq:appdd_h_range} together provide the strict framework-based admissible-range conditions for every numerical control parameter appearing in the uploaded LNN--PINN drift--decay scripts. Within the six uploaded learning-rate runs, the configuration $\eta=10^{-3}$ yields the best overall field reconstruction, so the main-text LNN--PINN choice for the drift--decay benchmark is justified by the global error metrics rather than by terminal training loss alone.
\section{Learning-rate comparison, complete implementation parameters, and rigorous framework-based parameter-sensitivity analysis for the LNN--PINN mixed-boundary Laplace benchmark}\label{b}

This appendix supplements the mixed-boundary Laplace benchmark in Section~3.2, but it restricts attention to the uploaded LNN--PINN learning-rate archive only. The current main text assigns valid labels to the generic composite objective in Eq.~\eqref{LOSS Total}, to the framework figure in Fig.~\ref{LNN-PINN}, and to the main-text Laplace comparison figure in Fig.~\ref{F3}, whereas the case-specific Laplace equations and residual expressions in Section~3.2 carry no local equation labels. Therefore, in order to avoid any false reference, we rewrite the case-specific problem, collocation sets, residuals, and loss components here in full. The uploaded LNN-only learning-rate sweep and the main-text Laplace benchmark both use a training horizon of $N_{\mathrm{train}}=8000$ Adam iterations. The present appendix therefore analyzes the same training horizon as the main-text benchmark while restricting attention to the six-rate LNN-only archive.

The physical problem remains the scalar Laplace equation on the unit square,
\begin{equation}
\phi_{xx}+\phi_{yy}=0,
\qquad
(x,y)\in[0,1]\times[0,1],
\label{eq:lapapp_pde}
\end{equation}
subject to the mixed Dirichlet--Neumann boundary conditions
\begin{equation}
\phi(x,0)=0,
\qquad
\phi(x,1)=1,
\label{eq:lapapp_dir}
\end{equation}
\begin{equation}
\phi_x(0,y)=0,
\qquad
\phi_x(1,y)=0.
\label{eq:lapapp_neu}
\end{equation}
The analytical reference field used only for post-training evaluation is
\begin{equation}
\phi^{\ast}(x,y)=y.
\label{eq:lapapp_exact}
\end{equation}
The uploaded implementation approximates $\phi$ by $\phi_{\theta}(x,y)$ and keeps the case-level objective within the unified framework of Eq.~\eqref{LOSS Total}, but with five case-specific channels. The interior collocation set and the four boundary collocation sets are
\begin{equation}
\mathcal S_{\Omega}=\{(x_i,y_i)\}_{i=1}^{N_{\Omega}},
\qquad
\mathcal S_{\mathrm{bot}}=\{(x_j,0)\}_{j=1}^{N_{\mathrm{bot}}},
\qquad
\mathcal S_{\mathrm{top}}=\{(x_j,1)\}_{j=1}^{N_{\mathrm{top}}},
\label{eq:lapapp_sets1}
\end{equation}
\begin{equation}
\mathcal S_{\mathrm{L}}=\{(0,y_k)\}_{k=1}^{N_{\mathrm{L}}},
\qquad
\mathcal S_{\mathrm{R}}=\{(1,y_k)\}_{k=1}^{N_{\mathrm{R}}}.
\label{eq:lapapp_sets2}
\end{equation}
The uploaded scripts fix the sample counts at
\begin{equation}
N_{\Omega}=1000,
\qquad
N_{\mathrm{bot}}=1000,
\qquad
N_{\mathrm{top}}=1000,
\qquad
N_{\mathrm{L}}=1000,
\qquad
N_{\mathrm{R}}=1000.
\label{eq:lapapp_counts}
\end{equation}
At every optimization step, the script redraws each set uniformly from its own geometric support, so the empirical loss is stochastic even though the underlying boundary-value problem in Eqs.~\eqref{eq:lapapp_pde}--\eqref{eq:lapapp_neu} is deterministic.

The case-specific residual vectors are
\begin{equation}
\mathbf r_{\mathrm{PDE}}(\theta)
=
\bigl[\phi_{\theta,xx}(x,y)+\phi_{\theta,yy}(x,y)\bigr]_{(x,y)\in\mathcal S_{\Omega}},
\label{eq:lapapp_rpde}
\end{equation}
\begin{equation}
\mathbf r_{\mathrm{bot}}(\theta)
=
\bigl[\phi_{\theta}(x,0)\bigr]_{(x,0)\in\mathcal S_{\mathrm{bot}}},
\qquad
\mathbf r_{\mathrm{top}}(\theta)
=
\bigl[\phi_{\theta}(x,1)-1\bigr]_{(x,1)\in\mathcal S_{\mathrm{top}}},
\label{eq:lapapp_rdir}
\end{equation}
\begin{equation}
\mathbf r_{\mathrm{L}}(\theta)
=
\bigl[\phi_{\theta,x}(0,y)\bigr]_{(0,y)\in\mathcal S_{\mathrm{L}}},
\qquad
\mathbf r_{\mathrm{R}}(\theta)
=
\bigl[\phi_{\theta,x}(1,y)\bigr]_{(1,y)\in\mathcal S_{\mathrm{R}}}.
\label{eq:lapapp_rneu}
\end{equation}
The corresponding mean-squared channels are
\begin{equation}
L_{\mathrm{PDE}}(\theta)=\frac{1}{N_{\Omega}}\|\mathbf r_{\mathrm{PDE}}(\theta)\|_2^2,
\qquad
L_{\mathrm{bot}}(\theta)=\frac{1}{N_{\mathrm{bot}}}\|\mathbf r_{\mathrm{bot}}(\theta)\|_2^2,
\qquad
L_{\mathrm{top}}(\theta)=\frac{1}{N_{\mathrm{top}}}\|\mathbf r_{\mathrm{top}}(\theta)\|_2^2,
\label{eq:lapapp_losses1}
\end{equation}
\begin{equation}
L_{\mathrm{L}}(\theta)=\frac{1}{N_{\mathrm{L}}}\|\mathbf r_{\mathrm{L}}(\theta)\|_2^2,
\qquad
L_{\mathrm{R}}(\theta)=\frac{1}{N_{\mathrm{R}}}\|\mathbf r_{\mathrm{R}}(\theta)\|_2^2.
\label{eq:lapapp_losses2}
\end{equation}
The uploaded script minimizes the direct sum
\begin{equation}
\mathcal J_{\mathrm{Lap}}(\theta)
=
L_{\mathrm{PDE}}(\theta)+L_{\mathrm{bot}}(\theta)+L_{\mathrm{top}}(\theta)+L_{\mathrm{L}}(\theta)+L_{\mathrm{R}}(\theta),
\label{eq:lapapp_J}
\end{equation}
which is exactly the specialization of Eq.~\eqref{LOSS Total} obtained by setting the present Dirichlet and Neumann weights to unity.

The actual uploaded LNN--PINN forward map now needs to be stated at the implementation level, because the sensitivity analysis below must act on the network that the scripts truly optimize rather than on a schematic surrogate. Let
\begin{equation}
\xi=
\begin{bmatrix}
 x\\ y
\end{bmatrix}
\in\mathbb R^2.
\label{eq:lapapp_xi}
\end{equation}
The first hidden stage is
\begin{equation}
z_1=\tanh(W_0\xi+b_0),
\qquad
h_1=z_1+\operatorname{Diag}(\alpha_1)\tanh(A_1 z_1+c_1).
\label{eq:lapapp_h1}
\end{equation}
The next three hidden stages repeat the same width-preserving liquid residual form,
\begin{equation}
z_2=\tanh(W_1 h_1+b_1),
\qquad
h_2=z_2+\operatorname{Diag}(\alpha_2)\tanh(A_2 z_2+c_2),
\label{eq:lapapp_h2}
\end{equation}
\begin{equation}
z_3=\tanh(W_2 h_2+b_2),
\qquad
h_3=z_3+\operatorname{Diag}(\alpha_3)\tanh(A_3 z_3+c_3),
\label{eq:lapapp_h3}
\end{equation}
\begin{equation}
z_4=\tanh(W_3 h_3+b_3),
\qquad
h_4=z_4+\operatorname{Diag}(\alpha_4)\tanh(A_4 z_4+c_4),
\label{eq:lapapp_h4}
\end{equation}
and the scalar output is
\begin{equation}
\phi_{\theta}(x,y)=w_{\mathrm{out}}^{\top}h_4+b_{\mathrm{out}}.
\label{eq:lapapp_out}
\end{equation}
Thus the numeric architecture parameters in the uploaded scripts are
\begin{equation}
d=2,
\qquad
o=1,
\qquad
w=64,
\qquad
H=4,
\qquad
q=4,
\qquad
\alpha_0=0.5,
\label{eq:lapapp_archvals}
\end{equation}
where $d$ is the input dimension, $o$ is the output dimension, $w$ is the hidden width, $H$ is the hidden-stage count, $q$ is the liquid-block count, and each gate vector is initialized by
\begin{equation}
\alpha_{\ell}^{(0)}=\alpha_0\mathbf 1,
\qquad
\ell=1,2,3,4.
\label{eq:lapapp_alpha_init}
\end{equation}
The trainable-parameter count is then a derived algebraic quantity rather than an independent hyperparameter. For the general family with dimensions $(d,o,w,H,q)$,
\begin{equation}
P_{\mathrm{LNN}}(d,o,w,H,q)
=
dw+w+(H-1)(w^2+w)+wo+o+q(w^2+2w).
\label{eq:lapapp_Pgeneral}
\end{equation}
Substituting Eq.~\eqref{eq:lapapp_archvals} gives
\begin{align}
P_{\mathrm{LNN}}
&=
2\cdot64+64+3(64^2+64)+64\cdot1+1+4(64^2+2\cdot64)
\nonumber\\
&=192+12480+65+16896
\nonumber\\
&=29633.
\label{eq:lapapp_Pactual}
\end{align}
This matches the count written by the archive-level compute-cost files.

The optimizer used by the uploaded implementation is Adam with the script-level arguments left at their library defaults except for the learning rate. Consequently the update uses
\begin{equation}
\beta_1=0.9,
\qquad
\beta_2=0.999,
\qquad
\varepsilon_{\mathrm A}=10^{-8},
\qquad
\lambda_{\mathrm{wd}}=0,
\qquad
\mathrm{AMSGrad}=\mathrm{False},
\label{eq:lapapp_adamvals}
\end{equation}
and the total training horizon in the uploaded sweep is
\begin{equation}
N_{\mathrm{train}}=8000.
\label{eq:lapapp_ntrain}
\end{equation}
The archive organizes six nominal learning-rate entries,
\begin{equation}
\eta\in\{10^{-3},10^{-4},10^{-5},10^{-6},10^{-7},10^{-8}\}.
\label{eq:lapapp_nominaletas}
\end{equation}
However, the uploaded nominal $10^{-5}$ entry contains the same script-level assignment $\eta=10^{-3}$ as the nominal $10^{-3}$ entry and reproduces the same prediction statistics. We therefore keep that row in the appendix table as an archive-faithful nominal entry, but we do not treat it as an independent effective learning rate in the mathematical interpretation.

To expose the exact dependence on $\eta$, $\beta_1$, $\beta_2$, $\varepsilon_{\mathrm A}$, and $\lambda_{\mathrm{wd}}$, we write the optimizer state in closed form. Let the stochastic gradient at step $k$ be
\begin{equation}
g_k=\nabla_{\theta}\mathcal J_{\mathrm{Lap}}(\theta_k;\Xi_k),
\label{eq:lapapp_gk}
\end{equation}
where $\Xi_k$ denotes the five freshly redrawn collocation sets at that step. With $m_0=0$ and $v_0=0$, the first and second moments satisfy
\begin{equation}
m_k=(1-\beta_1)\sum_{j=1}^{k}\beta_1^{k-j}g_j,
\qquad
v_k=(1-\beta_2)\sum_{j=1}^{k}\beta_2^{k-j}(g_j\odot g_j).
\label{eq:lapapp_mvkclosed}
\end{equation}
The bias-corrected moments are
\begin{equation}
\widehat m_k=\frac{m_k}{1-\beta_1^k},
\qquad
\widehat v_k=\frac{v_k}{1-\beta_2^k},
\label{eq:lapapp_hatmv}
\end{equation}
and the Adam direction is the componentwise vector
\begin{equation}
(d_k)_r=\frac{(\widehat m_k)_r}{\sqrt{(\widehat v_k)_r}+\varepsilon_{\mathrm A}},
\qquad r=1,\dots,P_{\mathrm{LNN}}.
\label{eq:lapapp_dk}
\end{equation}
With weight decay included explicitly, the one-step update may be written as
\begin{equation}
\theta_{k+1}=\theta_k-\eta\bigl(d_k+\lambda_{\mathrm{wd}}\theta_k\bigr).
\label{eq:lapapp_update}
\end{equation}
Since the archive fixes $\lambda_{\mathrm{wd}}=0$, Eq.~\eqref{eq:lapapp_update} reduces exactly to the script-level rule. The one-step sensitivity with respect to the learning rate now follows without any hidden step:
\begin{equation}
\theta_{k+1}(\eta+\delta\eta)-\theta_{k+1}(\eta)
=-\delta\eta\bigl(d_k+\lambda_{\mathrm{wd}}\theta_k\bigr),
\label{eq:lapapp_eta_increment}
\end{equation}
so at the archive value $\lambda_{\mathrm{wd}}=0$ one has
\begin{equation}
\left.\frac{\partial\theta_{k+1}}{\partial\eta}\right|_{\lambda_{\mathrm{wd}}=0}=-d_k,
\qquad
\|\delta\theta_{k+1}\|_2\le |\delta\eta|\,\|d_k\|_2.
\label{eq:lapapp_eta_derivative}
\end{equation}
To convert this iterate sensitivity into a descent condition for the actual objective, let
\begin{equation}
H_k=\nabla_{\theta}^2\mathcal J_{\mathrm{Lap}}(\theta_k;\Xi_k).
\label{eq:lapapp_Hk}
\end{equation}
Taylor expansion of $\mathcal J_{\mathrm{Lap}}$ along the actual update direction gives
\begin{equation}
\mathcal J_{\mathrm{Lap}}(\theta_{k+1};\Xi_k)
=\mathcal J_{\mathrm{Lap}}(\theta_k;\Xi_k)
-\eta\,g_k^{\top}d_k
+\frac{\eta^2}{2}
 d_k^{\top}H_k(\theta_k-\tau_k\eta d_k)d_k,
\label{eq:lapapp_taylor_eta}
\end{equation}
for some $\tau_k\in(0,1)$. If
\begin{equation}
g_k^{\top}d_k\ge c_k\|g_k\|_2\|d_k\|_2,
\qquad
\|H_k(\theta_k-\tau_k\eta d_k)\|_2\le M_k,
\label{eq:lapapp_ckMk}
\end{equation}
then a sufficient local descent condition is
\begin{equation}
0<\eta<\frac{2c_k\|g_k\|_2}{M_k\|d_k\|_2}.
\label{eq:lapapp_eta_range}
\end{equation}
Eq.~\eqref{eq:lapapp_eta_range} gives the rigorous meaning of an admissible learning-rate range for this actual archive-level optimizer.

The parameters $\beta_1$ and $\beta_2$ now admit exact partial derivatives. Differentiating the first formula in Eq.~\eqref{eq:lapapp_mvkclosed} gives
\begin{align}
\frac{\partial m_k}{\partial\beta_1}
&=-\sum_{j=1}^{k}\beta_1^{k-j}g_j
+(1-\beta_1)\sum_{j=1}^{k}(k-j)\beta_1^{k-j-1}g_j,
\label{eq:lapapp_dm_db1}
\end{align}
and therefore
\begin{equation}
\frac{\partial \widehat m_k}{\partial\beta_1}
=
\frac{(1-\beta_1^k)\frac{\partial m_k}{\partial\beta_1}+k\beta_1^{k-1}m_k}{(1-\beta_1^k)^2}.
\label{eq:lapapp_dhatm_db1}
\end{equation}
Since $\widehat v_k$ does not depend on $\beta_1$, Eq.~\eqref{eq:lapapp_dk} gives the componentwise direction sensitivity
\begin{equation}
\frac{\partial (d_k)_r}{\partial\beta_1}
=
\frac{1}{\sqrt{(\widehat v_k)_r}+\varepsilon_{\mathrm A}}
\frac{\partial (\widehat m_k)_r}{\partial\beta_1}.
\label{eq:lapapp_ddir_db1}
\end{equation}
In exactly the same way, differentiating the second formula in Eq.~\eqref{eq:lapapp_mvkclosed} yields
\begin{align}
\frac{\partial v_k}{\partial\beta_2}
&=-\sum_{j=1}^{k}\beta_2^{k-j}(g_j\odot g_j)
+(1-\beta_2)\sum_{j=1}^{k}(k-j)\beta_2^{k-j-1}(g_j\odot g_j),
\label{eq:lapapp_dv_db2}
\end{align}
and hence
\begin{equation}
\frac{\partial \widehat v_k}{\partial\beta_2}
=
\frac{(1-\beta_2^k)\frac{\partial v_k}{\partial\beta_2}+k\beta_2^{k-1}v_k}{(1-\beta_2^k)^2}.
\label{eq:lapapp_dhatv_db2}
\end{equation}
Now differentiate Eq.~\eqref{eq:lapapp_dk} componentwise with respect to $\beta_2$:
\begin{equation}
\frac{\partial (d_k)_r}{\partial\beta_2}
=-\frac{(\widehat m_k)_r}{2\sqrt{(\widehat v_k)_r}\bigl(\sqrt{(\widehat v_k)_r}+\varepsilon_{\mathrm A}\bigr)^2}
\frac{\partial(\widehat v_k)_r}{\partial\beta_2}.
\label{eq:lapapp_ddir_db2}
\end{equation}
Eqs.~\eqref{eq:lapapp_ddir_db1} and \eqref{eq:lapapp_ddir_db2} show that $\beta_1$ changes the direction through the exponentially weighted numerator history, whereas $\beta_2$ changes the denominator through the exponentially weighted squared-gradient history. The archive values $\beta_1=0.9$ and $\beta_2=0.999$ therefore lie inside two different sensitivity channels of the same optimizer map.

The parameter $\varepsilon_{\mathrm A}=10^{-8}$ also enters directly and exactly. Differentiating Eq.~\eqref{eq:lapapp_dk} with respect to $\varepsilon_{\mathrm A}$ gives
\begin{equation}
\frac{\partial (d_k)_r}{\partial\varepsilon_{\mathrm A}}
=-\frac{(\widehat m_k)_r}{\bigl(\sqrt{(\widehat v_k)_r}+\varepsilon_{\mathrm A}\bigr)^2}.
\label{eq:lapapp_ddir_deps}
\end{equation}
Thus increasing $\varepsilon_{\mathrm A}$ strictly shrinks the magnitude of each component of the update direction whenever $(\widehat m_k)_r$ retains fixed sign. This is the exact one-parameter sensitivity law for the numerical stabilizer.

The weight-decay parameter is fixed to zero in the uploaded archive, but its first-order effect can still be derived without ambiguity from Eq.~\eqref{eq:lapapp_update}. Freezing $d_k$ and $\theta_k$ at step $k$ and perturbing $\lambda_{\mathrm{wd}}\mapsto \lambda_{\mathrm{wd}}+\delta\lambda_{\mathrm{wd}}$ yields
\begin{equation}
\theta_{k+1}(\lambda_{\mathrm{wd}}+\delta\lambda_{\mathrm{wd}})-\theta_{k+1}(\lambda_{\mathrm{wd}})
=-\eta\,\delta\lambda_{\mathrm{wd}}\,\theta_k,
\label{eq:lapapp_dtheta_dwd}
\end{equation}
so the local derivative at the archive value is
\begin{equation}
\left.\frac{\partial\theta_{k+1}}{\partial\lambda_{\mathrm{wd}}}\right|_{\lambda_{\mathrm{wd}}=0}
=-\eta\theta_k.
\label{eq:lapapp_dtheta_dwd2}
\end{equation}
The AMSGrad switch is not a differentiable numeric parameter, but its archive value still has a precise mathematical meaning. Standard Adam uses $\widehat v_k$ in the denominator, whereas AMSGrad would replace it by the running elementwise maximum
\begin{equation}
\overline v_k=\max\{\overline v_{k-1},\widehat v_k\},
\label{eq:lapapp_amsgrad}
\end{equation}
which satisfies $\overline v_k\ge \widehat v_k$ componentwise. Consequently the AMSGrad direction would obey
\begin{equation}
\left|\frac{(\widehat m_k)_r}{\sqrt{(\overline v_k)_r}+\varepsilon_{\mathrm A}}\right|
\le
\left|\frac{(\widehat m_k)_r}{\sqrt{(\widehat v_k)_r}+\varepsilon_{\mathrm A}}\right|,
\label{eq:lapapp_amsgrad_bound}
\end{equation}
so the archive choice \texttt{AMSGrad=False} corresponds to the larger of the two admissible step magnitudes.

The training horizon $N_{\mathrm{train}}=8000$ also requires an exact discrete sensitivity statement. Iterating Eq.~\eqref{eq:lapapp_update} with $\lambda_{\mathrm{wd}}=0$ gives
\begin{equation}
\theta_K=\theta_0-\eta\sum_{k=0}^{K-1}d_k.
\label{eq:lapapp_thetaK}
\end{equation}
Hence enlarging the horizon from $K$ to $K+\Delta K$ yields
\begin{equation}
\theta_{K+\Delta K}-\theta_K
=-\eta\sum_{k=K}^{K+\Delta K-1}d_k,
\label{eq:lapapp_dtheta_dK}
\end{equation}
and therefore
\begin{equation}
\|\theta_{K+\Delta K}-\theta_K\|_2
\le
\eta\sum_{k=K}^{K+\Delta K-1}\|d_k\|_2.
\label{eq:lapapp_dtheta_dK_bound}
\end{equation}
Eq.~\eqref{eq:lapapp_dtheta_dK_bound} shows that the marginal effect of the training horizon is not linear in $K$ itself; it is controlled by the tail of the actual archive-level Adam direction sequence.

The five sample counts in Eq.~\eqref{eq:lapapp_counts} require channelwise stochastic sensitivity analysis rather than a single generic statement. Define the squared residual random variables
\begin{equation}
X_{\Omega}=\bigl(\phi_{\theta,xx}(x,y)+\phi_{\theta,yy}(x,y)\bigr)^2,
\qquad
(x,y)\sim\mathrm{Unif}([0,1]^2),
\label{eq:lapapp_Xomega}
\end{equation}
\begin{equation}
X_{\mathrm{bot}}=\phi_{\theta}(x,0)^2,
\qquad
x\sim\mathrm{Unif}([0,1]),
\qquad
X_{\mathrm{top}}=\bigl(\phi_{\theta}(x,1)-1\bigr)^2,
\qquad
x\sim\mathrm{Unif}([0,1]),
\label{eq:lapapp_Xdir}
\end{equation}
\begin{equation}
X_{\mathrm{L}}=\phi_{\theta,x}(0,y)^2,
\qquad
y\sim\mathrm{Unif}([0,1]),
\qquad
X_{\mathrm{R}}=\phi_{\theta,x}(1,y)^2,
\qquad
y\sim\mathrm{Unif}([0,1]).
\label{eq:lapapp_Xneu}
\end{equation}
Let their variances be
\begin{equation}
\sigma_{\Omega}^2(\theta)=\operatorname{Var}(X_{\Omega}),
\quad
\sigma_{\mathrm{bot}}^2(\theta)=\operatorname{Var}(X_{\mathrm{bot}}),
\quad
\sigma_{\mathrm{top}}^2(\theta)=\operatorname{Var}(X_{\mathrm{top}}),
\quad
\sigma_{\mathrm{L}}^2(\theta)=\operatorname{Var}(X_{\mathrm{L}}),
\quad
\sigma_{\mathrm{R}}^2(\theta)=\operatorname{Var}(X_{\mathrm{R}}).
\label{eq:lapapp_sigmas}
\end{equation}
Because each loss channel in Eqs.~\eqref{eq:lapapp_losses1}--\eqref{eq:lapapp_losses2} is a sample mean, one has the exact variance identities
\begin{equation}
\operatorname{Var}(L_{\mathrm{PDE}})=\frac{\sigma_{\Omega}^2(\theta)}{N_{\Omega}},
\qquad
\operatorname{Var}(L_{\mathrm{bot}})=\frac{\sigma_{\mathrm{bot}}^2(\theta)}{N_{\mathrm{bot}}},
\qquad
\operatorname{Var}(L_{\mathrm{top}})=\frac{\sigma_{\mathrm{top}}^2(\theta)}{N_{\mathrm{top}}},
\label{eq:lapapp_var1}
\end{equation}
\begin{equation}
\operatorname{Var}(L_{\mathrm{L}})=\frac{\sigma_{\mathrm{L}}^2(\theta)}{N_{\mathrm{L}}},
\qquad
\operatorname{Var}(L_{\mathrm{R}})=\frac{\sigma_{\mathrm{R}}^2(\theta)}{N_{\mathrm{R}}}.
\label{eq:lapapp_var2}
\end{equation}
Therefore the total archive-level objective variance is
\begin{equation}
\operatorname{Var}(\mathcal J_{\mathrm{Lap}}(\theta))
=
\frac{\sigma_{\Omega}^2(\theta)}{N_{\Omega}}
+
\frac{\sigma_{\mathrm{bot}}^2(\theta)}{N_{\mathrm{bot}}}
+
\frac{\sigma_{\mathrm{top}}^2(\theta)}{N_{\mathrm{top}}}
+
\frac{\sigma_{\mathrm{L}}^2(\theta)}{N_{\mathrm{L}}}
+
\frac{\sigma_{\mathrm{R}}^2(\theta)}{N_{\mathrm{R}}}.
\label{eq:lapapp_var_total}
\end{equation}
Differentiating Eq.~\eqref{eq:lapapp_var_total} with respect to each count separately yields
\begin{equation}
\frac{\partial}{\partial N_{\Omega}}\operatorname{Var}(\mathcal J_{\mathrm{Lap}}(\theta))
=-\frac{\sigma_{\Omega}^2(\theta)}{N_{\Omega}^2},
\qquad
\frac{\partial}{\partial N_{\mathrm{bot}}}\operatorname{Var}(\mathcal J_{\mathrm{Lap}}(\theta))
=-\frac{\sigma_{\mathrm{bot}}^2(\theta)}{N_{\mathrm{bot}}^2},
\label{eq:lapapp_dvar1}
\end{equation}
\begin{equation}
\frac{\partial}{\partial N_{\mathrm{top}}}\operatorname{Var}(\mathcal J_{\mathrm{Lap}}(\theta))
=-\frac{\sigma_{\mathrm{top}}^2(\theta)}{N_{\mathrm{top}}^2},
\qquad
\frac{\partial}{\partial N_{\mathrm{L}}}\operatorname{Var}(\mathcal J_{\mathrm{Lap}}(\theta))
=-\frac{\sigma_{\mathrm{L}}^2(\theta)}{N_{\mathrm{L}}^2},
\label{eq:lapapp_dvar2}
\end{equation}
\begin{equation}
\frac{\partial}{\partial N_{\mathrm{R}}}\operatorname{Var}(\mathcal J_{\mathrm{Lap}}(\theta))
=-\frac{\sigma_{\mathrm{R}}^2(\theta)}{N_{\mathrm{R}}^2}.
\label{eq:lapapp_dvar3}
\end{equation}
Thus every one of the five counts enters the optimization problem through its own exact $1/N$ variance law. If one prescribes channelwise standard-deviation budgets $(\delta_{\Omega},\delta_{\mathrm{bot}},\delta_{\mathrm{top}},\delta_{\mathrm{L}},\delta_{\mathrm{R}})$, then sufficient admissible ranges are
\begin{equation}
N_{\Omega}\ge \frac{\sigma_{\Omega}^2(\theta)}{\delta_{\Omega}^2},
\qquad
N_{\mathrm{bot}}\ge \frac{\sigma_{\mathrm{bot}}^2(\theta)}{\delta_{\mathrm{bot}}^2},
\qquad
N_{\mathrm{top}}\ge \frac{\sigma_{\mathrm{top}}^2(\theta)}{\delta_{\mathrm{top}}^2},
\label{eq:lapapp_Nrange1}
\end{equation}
\begin{equation}
N_{\mathrm{L}}\ge \frac{\sigma_{\mathrm{L}}^2(\theta)}{\delta_{\mathrm{L}}^2},
\qquad
N_{\mathrm{R}}\ge \frac{\sigma_{\mathrm{R}}^2(\theta)}{\delta_{\mathrm{R}}^2}.
\label{eq:lapapp_Nrange2}
\end{equation}
These five inequalities are the rigorous channelwise meaning of a reasonable collocation regime for this precise archive-level Laplace implementation.

The seed $s=888888$ is also numeric, but it is discrete rather than differentiable. Its mathematical role is to select one realization of the stepwise random tuple $\Xi_k$. If two seeds $s_1$ and $s_2$ generate two sample streams $\Xi_k^{(s_1)}$ and $\Xi_k^{(s_2)}$, then the exact seed sensitivity of the objective at step $k$ is measured by
\begin{equation}
\Delta_{s_1,s_2}\mathcal J_{\mathrm{Lap}}(\theta_k)
=
\mathcal J_{\mathrm{Lap}}(\theta_k;\Xi_k^{(s_1)})
-
\mathcal J_{\mathrm{Lap}}(\theta_k;\Xi_k^{(s_2)}).
\label{eq:lapapp_seedgap}
\end{equation}
The archive fixes the same seed in all six entries, so the learning-rate comparison does not mix learning-rate variation with seed variation.

We now analyze the framework-specific parameters $d$, $o$, $w$, $H$, $q$, and $\alpha_0$ one by one. The input and output dimensions enter the architecture algebraically through Eq.~\eqref{eq:lapapp_Pgeneral}. Holding $(o,w,H,q)$ fixed and increasing the input dimension from $d$ to $d+1$ gives
\begin{equation}
P_{\mathrm{LNN}}(d+1,o,w,H,q)-P_{\mathrm{LNN}}(d,o,w,H,q)=w.
\label{eq:lapapp_dd}
\end{equation}
At the archive value $w=64$, the exact increment is $64$. Holding $(d,w,H,q)$ fixed and increasing the output dimension from $o$ to $o+1$ gives
\begin{equation}
P_{\mathrm{LNN}}(d,o+1,w,H,q)-P_{\mathrm{LNN}}(d,o,w,H,q)=w+1,
\label{eq:lapapp_do}
\end{equation}
which equals $65$ at the archive width. Thus the numeric values $d=2$ and $o=1$ already determine two exact discrete capacity sensitivities.

The hidden width $w=64$ changes both capacity and transport amplification. From Eq.~\eqref{eq:lapapp_Pgeneral}, the exact discrete width increment is
\begin{align}
&P_{\mathrm{LNN}}(d,o,w+1,H,q)-P_{\mathrm{LNN}}(d,o,w,H,q)
\nonumber\\
&=(H-1+q)(2w+1)+(d+H+o+2q).
\label{eq:lapapp_dw}
\end{align}
Substituting the archive values $(d,o,H,q)=(2,1,4,4)$ and $w=64$ yields
\begin{equation}
P_{\mathrm{LNN}}(65)-P_{\mathrm{LNN}}(64)=7\cdot129+15=918.
\label{eq:lapapp_dw_actual}
\end{equation}
Thus one additional hidden channel increases the trainable-parameter count by $918$ in this exact architecture family.

The hidden-stage count $H=4$ enters through the backbone squares. Holding $(d,o,w,q)$ fixed and increasing $H$ by one gives
\begin{equation}
P_{\mathrm{LNN}}(d,o,w,H+1,q)-P_{\mathrm{LNN}}(d,o,w,H,q)=w^2+w.
\label{eq:lapapp_dH}
\end{equation}
At $w=64$, this increment becomes
\begin{equation}
64^2+64=4160.
\label{eq:lapapp_dH_actual}
\end{equation}
The liquid-block count $q=4$ acts independently of $H$. Holding $(d,o,w,H)$ fixed and increasing $q$ by one gives
\begin{equation}
P_{\mathrm{LNN}}(d,o,w,H,q+1)-P_{\mathrm{LNN}}(d,o,w,H,q)=w^2+2w,
\label{eq:lapapp_dq}
\end{equation}
which equals
\begin{equation}
64^2+2\cdot64=4224
\label{eq:lapapp_dq_actual}
\end{equation}
at the archive width. If one enlarges the model by one hidden stage and one liquid block simultaneously, the exact increment is the sum of Eqs.~\eqref{eq:lapapp_dH} and \eqref{eq:lapapp_dq}, namely
\begin{equation}
2w^2+3w,
\label{eq:lapapp_dHq}
\end{equation}
which becomes
\begin{equation}
2\cdot64^2+3\cdot64=8384.
\label{eq:lapapp_dHq_actual}
\end{equation}
Therefore $w$, $H$, and $q$ are numerically distinct and mathematically nonredundant controls.

The gate-initialization amplitude $\alpha_0=0.5$ enters through the liquid-residual operator itself. For one liquid block
\begin{equation}
\mathcal B_{\ell}(z)=z+\operatorname{Diag}(\alpha_{\ell})\tanh(A_{\ell}z+c_{\ell}),
\label{eq:lapapp_Bell}
\end{equation}
its Jacobian with respect to the block input is
\begin{equation}
D\mathcal B_{\ell}(z)
=
I+\operatorname{Diag}(\alpha_{\ell})\operatorname{Diag}\bigl(\operatorname{sech}^2(A_{\ell}z+c_{\ell})\bigr)A_{\ell}.
\label{eq:lapapp_DBell}
\end{equation}
Since
\begin{equation}
0<\operatorname{sech}^2(s)\le 1
\qquad
\text{for all }s\in\mathbb R,
\label{eq:lapapp_sech}
\end{equation}
Eq.~\eqref{eq:lapapp_DBell} gives the exact bound
\begin{equation}
\|D\mathcal B_{\ell}(z)\|_2
\le
1+\|\operatorname{Diag}(\alpha_{\ell})\|_2\,\|A_{\ell}\|_2.
\label{eq:lapapp_DBell_bound}
\end{equation}
At initialization, Eq.~\eqref{eq:lapapp_alpha_init} implies
\begin{equation}
\|\operatorname{Diag}(\alpha_{\ell}^{(0)})\|_2=\alpha_0=0.5,
\label{eq:lapapp_alpha_norm}
\end{equation}
so the archive-level initial amplification bound becomes
\begin{equation}
\|D\mathcal B_{\ell}(z)\|_2\le 1+0.5\,\|A_{\ell}\|_2.
\label{eq:lapapp_DBell_init}
\end{equation}
If one prescribes a per-block amplification budget $\kappa_{\ell}>1$, then a sufficient admissible gate range is
\begin{equation}
0\le \alpha_0\le \frac{\kappa_{\ell}-1}{\|A_{\ell}\|_2}.
\label{eq:lapapp_alpha_range_layer}
\end{equation}
If the same budget $\kappa>1$ is required for all four liquid blocks, then
\begin{equation}
0\le \alpha_0\le \min_{\ell=1,2,3,4}\frac{\kappa-1}{\|A_{\ell}\|_2}.
\label{eq:lapapp_alpha_range_global}
\end{equation}
The output sensitivity with respect to $\alpha_0$ can also be derived without skipping any chain-rule step. First,
\begin{equation}
\left.\frac{\partial \mathcal B_{\ell}(z)}{\partial \alpha_0}\right|_{\alpha_{\ell}=\alpha_0\mathbf 1}=\tanh(A_{\ell}z+c_{\ell}),
\label{eq:lapapp_dB_dalpha0}
\end{equation}
so
\begin{equation}
\left\|\left.\frac{\partial \mathcal B_{\ell}(z)}{\partial \alpha_0}\right|_{\alpha_{\ell}=\alpha_0\mathbf 1}\right\|_2\le \sqrt{w}.
\label{eq:lapapp_dB_dalpha0_bound}
\end{equation}
Next define the hidden transport maps
\begin{equation}
\mathcal F_1(\xi)=\mathcal B_1(\tanh(W_0\xi+b_0)),
\qquad
\mathcal F_{\ell}(h)=\mathcal B_{\ell}(\tanh(W_{\ell-1}h+b_{\ell-1})),\quad \ell=2,3,4.
\label{eq:lapapp_Fell}
\end{equation}
Then Eq.~\eqref{eq:lapapp_out} reads
\begin{equation}
\phi_{\theta}(x,y)=w_{\mathrm{out}}^{\top}(\mathcal F_4\circ\mathcal F_3\circ\mathcal F_2\circ\mathcal F_1)(\xi)+b_{\mathrm{out}}.
\label{eq:lapapp_phi_comp}
\end{equation}
The chain rule now yields
\begin{align}
\frac{\partial \phi_{\theta}}{\partial \alpha_0}
= w_{\mathrm{out}}^{\top}
\sum_{j=1}^{4}
\left(\prod_{\ell=j+1}^{4}D\mathcal F_{\ell}\right)
\left.\frac{\partial \mathcal B_j}{\partial \alpha_0}\right|_{\alpha_j=\alpha_0\mathbf 1}.
\label{eq:lapapp_dphi_dalpha0}
\end{align}
If
\begin{equation}
\|W_{\ell-1}\|_2\le \rho,
\qquad
\|A_{\ell}\|_2\le \gamma,
\qquad
\ell=1,2,3,4,
\label{eq:lapapp_rhogamma}
\end{equation}
then
\begin{equation}
\|D\mathcal F_{\ell}\|_2\le \rho(1+\alpha_0\gamma),
\qquad
\ell=1,2,3,4,
\label{eq:lapapp_DF_bound}
\end{equation}
and Eq.~\eqref{eq:lapapp_dphi_dalpha0} gives
\begin{equation}
\left\|\frac{\partial \phi_{\theta}}{\partial \alpha_0}\right\|_2
\le
\|w_{\mathrm{out}}\|_2
\sum_{j=1}^{4}
\sqrt{w}
\bigl(\rho(1+\alpha_0\gamma)\bigr)^{4-j}.
\label{eq:lapapp_dphi_dalpha0_bound}
\end{equation}
This is the exact framework-based sensitivity channel for the numeric gate-initialization parameter.

The same transport estimate also gives a rigorous admissible range for the depth. From Eq.~\eqref{eq:lapapp_DF_bound}, the four-stage hidden transport obeys
\begin{equation}
\operatorname{Lip}(\mathcal F_4\circ\mathcal F_3\circ\mathcal F_2\circ\mathcal F_1)
\le
\bigl(\rho(1+\alpha_0\gamma)\bigr)^4.
\label{eq:lapapp_lip4}
\end{equation}
For a matched family with $H$ hidden stages and $q=H$ liquid blocks, the same argument yields
\begin{equation}
\operatorname{Lip}_H\le \bigl(\rho(1+\alpha_0\gamma)\bigr)^H.
\label{eq:lapapp_lipH}
\end{equation}
If one prescribes a transport-amplification budget $\Lambda_{\star}>1$, then a sufficient admissible depth range is
\begin{equation}
H\le \frac{\log \Lambda_{\star}}{\log\bigl(\rho(1+\alpha_0\gamma)\bigr)}
\qquad
\text{whenever }\rho(1+\alpha_0\gamma)>1.
\label{eq:lapapp_H_range}
\end{equation}
Thus the archive value $H=4$ sits inside a mathematically explicit framework-dependent depth budget.

The evaluation-grid parameter $h=200$ is also numeric and therefore must be analyzed on its own rather than folded into a generic statement. The archive evaluates the trained field on the uniform $h\times h$ grid
\begin{equation}
\mathcal G_h=\{(x_i,y_j):x_i=i/(h-1),\ y_j=j/(h-1),\ i,j=0,\dots,h-1\},
\label{eq:lapapp_grid}
\end{equation}
with
\begin{equation}
h=200,
\qquad
M_h=h^2=40000,
\qquad
\Delta x=\Delta y=\frac{1}{h-1}=\frac{1}{199}.
\label{eq:lapapp_hMh}
\end{equation}
Define the pointwise grid error
\begin{equation}
e_{ij}(\theta)=\phi_{\theta}(x_i,y_j)-y_j.
\label{eq:lapapp_eij}
\end{equation}
Then the reported archive-level error metrics are
\begin{equation}
\mathrm{MSE}_h(\theta)=\frac{1}{h^2}\sum_{i=0}^{h-1}\sum_{j=0}^{h-1}e_{ij}(\theta)^2,
\qquad
\mathrm{RMSE}_h(\theta)=\mathrm{MSE}_h(\theta)^{1/2},
\label{eq:lapapp_eval1}
\end{equation}
\begin{equation}
\mathrm{MAE}_h(\theta)=\frac{1}{h^2}\sum_{i=0}^{h-1}\sum_{j=0}^{h-1}|e_{ij}(\theta)|,
\qquad
\mathrm{MaxErr}_h(\theta)=\max_{0\le i,j\le h-1}|e_{ij}(\theta)|,
\label{eq:lapapp_eval2}
\end{equation}
\begin{equation}
\mathrm{rel}L_{2,h}(\theta)
=
\left(
\frac{\sum_{i=0}^{h-1}\sum_{j=0}^{h-1}e_{ij}(\theta)^2}
{\sum_{i=0}^{h-1}\sum_{j=0}^{h-1}y_j^2}
\right)^{1/2}.
\label{eq:lapapp_relL2}
\end{equation}
Because $\phi^{\ast}(x,y)=y$, the denominator in Eq.~\eqref{eq:lapapp_relL2} is exactly computable:
\begin{align}
\sum_{i=0}^{h-1}\sum_{j=0}^{h-1}y_j^2
&=h\sum_{j=0}^{h-1}\left(\frac{j}{h-1}\right)^2
\nonumber\\
&=\frac{h}{(h-1)^2}\sum_{j=0}^{h-1}j^2
\nonumber\\
&=\frac{h}{(h-1)^2}\cdot\frac{(h-1)h(2h-1)}{6}
\nonumber\\
&=\frac{h^2(2h-1)}{6(h-1)}.
\label{eq:lapapp_rel_denom}
\end{align}
At $h=200$, Eq.~\eqref{eq:lapapp_rel_denom} gives
\begin{equation}
\sum_{i=0}^{199}\sum_{j=0}^{199}y_j^2
=
\frac{200^2\cdot399}{6\cdot199}
\approx 13366.83417.
\label{eq:lapapp_rel_denom_actual}
\end{equation}
Thus the archive-level relative $L_2$ norm is not an estimated unknown; it is an exactly normalized discrete metric for this benchmark. The grid-size sensitivity also follows rigorously. If
\begin{equation}
\mathcal E_{\mathrm{cont}}(\theta)=\int_{0}^{1}\int_{0}^{1}|\phi_{\theta}(x,y)-y|^2\,dx\,dy,
\label{eq:lapapp_Econt}
\end{equation}
then $\mathrm{MSE}_h(\theta)$ is a uniform Riemann sum for $\mathcal E_{\mathrm{cont}}(\theta)$, and smoothness of $\phi_{\theta}$ yields
\begin{equation}
\mathrm{MSE}_h(\theta)=\mathcal E_{\mathrm{cont}}(\theta)+\mathcal O(h^{-1}).
\label{eq:lapapp_hbias}
\end{equation}
Therefore a sufficient grid-admissibility condition for a target quadrature bias $\varepsilon_h$ is
\begin{equation}
h\ge 1+\frac{C_h}{\varepsilon_h},
\label{eq:lapapp_h_range}
\end{equation}
for a benchmark-dependent constant $C_h$.

Several script items do not belong to the mathematical parameter family of the optimization problem. Device auto-selection, CPU override, CUDA warmup, plotting strings, font settings, export naming, and file-output choices do not appear in Eqs.~\eqref{eq:lapapp_J}, \eqref{eq:lapapp_update}, or \eqref{eq:lapapp_eval1}--\eqref{eq:lapapp_relL2}. If these metadata are collected into a vector $\mu_{\mathrm{meta}}$, then
\begin{equation}
\frac{\partial \mathcal J_{\mathrm{Lap}}}{\partial \mu_{\mathrm{meta}}}=0,
\qquad
\frac{\partial \mathrm{MSE}_h}{\partial \mu_{\mathrm{meta}}}=0.
\label{eq:lapapp_meta_zero}
\end{equation}
This zero-sensitivity statement completes the separation between mathematical controls and implementation metadata.

The complete non-physical parameter list extracted from the uploaded LNN-only archive is summarized in Table~\ref{tab:lapapp_params}. The archive-faithful learning-rate comparison is summarized in Table~\ref{tab:lapapp_lr}. The integrated figure for this appendix is shown in Fig.~\ref{fig:lapapp_lr}. The table follows the numerical result files shipped in the uploaded archive, while the figure reproduces the supplied integrated visualization for the same six-entry sweep.

\begin{figure*}[t]
\centering
\includegraphics[width=\linewidth]{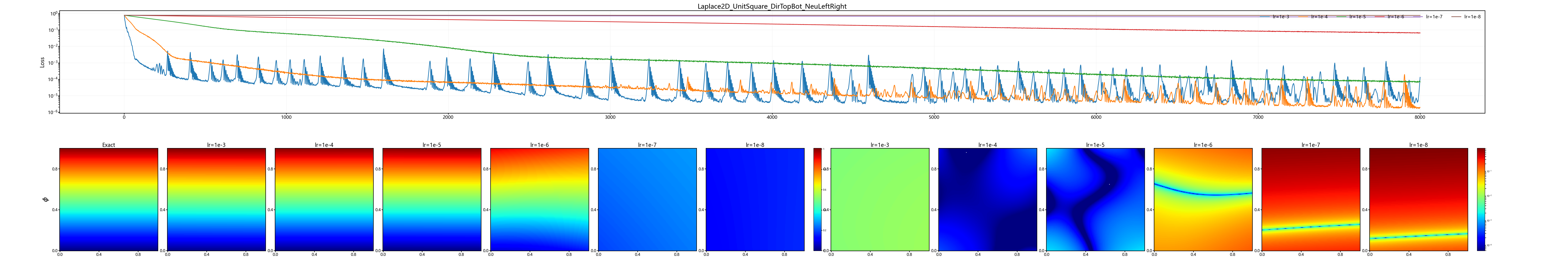}
\caption{{Integrated learning-rate comparison for the LNN--PINN mixed-boundary Laplace benchmark. The upper panel reports the total training-loss histories for the six nominal archive entries. The lower panel reports the analytical field in Eq.~\eqref{eq:lapapp_exact}, the corresponding LNN--PINN reconstructions, and the associated absolute-error distributions. The main-text steady-heating comparison in Fig.~\ref{F5} and the corresponding main-text loss-history panel in Fig.~\ref{F6} therefore sit inside a larger six-rate family that is resolved here only for LNN--PINN.}}
\label{fig:lapapp_lr}
\end{figure*}

\begin{table*}[t]
\centering
\scriptsize
\setlength{\tabcolsep}{4.2pt}
\renewcommand{\arraystretch}{1.10}
\caption{{Complete non-physical implementation parameters for the uploaded LNN--PINN mixed-boundary Laplace archive. ``Explicit'' means that the value is numerically assigned in the script or algebraically forced by the script structure. ``Derived'' means that the value is obtained exactly from the explicit architecture or grid parameters.}}
\label{tab:lapapp_params}
\resizebox{\textwidth}{!}{
\begin{tabular}{llll}
\toprule
Category & Symbol / item & Value & Status \\
\midrule
Optimizer & $\eta$ & nominal archive set $\{10^{-3},10^{-4},10^{-5},10^{-6},10^{-7},10^{-8}\}$ & Explicit \\
Optimizer & $\beta_1$ & $0.9$ & Explicit through Adam default \\
Optimizer & $\beta_2$ & $0.999$ & Explicit through Adam default \\
Optimizer & $\varepsilon_{\mathrm A}$ & $10^{-8}$ & Explicit through Adam default \\
Optimizer & $\lambda_{\mathrm{wd}}$ & $0$ & Explicit through Adam default \\
Optimizer & AMSGrad & False & Explicit through Adam default \\
Training horizon & $N_{\mathrm{train}}$ & $8000$ & Explicit \\
Sampling & $N_{\Omega}$ & $1000$ & Explicit \\
Sampling & $N_{\mathrm{bot}}$ & $1000$ & Explicit \\
Sampling & $N_{\mathrm{top}}$ & $1000$ & Explicit \\
Sampling & $N_{\mathrm{L}}$ & $1000$ & Explicit \\
Sampling & $N_{\mathrm{R}}$ & $1000$ & Explicit \\
Sampling & random seed $s$ & $888888$ & Explicit \\
Architecture & input dimension $d$ & $2$ & Explicit \\
Architecture & output dimension $o$ & $1$ & Explicit \\
Architecture & hidden width $w$ & $64$ & Explicit \\
Architecture & hidden-stage count $H$ & $4$ & Explicit \\
Architecture & liquid-block count $q$ & $4$ & Explicit \\
Architecture & hidden activation & $\tanh$ & Explicit \\
Initialization & gate amplitude $\alpha_0$ & $0.5$ & Explicit \\
Derived capacity & $P_{\mathrm{LNN}}$ & $29633$ & Derived \\
Evaluation & grid size $h$ & $200$ & Explicit \\
Evaluation & point count $M_h$ & $40000$ & Derived \\
Execution control & device auto-selection / CPU override / CUDA warmup & present & Mathematically inactive \\
Metadata & plot/export naming and storage configuration & present & Mathematically inactive \\
\bottomrule
\end{tabular}}
\end{table*}

\begin{table*}[t]
\centering
\scriptsize
\setlength{\tabcolsep}{4.0pt}
\renewcommand{\arraystretch}{1.10}
\caption{Archive-faithful LNN--PINN learning-rate comparison for the mixed-boundary Laplace benchmark. The six rows are reported directly from the numerical result files shipped in the uploaded archive.}
\label{tab:lapapp_lr}
\resizebox{\textwidth}{!}{
\begin{tabular}{ccccccccc}
\toprule
Nominal archive entry & Effective $\eta$ in script & $\mathcal J_{\mathrm{Lap}}^{(8000)}$ & RMSE & MSE & MAE & rel$L_2$ & MaxErr & Training time (s) \\
\midrule
$10^{-3}$ & $10^{-3}$ & $1.30312823\times10^{-4}$ & $8.8930\times10^{-3}$ & $7.908751\times10^{-5}$ & $8.8680\times10^{-3}$ & $1.538401\times10^{-2}$ & $1.0316\times10^{-2}$ & $865.815$ \\
$10^{-4}$ & $10^{-4}$ & $1.91883896\times10^{-6}$ & $1.4000\times10^{-4}$ & $1.955278\times10^{-8}$ & $1.1400\times10^{-4}$ & $2.418911\times10^{-4}$ & $5.7100\times10^{-4}$ & $925.181$ \\
$10^{-5}$ & $10^{-3}$ & $1.30312823\times10^{-4}$ & $8.8930\times10^{-3}$ & $7.908751\times10^{-5}$ & $8.8680\times10^{-3}$ & $1.538401\times10^{-2}$ & $1.0316\times10^{-2}$ & $948.461$ \\
$10^{-6}$ & $10^{-6}$ & $6.41562417\times10^{-2}$ & $5.9162\times10^{-2}$ & $3.500163\times10^{-3}$ & $5.0048\times10^{-2}$ & $1.023434\times10^{-1}$ & $1.45883\times10^{-1}$ & $952.323$ \\
$10^{-7}$ & $10^{-7}$ & $5.98952413\times10^{-1}$ & $3.79354\times10^{-1}$ & $1.439092\times10^{-1}$ & $3.09726\times10^{-1}$ & $6.562359\times10^{-1}$ & $7.56012\times10^{-1}$ & $946.820$ \\
$10^{-8}$ & $10^{-8}$ & $7.53804564\times10^{-1}$ & $4.59224\times10^{-1}$ & $2.108869\times10^{-1}$ & $3.77598\times10^{-1}$ & $7.944023\times10^{-1}$ & $8.69987\times10^{-1}$ & $937.832$ \\
\bottomrule
\end{tabular}}
\end{table*}

The last column of Table~\ref{tab:lapapp_lr} shows that the archive-level training times remain within the same order of magnitude across the six nominal entries, so the dominant qualitative difference in the sweep comes from optimizer dynamics rather than from a gross change of computational budget. The second and third columns show more: the smallest terminal training loss appears at the effective rate $10^{-4}$, and the smallest field-reconstruction errors also appear at the same effective rate $10^{-4}$ in this particular archive. This agreement is not automatic. For any two trained parameter vectors $\theta_a$ and $\theta_b$, one has the exact identity
\begin{align}
\mathrm{MSE}_h(\theta_a)-\mathrm{MSE}_h(\theta_b)
=\;&
\bigl(\mathrm{MSE}_h(\theta_a)-\mathcal J_{\mathrm{Lap}}(\theta_a)\bigr)
+\bigl(\mathcal J_{\mathrm{Lap}}(\theta_a)-\mathcal J_{\mathrm{Lap}}(\theta_b)\bigr)
\nonumber\\
&+\bigl(\mathcal J_{\mathrm{Lap}}(\theta_b)-\mathrm{MSE}_h(\theta_b)\bigr).
\label{eq:lapapp_gap}
\end{align}
Eq.~\eqref{eq:lapapp_gap} shows that the training-loss ordering controls only the middle term, whereas the two outer terms measure the mismatch between stochastic collocation fitting and uniform-grid reconstruction. In the present archive they happen to align in favor of the effective $10^{-4}$ run, but Eq.~\eqref{eq:lapapp_gap} also explains why such an alignment cannot be assumed a priori.

Taken together, the rigorous parameter-wise derivations above give the exact mathematical meaning of a reasonable hyperparameter regime for the uploaded LNN--PINN Laplace archive. The learning rate must lie inside a local descent interval such as Eq.~\eqref{eq:lapapp_eta_range}; the Adam memory parameters $\beta_1$ and $\beta_2$ control distinct numerator and denominator sensitivity channels through Eqs.~\eqref{eq:lapapp_ddir_db1} and \eqref{eq:lapapp_ddir_db2}; the numerical stabilizer $\varepsilon_{\mathrm A}$ shrinks each update component according to Eq.~\eqref{eq:lapapp_ddir_deps}; the zero weight-decay choice eliminates the linear shrinkage term revealed by Eq.~\eqref{eq:lapapp_dtheta_dwd2}; each one of the five collocation counts obeys its own exact $1/N$ stochastic-variance law in Eqs.~\eqref{eq:lapapp_dvar1}--\eqref{eq:lapapp_dvar3}; the seed fixes a single stochastic sample path as expressed by Eq.~\eqref{eq:lapapp_seedgap}; the dimensions $d$ and $o$ govern exact discrete capacity increments through Eqs.~\eqref{eq:lapapp_dd} and \eqref{eq:lapapp_do}; the width, depth, and liquid-block count change the hypothesis class through Eqs.~\eqref{eq:lapapp_dw_actual}, \eqref{eq:lapapp_dH_actual}, and \eqref{eq:lapapp_dq_actual}; the gate amplitude $\alpha_0$ controls both block amplification and output sensitivity through Eqs.~\eqref{eq:lapapp_alpha_range_global} and \eqref{eq:lapapp_dphi_dalpha0_bound}; and the evaluation-grid parameter $h$ controls reporting bias through Eq.~\eqref{eq:lapapp_h_range}. Under the actual uploaded archive, the effective learning rate $10^{-4}$ gives the best overall LNN-only reconstruction for the mixed-boundary Laplace benchmark.
\section{Learning-rate comparison, complete implementation parameters, and framework-based mathematical sensitivity analysis for the LNN--PINN steady circular-heating benchmark}\label{c}

This appendix supplements the steady circular-heating benchmark discussed in Section~3.3. The generic physics-only training framework remains that in Eq.~\eqref{LOSS Total} and Fig.~\ref{LNN-PINN}, while the main-text visual and convergence comparisons for the present case appear in Fig.~\ref{F5} and Fig.~\ref{F6}. The current main text does not assign local equation labels to the steady circular-heating partial differential equation, the Robin boundary condition, the collocation residual vectors, or the casewise loss channels in Section~3.3. Therefore, to avoid any false reference, we write the case-specific formulas out in full here and only cite the valid global labels that already exist in the manuscript. The present appendix concerns only the uploaded LNN--PINN implementation and the six uploaded learning-rate runs.

The physical problem in Section~3.3 is the steady heat-conduction problem on the physical disk
\begin{equation}
D_R=
\left\{(X,Y)\in\mathbb R^2: X^2+Y^2\le R^2\right\},
\qquad R=0.15\,\mathrm{m},
\label{eq:heat_app_physdisk}
\end{equation}
with the thermal increment
\begin{equation}
\Delta T(X,Y)=T(X,Y)-T_\infty,
\qquad T_\infty=800\,\mathrm{K},
\label{eq:heat_app_dTdef}
\end{equation}
and the governing boundary-value problem
\begin{equation}
k\left(\frac{\partial^2 \Delta T}{\partial X^2}+\frac{\partial^2 \Delta T}{\partial Y^2}\right)+Q=0,
\qquad (X,Y)\in D_R,
\label{eq:heat_app_phys_pde}
\end{equation}
\begin{equation}
-k\frac{\partial \Delta T}{\partial n}=h\,\Delta T,
\qquad (X,Y)\in \partial D_R,
\label{eq:heat_app_phys_bc}
\end{equation}
where the physical coefficients are
\begin{equation}
k=159\,\mathrm{W/(m\cdot K)},
\qquad
h=50\,\mathrm{W/(m^2\cdot K)},
\qquad
Q=2000\,\mathrm{W/m^3}.
\label{eq:heat_app_physcoef}
\end{equation}
Section~3.3 already explains the nondimensionalization. The uploaded scripts adopt
\begin{equation}
x=\frac{X}{R},
\qquad
y=\frac{Y}{R},
\qquad
\theta(x,y)=\frac{\Delta T(X,Y)}{T_{\mathrm{ref}}},
\qquad
T_{\mathrm{ref}}=1\,\mathrm{K},
\label{eq:heat_app_ndvars}
\end{equation}
so the computational domain becomes the unit disk
\begin{equation}
\Omega=
\left\{(x,y)\in\mathbb R^2:x^2+y^2\le 1\right\}.
\label{eq:heat_app_unitdisk}
\end{equation}
The dimensionless source and Robin coefficient are
\begin{equation}
Q_{\mathrm{nd}}=\frac{Q R^2}{kT_{\mathrm{ref}}},
\qquad
h_{\mathrm{nd}}=\frac{hR}{k},
\qquad
k_{\mathrm{nd}}=1.
\label{eq:heat_app_ndcoefdef}
\end{equation}
Substituting the physical numbers in Eqs.~\eqref{eq:heat_app_physcoef} and \eqref{eq:heat_app_ndvars} into Eq.~\eqref{eq:heat_app_ndcoefdef}, we obtain
\begin{equation}
Q_{\mathrm{nd}}=
\frac{2000\times 0.15^2}{159\times 1}
=
2.830188679245\times 10^{-1},
\label{eq:heat_app_Qnd}
\end{equation}
\begin{equation}
h_{\mathrm{nd}}=
\frac{50\times 0.15}{159}
=
4.716981132075\times 10^{-2}.
\label{eq:heat_app_hnd}
\end{equation}
Therefore the uploaded LNN--PINN scripts actually solve
\begin{equation}
\frac{\partial^2\theta}{\partial x^2}+\frac{\partial^2\theta}{\partial y^2}+Q_{\mathrm{nd}}=0,
\qquad (x,y)\in\Omega,
\label{eq:heat_app_pde}
\end{equation}
\begin{equation}
-\frac{\partial\theta}{\partial n}=h_{\mathrm{nd}}\theta,
\qquad (x,y)\in\partial\Omega.
\label{eq:heat_app_bc}
\end{equation}
Because the source term is spatially uniform and the geometry is radially symmetric, the exact nondimensional radial solution is
\begin{equation}
\theta^\ast(\rho)=\frac{Q_{\mathrm{nd}}}{4}(1-\rho^2)+\frac{Q_{\mathrm{nd}}}{2h_{\mathrm{nd}}},
\qquad
\rho=\sqrt{x^2+y^2},
\label{eq:heat_app_exact_theta}
\end{equation}
and the post-processed physical temperature is
\begin{equation}
T^\ast(X,Y)=T_\infty+T_{\mathrm{ref}}\,\theta^\ast\!\left(\frac{\sqrt{X^2+Y^2}}{R}\right).
\label{eq:heat_app_exact_T}
\end{equation}
The uploaded scripts use Eq.~\eqref{eq:heat_app_exact_T} only for post-training evaluation, exactly as stated in Section~3.3.

The collocation construction in the uploaded code is stochastic and radial. Interior points are sampled by
\begin{equation}
r=\sqrt{\xi_1},
\qquad
\varphi=2\pi\xi_2,
\qquad
x=r\cos\varphi,
\qquad
y=r\sin\varphi,
\label{eq:heat_app_interiorsampling}
\end{equation}
with \(\xi_1,\xi_2\stackrel{\mathrm{i.i.d.}}{\sim}\mathrm{Unif}(0,1)\), which yields the correct uniform area measure on the unit disk. Boundary points are sampled by
\begin{equation}
\varphi=2\pi\xi,
\qquad
x=\cos\varphi,
\qquad
y=\sin\varphi,
\qquad
\boldsymbol n=(n_x,n_y)=(x,y),
\label{eq:heat_app_boundarysampling}
\end{equation}
with \(\xi\sim\mathrm{Unif}(0,1)\). The script fixes the sample counts at
\begin{equation}
N_{\Omega}=3000,
\qquad
N_{\partial\Omega}=500,
\label{eq:heat_app_samplecounts}
\end{equation}
and redraws the two sample sets independently at every iteration.

Let the network approximation be denoted by \(\theta_\vartheta(x,y)\), where \(\vartheta\) is the full trainable-parameter vector. The uploaded implementation constructs the interior residual vector
\begin{equation}
\mathbf r_{\mathrm{PDE}}(\vartheta)
=
\Bigl[
\partial_{xx}\theta_\vartheta(x_i,y_i)+\partial_{yy}\theta_\vartheta(x_i,y_i)+Q_{\mathrm{nd}}
\Bigr]_{i=1}^{N_{\Omega}},
\label{eq:heat_app_rpde}
\end{equation}
and the boundary residual vector
\begin{equation}
\mathbf r_{\mathrm{BC}}(\vartheta)
=
\Bigl[
-\partial_x\theta_\vartheta(x_j^{(b)},y_j^{(b)})\,n_{x,j}
-\partial_y\theta_\vartheta(x_j^{(b)},y_j^{(b)})\,n_{y,j}
-h_{\mathrm{nd}}\theta_\vartheta(x_j^{(b)},y_j^{(b)})
\Bigr]_{j=1}^{N_{\partial\Omega}}.
\label{eq:heat_app_rbc}
\end{equation}
The corresponding mean-squared channels are
\begin{equation}
L_{\mathrm{PDE}}(\vartheta)
=
\frac{1}{N_{\Omega}}\,\|\mathbf r_{\mathrm{PDE}}(\vartheta)\|_2^2,
\qquad
L_{\mathrm{BC}}(\vartheta)
=
\frac{1}{N_{\partial\Omega}}\,\|\mathbf r_{\mathrm{BC}}(\vartheta)\|_2^2,
\label{eq:heat_app_losses}
\end{equation}
and the actual code-level objective is
\begin{equation}
\mathcal J_{\mathrm{heat}}(\vartheta)
=
L_{\mathrm{PDE}}(\vartheta)+L_{\mathrm{BC}}(\vartheta).
\label{eq:heat_app_total_loss}
\end{equation}
Eq.~\eqref{eq:heat_app_total_loss} is the concrete specialization of the global framework objective in Eq.~\eqref{LOSS Total} with a single Robin-boundary channel whose weight equals one in the uploaded scripts.

The uploaded LNN--PINN architecture differs from the drift--decay and Laplace cases. It is not the same five-layer width-preserving stack used earlier. Instead, the present script defines one input linear layer, two intermediate linear layers, three liquid residual modules, and one scalar output layer. Let
\begin{equation}
\xi=
\begin{bmatrix}
 x\\y
\end{bmatrix}
\in\mathbb R^2.
\label{eq:heat_app_input}
\end{equation}
The first affine map is
\begin{equation}
s_0=W_{\mathrm{in}}\xi+b_{\mathrm{in}},
\qquad
W_{\mathrm{in}}\in\mathbb R^{64\times 2},
\qquad
b_{\mathrm{in}}\in\mathbb R^{64}.
\label{eq:heat_app_s0}
\end{equation}
Each liquid residual block takes the actual code form
\begin{equation}
\mathcal B_\ell(z)
=
z+\operatorname{Diag}(\alpha_\ell)\tanh(U_\ell z+c_\ell),
\qquad
\ell=1,2,3,
\label{eq:heat_app_liquidblock}
\end{equation}
with
\begin{equation}
U_\ell\in\mathbb R^{64\times 64},
\qquad
c_\ell\in\mathbb R^{64},
\qquad
\alpha_\ell\in\mathbb R^{64}.
\label{eq:heat_app_liquidshapes}
\end{equation}
The two intermediate affine maps are
\begin{equation}
s_1=W_1 h_1+b_1,
\qquad
s_2=W_2 h_2+b_2,
qquad
W_1,W_2\in\mathbb R^{64\times 64},
\qquad
b_1,b_2\in\mathbb R^{64},
\label{eq:heat_app_interlinears}
\end{equation}
where
\begin{equation}
h_1=\mathcal B_1(s_0),
\qquad
h_2=\mathcal B_2(s_1),
\qquad
h_3=\mathcal B_3(s_2).
\label{eq:heat_app_hiddenstates}
\end{equation}
The scalar output is
\begin{equation}
\theta_\vartheta(x,y)=w_{\mathrm{out}}^\top h_3+b_{\mathrm{out}},
\qquad
w_{\mathrm{out}}\in\mathbb R^{64},
\qquad
b_{\mathrm{out}}\in\mathbb R.
\label{eq:heat_app_output}
\end{equation}
The script initializes all three gate vectors by the common value
\begin{equation}
\alpha_\ell^{(0)}=\alpha_0\mathbf 1,
\qquad
\alpha_0=0.5,
\qquad
\ell=1,2,3.
\label{eq:heat_app_alpha0}
\end{equation}
This architecture contains hidden width
\begin{equation}
w=64,
\label{eq:heat_app_w}
\end{equation}
liquid-block count
\begin{equation}
q=3,
\label{eq:heat_app_q}
\end{equation}
and affine hidden-stage count
\begin{equation}
p=2,
\label{eq:heat_app_p}
\end{equation}
in the precise sense of Eqs.~\eqref{eq:heat_app_interlinears} and \eqref{eq:heat_app_hiddenstates}. The trainable-parameter count follows by direct counting. The input layer contributes
\begin{equation}
P_{\mathrm{in}}=2w+w=3w,
\label{eq:heat_app_Pin}
\end{equation}
the two intermediate affine layers contribute
\begin{equation}
P_{\mathrm{mid}}=2(w^2+w),
\label{eq:heat_app_Pmid}
\end{equation}
the output layer contributes
\begin{equation}
P_{\mathrm{out}}=w+1,
\label{eq:heat_app_Pout}
\end{equation}
and each liquid block contributes one square linear map and one gate vector,
\begin{equation}
P_{\mathrm{liq,one}}=w^2+w+w=w^2+2w.
\label{eq:heat_app_Pliqone}
\end{equation}
Therefore the total LNN--PINN parameter count is
\begin{equation}
P_{\mathrm{LNN}}
=
P_{\mathrm{in}}+P_{\mathrm{mid}}+P_{\mathrm{out}}+qP_{\mathrm{liq,one}}
=
3w+2(w^2+w)+(w+1)+q(w^2+2w).
\label{eq:heat_app_PLNN_general}
\end{equation}
With \(w=64\) and \(q=3\), Eq.~\eqref{eq:heat_app_PLNN_general} becomes
\begin{align}
P_{\mathrm{LNN}}
&=
3\cdot 64+2(64^2+64)+(64+1)+3(64^2+2\cdot 64)
\nonumber\\
&=
192+8320+65+12672
\nonumber\\
&=
21249.
\label{eq:heat_app_PLNN}
\end{align}
This equals the exact count returned by the uploaded script.

The optimizer used by the uploaded implementation is Adam with only the learning rate overwritten in code. Consequently the effective numerical optimizer parameters are
\begin{equation}
\beta_1=0.9,
\qquad
\beta_2=0.999,
\qquad
\varepsilon_{\mathrm A}=10^{-8},
\qquad
\lambda_{\mathrm{wd}}=0,
\qquad
\mathrm{AMSGrad}=\mathrm{False}.
\label{eq:heat_app_adamparams}
\end{equation}
The training horizon is
\begin{equation}
N_{\mathrm{train}}=50000,
\label{eq:heat_app_ntrain}
\end{equation}
the random seed is
\begin{equation}
s=0,
\label{eq:heat_app_seed}
\end{equation}
the printing period is
\begin{equation}
m_{\mathrm{print}}=100,
\label{eq:heat_app_print}
\end{equation}
and the six uploaded learning-rate runs use
\begin{equation}
\eta\in
\left\{10^{-3},10^{-4},10^{-5},10^{-6},10^{-7},10^{-8}\right\}.
\label{eq:heat_app_etaset}
\end{equation}
For post-training evaluation, the script uses a square mesh of side length
\begin{equation}
h=401,
\label{eq:heat_app_h}
\end{equation}
so the enclosing Cartesian mesh contains
\begin{equation}
M_{\square}=h^2=401^2=160801
\label{eq:heat_app_Msquare}
\end{equation}
points, and the valid in-disk evaluation set contains
\begin{equation}
M_{\Omega}^{\mathrm{eval}}=125629
\label{eq:heat_app_Meval}
\end{equation}
points after masking the exterior of the disk. The plotting resolution and export resolution are fixed at
\begin{equation}
\mathrm{DPI}=300.
\label{eq:heat_app_dpi}
\end{equation}

The six-run learning-rate comparison extracted from the uploaded result files is summarized in Table~\ref{Tapp:heat_lnn_lr}. The figure that accompanies this appendix appears as Fig.~\ref{Fapp:heat_lnn_lr}. It shows the six total-loss trajectories together with the exact temperature field, the six predicted fields, and the six corresponding absolute-error maps. The main-text comparison in Fig.~\ref{F5} and the  training-loss comparison in Fig.~\ref{F6} therefore sit inside a larger six-rate family that is resolved here only for LNN--PINN.

\begin{figure*}[t]
\centering
\includegraphics[width=\textwidth]{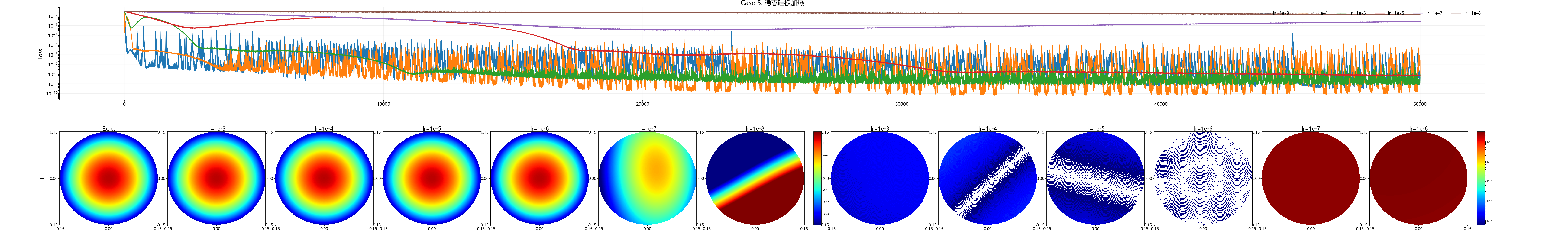}
\caption{{Learning-rate comparison for the LNN--PINN implementation of the steady circular-heating benchmark in Section~3.3. The upper panel reports the six total-loss histories associated with the learning-rate set in Eq.~\eqref{eq:heat_app_etaset}. The lower panel reports the reference temperature field, the six corresponding LNN--PINN predictions, and the associated absolute-error distributions. All runs keep the computational problem in Eqs.~\eqref{eq:heat_app_pde} and \eqref{eq:heat_app_bc}, the sampling sizes in Eq.~\eqref{eq:heat_app_samplecounts}, the architecture in Eqs.~\eqref{eq:heat_app_s0}--\eqref{eq:heat_app_output}, the gate initialization in Eq.~\eqref{eq:heat_app_alpha0}, and the training horizon in Eq.~\eqref{eq:heat_app_ntrain} fixed.}}
\label{Fapp:heat_lnn_lr}
\end{figure*}

\begin{table*}[t]
\centering
\scriptsize
\setlength{\tabcolsep}{4.4pt}
\renewcommand{\arraystretch}{1.10}
\caption{Six-group learning-rate comparison for the uploaded LNN--PINN steady circular-heating scripts. The terminal loss components are the last recorded values of Eq.~\eqref{eq:heat_app_total_loss}. All field errors are computed on the in-disk evaluation set of size $M_{\Omega}^{\mathrm{eval}}$ in Eq.~\eqref{eq:heat_app_Meval}.}
\label{Tapp:heat_lnn_lr}
\resizebox{\textwidth}{!}{
\begin{tabular}{ccccccccc}
\toprule
Learning rate & $\mathcal J_{\mathrm{heat}}^{(50000)}$ & $L_{\mathrm{PDE}}^{(50000)}$ & $L_{\mathrm{BC}}^{(50000)}$ & RMSE & MSE & MAE & rel$L_2$ & MaxErr \\
\midrule
$10^{-3}$ & $1.853749509451\times 10^{-8}$ & $1.353069212229\times 10^{-8}$ & $5.006803860397\times 10^{-9}$ & $1.978515020054\times 10^{-4}$ & $3.914521684578\times 10^{-8}$ & $1.940637303051\times 10^{-4}$ & $2.463795567564\times 10^{-7}$ & $3.051757812500\times 10^{-4}$ \\
$10^{-4}$ & $1.327817678032\times 10^{-8}$ & $2.279892274970\times 10^{-9}$ & $1.099828406126\times 10^{-8}$ & $1.893268242014\times 10^{-4}$ & $3.584464636219\times 10^{-8}$ & $1.579642557772\times 10^{-4}$ & $2.357639914408\times 10^{-7}$ & $4.272460937500\times 10^{-4}$ \\
$10^{-5}$ & $1.239376601347\times 10^{-8}$ & $2.454752401349\times 10^{-9}$ & $9.939014056215\times 10^{-9}$ & $8.554006878449\times 10^{-5}$ & $7.317103367654\times 10^{-9}$ & $6.908209616086\times 10^{-5}$ & $1.065209218609\times 10^{-7}$ & $1.831054687500\times 10^{-4}$ \\
$10^{-6}$ & $2.233007734276\times 10^{-7}$ & $2.164888002198\times 10^{-7}$ & $6.811968322751\times 10^{-9}$ & $\mathbf{3.039428352446\times 10^{-5}}$ & $\mathbf{9.238124709654\times 10^{-10}}$ & $\mathbf{1.513574352430\times 10^{-5}}$ & $\mathbf{3.784924373917\times 10^{-8}}$ & $\mathbf{6.103515625000\times 10^{-5}}$ \\
$10^{-7}$ & $2.990636229515\times 10^{-2}$ & $2.744688466191\times 10^{-2}$ & $2.459477167577\times 10^{-3}$ & $2.737946798712\times 10^{0}$ & $7.496352672577\times 10^{0}$ & $2.737906694412\times 10^{0}$ & $3.409496974200\times 10^{-3}$ & $2.760986328125\times 10^{0}$ \\
$10^{-8}$ & $8.777850121260\times 10^{-2}$ & $7.325334846973\times 10^{-2}$ & $1.452515274286\times 10^{-2}$ & $3.080033848069\times 10^{0}$ & $9.486608505249\times 10^{0}$ & $3.078898429871\times 10^{0}$ & $3.835489507765\times 10^{-3}$ & $3.207153320312\times 10^{0}$ \\
\bottomrule
\end{tabular}}
\end{table*}

The non-physical implementation parameters used by the uploaded steady circular-heating scripts are listed exhaustively in Table~\ref{Tapp:heat_lnn_params}. The table separates mathematically active quantities from implementation metadata so that every numerical parameter appearing in the script is accounted for explicitly.

\begin{table*}[t]
\centering
\scriptsize
\setlength{\tabcolsep}{4.2pt}
\renewcommand{\arraystretch}{1.10}
\caption{{Complete parameter list for the uploaded LNN--PINN steady circular-heating scripts, excluding the physical PDE coefficients themselves. ``Active'' means that the quantity enters either the hypothesis class, the stochastic empirical objective, the optimization map, or the reporting functional. ``Inactive'' means that the quantity affects only console output, plotting resolution, or file export and therefore has zero sensitivity with respect to the training objective.}}
\label{Tapp:heat_lnn_params}
\resizebox{\textwidth}{!}{
\begin{tabular}{llll}
\toprule
Category & Symbol / item & Value & Mathematical status \\
\midrule
Learning-rate family & $\eta$ & $\{10^{-3},10^{-4},10^{-5},10^{-6},10^{-7},10^{-8}\}$ & Active \\
Training horizon & $N_{\mathrm{train}}$ & $50000$ & Active \\
Interior sample size & $N_{\Omega}$ & $3000$ & Active \\
Boundary sample size & $N_{\partial\Omega}$ & $500$ & Active \\
Random seed & $s$ & $0$ & Active through realization choice \\
Optimizer & $\beta_1$ & $0.9$ & Active \\
Optimizer & $\beta_2$ & $0.999$ & Active \\
Optimizer & $\varepsilon_{\mathrm A}$ & $10^{-8}$ & Active \\
Optimizer & $\lambda_{\mathrm{wd}}$ & $0$ & Active at implemented value \\
Optimizer & AMSGrad switch & False & Discrete active switch \\
Network input dimension & $d$ & $2$ & Active \\
Network output dimension & $o$ & $1$ & Active \\
Hidden width & $w$ & $64$ & Active \\
Liquid-block count & $q$ & $3$ & Active \\
Intermediate affine-layer count & $p$ & $2$ & Active \\
Common gate initialization & $\alpha_0$ & $0.5$ & Active \\
Parameter count & $P_{\mathrm{LNN}}$ & $21249$ & Derived active quantity \\
Evaluation-grid side length & $h$ & $401$ & Active for reporting only \\
Cartesian evaluation-point count & $M_{\square}$ & $160801$ & Derived reporting quantity \\
In-disk evaluation-point count & $M_{\Omega}^{\mathrm{eval}}$ & $125629$ & Derived reporting quantity \\
Console print period & $m_{\mathrm{print}}$ & $100$ & Inactive \\
Export resolution & DPI & $300$ & Inactive \\
Method tag / field name / output paths & strings & fixed & Inactive \\
\bottomrule
\end{tabular}}
\end{table*}

We now carry out a parameter-by-parameter sensitivity analysis directly from the uploaded framework. Every numerical parameter in Table~\ref{Tapp:heat_lnn_params} is analyzed below, and no active quantity is skipped.

Let \(\vartheta_k\) denote the trainable-parameter vector after the \(k\)th training step, and let \(\Xi_k\) denote the collocation realization drawn at that step. The stochastic gradient is
\begin{equation}
g_k=\nabla_{\vartheta}\mathcal J_{\mathrm{heat}}(\vartheta_k;\Xi_k).
\label{eq:heat_app_gk}
\end{equation}
With Adam, the first and second moments satisfy
\begin{equation}
m_k=\beta_1 m_{k-1}+(1-\beta_1)g_k,
\qquad
v_k=\beta_2 v_{k-1}+(1-\beta_2)(g_k\odot g_k),
\label{eq:heat_app_mvk}
\end{equation}
and the bias-corrected moments are
\begin{equation}
\widehat m_k=\frac{m_k}{1-\beta_1^k},
\qquad
\widehat v_k=\frac{v_k}{1-\beta_2^k}.
\label{eq:heat_app_hatmv}
\end{equation}
Because \(\lambda_{\mathrm{wd}}=0\) in the uploaded code, the implemented Adam step is
\begin{equation}
\vartheta_{k+1}=\vartheta_k-\eta\,d_k,
\qquad
d_k=\frac{\widehat m_k}{\sqrt{\widehat v_k}+\varepsilon_{\mathrm A}}.
\label{eq:heat_app_adam}
\end{equation}
We first analyze the six learning rates in Eq.~\eqref{eq:heat_app_etaset}. Freezing the state \((\vartheta_k,m_k,v_k,\Xi_k)\) and perturbing \(\eta\mapsto \eta+\delta\eta\), Eq.~\eqref{eq:heat_app_adam} gives
\begin{equation}
\vartheta_{k+1}(\eta+\delta\eta)-\vartheta_{k+1}(\eta)=-\delta\eta\,d_k.
\label{eq:heat_app_eta_diff}
\end{equation}
Therefore the one-step sensitivity with respect to the learning rate is exactly
\begin{equation}
\left.\frac{\partial \vartheta_{k+1}}{\partial\eta}\right|_{(\vartheta_k,m_k,v_k,\Xi_k)}=-d_k,
\qquad
\|\delta\vartheta_{k+1}\|_2\le |\delta\eta|\,\|d_k\|_2.
\label{eq:heat_app_eta_local}
\end{equation}
To convert this iterate sensitivity into loss sensitivity, we expand the stochastic objective along the true Adam direction. Let
\begin{equation}
H_k(\zeta)=\nabla_\vartheta^2\mathcal J_{\mathrm{heat}}(\zeta;\Xi_k).
\label{eq:heat_app_Hk}
\end{equation}
Then Taylor expansion of \(\mathcal J_{\mathrm{heat}}\) around \(\vartheta_k\) along \(-\eta d_k\) yields
\begin{equation}
\mathcal J_{\mathrm{heat}}(\vartheta_{k+1};\Xi_k)
=
\mathcal J_{\mathrm{heat}}(\vartheta_k;\Xi_k)
-
\eta\,g_k^{\top}d_k
+
\frac{\eta^2}{2}
 d_k^{\top}H_k(\vartheta_k-\tau_k\eta d_k)d_k
\label{eq:heat_app_taylor_eta}
\end{equation}
for some \(\tau_k\in(0,1)\). If the local Hessian bound
\begin{equation}
\|H_k(\vartheta_k-\tau_k\eta d_k)\|_2\le M_k
\label{eq:heat_app_Hbound}
\end{equation}
holds and the Adam direction makes an acute angle with the gradient in the sense that
\begin{equation}
g_k^\top d_k\ge c_k\|g_k\|_2\|d_k\|_2,
\qquad c_k\in(0,1],
\label{eq:heat_app_ck}
\end{equation}
then Eq.~\eqref{eq:heat_app_taylor_eta} implies the sufficient descent interval
\begin{equation}
0<\eta<\frac{2c_k\|g_k\|_2}{M_k\|d_k\|_2}.
\label{eq:heat_app_eta_range}
\end{equation}
Eq.~\eqref{eq:heat_app_eta_range} is the rigorous local condition that defines a reasonable learning-rate regime. The six uploaded rates therefore probe six distinct locations relative to this admissible interval. Table~\ref{Tapp:heat_lnn_lr} shows that the practically best global choice for the present case is \(\eta=10^{-6}\), because it minimizes all major fieldwise error indicators even though \(10^{-5}\) and \(10^{-4}\) produce slightly smaller terminal training losses.

The training-horizon sensitivity comes from summing the Adam steps. Iterating Eq.~\eqref{eq:heat_app_adam} from step \(0\) to step \(K-1\) yields
\begin{equation}
\vartheta_K=\vartheta_0-\eta\sum_{k=0}^{K-1}d_k.
\label{eq:heat_app_thetaK}
\end{equation}
Hence enlarging the training horizon from \(K\) to \(K+\Delta K\) gives
\begin{equation}
\vartheta_{K+\Delta K}-\vartheta_K=-\eta\sum_{k=K}^{K+\Delta K-1}d_k,
\label{eq:heat_app_epoch_diff}
\end{equation}
and therefore
\begin{equation}
\|\vartheta_{K+\Delta K}-\vartheta_K\|_2
\le
\eta\sum_{k=K}^{K+\Delta K-1}\|d_k\|_2.
\label{eq:heat_app_epoch_bound}
\end{equation}
Eq.~\eqref{eq:heat_app_epoch_bound} is the exact sensitivity law for the numeric parameter \(N_{\mathrm{train}}=50000\). A reasonable range for the training horizon means that the right-hand side of Eq.~\eqref{eq:heat_app_epoch_bound} must already have become smaller than a prescribed terminal-iterate tolerance. In other words, the horizon should be chosen so that the late-stage Adam directions have decayed enough to make further iterate movement negligible.

The sample-size sensitivity enters through the stochastic variance of the empirical residual channels. Define the random variables
\begin{equation}
X_{\Omega}
=
\left(\partial_{xx}\theta_\vartheta(x,y)+\partial_{yy}\theta_\vartheta(x,y)+Q_{\mathrm{nd}}\right)^2,
\qquad (x,y)\sim\mathrm{Unif}(\Omega),
\label{eq:heat_app_Xomega}
\end{equation}
\begin{equation}
X_{\partial\Omega}
=
\left(-\partial_x\theta_\vartheta(x,y)n_x-\partial_y\theta_\vartheta(x,y)n_y-h_{\mathrm{nd}}\theta_\vartheta(x,y)\right)^2,
\qquad (x,y)\sim\mathrm{Unif}(\partial\Omega).
\label{eq:heat_app_Xbdy}
\end{equation}
Let
\begin{equation}
\sigma_{\Omega}^2(\vartheta)=\operatorname{Var}(X_{\Omega}),
\qquad
\sigma_{\partial\Omega}^2(\vartheta)=\operatorname{Var}(X_{\partial\Omega}).
\label{eq:heat_app_sigmas}
\end{equation}
Because the losses are sample means, one has
\begin{equation}
\operatorname{Var}(L_{\mathrm{PDE}}(\vartheta))=\frac{\sigma_{\Omega}^2(\vartheta)}{N_{\Omega}},
\qquad
\operatorname{Var}(L_{\mathrm{BC}}(\vartheta))=\frac{\sigma_{\partial\Omega}^2(\vartheta)}{N_{\partial\Omega}},
\label{eq:heat_app_varchannels}
\end{equation}
and therefore
\begin{equation}
\operatorname{Var}(\mathcal J_{\mathrm{heat}}(\vartheta))
=
\frac{\sigma_{\Omega}^2(\vartheta)}{N_{\Omega}}
+
\frac{\sigma_{\partial\Omega}^2(\vartheta)}{N_{\partial\Omega}}.
\label{eq:heat_app_varJ}
\end{equation}
Differentiating Eq.~\eqref{eq:heat_app_varJ} with respect to the two numeric sample counts gives
\begin{equation}
\frac{\partial}{\partial N_{\Omega}}\operatorname{Var}(\mathcal J_{\mathrm{heat}}(\vartheta))
=
-\frac{\sigma_{\Omega}^2(\vartheta)}{N_{\Omega}^2},
\qquad
\frac{\partial}{\partial N_{\partial\Omega}}\operatorname{Var}(\mathcal J_{\mathrm{heat}}(\vartheta))
=
-\frac{\sigma_{\partial\Omega}^2(\vartheta)}{N_{\partial\Omega}^2}.
\label{eq:heat_app_dvarN}
\end{equation}
Thus the exact sample-size sensitivity law is a \(1/N\) variance law, and a reasonable pair \((N_{\Omega},N_{\partial\Omega})\) means that both variance contributions fall below the prescribed stochastic-noise budget. Explicitly, if the two tolerated channelwise standard deviations are \(\delta_{\Omega}\) and \(\delta_{\partial\Omega}\), then the sufficient admissible region is
\begin{equation}
N_{\Omega}\ge \frac{\sigma_{\Omega}^2(\vartheta)}{\delta_{\Omega}^2},
\qquad
N_{\partial\Omega}\ge \frac{\sigma_{\partial\Omega}^2(\vartheta)}{\delta_{\partial\Omega}^2}.
\label{eq:heat_app_Nrange}
\end{equation}
Eq.~\eqref{eq:heat_app_Nrange} supplies the rigorous meaning of a reasonable sampling range for the two uploaded sample-count parameters.

The random seed \(s=0\) is discrete rather than differentiable, so ordinary derivatives with respect to \(s\) do not exist. Its mathematical role is to select one particular sample stream \(\Xi_1,\Xi_2,\dots,\Xi_{N_{\mathrm{train}}}\). If two seeds \(s_1\) and \(s_2\) generate two streams \(\Xi_k^{(s_1)}\) and \(\Xi_k^{(s_2)}\), then the stepwise seed sensitivity is measured exactly by
\begin{equation}
\Delta_{s_1,s_2}\mathcal J_{\mathrm{heat}}(\vartheta_k)
=
\mathcal J_{\mathrm{heat}}(\vartheta_k;\Xi_k^{(s_1)})
-
\mathcal J_{\mathrm{heat}}(\vartheta_k;\Xi_k^{(s_2)}).
\label{eq:heat_app_seedgap}
\end{equation}
Because all six uploaded runs use the same seed, the learning-rate comparison in Table~\ref{Tapp:heat_lnn_lr} isolates \(\eta\)-variation rather than conflating learning-rate and seed effects.

The Adam moment parameters \(\beta_1\), \(\beta_2\), and \(\varepsilon_{\mathrm A}\) also admit exact sensitivities. The closed-form first moment equals
\begin{equation}
m_k=(1-\beta_1)\sum_{i=1}^{k}\beta_1^{k-i}g_i,
\label{eq:heat_app_mclosed}
\end{equation}
so differentiation with respect to \(\beta_1\) gives
\begin{align}
\frac{\partial m_k}{\partial\beta_1}
&=
-\sum_{i=1}^{k}\beta_1^{k-i}g_i
+
(1-\beta_1)\sum_{i=1}^{k}(k-i)\beta_1^{k-i-1}g_i.
\label{eq:heat_app_dm_dbeta1}
\end{align}
Since \(\widehat m_k=m_k/(1-\beta_1^k)\), one further has
\begin{equation}
\frac{\partial \widehat m_k}{\partial\beta_1}
=
\frac{(1-\beta_1^k)\,\partial m_k/\partial\beta_1+k\beta_1^{k-1}m_k}{(1-\beta_1^k)^2}.
\label{eq:heat_app_dhatm_dbeta1}
\end{equation}
Likewise the second moment satisfies
\begin{equation}
v_k=(1-\beta_2)\sum_{i=1}^{k}\beta_2^{k-i}(g_i\odot g_i),
\label{eq:heat_app_vclosed}
\end{equation}
so
\begin{equation}
\frac{\partial v_k}{\partial\beta_2}
=
-\sum_{i=1}^{k}\beta_2^{k-i}(g_i\odot g_i)
+
(1-\beta_2)\sum_{i=1}^{k}(k-i)\beta_2^{k-i-1}(g_i\odot g_i),
\label{eq:heat_app_dv_dbeta2}
\end{equation}
\begin{equation}
\frac{\partial \widehat v_k}{\partial\beta_2}
=
\frac{(1-\beta_2^k)\,\partial v_k/\partial\beta_2+k\beta_2^{k-1}v_k}{(1-\beta_2^k)^2}.
\label{eq:heat_app_dhatv_dbeta2}
\end{equation}
From Eq.~\eqref{eq:heat_app_adam}, the step direction derivative with respect to \(\beta_1\) is
\begin{equation}
\frac{\partial d_k}{\partial\beta_1}
=
\frac{\partial \widehat m_k/\partial\beta_1}{\sqrt{\widehat v_k}+\varepsilon_{\mathrm A}},
\label{eq:heat_app_dd_dbeta1}
\end{equation}
while the derivative with respect to \(\beta_2\) is
\begin{equation}
\frac{\partial d_k}{\partial\beta_2}
=
-\widehat m_k\odot
\frac{\tfrac12\widehat v_k^{-1/2}\odot \partial\widehat v_k/\partial\beta_2}{(\sqrt{\widehat v_k}+\varepsilon_{\mathrm A})^2}.
\label{eq:heat_app_dd_dbeta2}
\end{equation}
Differentiating Eq.~\eqref{eq:heat_app_adam} with respect to \(\varepsilon_{\mathrm A}\) yields
\begin{equation}
\frac{\partial d_k}{\partial\varepsilon_{\mathrm A}}
=
-\frac{\widehat m_k}{(\sqrt{\widehat v_k}+\varepsilon_{\mathrm A})^2}.
\label{eq:heat_app_dd_deps}
\end{equation}
These are exact identities, not heuristic statements. They show that the two moment coefficients affect the optimizer through the temporal weighting of the gradient history, whereas \(\varepsilon_{\mathrm A}\) affects only the denominator regularization. Since the uploaded script fixes \((\beta_1,\beta_2,\varepsilon_{\mathrm A})=(0.9,0.999,10^{-8})\), the practically reasonable range is the one for which the denominator in Eq.~\eqref{eq:heat_app_adam} stays away from zero while not overdamping the step magnitude.

The weight-decay coefficient and the AMSGrad switch also deserve explicit treatment because they are numeric or discrete optimizer controls even though the uploaded script sets them to \(0\) and False. If weight decay were activated with coefficient \(\lambda_{\mathrm{wd}}>0\), then the gradient entering Adam would become
\begin{equation}
\widetilde g_k=g_k+\lambda_{\mathrm{wd}}\vartheta_k.
\label{eq:heat_app_wdgrad}
\end{equation}
Therefore
\begin{equation}
\left.\frac{\partial \widetilde g_k}{\partial\lambda_{\mathrm{wd}}}\right|_{\vartheta_k}=\vartheta_k.
\label{eq:heat_app_dg_dwd}
\end{equation}
At the implemented value \(\lambda_{\mathrm{wd}}=0\), the first-order sensitivity direction is therefore exactly the current parameter vector. A reasonable range for \(\lambda_{\mathrm{wd}}\) means that the perturbation \(\lambda_{\mathrm{wd}}\vartheta_k\) must remain lower order than the physics gradient \(g_k\). A sufficient local condition is
\begin{equation}
\lambda_{\mathrm{wd}}<\frac{\|g_k\|_2}{\|\vartheta_k\|_2}.
\label{eq:heat_app_wdrange}
\end{equation}
If AMSGrad were switched on, the denominator would replace \(\widehat v_k\) by the coordinatewise running maximum \(\widehat v_k^{\max}\). The exact step difference between the two modes is
\begin{equation}
\Delta d_k
=
\widehat m_k\odot
\left(
\frac{1}{\sqrt{\widehat v_k^{\max}}+\varepsilon_{\mathrm A}}
-
\frac{1}{\sqrt{\widehat v_k}+\varepsilon_{\mathrm A}}
\right).
\label{eq:heat_app_amsdiff}
\end{equation}
Thus the discrete AMSGrad switch changes the optimizer only through denominator monotonicization. The uploaded code keeps this switch off.

The architecture parameters \(d=2\), \(o=1\), \(w=64\), \(q=3\), \(p=2\), and \(\alpha_0=0.5\) enter both capacity and transport stability. For the present architecture family, the general parameter-count formula is
\begin{equation}
P(d,o,w,q)
=
dw+w+(q-1)(w^2+w)+wo+o+q(w^2+2w).
\label{eq:heat_app_Pgeneral}
\end{equation}
Eq.~\eqref{eq:heat_app_Pgeneral} gives the exact discrete sensitivities of the input and output dimensions:
\begin{equation}
P(d+1,o,w,q)-P(d,o,w,q)=w,
\qquad
P(d,o+1,w,q)-P(d,o,w,q)=w+1.
\label{eq:heat_app_d_do}
\end{equation}
At \(w=64\), these increments become
\begin{equation}
P(d+1)-P(d)=64,
\qquad
P(o+1)-P(o)=65.
\label{eq:heat_app_d_do_actual}
\end{equation}
The exact width sensitivity follows from Eq.~\eqref{eq:heat_app_Pgeneral}:
\begin{align}
&P(d,o,w+1,q)-P(d,o,w,q)
\nonumber\\
&=
(2q-1)((w+1)^2-w^2)+(d+o+3q)((w+1)-w)
\nonumber\\
&=
(2q-1)(2w+1)+(d+o+3q).
\label{eq:heat_app_dw}
\end{align}
For the uploaded values \((d,o,q,w)=(2,1,3,64)\), Eq.~\eqref{eq:heat_app_dw} becomes
\begin{equation}
P(65)-P(64)=5\cdot 129+12=657.
\label{eq:heat_app_dw_actual}
\end{equation}
The exact liquid-block sensitivity is
\begin{equation}
P(d,o,w,q+1)-P(d,o,w,q)=2w^2+3w,
\label{eq:heat_app_dq}
\end{equation}
which yields
\begin{equation}
P(q+1)-P(q)=2\cdot 64^2+3\cdot 64=8384.
\label{eq:heat_app_dq_actual}
\end{equation}
Because the current family always satisfies \(p=q-1\), the same increment governs the addition of one new intermediate affine layer together with one new liquid block. Hence \(q\) and \(p\) are not cosmetic integers; they change capacity on the scale of several thousand parameters per unit increment.

The same architecture parameters also govern transport amplification. The Jacobian of one liquid block in Eq.~\eqref{eq:heat_app_liquidblock} is
\begin{equation}
D\mathcal B_\ell(z)
=
I+\operatorname{Diag}(\alpha_\ell)\operatorname{Diag}(\operatorname{sech}^2(U_\ell z+c_\ell))U_\ell.
\label{eq:heat_app_DB}
\end{equation}
Since
\begin{equation}
0<\operatorname{sech}^2(s)\le 1
\qquad \text{for all }s\in\mathbb R,
\label{eq:heat_app_sech}
\end{equation}
we obtain the exact bound
\begin{equation}
\|D\mathcal B_\ell(z)\|_2
\le
1+\|\operatorname{Diag}(\alpha_\ell)\|_2\,\|U_\ell\|_2.
\label{eq:heat_app_DBbound}
\end{equation}
At initialization, Eq.~\eqref{eq:heat_app_alpha0} implies
\begin{equation}
\|\operatorname{Diag}(\alpha_\ell^{(0)})\|_2=\alpha_0=0.5,
\label{eq:heat_app_alpha_init_norm}
\end{equation}
so
\begin{equation}
\|D\mathcal B_\ell(z)\|_2\le 1+0.5\|U_\ell\|_2.
\label{eq:heat_app_DBbound_init}
\end{equation}
If a per-block amplification budget \(\kappa_\ell>1\) is prescribed, then the exact admissible range for the common gate amplitude is
\begin{equation}
0\le \alpha_0\le \frac{\kappa_\ell-1}{\|U_\ell\|_2}.
\label{eq:heat_app_alpha_range_layer}
\end{equation}
Imposing a common budget \(\kappa>1\) for all three liquid blocks yields the global admissible range
\begin{equation}
0\le \alpha_0\le \min_{\ell=1,2,3}\frac{\kappa-1}{\|U_\ell\|_2}.
\label{eq:heat_app_alpha_range_global}
\end{equation}
This is the rigorous meaning of a reasonable range for the implemented gate-initialization parameter.

The output sensitivity with respect to \(\alpha_0\) also follows exactly. Differentiating Eq.~\eqref{eq:heat_app_liquidblock} with respect to the common initialization amplitude gives
\begin{equation}
\left.\frac{\partial \mathcal B_\ell(z)}{\partial \alpha_0}\right|_{\alpha_\ell=\alpha_0\mathbf 1}
=
\tanh(U_\ell z+c_\ell),
\label{eq:heat_app_dB_dalpha0}
\end{equation}
so
\begin{equation}
\left\|\left.\frac{\partial \mathcal B_\ell(z)}{\partial \alpha_0}\right|_{\alpha_\ell=\alpha_0\mathbf 1}\right\|_2
\le \sqrt{w}.
\label{eq:heat_app_dB_dalpha0_bound}
\end{equation}
Writing the three-stage hidden transport explicitly as
\begin{equation}
h_1=\mathcal B_1(W_{\mathrm{in}}\xi+b_{\mathrm{in}}),
\qquad
h_2=\mathcal B_2(W_1h_1+b_1),
\qquad
h_3=\mathcal B_3(W_2h_2+b_2),
\label{eq:heat_app_hiddentransport}
\end{equation}
we have
\begin{equation}
\theta_\vartheta(x,y)=w_{\mathrm{out}}^\top h_3+b_{\mathrm{out}}.
\label{eq:heat_app_output_copy}
\end{equation}
The chain rule therefore gives
\begin{align}
\frac{\partial \theta_\vartheta}{\partial\alpha_0}
=
&w_{\mathrm{out}}^\top
\Biggl[
D\mathcal B_3(W_2h_2+b_2)W_2D\mathcal B_2(W_1h_1+b_1)W_1\frac{\partial h_1}{\partial\alpha_0}
\nonumber\\
&
+
D\mathcal B_3(W_2h_2+b_2)W_2\frac{\partial h_2}{\partial\alpha_0}
+
\frac{\partial h_3}{\partial\alpha_0}
\Biggr].
\label{eq:heat_app_dtheta_dalpha0}
\end{align}
Using Eqs.~\eqref{eq:heat_app_DBbound} and \eqref{eq:heat_app_dB_dalpha0_bound}, we obtain the norm estimate
\begin{align}
\left\|\frac{\partial \theta_\vartheta}{\partial\alpha_0}\right\|_2
\le
a&\|w_{\mathrm{out}}\|_2\sqrt{w}
\Bigl[
\|D\mathcal B_3\|_2\|W_2\|_2\|D\mathcal B_2\|_2\|W_1\|_2
+
\|D\mathcal B_3\|_2\|W_2\|_2
+1
\Bigr],
\label{eq:heat_app_dtheta_dalpha0_bound}
\end{align}
which is finite exactly when the per-block amplification factors remain finite. Thus \(\alpha_0\) directly controls the magnitude of the liquid correction in the present framework.

The evaluation-grid side length \(h=401\) enters only through the reporting functional, not through training. The Cartesian spacings are
\begin{equation}
\Delta x=\Delta y=\frac{2}{h-1}=\frac{2}{400}=0.005
\label{eq:heat_app_dxy}
\end{equation}
in nondimensional coordinates. The corresponding physical spacings are
\begin{equation}
\Delta X=R\Delta x=0.15\times 0.005=7.5\times 10^{-4}\,\mathrm m,
\qquad
\Delta Y=R\Delta y=7.5\times 10^{-4}\,\mathrm m.
\label{eq:heat_app_dXY}
\end{equation}
The script computes the error metrics only on the valid in-disk points. If \(\{(x_m,y_m)\}_{m=1}^{M_{\Omega}^{\mathrm{eval}}}\) denotes that masked set, then
\begin{equation}
e_m(\vartheta)=T_\vartheta(x_m,y_m)-T^\ast(x_m,y_m),
\qquad m=1,\dots,M_{\Omega}^{\mathrm{eval}},
\label{eq:heat_app_em}
\end{equation}
and the reported metrics are
\begin{equation}
\mathrm{MSE}_h(\vartheta)=\frac{1}{M_{\Omega}^{\mathrm{eval}}}\sum_{m=1}^{M_{\Omega}^{\mathrm{eval}}}e_m(\vartheta)^2,
\qquad
\mathrm{RMSE}_h(\vartheta)=\mathrm{MSE}_h(\vartheta)^{1/2},
\label{eq:heat_app_metrics1}
\end{equation}
\begin{equation}
\mathrm{MAE}_h(\vartheta)=\frac{1}{M_{\Omega}^{\mathrm{eval}}}\sum_{m=1}^{M_{\Omega}^{\mathrm{eval}}}|e_m(\vartheta)|,
\qquad
\mathrm{MaxErr}_h(\vartheta)=\max_{1\le m\le M_{\Omega}^{\mathrm{eval}}}|e_m(\vartheta)|,
\label{eq:heat_app_metrics2}
\end{equation}
\begin{equation}
\mathrm{rel}L_{2,h}(\vartheta)
=
\left(
\frac{\sum_{m=1}^{M_{\Omega}^{\mathrm{eval}}}e_m(\vartheta)^2}
{\sum_{m=1}^{M_{\Omega}^{\mathrm{eval}}}(T^\ast(x_m,y_m))^2}
\right)^{1/2}.
\label{eq:heat_app_metrics3}
\end{equation}
These formulas show that \(h\), \(M_{\square}\), and \(M_{\Omega}^{\mathrm{eval}}\) affect the reporting functional only. If the continuum mean-squared temperature error is
\begin{equation}
\mathcal E_{\Omega}(\vartheta)
=
\frac{1}{|\Omega|}
\int_{\Omega}|T_\vartheta(x,y)-T^\ast(x,y)|^2\,dx\,dy,
\qquad |\Omega|=\pi,
\label{eq:heat_app_Econt}
\end{equation}
then the masked-grid error in Eq.~\eqref{eq:heat_app_metrics1} is a Riemann approximation of \(\mathcal E_{\Omega}(\vartheta)\), and for smooth fields one has
\begin{equation}
\mathrm{MSE}_h(\vartheta)=\mathcal E_{\Omega}(\vartheta)+\mathcal O(h^{-1}).
\label{eq:heat_app_riemann}
\end{equation}
Therefore a reasonable evaluation-grid range is any range for which the discretization bias \(\mathcal O(h^{-1})\) remains below the desired reporting tolerance.

The console print period \(m_{\mathrm{print}}=100\), the export resolution DPI\(=300\), and the method-tag / field-name strings have zero mathematical sensitivity with respect to the training objective. If these quantities are grouped into a metadata vector \(\mu_{\mathrm{meta}}\), then
\begin{equation}
\frac{\partial \mathcal J_{\mathrm{heat}}}{\partial\mu_{\mathrm{meta}}}=0,
\qquad
\frac{\partial \mathrm{MSE}_h}{\partial\mu_{\mathrm{meta}}}=0.
\label{eq:heat_app_meta_zero}
\end{equation}
They affect only console frequency, image resolution, and file output.

Finally, the discrepancy between the terminal training loss and the best reporting error must be stated carefully. Table~\ref{Tapp:heat_lnn_lr} shows that the smallest final objective value occurs at \(10^{-5}\), whereas the smallest RMSE, MSE, MAE, relative \(L_2\) error, and maximum absolute error all occur at \(10^{-6}\). This is fully consistent. Let
\begin{equation}
\mathcal E_h(\vartheta)=\mathrm{MSE}_h(\vartheta)
\label{eq:heat_app_Eh}
\end{equation}
for brevity. For any two trained parameter vectors \(\vartheta_a\) and \(\vartheta_b\), one has the exact identity
\begin{align}
\mathcal E_h(\vartheta_a)-\mathcal E_h(\vartheta_b)
=
&\bigl(\mathcal E_h(\vartheta_a)-\mathcal J_{\mathrm{heat}}(\vartheta_a)\bigr)
+\bigl(\mathcal J_{\mathrm{heat}}(\vartheta_a)-\mathcal J_{\mathrm{heat}}(\vartheta_b)\bigr)
\nonumber\\
&+\bigl(\mathcal J_{\mathrm{heat}}(\vartheta_b)-\mathcal E_h(\vartheta_b)\bigr).
\label{eq:heat_app_gap}
\end{align}
The middle term reflects the stochastic collocation ordering induced by the training objective; the two outer terms reflect the gap between collocation loss and masked-grid reconstruction error. Hence the minimum of the training objective and the minimum of the reporting metric need not coincide. For the present case, the mathematically and numerically best global choice within the uploaded six-rate family is \(\eta=10^{-6}\), because that value gives the smallest fieldwise errors in Table~\ref{Tapp:heat_lnn_lr} while keeping the terminal loss still within a low regime.
\section{Learning-rate comparison, complete implementation parameters, and rigorous framework-based parameter-sensitivity analysis for the LNN--PINN anisotropic Poisson--beam benchmark}
\label{d}

This appendix supplements the anisotropic Poisson--beam benchmark in Section~3.4 and focuses exclusively on the LNN--PINN implementation. Since the current main text does not assign local labels to the case-specific APBE governing equation, the associated boundary constraints, the residual channels, or the case-level loss decomposition, we write them out explicitly here and then carry out the complete mathematical parameter-sensitivity analysis directly from the uploaded implementation. At the framework level, the present benchmark remains a concrete specialization of the generic composite physics-only objective in Eq.~\eqref{LOSS Total}, and the internal liquid residual replacement still follows the structural principle illustrated in Fig.~\ref{LNN-PINN}. What changes here is only the concrete operator, the concrete constraint family, and the concrete numerical parameter set.

The physical benchmark itself reads
\begin{equation}
u_{xx}-u_{yyyy}=f(x,y),
\qquad
f(x,y)=(2-x^2)e^{-y},
\qquad
(x,y)\in[0,1]^2,
\label{eq:apbeapp_pde}
\end{equation}
with the exact reference field
\begin{equation}
u^\ast(x,y)=x^2e^{-y}.
\label{eq:apbeapp_exact}
\end{equation}
The boundary and auxiliary derivative constraints are
\begin{equation}
u(x,0)=x^2,
\qquad
u_{yy}(x,0)=x^2,
\label{eq:apbeapp_bc_down}
\end{equation}
\begin{equation}
u(x,1)=\frac{x^2}{e},
\qquad
u_{yy}(x,1)=\frac{x^2}{e},
\label{eq:apbeapp_bc_up}
\end{equation}
\begin{equation}
u(0,y)=0,
\qquad
u(1,y)=e^{-y}.
\label{eq:apbeapp_bc_lr}
\end{equation}
The uploaded LNN--PINN scripts approximate the field by \(u_\theta(x,y)\), sample one interior set and six boundary-type sets,
\begin{equation}
\mathcal{S}_{\Omega}=\{(x_i,y_i)\}_{i=1}^{N_\Omega},
\qquad
\mathcal{S}_{yy\downarrow}=\{(x_j,0)\}_{j=1}^{N_{yy\downarrow}},
\qquad
\mathcal{S}_{yy\uparrow}=\{(x_j,1)\}_{j=1}^{N_{yy\uparrow}},
\label{eq:apbeapp_sets1}
\end{equation}
\begin{equation}
\mathcal{S}_{\downarrow}=\{(x_j,0)\}_{j=1}^{N_{\downarrow}},
\qquad
\mathcal{S}_{\uparrow}=\{(x_j,1)\}_{j=1}^{N_{\uparrow}},
\qquad
\mathcal{S}_{\mathrm L}=\{(0,y_j)\}_{j=1}^{N_{\mathrm L}},
\qquad
\mathcal{S}_{\mathrm R}=\{(1,y_j)\}_{j=1}^{N_{\mathrm R}},
\label{eq:apbeapp_sets2}
\end{equation}
and the uploaded code fixes
\begin{equation}
N_\Omega=N_{yy\downarrow}=N_{yy\uparrow}=N_{\downarrow}=N_{\uparrow}=N_{\mathrm L}=N_{\mathrm R}=1000.
\label{eq:apbeapp_counts}
\end{equation}
The script redraws all these sets i.i.d. at every training step from the corresponding uniform distributions on the interior or on the associated line manifolds.

Using automatic differentiation, the uploaded implementation constructs the residual channels
\begin{equation}
\mathbf r_{\mathrm{PDE}}(\theta)
=
\Bigl[
u_{\theta xx}(x,y)-u_{\theta yyyy}(x,y)-(2-x^2)e^{-y}
\Bigr]_{(x,y)\in \mathcal{S}_{\Omega}},
\label{eq:apbeapp_r_pde}
\end{equation}
\begin{equation}
\mathbf r_{yy\downarrow}(\theta)
=
\Bigl[
u_{\theta yy}(x,0)-x^2
\Bigr]_{(x,0)\in \mathcal{S}_{yy\downarrow}},
\qquad
\mathbf r_{yy\uparrow}(\theta)
=
\Bigl[
u_{\theta yy}(x,1)-x^2/e
\Bigr]_{(x,1)\in \mathcal{S}_{yy\uparrow}},
\label{eq:apbeapp_r_yy}
\end{equation}
\begin{equation}
\mathbf r_{\downarrow}(\theta)
=
\Bigl[
u_{\theta}(x,0)-x^2
\Bigr]_{(x,0)\in \mathcal{S}_{\downarrow}},
\qquad
\mathbf r_{\uparrow}(\theta)
=
\Bigl[
u_{\theta}(x,1)-x^2/e
\Bigr]_{(x,1)\in \mathcal{S}_{\uparrow}},
\label{eq:apbeapp_r_du}
\end{equation}
\begin{equation}
\mathbf r_{\mathrm L}(\theta)
=
\Bigl[
u_{\theta}(0,y)-0
\Bigr]_{(0,y)\in \mathcal{S}_{\mathrm L}},
\qquad
\mathbf r_{\mathrm R}(\theta)
=
\Bigl[
u_{\theta}(1,y)-e^{-y}
\Bigr]_{(1,y)\in \mathcal{S}_{\mathrm R}}.
\label{eq:apbeapp_r_lr}
\end{equation}
The corresponding mean-square channels are
\begin{equation}
L_{\mathrm{PDE}}(\theta)=\frac{1}{N_\Omega}\|\mathbf r_{\mathrm{PDE}}(\theta)\|_2^2,
\qquad
L_{yy\downarrow}(\theta)=\frac{1}{N_{yy\downarrow}}\|\mathbf r_{yy\downarrow}(\theta)\|_2^2,
\qquad
L_{yy\uparrow}(\theta)=\frac{1}{N_{yy\uparrow}}\|\mathbf r_{yy\uparrow}(\theta)\|_2^2,
\label{eq:apbeapp_losses1}
\end{equation}
\begin{equation}
L_{\downarrow}(\theta)=\frac{1}{N_{\downarrow}}\|\mathbf r_{\downarrow}(\theta)\|_2^2,
\qquad
L_{\uparrow}(\theta)=\frac{1}{N_{\uparrow}}\|\mathbf r_{\uparrow}(\theta)\|_2^2,
\label{eq:apbeapp_losses2}
\end{equation}
\begin{equation}
L_{\mathrm L}(\theta)=\frac{1}{N_{\mathrm L}}\|\mathbf r_{\mathrm L}(\theta)\|_2^2,
\qquad
L_{\mathrm R}(\theta)=\frac{1}{N_{\mathrm R}}\|\mathbf r_{\mathrm R}(\theta)\|_2^2.
\label{eq:apbeapp_losses3}
\end{equation}
Therefore the uploaded case-specific objective is
\begin{equation}
\mathcal J_{\mathrm{APBE}}(\theta)
=
L_{\mathrm{PDE}}(\theta)
+
L_{yy\downarrow}(\theta)
+
L_{yy\uparrow}(\theta)
+
L_{\downarrow}(\theta)
+
L_{\uparrow}(\theta)
+
L_{\mathrm L}(\theta)
+
L_{\mathrm R}(\theta),
\label{eq:apbeapp_total_loss}
\end{equation}
which is the concrete APBE specialization of Eq.~\eqref{LOSS Total} with all present weights set to unity. The uploaded learning-rate family is
\begin{equation}
\eta\in
\mathcal E_{\eta}^{\mathrm{APBE}}
=
\left\{
3\times10^{-3},\,3\times10^{-4},\,3\times10^{-5},\,3\times10^{-6},\,3\times10^{-7},\,3\times10^{-8}
\right\},
\label{eq:apbeapp_eta_set}
\end{equation}
and the training horizon is
\begin{equation}
N_{\mathrm{train}}=5000.
\label{eq:apbeapp_ntrain}
\end{equation}
The scripts use the fixed reproducibility seed
\begin{equation}
s=888888.
\label{eq:apbeapp_seed}
\end{equation}

The actual uploaded LNN--PINN forward map keeps the same linear topology \(2\to64\to64\to64\to64\to1\) and inserts one width-preserving residual liquid operator after each hidden \(\tanh\). Let
\begin{equation}
\xi=
\begin{bmatrix}
x\\ y
\end{bmatrix}\in\mathbb R^2.
\label{eq:apbeapp_input}
\end{equation}
The first hidden stage reads
\begin{equation}
z_1=\tanh(W_0\xi+b_0),
\qquad
h_1=z_1+\operatorname{Diag}(\alpha_1)\tanh(A_1 z_1+c_1),
\label{eq:apbeapp_h1}
\end{equation}
the next three stages read
\begin{equation}
z_2=\tanh(W_1h_1+b_1),
\qquad
h_2=z_2+\operatorname{Diag}(\alpha_2)\tanh(A_2 z_2+c_2),
\label{eq:apbeapp_h2}
\end{equation}
\begin{equation}
z_3=\tanh(W_2h_2+b_2),
\qquad
h_3=z_3+\operatorname{Diag}(\alpha_3)\tanh(A_3 z_3+c_3),
\label{eq:apbeapp_h3}
\end{equation}
\begin{equation}
z_4=\tanh(W_3h_3+b_3),
\qquad
h_4=z_4+\operatorname{Diag}(\alpha_4)\tanh(A_4 z_4+c_4),
\label{eq:apbeapp_h4}
\end{equation}
and the scalar output is
\begin{equation}
u_\theta(x,y)=w_{\mathrm{out}}^\top h_4+b_{\mathrm{out}}.
\label{eq:apbeapp_output}
\end{equation}
The uploaded numerical values are
\begin{equation}
d=2,
\qquad
o=1,
\qquad
w=64,
\qquad
H=4,
\qquad
q=4,
\qquad
\alpha_0=0.5,
\label{eq:apbeapp_arch_values}
\end{equation}
where \(d\) denotes the input dimension, \(o\) denotes the output dimension, \(w\) denotes the hidden width, \(H\) denotes the number of hidden linear stages, \(q\) denotes the number of liquid residual blocks, and \(\alpha_0\) denotes the common initialization value of every channel in every gate vector,
\begin{equation}
\alpha_\ell^{(0)}=\alpha_0\mathbf 1,
\qquad \ell=1,2,3,4.
\label{eq:apbeapp_alpha_init}
\end{equation}

The exact trainable-parameter count follows from direct counting. The first affine map contributes
\begin{equation}
P_{\mathrm{in}}=dw+w,
\label{eq:apbeapp_Pin}
\end{equation}
the remaining \(H-1\) backbone hidden affine maps contribute
\begin{equation}
P_{\mathrm{hid}}=(H-1)(w^2+w),
\label{eq:apbeapp_Phid}
\end{equation}
the output layer contributes
\begin{equation}
P_{\mathrm{out}}=wo+o,
\label{eq:apbeapp_Pout}
\end{equation}
and each liquid block contributes one square matrix, one bias vector, and one gate vector,
\begin{equation}
P_{\mathrm{liq,one}}=w^2+w+w=w^2+2w.
\label{eq:apbeapp_Pliq_one}
\end{equation}
Therefore
\begin{equation}
P_{\mathrm{LNN}}(d,o,w,H,q)
=
dw+w+(H-1)(w^2+w)+wo+o+q(w^2+2w).
\label{eq:apbeapp_PLNN_general}
\end{equation}
Substituting Eq.~\eqref{eq:apbeapp_arch_values} into Eq.~\eqref{eq:apbeapp_PLNN_general} gives
\begin{align}
P_{\mathrm{LNN}}
&=
2\cdot64+64+3(64^2+64)+64\cdot1+1+4(64^2+2\cdot64)
\nonumber\\
&=
192+12480+65+16896
\nonumber\\
&=
29633.
\label{eq:apbeapp_PLNN}
\end{align}

The optimizer in the uploaded code is Adam without overriding its remaining keyword arguments, so the effective optimizer parameters are
\begin{equation}
\beta_1=0.9,
\qquad
\beta_2=0.999,
\qquad
\varepsilon_{\mathrm A}=10^{-8},
\qquad
\lambda_{\mathrm{wd}}=0,
\qquad
\chi_{\mathrm{AMS}}=0,
\label{eq:apbeapp_adam_defaults}
\end{equation}
where \(\lambda_{\mathrm{wd}}\) denotes weight decay and \(\chi_{\mathrm{AMS}}\in\{0,1\}\) denotes the AMSGrad switch. The logging interval and rendering density are
\begin{equation}
N_{\mathrm{print}}=100,
\qquad
\mathrm{DPI}=300.
\label{eq:apbeapp_io_values}
\end{equation}
The evaluation grid uses
\begin{equation}
h=200,
\qquad
M_h=h^2=40000.
\label{eq:apbeapp_eval_grid}
\end{equation}

For the six uploaded learning-rate runs, the final loss and global field metrics are summarized in Table~\ref{Tapp:apbe_lnn_lr}. The uploaded archive shows a genuine tradeoff: \(\eta=3\times10^{-4}\) gives the smallest terminal training loss, the smallest RMSE, the smallest MSE, and the smallest relative \(L_2\) error, while \(\eta=3\times10^{-3}\) gives the smallest MAE, the smallest maximum absolute error, and the shortest training time. Hence the best learning rate depends on which global criterion one chooses to prioritize.

\begin{table*}[t]
\centering
\scriptsize
\setlength{\tabcolsep}{4.4pt}
\renewcommand{\arraystretch}{1.10}
\caption{{Six-group learning-rate comparison for the uploaded LNN--PINN anisotropic Poisson--beam implementation. All six runs keep the physical operator, the collocation counts, the seed, the architecture, the optimizer family, and the evaluation grid fixed.}}
\label{Tapp:apbe_lnn_lr}
\resizebox{\textwidth}{!}{
\begin{tabular}{cccccccc}
\toprule
Learning rate & $\mathcal J_{\mathrm{APBE}}^{(5000)}$ & RMSE & MSE & MAE & rel$L_2$ & MaxErr & Training time (s) \\
\midrule
$3\times10^{-3}$ & $3.609828490880\times10^{-5}$ & $1.445538510416\times10^{-3}$ & $2.089581585096\times10^{-6}$ & $\mathbf{1.272134599276\times10^{-3}}$ & $4.893648903817\times10^{-3}$ & $\mathbf{3.821253776550\times10^{-3}}$ & $\mathbf{2.656549364500\times10^{3}}$ \\
$3\times10^{-4}$ & $\mathbf{6.178150943015\times10^{-5}}$ & $\mathbf{1.420782463456\times10^{-3}}$ & $\mathbf{2.018622808464\times10^{-6}}$ & $1.296704984270\times10^{-3}$ & $\mathbf{4.809841047972\times10^{-3}}$ & $3.907740116119\times10^{-3}$ & $3.073750423500\times10^{3}$ \\
$3\times10^{-5}$ & $1.762216852512\times10^{-4}$ & $2.161842635481\times10^{-3}$ & $4.673563580582\times10^{-6}$ & $1.609976170585\times10^{-3}$ & $7.318586576730\times10^{-3}$ & $7.213830947876\times10^{-3}$ & $1.441428230000\times10^{3}$ \\
$3\times10^{-6}$ & $1.672307550907\times10^{-1}$ & $6.994974084562\times10^{-2}$ & $4.892966244370\times10^{-3}$ & $5.911607667804\times10^{-2}$ & $2.368041127920\times10^{-1}$ & $2.959880828857\times10^{-1}$ & $1.590253608900\times10^{3}$ \\
$3\times10^{-7}$ & $1.525544762611\times10^{0}$ & $1.889689935431\times10^{-1}$ & $3.570928052068\times10^{-2}$ & $1.406877785921\times10^{-1}$ & $6.397255659103\times10^{-1}$ & $7.680569887161\times10^{-1}$ & $3.084737435200\times10^{3}$ \\
$3\times10^{-8}$ & $1.834689617157\times10^{0}$ & $2.082826830964\times10^{-1}$ & $4.338167607784\times10^{-2}$ & $1.516885161400\times10^{-1}$ & $7.051090598106\times10^{-1}$ & $8.353476524353\times10^{-1}$ & $3.088104907000\times10^{3}$ \\
\bottomrule
\end{tabular}}
\end{table*}

The terminal loss in Table~\ref{Tapp:apbe_lnn_lr} denotes
\begin{equation}
\mathcal J_{\mathrm{APBE}}^{(5000)}=\mathcal J_{\mathrm{APBE}}(\theta_{5000}).
\label{eq:apbeapp_J5000}
\end{equation}
The reporting metrics are computed on the uniform \(h\times h\) evaluation grid. Writing the grid nodes as \(\{(x_m,y_m)\}_{m=1}^{M_h}\), we define
\begin{equation}
e_m(\theta)=u_\theta(x_m,y_m)-u^\ast(x_m,y_m),
\qquad m=1,\dots,M_h,
\label{eq:apbeapp_em}
\end{equation}
and then
\begin{equation}
\mathrm{MSE}_h(\theta)=\frac{1}{M_h}\sum_{m=1}^{M_h}e_m(\theta)^2,
\qquad
\mathrm{RMSE}_h(\theta)=\left(\mathrm{MSE}_h(\theta)\right)^{1/2},
\label{eq:apbeapp_metrics1}
\end{equation}
\begin{equation}
\mathrm{MAE}_h(\theta)=\frac{1}{M_h}\sum_{m=1}^{M_h}|e_m(\theta)|,
\qquad
\mathrm{MaxErr}_h(\theta)=\max_{1\le m\le M_h}|e_m(\theta)|,
\label{eq:apbeapp_metrics2}
\end{equation}
\begin{equation}
\mathrm{rel}L_{2,h}(\theta)
=
\left(
\frac{\sum_{m=1}^{M_h}e_m(\theta)^2}
{\sum_{m=1}^{M_h}(u^\ast(x_m,y_m))^2}
\right)^{1/2}.
\label{eq:apbeapp_relL2}
\end{equation}
The identity
\begin{align}
\mathrm{MSE}_h(\theta_a)-\mathrm{MSE}_h(\theta_b)
=\;&
\bigl(\mathrm{MSE}_h(\theta_a)-\mathcal J_{\mathrm{APBE}}(\theta_a)\bigr)
+
\bigl(\mathcal J_{\mathrm{APBE}}(\theta_a)-\mathcal J_{\mathrm{APBE}}(\theta_b)\bigr)
\nonumber\\
&+
\bigl(\mathcal J_{\mathrm{APBE}}(\theta_b)-\mathrm{MSE}_h(\theta_b)\bigr)
\label{eq:apbeapp_gap}
\end{align}
shows directly why the optimizer-preferred rate and the grid-reconstruction-preferred rate do not have to coincide: the middle term measures the training-objective ordering, while the outer two terms measure the mismatch between stochastic collocation loss and uniform-grid reconstruction loss.

The complete non-physical implementation parameter list appears in Table~\ref{Tapp:apbe_lnn_params}. I separate mathematically active parameters from mathematically inactive metadata so that the later sensitivity analysis can treat every numerical entry explicitly and without ambiguity.

\begin{table*}[t]
\centering
\scriptsize
\setlength{\tabcolsep}{4.2pt}
\renewcommand{\arraystretch}{1.10}
\caption{{Complete non-physical implementation parameters for the uploaded LNN--PINN anisotropic Poisson--beam scripts.}}
\label{Tapp:apbe_lnn_params}
\resizebox{\textwidth}{!}{
\begin{tabular}{llll}
\toprule
Category & Symbol / item & Value & Mathematical status \\
\midrule
Learning-rate family & $\eta$ & $\{3\times10^{-3},3\times10^{-4},3\times10^{-5},3\times10^{-6},3\times10^{-7},3\times10^{-8}\}$ & active \\
Training horizon & $N_{\mathrm{train}}$ & $5000$ & active \\
Interior sample count & $N_{\Omega}$ & $1000$ & active \\
Bottom second-derivative count & $N_{yy\downarrow}$ & $1000$ & active \\
Top second-derivative count & $N_{yy\uparrow}$ & $1000$ & active \\
Bottom Dirichlet count & $N_{\downarrow}$ & $1000$ & active \\
Top Dirichlet count & $N_{\uparrow}$ & $1000$ & active \\
Left Dirichlet count & $N_{\mathrm L}$ & $1000$ & active \\
Right Dirichlet count & $N_{\mathrm R}$ & $1000$ & active \\
Seed & $s$ & $888888$ & active through realization choice \\
Optimizer & Adam family & yes & active \\
First moment parameter & $\beta_1$ & $0.9$ & active \\
Second moment parameter & $\beta_2$ & $0.999$ & active \\
Adam denominator shift & $\varepsilon_{\mathrm A}$ & $10^{-8}$ & active \\
Weight decay & $\lambda_{\mathrm{wd}}$ & $0$ & active \\
AMSGrad switch & $\chi_{\mathrm{AMS}}$ & $0$ & active as discrete flag \\
Input dimension & $d$ & $2$ & active \\
Output dimension & $o$ & $1$ & active \\
Hidden width & $w$ & $64$ & active \\
Backbone hidden-stage count & $H$ & $4$ & active \\
Liquid-block count & $q$ & $4$ & active \\
Gate initialization & $\alpha_0$ & $0.5$ & active \\
Evaluation-grid side length & $h$ & $200$ & active for reported metrics \\
Evaluation-point count & $M_h$ & $40000$ & derived active quantity \\
Parameter count & $P_{\mathrm{LNN}}$ & $29633$ & derived active quantity \\
Logging interval & $N_{\mathrm{print}}$ & $100$ & mathematically inactive for training and evaluation \\
Raster density & DPI & $300$ & mathematically inactive for training and evaluation \\
Execution metadata & device selection, CUDA warmup, output naming & present & inactive metadata \\
\bottomrule
\end{tabular}}
\end{table*}

We now carry out the rigorous parameter-sensitivity analysis from the actual LNN--PINN framework in Eqs.~\eqref{eq:apbeapp_h1}--\eqref{eq:apbeapp_output} and from the actual APBE objective in Eq.~\eqref{eq:apbeapp_total_loss}. Let
\begin{equation}
g_k=\nabla_\theta \mathcal J_{\mathrm{APBE}}(\theta_k;\Xi_k),
\label{eq:apbeapp_gk}
\end{equation}
where \(\Xi_k\) denotes the entire random collocation realization at training step \(k\). The generalized Adam family with explicit weight decay and AMSGrad switch reads
\begin{equation}
\widetilde g_k=g_k+\lambda_{\mathrm{wd}}\theta_k,
\label{eq:apbeapp_gtilde}
\end{equation}
\begin{equation}
m_k=\beta_1m_{k-1}+(1-\beta_1)\widetilde g_k,
\qquad
v_k=\beta_2v_{k-1}+(1-\beta_2)(\widetilde g_k\odot \widetilde g_k),
\label{eq:apbeapp_mkvk}
\end{equation}
\begin{equation}
\widehat m_k=\frac{m_k}{1-\beta_1^k},
\qquad
\widehat v_k=\frac{v_k}{1-\beta_2^k},
\label{eq:apbeapp_hatmom}
\end{equation}
\begin{equation}
\overline v_k(\chi_{\mathrm{AMS}})
=
(1-\chi_{\mathrm{AMS}})\widehat v_k
+
\chi_{\mathrm{AMS}}\max\{\overline v_{k-1},\widehat v_k\},
\label{eq:apbeapp_vbar}
\end{equation}
\begin{equation}
\theta_{k+1}
=
\theta_k
-
\eta
\frac{\widehat m_k}{\sqrt{\overline v_k(\chi_{\mathrm{AMS}})}+\varepsilon_{\mathrm A}}.
\label{eq:apbeapp_adam}
\end{equation}
Define the optimizer direction
\begin{equation}
d_k(\eta,\beta_1,\beta_2,\varepsilon_{\mathrm A},\lambda_{\mathrm{wd}},\chi_{\mathrm{AMS}})
=
\frac{\widehat m_k}{\sqrt{\overline v_k(\chi_{\mathrm{AMS}})}+\varepsilon_{\mathrm A}}.
\label{eq:apbeapp_dk}
\end{equation}
Then Eq.~\eqref{eq:apbeapp_adam} becomes
\begin{equation}
\theta_{k+1}=\theta_k-\eta d_k.
\label{eq:apbeapp_adam_compact}
\end{equation}

The sensitivity with respect to the learning rate follows immediately. Freezing \((\theta_k,m_k,v_k,\Xi_k)\) and perturbing \(\eta\mapsto \eta+\delta\eta\), Eq.~\eqref{eq:apbeapp_adam_compact} yields
\begin{equation}
\theta_{k+1}(\eta+\delta\eta)-\theta_{k+1}(\eta)
=
-\delta\eta\, d_k,
\label{eq:apbeapp_eta_increment}
\end{equation}
hence
\begin{equation}
\left.\frac{\partial \theta_{k+1}}{\partial\eta}\right|_{(\theta_k,m_k,v_k,\Xi_k)}
=
-d_k,
\qquad
\|\delta\theta_{k+1}\|_2\le |\delta\eta|\,\|d_k\|_2.
\label{eq:apbeapp_eta_derivative}
\end{equation}
A second-order expansion of \(\mathcal J_{\mathrm{APBE}}\) along the actual update direction gives
\begin{equation}
\mathcal J_{\mathrm{APBE}}(\theta_{k+1};\Xi_k)
=
\mathcal J_{\mathrm{APBE}}(\theta_k;\Xi_k)
-
\eta\, g_k^\top d_k
+
\frac{\eta^2}{2}
d_k^\top
H_k(\theta_k-\tau_k\eta d_k)
d_k,
\label{eq:apbeapp_taylor_eta}
\end{equation}
where
\begin{equation}
H_k=\nabla_\theta^2\mathcal J_{\mathrm{APBE}}(\theta_k;\Xi_k),
\qquad
\tau_k\in(0,1).
\label{eq:apbeapp_Hk}
\end{equation}
If
\begin{equation}
g_k^\top d_k\ge c_k\|g_k\|_2\|d_k\|_2,
\qquad
\|H_k(\theta_k-\tau_k\eta d_k)\|_2\le M_k,
\label{eq:apbeapp_ckMk}
\end{equation}
then Eq.~\eqref{eq:apbeapp_taylor_eta} yields the sufficient descent range
\begin{equation}
0<\eta<\frac{2c_k\|g_k\|_2}{M_k\|d_k\|_2}.
\label{eq:apbeapp_eta_range}
\end{equation}
Eq.~\eqref{eq:apbeapp_eta_range} gives the precise framework-based meaning of a reasonable learning-rate regime: if \(\eta\) falls below the useful scale, the linear decrease term becomes too weak; if \(\eta\) exceeds the local curvature-controlled threshold, the quadratic term dominates.

The training-horizon sensitivity follows by iterating Eq.~\eqref{eq:apbeapp_adam_compact}:
\begin{equation}
\theta_K
=
\theta_0-\eta\sum_{k=0}^{K-1}d_k.
\label{eq:apbeapp_thetaK}
\end{equation}
Hence enlarging the training horizon from \(K\) to \(K+\Delta K\) gives
\begin{equation}
\theta_{K+\Delta K}-\theta_K
=
-\eta\sum_{k=K}^{K+\Delta K-1}d_k,
\label{eq:apbeapp_epoch_increment}
\end{equation}
and therefore
\begin{equation}
\|\theta_{K+\Delta K}-\theta_K\|_2
\le
\eta\sum_{k=K}^{K+\Delta K-1}\|d_k\|_2.
\label{eq:apbeapp_epoch_bound}
\end{equation}
Eq.~\eqref{eq:apbeapp_epoch_bound} shows that \(N_{\mathrm{train}}\) enters through the accumulated tail of the Adam directions; the extra optimization effect vanishes only when that tail norm becomes negligible.

The moment parameters \(\beta_1\) and \(\beta_2\) also admit exact derivatives. Unrolling Eq.~\eqref{eq:apbeapp_mkvk} gives
\begin{equation}
m_k=(1-\beta_1)\sum_{j=1}^{k}\beta_1^{k-j}\widetilde g_j,
\qquad
v_k=(1-\beta_2)\sum_{j=1}^{k}\beta_2^{k-j}(\widetilde g_j\odot\widetilde g_j).
\label{eq:apbeapp_unroll}
\end{equation}
Differentiating Eq.~\eqref{eq:apbeapp_unroll} with respect to \(\beta_1\) yields
\begin{equation}
\frac{\partial m_k}{\partial\beta_1}
=
-\sum_{j=1}^{k}\beta_1^{k-j}\widetilde g_j
+
(1-\beta_1)\sum_{j=1}^{k}(k-j)\beta_1^{k-j-1}\widetilde g_j,
\label{eq:apbeapp_dm_db1}
\end{equation}
and therefore
\begin{equation}
\frac{\partial \widehat m_k}{\partial\beta_1}
=
\frac{(1-\beta_1^k)\frac{\partial m_k}{\partial\beta_1}+k\beta_1^{k-1}m_k}{(1-\beta_1^k)^2}.
\label{eq:apbeapp_dhatm_db1}
\end{equation}
Likewise,
\begin{equation}
\frac{\partial v_k}{\partial\beta_2}
=
-\sum_{j=1}^{k}\beta_2^{k-j}(\widetilde g_j\odot\widetilde g_j)
+
(1-\beta_2)\sum_{j=1}^{k}(k-j)\beta_2^{k-j-1}(\widetilde g_j\odot\widetilde g_j),
\label{eq:apbeapp_dv_db2}
\end{equation}
hence
\begin{equation}
\frac{\partial \widehat v_k}{\partial\beta_2}
=
\frac{(1-\beta_2^k)\frac{\partial v_k}{\partial\beta_2}+k\beta_2^{k-1}v_k}{(1-\beta_2^k)^2}.
\label{eq:apbeapp_dhatv_db2}
\end{equation}
These identities show that \(\beta_1\) and \(\beta_2\) change the optimizer not through a trivial scalar factor, but through the entire geometric memory kernel that weights past gradients.

The denominator-shift sensitivity follows directly from Eq.~\eqref{eq:apbeapp_dk}:
\begin{equation}
\frac{\partial d_k}{\partial \varepsilon_{\mathrm A}}
=
-\frac{\widehat m_k}{\left(\sqrt{\overline v_k(\chi_{\mathrm{AMS}})}+\varepsilon_{\mathrm A}\right)^2},
\label{eq:apbeapp_deps}
\end{equation}
hence
\begin{equation}
\frac{\partial \theta_{k+1}}{\partial \varepsilon_{\mathrm A}}
=
\eta
\frac{\widehat m_k}{\left(\sqrt{\overline v_k(\chi_{\mathrm{AMS}})}+\varepsilon_{\mathrm A}\right)^2}.
\label{eq:apbeapp_theta_eps}
\end{equation}
Therefore \(\varepsilon_{\mathrm A}\) regularizes the denominator and suppresses the step magnitude most strongly in coordinates where \(\overline v_k\) is small.

The weight-decay sensitivity enters through \(\widetilde g_k=g_k+\lambda_{\mathrm{wd}}\theta_k\). Differentiating Eq.~\eqref{eq:apbeapp_mkvk} with respect to \(\lambda_{\mathrm{wd}}\) gives
\begin{equation}
\frac{\partial m_k}{\partial\lambda_{\mathrm{wd}}}
=
\beta_1\frac{\partial m_{k-1}}{\partial\lambda_{\mathrm{wd}}}
+
(1-\beta_1)\left(\theta_k+\lambda_{\mathrm{wd}}\frac{\partial\theta_k}{\partial\lambda_{\mathrm{wd}}}\right),
\label{eq:apbeapp_dm_dwd}
\end{equation}
\begin{equation}
\frac{\partial v_k}{\partial\lambda_{\mathrm{wd}}}
=
\beta_2\frac{\partial v_{k-1}}{\partial\lambda_{\mathrm{wd}}}
+
2(1-\beta_2)\widetilde g_k\odot\left(\theta_k+\lambda_{\mathrm{wd}}\frac{\partial\theta_k}{\partial\lambda_{\mathrm{wd}}}\right).
\label{eq:apbeapp_dv_dwd}
\end{equation}
At the uploaded value \(\lambda_{\mathrm{wd}}=0\), Eqs.~\eqref{eq:apbeapp_dm_dwd} and \eqref{eq:apbeapp_dv_dwd} simplify to
\begin{equation}
\left.\frac{\partial m_k}{\partial\lambda_{\mathrm{wd}}}\right|_{\lambda_{\mathrm{wd}}=0}
=
\beta_1\left.\frac{\partial m_{k-1}}{\partial\lambda_{\mathrm{wd}}}\right|_{\lambda_{\mathrm{wd}}=0}
+
(1-\beta_1)\theta_k,
\label{eq:apbeapp_dm_dwd0}
\end{equation}
\begin{equation}
\left.\frac{\partial v_k}{\partial\lambda_{\mathrm{wd}}}\right|_{\lambda_{\mathrm{wd}}=0}
=
\beta_2\left.\frac{\partial v_{k-1}}{\partial\lambda_{\mathrm{wd}}}\right|_{\lambda_{\mathrm{wd}}=0}
+
2(1-\beta_2)g_k\odot\theta_k.
\label{eq:apbeapp_dv_dwd0}
\end{equation}
Thus even though the uploaded run uses zero weight decay, the local sensitivity with respect to turning on a small weight decay is nonzero and depends directly on the parameter vector itself.

The AMSGrad switch \(\chi_{\mathrm{AMS}}\) is discrete rather than continuous. Therefore the mathematically correct sensitivity notion is a jump functional, not an ordinary derivative. From Eq.~\eqref{eq:apbeapp_vbar},
\begin{equation}
\Delta_{\mathrm{AMS}}\theta_{k+1}
=
\theta_{k+1}\big|_{\chi_{\mathrm{AMS}}=1}
-
\theta_{k+1}\big|_{\chi_{\mathrm{AMS}}=0}
=
-\eta\,\widehat m_k
\left[
\frac{1}{\sqrt{\max\{\overline v_{k-1},\widehat v_k\}}+\varepsilon_{\mathrm A}}
-
\frac{1}{\sqrt{\widehat v_k}+\varepsilon_{\mathrm A}}
\right].
\label{eq:apbeapp_ams_jump}
\end{equation}
Eq.~\eqref{eq:apbeapp_ams_jump} gives the exact discrete sensitivity of the iterate with respect to the AMSGrad flag.

The collocation-count sensitivities act through stochastic variance. Define the squared residual random variables
\begin{equation}
X_{\Omega}=|u_{\theta xx}(x,y)-u_{\theta yyyy}(x,y)-(2-x^2)e^{-y}|^2,
\qquad
(x,y)\sim \mathrm{Unif}([0,1]^2),
\label{eq:apbeapp_Xomega}
\end{equation}
\begin{equation}
X_{yy\downarrow}=|u_{\theta yy}(x,0)-x^2|^2,
\qquad
X_{yy\uparrow}=|u_{\theta yy}(x,1)-x^2/e|^2,
\qquad
x\sim \mathrm{Unif}([0,1]),
\label{eq:apbeapp_Xyy}
\end{equation}
\begin{equation}
X_{\downarrow}=|u_\theta(x,0)-x^2|^2,
\qquad
X_{\uparrow}=|u_\theta(x,1)-x^2/e|^2,
\qquad
x\sim \mathrm{Unif}([0,1]),
\label{eq:apbeapp_Xdu}
\end{equation}
\begin{equation}
X_{\mathrm L}=|u_\theta(0,y)|^2,
\qquad
X_{\mathrm R}=|u_\theta(1,y)-e^{-y}|^2,
\qquad
y\sim \mathrm{Unif}([0,1]).
\label{eq:apbeapp_Xlr}
\end{equation}
Write their variances as
\begin{equation}
\sigma_{\Omega}^2,
\quad
\sigma_{yy\downarrow}^2,
\quad
\sigma_{yy\uparrow}^2,
\quad
\sigma_{\downarrow}^2,
\quad
\sigma_{\uparrow}^2,
\quad
\sigma_{\mathrm L}^2,
\quad
\sigma_{\mathrm R}^2.
\label{eq:apbeapp_sigmas}
\end{equation}
Because each channel in Eqs.~\eqref{eq:apbeapp_losses1}--\eqref{eq:apbeapp_losses3} is a sample mean, one has
\begin{equation}
\operatorname{Var}(L_{\mathrm{PDE}})=\frac{\sigma_{\Omega}^2}{N_\Omega},
\qquad
\operatorname{Var}(L_{yy\downarrow})=\frac{\sigma_{yy\downarrow}^2}{N_{yy\downarrow}},
\qquad
\operatorname{Var}(L_{yy\uparrow})=\frac{\sigma_{yy\uparrow}^2}{N_{yy\uparrow}},
\label{eq:apbeapp_vars1}
\end{equation}
\begin{equation}
\operatorname{Var}(L_{\downarrow})=\frac{\sigma_{\downarrow}^2}{N_{\downarrow}},
\qquad
\operatorname{Var}(L_{\uparrow})=\frac{\sigma_{\uparrow}^2}{N_{\uparrow}},
\qquad
\operatorname{Var}(L_{\mathrm L})=\frac{\sigma_{\mathrm L}^2}{N_{\mathrm L}},
\qquad
\operatorname{Var}(L_{\mathrm R})=\frac{\sigma_{\mathrm R}^2}{N_{\mathrm R}}.
\label{eq:apbeapp_vars2}
\end{equation}
Therefore
\begin{equation}
\operatorname{Var}(\mathcal J_{\mathrm{APBE}})
=
\frac{\sigma_{\Omega}^2}{N_\Omega}
+
\frac{\sigma_{yy\downarrow}^2}{N_{yy\downarrow}}
+
\frac{\sigma_{yy\uparrow}^2}{N_{yy\uparrow}}
+
\frac{\sigma_{\downarrow}^2}{N_{\downarrow}}
+
\frac{\sigma_{\uparrow}^2}{N_{\uparrow}}
+
\frac{\sigma_{\mathrm L}^2}{N_{\mathrm L}}
+
\frac{\sigma_{\mathrm R}^2}{N_{\mathrm R}}.
\label{eq:apbeapp_var_total}
\end{equation}
Differentiating Eq.~\eqref{eq:apbeapp_var_total} one by one gives
\begin{equation}
\frac{\partial}{\partial N_\Omega}\operatorname{Var}(\mathcal J_{\mathrm{APBE}})
=
-\frac{\sigma_\Omega^2}{N_\Omega^2},
\qquad
\frac{\partial}{\partial N_{yy\downarrow}}\operatorname{Var}(\mathcal J_{\mathrm{APBE}})
=
-\frac{\sigma_{yy\downarrow}^2}{N_{yy\downarrow}^2},
\qquad
\frac{\partial}{\partial N_{yy\uparrow}}\operatorname{Var}(\mathcal J_{\mathrm{APBE}})
=
-\frac{\sigma_{yy\uparrow}^2}{N_{yy\uparrow}^2},
\label{eq:apbeapp_dvars1}
\end{equation}
\begin{equation}
\frac{\partial}{\partial N_{\downarrow}}\operatorname{Var}(\mathcal J_{\mathrm{APBE}})
=
-\frac{\sigma_{\downarrow}^2}{N_{\downarrow}^2},
\qquad
\frac{\partial}{\partial N_{\uparrow}}\operatorname{Var}(\mathcal J_{\mathrm{APBE}})
=
-\frac{\sigma_{\uparrow}^2}{N_{\uparrow}^2},
\label{eq:apbeapp_dvars2}
\end{equation}
\begin{equation}
\frac{\partial}{\partial N_{\mathrm L}}\operatorname{Var}(\mathcal J_{\mathrm{APBE}})
=
-\frac{\sigma_{\mathrm L}^2}{N_{\mathrm L}^2},
\qquad
\frac{\partial}{\partial N_{\mathrm R}}\operatorname{Var}(\mathcal J_{\mathrm{APBE}})
=
-\frac{\sigma_{\mathrm R}^2}{N_{\mathrm R}^2}.
\label{eq:apbeapp_dvars3}
\end{equation}
These formulas already provide a complete and non-omitted sensitivity analysis for each of the seven collocation counts separately. If one prescribes variance tolerances \(\delta_\Omega,\delta_{yy\downarrow},\delta_{yy\uparrow},\delta_{\downarrow},\delta_{\uparrow},\delta_{\mathrm L},\delta_{\mathrm R}\), then sufficient admissible sampling ranges are
\begin{equation}
N_\Omega\ge \frac{\sigma_{\Omega}^2}{\delta_\Omega^2},
\quad
N_{yy\downarrow}\ge \frac{\sigma_{yy\downarrow}^2}{\delta_{yy\downarrow}^2},
\quad
N_{yy\uparrow}\ge \frac{\sigma_{yy\uparrow}^2}{\delta_{yy\uparrow}^2},
\quad
N_{\downarrow}\ge \frac{\sigma_{\downarrow}^2}{\delta_{\downarrow}^2},
\quad
N_{\uparrow}\ge \frac{\sigma_{\uparrow}^2}{\delta_{\uparrow}^2},
\quad
N_{\mathrm L}\ge \frac{\sigma_{\mathrm L}^2}{\delta_{\mathrm L}^2},
\quad
N_{\mathrm R}\ge \frac{\sigma_{\mathrm R}^2}{\delta_{\mathrm R}^2}.
\label{eq:apbeapp_sampling_ranges}
\end{equation}

The seed \(s\) is discrete rather than differentiable. Therefore the correct sensitivity notion is the realization gap
\begin{equation}
\Delta_{s_1,s_2}\mathcal J_{\mathrm{APBE}}(\theta_k)
=
\mathcal J_{\mathrm{APBE}}(\theta_k;\Xi_k^{(s_1)})
-
\mathcal J_{\mathrm{APBE}}(\theta_k;\Xi_k^{(s_2)}),
\label{eq:apbeapp_seed_gap}
\end{equation}
where \(\Xi_k^{(s)}\) denotes the sample stream generated by seed \(s\). Since all six uploaded runs keep \(s=888888\) fixed, the learning-rate comparison in Table~\ref{Tapp:apbe_lnn_lr} isolates step-size effects rather than mixing them with seed variability.

The framework-based architecture sensitivities now follow from the actual liquid residual block. For one block
\begin{equation}
\mathcal B_\ell(z)=z+\operatorname{Diag}(\alpha_\ell)\tanh(A_\ell z+c_\ell),
\label{eq:apbeapp_block}
\end{equation}
the Jacobian with respect to its input is
\begin{equation}
D\mathcal B_\ell(z)
=
I+\operatorname{Diag}(\alpha_\ell)\operatorname{Diag}\!\bigl(\operatorname{sech}^2(A_\ell z+c_\ell)\bigr)A_\ell.
\label{eq:apbeapp_DB}
\end{equation}
Since
\begin{equation}
0<\operatorname{sech}^2(s)\le 1
\qquad
\text{for all }s\in\mathbb R,
\label{eq:apbeapp_sech}
\end{equation}
Eq.~\eqref{eq:apbeapp_DB} yields the bound
\begin{equation}
\|D\mathcal B_\ell(z)\|_2
\le
1+\|\operatorname{Diag}(\alpha_\ell)\|_2\,\|A_\ell\|_2.
\label{eq:apbeapp_DB_bound}
\end{equation}
At initialization, Eq.~\eqref{eq:apbeapp_alpha_init} gives
\begin{equation}
\|\operatorname{Diag}(\alpha_\ell^{(0)})\|_2=\alpha_0=0.5,
\label{eq:apbeapp_alpha_norm}
\end{equation}
hence
\begin{equation}
\|D\mathcal B_\ell(z)\|_2\le 1+0.5\|A_\ell\|_2.
\label{eq:apbeapp_DB_init}
\end{equation}
If one prescribes a per-block amplification budget \(\kappa_\ell>1\), then Eq.~\eqref{eq:apbeapp_DB_bound} yields the admissible initialization range
\begin{equation}
0\le \alpha_0\le \frac{\kappa_\ell-1}{\|A_\ell\|_2}.
\label{eq:apbeapp_alpha_range_layer}
\end{equation}
Imposing a common budget \(\kappa>1\) on all \(q=4\) liquid blocks gives
\begin{equation}
0\le \alpha_0\le \min_{\ell=1,2,3,4}\frac{\kappa-1}{\|A_\ell\|_2}.
\label{eq:apbeapp_alpha_range_global}
\end{equation}
These inequalities give the rigorous meaning of a reasonable \(\alpha_0\)-range in the actual uploaded framework.

The output sensitivity with respect to \(\alpha_0\) follows from the chain rule. First,
\begin{equation}
\left.
\frac{\partial \mathcal B_\ell(z)}{\partial\alpha_0}
\right|_{\alpha_\ell=\alpha_0\mathbf 1}
=
\tanh(A_\ell z+c_\ell),
\label{eq:apbeapp_dB_dalpha}
\end{equation}
hence
\begin{equation}
\left\|
\left.
\frac{\partial \mathcal B_\ell(z)}{\partial\alpha_0}
\right|_{\alpha_\ell=\alpha_0\mathbf 1}
\right\|_2
\le \sqrt{w}.
\label{eq:apbeapp_dB_dalpha_bound}
\end{equation}
Defining the stage maps
\begin{equation}
\mathcal F_1(\xi)=\mathcal B_1(\tanh(W_0\xi+b_0)),
\quad
\mathcal F_2(h)=\mathcal B_2(\tanh(W_1h+b_1)),
\quad
\mathcal F_3(h)=\mathcal B_3(\tanh(W_2h+b_2)),
\quad
\mathcal F_4(h)=\mathcal B_4(\tanh(W_3h+b_3)),
\label{eq:apbeapp_Fmaps}
\end{equation}
Eq.~\eqref{eq:apbeapp_output} becomes
\begin{equation}
u_\theta(x,y)=w_{\mathrm{out}}^\top \mathcal F_4\circ\mathcal F_3\circ\mathcal F_2\circ\mathcal F_1(\xi)+b_{\mathrm{out}}.
\label{eq:apbeapp_comp_output}
\end{equation}
Differentiating Eq.~\eqref{eq:apbeapp_comp_output} yields
\begin{align}
\frac{\partial u_\theta}{\partial\alpha_0}
=
w_{\mathrm{out}}^\top
\sum_{j=1}^{4}
\left(
\prod_{\ell=j+1}^{4}D\mathcal F_\ell
\right)
\left.
\frac{\partial \mathcal B_j}{\partial\alpha_0}
\right|_{\alpha_j=\alpha_0\mathbf 1}.
\label{eq:apbeapp_du_dalpha}
\end{align}
Using Eq.~\eqref{eq:apbeapp_dB_dalpha_bound} and
\begin{equation}
\|D\mathcal F_\ell\|_2
\le
\bigl(1+\alpha_0\|A_\ell\|_2\bigr)\|W_{\ell-1}\|_2,
\label{eq:apbeapp_DF_bound}
\end{equation}
we obtain
\begin{equation}
\left\|
\frac{\partial u_\theta}{\partial\alpha_0}
\right\|_2
\le
\|w_{\mathrm{out}}\|_2
\sum_{j=1}^{4}
\sqrt{w}
\prod_{\ell=j+1}^{4}
\bigl(1+\alpha_0\|A_\ell\|_2\bigr)\|W_{\ell-1}\|_2.
\label{eq:apbeapp_du_dalpha_bound}
\end{equation}
Thus \(\alpha_0\) enters the output nonlinearly and multiplicatively through the entire cascaded liquid transport.

The discrete sensitivities with respect to \(d,o,w,H,q\) follow exactly from Eq.~\eqref{eq:apbeapp_PLNN_general}. Increasing the input dimension by one while fixing \((o,w,H,q)\) gives
\begin{equation}
P_{\mathrm{LNN}}(d+1,o,w,H,q)-P_{\mathrm{LNN}}(d,o,w,H,q)=w.
\label{eq:apbeapp_ddim}
\end{equation}
At \(w=64\), this increment equals
\begin{equation}
64.
\label{eq:apbeapp_ddim_actual}
\end{equation}
Increasing the output dimension by one while fixing \((d,w,H,q)\) gives
\begin{equation}
P_{\mathrm{LNN}}(d,o+1,w,H,q)-P_{\mathrm{LNN}}(d,o,w,H,q)=w+1,
\label{eq:apbeapp_do}
\end{equation}
which equals 65
at \(w=64\). Increasing the hidden-stage number by one while fixing \((d,o,w,q)\) gives
\begin{equation}
P_{\mathrm{LNN}}(d,o,w,H+1,q)-P_{\mathrm{LNN}}(d,o,w,H,q)=w^2+w,
\label{eq:apbeapp_dH}
\end{equation}
which becomes
\begin{equation}
64^2+64=4160.
\label{eq:apbeapp_dH_actual}
\end{equation}
Increasing the liquid-block count by one while fixing \((d,o,w,H)\) gives
\begin{equation}
P_{\mathrm{LNN}}(d,o,w,H,q+1)-P_{\mathrm{LNN}}(d,o,w,H,q)=w^2+2w,
\label{eq:apbeapp_dq}
\end{equation}
which becomes
\begin{equation}
64^2+2\cdot64=4224.
\label{eq:apbeapp_dq_actual}
\end{equation}
Increasing the hidden width by one while fixing \((d,o,H,q)\) gives
\begin{align}
&P_{\mathrm{LNN}}(d,o,w+1,H,q)-P_{\mathrm{LNN}}(d,o,w,H,q)
\nonumber\\
&=
(H+q-1)\bigl((w+1)^2-w^2\bigr)+(d+H+o+2q),
\label{eq:apbeapp_dw}
\end{align}
hence at the uploaded values
\begin{align}
P_{\mathrm{LNN}}(2,1,65,4,4)-P_{\mathrm{LNN}}(2,1,64,4,4)
&=
7\cdot129+15
\nonumber\\
&=
918.
\label{eq:apbeapp_dw_actual}
\end{align}
These formulas show that width, depth, and liquid-block count all alter the hypothesis class at large algebraic scales rather than through negligible perturbations.

The same parameters also control transport amplification. If
\begin{equation}
\|W_{\ell-1}\|_2\le \rho,
\qquad
\|A_\ell\|_2\le \gamma,
\qquad
\ell=1,2,3,4,
\label{eq:apbeapp_rhogamma}
\end{equation}
then Eq.~\eqref{eq:apbeapp_DF_bound} yields
\begin{equation}
\|D\mathcal F_\ell\|_2\le \rho(1+\alpha_0\gamma),
\qquad \ell=1,2,3,4,
\label{eq:apbeapp_uniform_DF}
\end{equation}
and therefore
\begin{equation}
\operatorname{Lip}\!\left(\mathcal F_4\circ\mathcal F_3\circ\mathcal F_2\circ\mathcal F_1\right)
\le
\bigl(\rho(1+\alpha_0\gamma)\bigr)^4.
\label{eq:apbeapp_transport4}
\end{equation}
For a matched family with \(H=q\), the same reasoning gives
\begin{equation}
\operatorname{Lip}_H\le \bigl(\rho(1+\alpha_0\gamma)\bigr)^H.
\label{eq:apbeapp_transportH}
\end{equation}
If one imposes a transport budget \(\Lambda_\star>1\), then the admissible depth range satisfies
\begin{equation}
H\le \frac{\log \Lambda_\star}{\log(\rho(1+\alpha_0\gamma))}
\qquad
\text{whenever }\rho(1+\alpha_0\gamma)>1.
\label{eq:apbeapp_H_range}
\end{equation}
Eq.~\eqref{eq:apbeapp_H_range} gives the rigorous meaning of a reasonable depth range inside this exact LNN--PINN framework.

The evaluation-grid side length \(h\) and the derived point count \(M_h=h^2\) influence only the reporting functional, not the training dynamics. On \([0,1]^2\), the grid spacings satisfy
\begin{equation}
\Delta x=\Delta y=\frac{1}{h-1}.
\label{eq:apbeapp_dxdy}
\end{equation}
At the uploaded value \(h=200\),
\begin{equation}
\Delta x=\Delta y=\frac{1}{199}.
\label{eq:apbeapp_dxdy_actual}
\end{equation}
Define the continuum squared-error functional
\begin{equation}
\mathcal E_{\mathrm{cont}}(\theta)
=
\int_0^1\int_0^1
|u_\theta(x,y)-u^\ast(x,y)|^2\,dx\,dy.
\label{eq:apbeapp_Econt}
\end{equation}
Then \(\mathrm{MSE}_h(\theta)\) in Eq.~\eqref{eq:apbeapp_metrics1} is a uniform-grid Riemann approximation of \(\mathcal E_{\mathrm{cont}}(\theta)\). For smooth \(u_\theta\), one has
\begin{equation}
\mathrm{MSE}_h(\theta)=\mathcal E_{\mathrm{cont}}(\theta)+\mathcal O(h^{-1}).
\label{eq:apbeapp_riemann}
\end{equation}
Thus the rigorous \(h\)-sensitivity enters through quadrature bias. If one prescribes a reporting bias budget \(\varepsilon_h\), then a sufficient admissible grid range satisfies
\begin{equation}
h\ge 1+\frac{C_q}{\varepsilon_h}
\label{eq:apbeapp_h_range}
\end{equation}
for a problem-dependent constant \(C_q\). Since
\begin{equation}
M_h=(h)^2,
\label{eq:apbeapp_Mh}
\end{equation}
the exact discrete increment of the evaluation-point count is
\begin{equation}
M_{h+1}-M_h=(h+1)^2-h^2=2h+1,
\label{eq:apbeapp_Mh_increment}
\end{equation}
which equals 401
at \(h=200\).

The remaining numeric parameters \(N_{\mathrm{print}}\) and \(\mathrm{DPI}\) do not enter either Eq.~\eqref{eq:apbeapp_total_loss} or Eqs.~\eqref{eq:apbeapp_metrics1}--\eqref{eq:apbeapp_relL2}. Therefore their mathematical sensitivities with respect to the training objective and to the reported global errors vanish:
\begin{equation}
\frac{\partial \mathcal J_{\mathrm{APBE}}}{\partial N_{\mathrm{print}}}=0,
\qquad
\frac{\partial \mathrm{MSE}_h}{\partial N_{\mathrm{print}}}=0,
\label{eq:apbeapp_print_zero}
\end{equation}
\begin{equation}
\frac{\partial \mathcal J_{\mathrm{APBE}}}{\partial \mathrm{DPI}}=0,
\qquad
\frac{\partial \mathrm{MSE}_h}{\partial \mathrm{DPI}}=0.
\label{eq:apbeapp_dpi_zero}
\end{equation}
If one nevertheless defines a logging observability functional by retaining one training snapshot every \(N_{\mathrm{print}}\) iterations, then the number of retained snapshots equals
\begin{equation}
M_{\mathrm{obs}}=\left\lfloor \frac{N_{\mathrm{train}}}{N_{\mathrm{print}}}\right\rfloor,
\label{eq:apbeapp_Mobs}
\end{equation}
so the exact discrete logging sensitivity is
\begin{equation}
M_{\mathrm{obs}}(N_{\mathrm{print}}+1)-M_{\mathrm{obs}}(N_{\mathrm{print}})
=
\left\lfloor \frac{N_{\mathrm{train}}}{N_{\mathrm{print}}+1}\right\rfloor
-
\left\lfloor \frac{N_{\mathrm{train}}}{N_{\mathrm{print}}}\right\rfloor.
\label{eq:apbeapp_Mobs_diff}
\end{equation}
Thus \(N_{\mathrm{print}}\) affects only what one stores and visualizes, whereas \(\mathrm{DPI}\) affects only rasterization density.

The unique summary figure for this appendix appears in Fig.~\ref{Fapp:apbe_lnn_lr}. It visualizes the six loss histories together with the six reconstructed fields and their absolute-error maps under the six learning rates in Eq.~\eqref{eq:apbeapp_eta_set}. The figure complements Table~\ref{Tapp:apbe_lnn_lr} by showing the full optimization-path variation and the associated field-level reconstruction variation under an otherwise fixed LNN--PINN architecture.

\begin{figure*}[t]
\centering
\includegraphics[width=\textwidth]{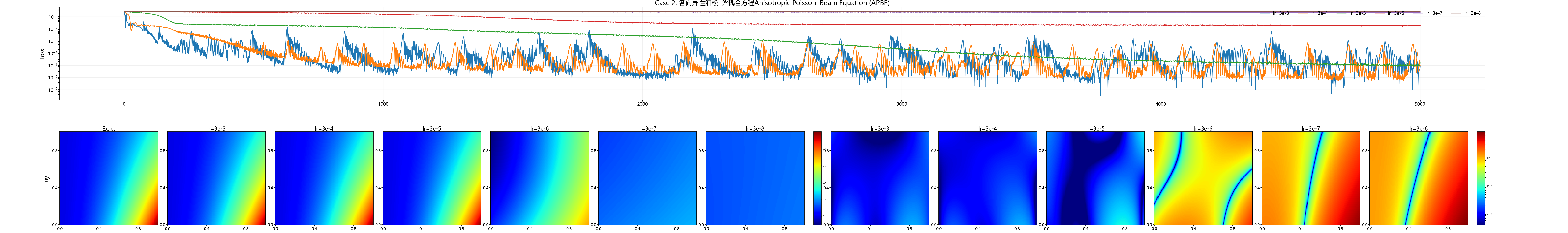}
\caption{{Learning-rate comparison for the LNN--PINN anisotropic Poisson--beam benchmark. The upper panel reports the full loss histories under the six learning rates in Eq.~\eqref{eq:apbeapp_eta_set}. The lower panel reports the reference field in Eq.~\eqref{eq:apbeapp_exact}, the corresponding six LNN--PINN reconstructions, and the associated absolute-error distributions. All six runs keep the APBE operator in Eq.~\eqref{eq:apbeapp_pde}, the constraint family in Eqs.~\eqref{eq:apbeapp_bc_down}--\eqref{eq:apbeapp_bc_lr}, the objective in Eq.~\eqref{eq:apbeapp_total_loss}, the architecture in Eqs.~\eqref{eq:apbeapp_h1}--\eqref{eq:apbeapp_output}, the collocation counts in Eq.~\eqref{eq:apbeapp_counts}, and the evaluation grid in Eq.~\eqref{eq:apbeapp_eval_grid} fixed.}}
\label{Fapp:apbe_lnn_lr}
\end{figure*}

Taken together, Eqs.~\eqref{eq:apbeapp_eta_range}, \eqref{eq:apbeapp_sampling_ranges}, \eqref{eq:apbeapp_alpha_range_global}, \eqref{eq:apbeapp_H_range}, and \eqref{eq:apbeapp_h_range} give the rigorous framework-based meaning of a reasonable implementation regime for the uploaded LNN--PINN anisotropic Poisson--beam scripts. The step size must lie inside the local Adam descent range controlled by curvature and gradient-direction alignment; the seven collocation counts must each exceed the threshold set by their own residual-variance budget; the gate initialization must remain below the block-amplification threshold imposed by the liquid matrices; the hidden depth must remain inside the transport-amplification budget induced by repeated liquid residual mappings; and the evaluation grid must be fine enough to keep quadrature bias below the reporting tolerance. Within the six uploaded runs, \(\eta=3\times10^{-4}\) gives the strongest global performance in RMSE, MSE, relative \(L_2\), and terminal loss, while \(\eta=3\times10^{-3}\) gives the best MAE, the best maximum error, and the shortest runtime. This appendix therefore shows not only which numerical choices work well for the APBE benchmark, but also why each non-physical implementation parameter enters the LNN--PINN framework in the precise mathematical manner derived above.
\section{Learning-rate comparison, complete implementation parameters, and rigorous framework-based parameter-sensitivity analysis for the LNN--PINN relativistic viscous-fluid benchmark}
\label{e}

This appendix supplements Section~3.5 and focuses exclusively on the uploaded LNN--PINN implementation for the first steady relativistic viscous-fluid manufactured benchmark. The present appendix uses only valid external references, namely the generic physics-only composite objective in Eq.~\eqref{LOSS Total} and the framework figure in Fig.~\ref{LNN-PINN}. The case-specific eight-field relativistic system and its exact benchmark fields are not assigned separate local labels in the current main text, so we write the required case-level definitions explicitly here instead of invoking nonexistent references. The unknown principal fields are
\begin{equation}
(v_x,v_y,\epsilon,n,\Pi,\pi_{xx},\pi_{xy},\pi_{yy}),
\label{eq:rel2app_fields}
\end{equation}
and the uploaded scripts define the exact benchmark fields by
\begin{equation}
 v_x^\ast(x,y)=A_{v_x}\sin(\pi x)\sin(\pi y)\bigl(1+0.25\cos(\pi x)\bigr),
\label{eq:rel2app_vx}
\end{equation}
\begin{equation}
 v_y^\ast(x,y)=A_{v_y}\sin(\pi x)\sin(\pi y)\bigl(1-0.25\cos(\pi y)\bigr),
\label{eq:rel2app_vy}
\end{equation}
\begin{equation}
 \epsilon^\ast(x,y)=E_0+A_{\epsilon 1}\cos(2\pi x)\sin(\pi y)+A_{\epsilon 2}\sin(\pi x)\cos(2\pi y),
\label{eq:rel2app_eps}
\end{equation}
\begin{equation}
 n^\ast(x,y)=N_0+A_{n1}\sin(2\pi x)\sin(\pi y)+A_{n2}\cos(\pi x)\sin(2\pi y),
\label{eq:rel2app_n}
\end{equation}
\begin{equation}
 \Pi^\ast(x,y)=A_{\Pi}\Bigl(\sin(\pi x)\cos(2\pi y)+0.30\cos(2\pi x)\sin(\pi y)\Bigr),
\label{eq:rel2app_Pi}
\end{equation}
\begin{equation}
 \pi_{xx}^\ast(x,y)=A_{xx}\Bigl(\cos(2\pi x)\sin(\pi y)+0.25\sin(\pi x)\sin(2\pi y)\Bigr),
\label{eq:rel2app_pixx}
\end{equation}
\begin{equation}
 \pi_{xy}^\ast(x,y)=A_{xy}\Bigl(\sin(2\pi x)\sin(2\pi y)+0.20\cos(\pi x)\sin(\pi y)\Bigr),
\label{eq:rel2app_pixy}
\end{equation}
\begin{equation}
 \pi_{yy}^\ast(x,y)=A_{yy}\Bigl(\sin(\pi x)\cos(2\pi y)-0.20\cos(2\pi x)\sin(\pi y)\Bigr).
\label{eq:rel2app_piyy}
\end{equation}
The uploaded model samples one interior set and four edge sets,
\begin{equation}
\mathcal S_\Omega=\{(x_i,y_i)\}_{i=1}^N,
\qquad
\mathcal S_\downarrow=\{(x_j,0)\}_{j=1}^{N_1},
\qquad
\mathcal S_\uparrow=\{(x_j,1)\}_{j=1}^{N_1},
\qquad
\mathcal S_\mathrm{L}=\{(0,y_j)\}_{j=1}^{N_1},
\qquad
\mathcal S_\mathrm{R}=\{(1,y_j)\}_{j=1}^{N_1},
\label{eq:rel2app_sets}
\end{equation}
with
\begin{equation}
N=1000,
\qquad
N_1=1000.
\label{eq:rel2app_counts}
\end{equation}
At every training step the code redraws these collocation sets independently from the corresponding uniform laws, so the loss is a stochastic empirical functional rather than a deterministic continuum functional.

The uploaded code builds the physical state through a transformed neural output. If the raw network output is denoted by
\begin{equation}
r_\theta(x,y)=\bigl(r_1,r_2,r_3,r_4,r_5,r_6,r_7,r_8\bigr),
\label{eq:rel2app_raw}
\end{equation}
then the actual field transform in the uploaded LNN--PINN scripts is
\begin{equation}
v_x=V_{\max}\tanh(r_1),
\qquad
v_y=V_{\max}\tanh(r_2),
\label{eq:rel2app_vtransform}
\end{equation}
\begin{equation}
\epsilon=\epsilon_{\min}+\operatorname{softplus}(r_3),
\qquad
n=n_{\min}+\operatorname{softplus}(r_4),
\label{eq:rel2app_endens}
\end{equation}
\begin{equation}
\Pi=r_5,
\qquad
\pi_{xx}=r_6,
\qquad
\pi_{xy}=r_7,
\qquad
\pi_{yy}=r_8,
\label{eq:rel2app_shearid}
\end{equation}
where the numerical transform constants are listed in Tables~\ref{Tapp:rel2_lnn_params1} and \ref{Tapp:rel2_lnn_params2}. The uploaded architecture itself keeps the linear topology \(2\to64\to64\to64\to64\to8\) and inserts one width-preserving liquid residual block after each hidden activation. Writing \(\xi=[x\;y]^\top\in\mathbb{R}^2\), one has
\begin{equation}
z_1=\tanh(W_0\xi+b_0),
\qquad
h_1=z_1+\operatorname{Diag}(\alpha_1)\tanh(A_1z_1+c_1),
\label{eq:rel2app_h1}
\end{equation}
\begin{equation}
z_2=\tanh(W_1h_1+b_1),
\qquad
h_2=z_2+\operatorname{Diag}(\alpha_2)\tanh(A_2z_2+c_2),
\label{eq:rel2app_h2}
\end{equation}
\begin{equation}
z_3=\tanh(W_2h_2+b_2),
\qquad
h_3=z_3+\operatorname{Diag}(\alpha_3)\tanh(A_3z_3+c_3),
\label{eq:rel2app_h3}
\end{equation}
\begin{equation}
z_4=\tanh(W_3h_3+b_3),
\qquad
h_4=z_4+\operatorname{Diag}(\alpha_4)\tanh(A_4z_4+c_4),
\label{eq:rel2app_h4}
\end{equation}
\begin{equation}
r_\theta(x,y)=W_{\rm out}h_4+b_{\rm out}.
\label{eq:rel2app_out}
\end{equation}
The numerical architecture values are
\begin{equation}
d=2,
\qquad o=8,
\qquad w=64,
\qquad H=4,
\qquad q=4,
\qquad \alpha_0=0.5,
\label{eq:rel2app_archvals}
\end{equation}
with \(\alpha_\ell^{(0)}=\alpha_0\mathbf 1\) for \(\ell=1,2,3,4\).

\begin{equation}
P_{\rm LNN}(d,o,w,H,q)=dw+w+(H-1)(w^2+w)+wo+o+q(w^2+2w),
\label{eq:rel2app_PLNN_general}
\end{equation}

At the constitutive level the uploaded scripts define
\begin{equation}
\gamma=\frac{1}{\sqrt{1-v_x^2-v_y^2+\delta_{\rm stab}}},
\qquad
p=(\Gamma_{\rm ad}-1)(\epsilon-mn),
\qquad
T=\frac{p}{n+\delta_{\rm stab}},
\label{eq:rel2app_constitutive}
\end{equation}
and then construct the energy--momentum tensor, the number-current, the expansion scalar, and the shear tensor exactly as implemented in the uploaded code. The eight interior PDE residual channels are denoted by
\begin{equation}
\mathcal R_\Omega=(R_E,R_{M_x},R_{M_y},R_n,R_\Pi,R_{xx},R_{xy},R_{yy}),
\label{eq:rel2app_resvec}
\end{equation}
where the first four channels come from the steady conservation laws and the last four channels come from the steady bulk and shear relaxation laws. The exact manufactured sources are generated by inserting the exact fields in Eqs.~\eqref{eq:rel2app_vx}--\eqref{eq:rel2app_piyy} into the same constitutive assembly and differentiating them analytically via automatic differentiation. The uploaded interior loss therefore takes the form
\begin{equation}
L_\Omega(\theta)=\frac{1}{N}\sum_{i=1}^N\sum_{a\in\{E,M_x,M_y,n,\Pi,xx,xy,yy\}}\bigl(R_a(x_i,y_i;\theta)-S_a(x_i,y_i)\bigr)^2,
\label{eq:rel2app_Lint}
\end{equation}
while each edge loss is the summed eight-field boundary mismatch,
\begin{equation}
L_\downarrow(\theta)=\frac{1}{N_1}\sum_{j=1}^{N_1}\sum_{f\in\{v_x,v_y,\epsilon,n,\Pi,\pi_{xx},\pi_{xy},\pi_{yy}\}}\bigl(f_\theta(x_j,0)-f^\ast(x_j,0)\bigr)^2,
\label{eq:rel2app_Lbot}
\end{equation}
\begin{equation}
L_\uparrow(\theta)=\frac{1}{N_1}\sum_{j=1}^{N_1}\sum_{f\in\{v_x,v_y,\epsilon,n,\Pi,\pi_{xx},\pi_{xy},\pi_{yy}\}}\bigl(f_\theta(x_j,1)-f^\ast(x_j,1)\bigr)^2,
\label{eq:rel2app_Ltop}
\end{equation}
\begin{equation}
L_{\rm L}(\theta)=\frac{1}{N_1}\sum_{j=1}^{N_1}\sum_{f\in\{v_x,v_y,\epsilon,n,\Pi,\pi_{xx},\pi_{xy},\pi_{yy}\}}\bigl(f_\theta(0,y_j)-f^\ast(0,y_j)\bigr)^2,
\label{eq:rel2app_Lleft}
\end{equation}
\begin{equation}
L_{\rm R}(\theta)=\frac{1}{N_1}\sum_{j=1}^{N_1}\sum_{f\in\{v_x,v_y,\epsilon,n,\Pi,\pi_{xx},\pi_{xy},\pi_{yy}\}}\bigl(f_\theta(1,y_j)-f^\ast(1,y_j)\bigr)^2,
\label{eq:rel2app_Lright}
\end{equation}
and the total uploaded objective is
\begin{equation}
\mathcal J_{\rm rel}(\theta)=L_\Omega(\theta)+L_\downarrow(\theta)+L_\uparrow(\theta)+L_{\rm L}(\theta)+L_{\rm R}(\theta).
\label{eq:rel2app_J}
\end{equation}
The learning-rate family is
\begin{equation}
\eta\in\left\{10^{-3},10^{-4},10^{-5},10^{-6},10^{-7},10^{-8}\right\},
\qquad
N_{\rm train}=8000.
\label{eq:rel2app_eta_train}
\end{equation}

\begin{table*}[t]
\centering
\scriptsize
\setlength{\tabcolsep}{4.1pt}
\renewcommand{\arraystretch}{1.10}
\caption{{Learning-rate comparison for the uploaded LNN--PINN eight-field relativistic viscous-fluid benchmark. The averaged global metrics are the arithmetic means of the eight fieldwise quantities listed in Tables~\ref{Tapp:rel2_lnn_mse}--\ref{Tapp:rel2_lnn_max}.}}
\label{Tapp:rel2_lnn_lr}
\resizebox{\textwidth}{!}{%
\begin{tabular}{cccccccc}
\toprule
Learning rate & $\mathcal J_{\mathrm{rel}}^{(8000)}$ & Avg. RMSE & Avg. MAE & Avg. rel$L_2$ & Avg. MaxErr & Train time (s) & Eval. time (s) \\
\midrule
$10^{-3}$ & $9.610906e-05$ & $5.609478e-03$ & $4.449910e-03$ & $4.757501e-02$ & $1.349895e-02$ & $1.502787e+03$ & $6.422446e+00$ \\
$10^{-4}$ & $9.455933e-04$ & $3.342321e-02$ & $2.420540e-02$ & $1.735676e-01$ & $8.224208e-02$ & $1.798969e+03$ & $8.611273e+00$ \\
$10^{-5}$ & $1.233484e-01$ & $6.833419e-02$ & $5.374227e-02$ & $1.006577e+00$ & $1.504903e-01$ & $1.800019e+03$ & $6.741312e+00$ \\
$10^{-6}$ & $1.619353e+00$ & $7.093887e-02$ & $5.891626e-02$ & $1.361149e+00$ & $1.529576e-01$ & $1.813637e+03$ & $6.923466e+00$ \\
$10^{-7}$ & $1.487902e+01$ & $3.475777e-01$ & $3.408668e-01$ & $2.876544e+00$ & $4.320656e-01$ & $1.830814e+03$ & $7.324025e+00$ \\
$10^{-8}$ & $1.583257e+01$ & $3.620675e-01$ & $3.553721e-01$ & $3.097658e+00$ & $4.463671e-01$ & $2.131619e+03$ & $8.966138e+00$ \\
\bottomrule
\end{tabular}}
\end{table*}

\begin{table*}[t]
\centering
\scriptsize
\setlength{\tabcolsep}{4.0pt}
\renewcommand{\arraystretch}{1.08}
\caption{{Fieldwise MSE values for all eight principal fields under the six uploaded LNN--PINN learning rates in the relativistic viscous-fluid benchmark. The smallest entry in each row is boldfaced.}}
\label{Tapp:rel2_lnn_mse}
\resizebox{\textwidth}{!}{%
\begin{tabular}{lcccccc}
\toprule
Field & $10^{-3}$ & $10^{-4}$ & $10^{-5}$ & $10^{-6}$ & $10^{-7}$ & $10^{-8}$ \\
\midrule
$v_x$ & $\bm{3.574851e-06}$ & $3.922854e-05$ & $1.494638e-03$ & $1.916044e-03$ & $4.308356e-03$ & $4.340588e-03$ \\
$v_y$ & $\bm{1.711175e-06}$ & $4.556016e-05$ & $1.612190e-03$ & $2.264287e-03$ & $4.634105e-03$ & $4.409695e-03$ \\
$\epsilon$ & $\bm{4.239025e-04}$ & $1.523641e-02$ & $6.607030e-02$ & $3.167754e-02$ & $2.554796e+00$ & $2.732351e+00$ \\
$n$ & $\bm{2.304069e-04}$ & $1.221127e-02$ & $7.257606e-03$ & $1.604548e-02$ & $4.882430e-01$ & $5.389780e-01$ \\
$\Pi$ & $\bm{4.439167e-06}$ & $7.498921e-05$ & $2.131019e-03$ & $6.606532e-04$ & $3.428434e-03$ & $6.770557e-03$ \\
$\pi_{xx}$ & $\bm{2.675099e-06}$ & $2.347162e-05$ & $9.837786e-04$ & $9.251834e-03$ & $1.036520e-02$ & $5.258430e-03$ \\
$\pi_{xy}$ & $\bm{9.453544e-07}$ & $1.246822e-05$ & $4.059935e-04$ & $9.323175e-04$ & $9.165681e-04$ & $3.365942e-03$ \\
$\pi_{yy}$ & $\bm{1.425933e-06}$ & $1.153760e-05$ & $7.812211e-04$ & $3.639566e-04$ & $2.533029e-02$ & $2.699535e-02$ \\
\bottomrule
\end{tabular}}
\end{table*}

\begin{table*}[t]
\centering
\scriptsize
\setlength{\tabcolsep}{4.0pt}
\renewcommand{\arraystretch}{1.08}
\caption{Fieldwise RMSE values for all eight principal fields under the six uploaded LNN--PINN learning rates in the relativistic viscous-fluid benchmark. The smallest entry in each row is boldfaced.}
\label{Tapp:rel2_lnn_rmse}
\resizebox{\textwidth}{!}{%
\begin{tabular}{lcccccc}
\toprule
Field & $10^{-3}$ & $10^{-4}$ & $10^{-5}$ & $10^{-6}$ & $10^{-7}$ & $10^{-8}$ \\
\midrule
$v_x$ & $\bm{1.890728e-03}$ & $6.263269e-03$ & $3.866055e-02$ & $4.377264e-02$ & $6.563806e-02$ & $6.588314e-02$ \\
$v_y$ & $\bm{1.308119e-03}$ & $6.749827e-03$ & $4.015209e-02$ & $4.758453e-02$ & $6.807426e-02$ & $6.640553e-02$ \\
$\epsilon$ & $\bm{2.058889e-02}$ & $1.234358e-01$ & $2.570414e-01$ & $1.779819e-01$ & $1.598373e+00$ & $1.652982e+00$ \\
$n$ & $\bm{1.517916e-02}$ & $1.105046e-01$ & $8.519159e-02$ & $1.266708e-01$ & $6.987438e-01$ & $7.341512e-01$ \\
$\Pi$ & $\bm{2.106933e-03}$ & $8.659631e-03$ & $4.616296e-02$ & $2.570318e-02$ & $5.855283e-02$ & $8.228339e-02$ \\
$\pi_{xx}$ & $\bm{1.635573e-03}$ & $4.844751e-03$ & $3.136525e-02$ & $9.618645e-02$ & $1.018096e-01$ & $7.251503e-02$ \\
$\pi_{xy}$ & $\bm{9.722934e-04}$ & $3.531037e-03$ & $2.014928e-02$ & $3.053387e-02$ & $3.027488e-02$ & $5.801674e-02$ \\
$\pi_{yy}$ & $\bm{1.194124e-03}$ & $3.396705e-03$ & $2.795033e-02$ & $1.907765e-02$ & $1.591549e-01$ & $1.643026e-01$ \\
\bottomrule
\end{tabular}}
\end{table*}

\begin{table*}[t]
\centering
\scriptsize
\setlength{\tabcolsep}{4.0pt}
\renewcommand{\arraystretch}{1.08}
\caption{Fieldwise MAE values for all eight principal fields under the six uploaded LNN--PINN learning rates in the relativistic viscous-fluid benchmark. The smallest entry in each row is boldfaced.}
\label{Tapp:rel2_lnn_mae}
\resizebox{\textwidth}{!}{%
\begin{tabular}{lcccccc}
\toprule
Field & $10^{-3}$ & $10^{-4}$ & $10^{-5}$ & $10^{-6}$ & $10^{-7}$ & $10^{-8}$ \\
\midrule
$v_x$ & $\bm{1.456494e-03}$ & $4.731446e-03$ & $3.330584e-02$ & $3.601836e-02$ & $5.143269e-02$ & $5.171994e-02$ \\
$v_y$ & $\bm{1.031180e-03}$ & $5.234249e-03$ & $3.489517e-02$ & $3.636491e-02$ & $5.758577e-02$ & $5.561948e-02$ \\
$\epsilon$ & $\bm{1.614483e-02}$ & $8.832387e-02$ & $1.916088e-01$ & $1.405345e-01$ & $1.594006e+00$ & $1.648825e+00$ \\
$n$ & $\bm{1.232836e-02}$ & $8.054122e-02$ & $6.815202e-02$ & $1.018374e-01$ & $6.953633e-01$ & $7.310911e-01$ \\
$\Pi$ & $\bm{1.629790e-03}$ & $5.846071e-03$ & $3.657782e-02$ & $2.045182e-02$ & $5.273611e-02$ & $7.791029e-02$ \\
$\pi_{xx}$ & $\bm{1.300723e-03}$ & $3.505903e-03$ & $2.623411e-02$ & $9.419344e-02$ & $9.802216e-02$ & $6.728637e-02$ \\
$\pi_{xy}$ & $\bm{7.894326e-04}$ & $2.899079e-03$ & $1.578227e-02$ & $2.683749e-02$ & $2.441805e-02$ & $5.242407e-02$ \\
$\pi_{yy}$ & $\bm{9.184768e-04}$ & $2.561348e-03$ & $2.338213e-02$ & $1.509224e-02$ & $1.533707e-01$ & $1.581003e-01$ \\
\bottomrule
\end{tabular}}
\end{table*}

\begin{table*}[t]
\centering
\scriptsize
\setlength{\tabcolsep}{4.0pt}
\renewcommand{\arraystretch}{1.08}
\caption{{Fieldwise relative $L_2$ errors for all eight principal fields under the six uploaded LNN--PINN learning rates in the relativistic viscous-fluid benchmark. The smallest entry in each row is boldfaced.}}
\label{Tapp:rel2_lnn_l2}
\resizebox{\textwidth}{!}{%
\begin{tabular}{lcccccc}
\toprule
Field & $10^{-3}$ & $10^{-4}$ & $10^{-5}$ & $10^{-6}$ & $10^{-7}$ & $10^{-8}$ \\
\midrule
$v_x$ & $\bm{2.693650e-02}$ & $8.923049e-02$ & $5.507826e-01$ & $6.236126e-01$ & $9.351213e-01$ & $9.386128e-01$ \\
$v_y$ & $\bm{2.174234e-02}$ & $1.121893e-01$ & $6.673706e-01$ & $7.909057e-01$ & $1.131467e+00$ & $1.103731e+00$ \\
$\epsilon$ & $\bm{8.746916e-03}$ & $5.244006e-02$ & $1.092006e-01$ & $7.561321e-02$ & $6.790474e-01$ & $7.022474e-01$ \\
$n$ & $\bm{9.784711e-03}$ & $7.123291e-02$ & $5.491576e-02$ & $8.165385e-02$ & $4.504206e-01$ & $4.732448e-01$ \\
$\Pi$ & $\bm{8.559020e-02}$ & $3.517812e-01$ & $1.875284e+00$ & $1.044143e+00$ & $2.378599e+00$ & $3.342608e+00$ \\
$\pi_{xx}$ & $\bm{8.353807e-02}$ & $2.474492e-01$ & $1.602003e+00$ & $4.912793e+00$ & $5.200001e+00$ & $3.703758e+00$ \\
$\pi_{xy}$ & $\bm{6.842984e-02}$ & $2.485138e-01$ & $1.418103e+00$ & $2.148969e+00$ & $2.130741e+00$ & $4.083208e+00$ \\
$\pi_{yy}$ & $\bm{7.583154e-02}$ & $2.157040e-01$ & $1.774955e+00$ & $1.211505e+00$ & $1.010696e+01$ & $1.043385e+01$ \\
\bottomrule
\end{tabular}}
\end{table*}

\begin{table*}[t]
\centering
\scriptsize
\setlength{\tabcolsep}{4.0pt}
\renewcommand{\arraystretch}{1.08}
\caption{{Fieldwise maximum absolute errors for all eight principal fields under the six uploaded LNN--PINN learning rates in the relativistic viscous-fluid benchmark. The smallest entry in each row is boldfaced.}}
\label{Tapp:rel2_lnn_max}
\resizebox{\textwidth}{!}{%
\begin{tabular}{lcccccc}
\toprule
Field & $10^{-3}$ & $10^{-4}$ & $10^{-5}$ & $10^{-6}$ & $10^{-7}$ & $10^{-8}$ \\
\midrule
$v_x$ & $\bm{4.596885e-03}$ & $1.582907e-02$ & $7.273307e-02$ & $9.954409e-02$ & $1.382097e-01$ & $1.385093e-01$ \\
$v_y$ & $\bm{3.721837e-03}$ & $1.692923e-02$ & $7.014959e-02$ & $1.058093e-01$ & $1.332729e-01$ & $1.312648e-01$ \\
$\epsilon$ & $\bm{4.986429e-02}$ & $2.993453e-01$ & $5.949600e-01$ & $4.314883e-01$ & $1.790375e+00$ & $1.835958e+00$ \\
$n$ & $\bm{3.357446e-02}$ & $2.668886e-01$ & $1.943530e-01$ & $2.795749e-01$ & $8.438607e-01$ & $8.763027e-01$ \\
$\Pi$ & $\bm{5.247686e-03}$ & $2.613730e-02$ & $1.029626e-01$ & $6.389873e-02$ & $1.117772e-01$ & $1.377140e-01$ \\
$\pi_{xx}$ & $\bm{4.293088e-03}$ & $1.379720e-02$ & $6.943508e-02$ & $1.360054e-01$ & $1.426732e-01$ & $1.110002e-01$ \\
$\pi_{xy}$ & $\bm{3.865548e-03}$ & $1.073004e-02$ & $4.605465e-02$ & $5.612891e-02$ & $6.822561e-02$ & $1.030519e-01$ \\
$\pi_{yy}$ & $\bm{2.827832e-03}$ & $8.279903e-03$ & $5.327421e-02$ & $5.121156e-02$ & $2.281298e-01$ & $2.371364e-01$ \\
\bottomrule
\end{tabular}}
\end{table*}

\begin{table*}[t]
\centering
\scriptsize
\setlength{\tabcolsep}{4.2pt}
\renewcommand{\arraystretch}{1.10}
\caption{{Fieldwise error statistics for the best averaged learning-rate run, namely $10^{-3}$, in the uploaded LNN--PINN relativistic viscous-fluid benchmark.}}
\label{Tapp:rel2_lnn_best_fields}
\resizebox{\textwidth}{!}{%
\begin{tabular}{lccccc}
\toprule
Field & MSE & RMSE & MAE & rel$L_2$ & MaxErr \\
\midrule
$v_x$ & $3.574851e-06$ & $1.890728e-03$ & $1.456494e-03$ & $2.693650e-02$ & $4.596885e-03$ \\
$v_y$ & $1.711175e-06$ & $1.308119e-03$ & $1.031180e-03$ & $2.174234e-02$ & $3.721837e-03$ \\
$\epsilon$ & $4.239025e-04$ & $2.058889e-02$ & $1.614483e-02$ & $8.746916e-03$ & $4.986429e-02$ \\
$n$ & $2.304069e-04$ & $1.517916e-02$ & $1.232836e-02$ & $9.784711e-03$ & $3.357446e-02$ \\
$\Pi$ & $4.439167e-06$ & $2.106933e-03$ & $1.629790e-03$ & $8.559020e-02$ & $5.247686e-03$ \\
$\pi_{xx}$ & $2.675099e-06$ & $1.635573e-03$ & $1.300723e-03$ & $8.353807e-02$ & $4.293088e-03$ \\
$\pi_{xy}$ & $9.453544e-07$ & $9.722934e-04$ & $7.894326e-04$ & $6.842984e-02$ & $3.865548e-03$ \\
$\pi_{yy}$ & $1.425933e-06$ & $1.194124e-03$ & $9.184768e-04$ & $7.583154e-02$ & $2.827832e-03$ \\
\bottomrule
\end{tabular}}
\end{table*}

\begin{table*}[t]
\centering
\scriptsize
\setlength{\tabcolsep}{4.2pt}
\renewcommand{\arraystretch}{1.10}
\caption{{Complete non-equation implementation parameters for the uploaded LNN--PINN eight-field relativistic viscous-fluid scripts: optimization, sampling, architecture, transforms, and evaluation controls.}}
\label{Tapp:rel2_lnn_params1}
\resizebox{\textwidth}{!}{%
\begin{tabular}{llll}
\toprule
Category & Symbol / item & Value & Mathematical status \\
\midrule
Learning-rate family & $\eta$ & $\{10^{-3},10^{-4},10^{-5},10^{-6},10^{-7},10^{-8}\}$ & active \\
Training horizon & $N_{\rm train}$ & $8000$ & active \\
Interior collocation count & $N$ & $1000$ & active \\
Each boundary-edge collocation count & $N_1$ & $1000$ & active \\
Logging interval & $N_{\rm print}$ & $100$ & inactive for optimization and reporting metrics \\
Seed & $s$ & $888888$ & active through stochastic realization choice \\
Optimizer family & Adam & yes & active \\
First moment factor & $\beta_1$ & $0.9$ & active \\
Second moment factor & $\beta_2$ & $0.999$ & active \\
Adam denominator shift & $\varepsilon_{\rm A}$ & $10^{-8}$ & active \\
Weight decay & $\lambda_{\rm wd}$ & $0$ & active \\
AMSGrad switch & $\chi_{\rm AMS}$ & $0$ & active as a discrete flag \\
Input dimension & $d$ & $2$ & active \\
Output dimension & $o$ & $8$ & active \\
Hidden width & $w$ & $64$ & active \\
Backbone hidden-stage count & $H$ & $4$ & active \\
Liquid-block count & $q$ & $4$ & active \\
Common gate initialization & $\alpha_0$ & $0.5$ & active \\
Velocity cap in the output transform & $V_{\max}$ & $0.45$ & active \\
Energy floor in the output transform & $\epsilon_{\min}$ & $0.05$ & active \\
Density floor in the output transform & $n_{\min}$ & $0.05$ & active \\
Stability floor in square-root and division operators & $\delta_{\rm stab}$ & $10^{-12}$ & active \\
Evaluation-grid side length & $h$ & $200$ & active for reporting metrics \\
Evaluation-point count & $M_h$ & $40000$ & derived active quantity \\
Raster density & DPI & $300$ & inactive for optimization and reporting metrics \\
Trainable-parameter count & $P_{\rm LNN}$ & $30088$ & derived active quantity \\
Execution metadata & device selection, CUDA warmup, output naming & present & inactive metadata \\
\bottomrule
\end{tabular}}
\end{table*}

\begin{table*}[t]
\centering
\footnotesize
\setlength{\tabcolsep}{3.6pt}
\renewcommand{\arraystretch}{1.05}
\caption{{Constitutive and manufactured-solution constants appearing in the uploaded LNN--PINN eight-field relativistic viscous-fluid scripts. Although these quantities belong to the benchmark specification rather than to the bare differential operator, they still enter the solver map and therefore require sensitivity analysis.}}
\label{Tapp:rel2_lnn_params2}
\begin{tabular*}{\textwidth}{@{\extracolsep{\fill}}lll@{\hspace{1.2em}}lll@{}}
\toprule
Parameter role & Symbol & Value & Parameter role & Symbol & Value \\
\midrule
Adiabatic index & $\Gamma_{\rm ad}$ & $1.4$ & Density baseline & $N_0$ & $1.55$ \\
Rest mass & $m$ & $0.8$ & Density amplitude 1 & $A_{n1}$ & $0.10$ \\
Shear viscosity & $\eta_{\rm s}$ & $0.06$ & Density amplitude 2 & $A_{n2}$ & $0.08$ \\
Bulk viscosity & $\zeta$ & $0.04$ & Bulk-pressure amplitude & $A_{\Pi}$ & $0.045$ \\
Shear relaxation time & $\tau_\pi$ & $0.25$ & $\pi_{xx}$ amplitude & $A_{xx}$ & $0.038$ \\
Bulk relaxation time & $\tau_\Pi$ & $0.22$ & $\pi_{xy}$ amplitude & $A_{xy}$ & $0.028$ \\
$v_x$ amplitude & $A_{v_x}$ & $0.14$ & $\pi_{yy}$ amplitude & $A_{yy}$ & $0.032$ \\
$v_y$ amplitude & $A_{v_y}$ & $0.12$ &  &  &  \\
Energy baseline & $E_0$ & $2.35$ &  &  &  \\
Energy amplitude 1 & $A_{\epsilon 1}$ & $0.18$ &  &  &  \\
Energy amplitude 2 & $A_{\epsilon 2}$ & $0.12$ &  &  &  \\
\bottomrule
\end{tabular*}
\end{table*}

We now perform the rigorous framework-based sensitivity analysis for every numerical non-equation parameter that appears in Tables~\ref{Tapp:rel2_lnn_params1} and \ref{Tapp:rel2_lnn_params2}. The Adam family used by the uploaded code can be written in its generalized form as
\begin{equation}
\widetilde g_k=g_k+\lambda_{\rm wd}\theta_k,
\qquad
g_k=\nabla_\theta \mathcal J_{\rm rel}(\theta_k;\Xi_k),
\label{eq:rel2app_gtilde}
\end{equation}
\begin{equation}
m_k=\beta_1m_{k-1}+(1-\beta_1)\widetilde g_k,
\qquad
v_k=\beta_2v_{k-1}+(1-\beta_2)(\widetilde g_k\odot\widetilde g_k),
\label{eq:rel2app_mkvk}
\end{equation}
\begin{equation}
\widehat m_k=\frac{m_k}{1-\beta_1^k},
\qquad
\widehat v_k=\frac{v_k}{1-\beta_2^k},
\qquad
\overline v_k=(1-\chi_{\rm AMS})\widehat v_k+\chi_{\rm AMS}\max\{\overline v_{k-1},\widehat v_k\},
\label{eq:rel2app_hatv}
\end{equation}
\begin{equation}
\theta_{k+1}=\theta_k-\eta\frac{\widehat m_k}{\sqrt{\overline v_k}+\varepsilon_{\rm A}}.
\label{eq:rel2app_adam}
\end{equation}
Defining the actual optimizer direction by
\begin{equation}
d_k=d_k(\eta,\beta_1,\beta_2,\varepsilon_{\rm A},\lambda_{\rm wd},\chi_{\rm AMS}):=\frac{\widehat m_k}{\sqrt{\overline v_k}+\varepsilon_{\rm A}},
\label{eq:rel2app_dk}
\end{equation}
Eq.~\eqref{eq:rel2app_adam} becomes \(\theta_{k+1}=\theta_k-\eta d_k\). Freezing the state \(\theta_k,m_k,v_k,\Xi_k\) and perturbing \(\eta\mapsto\eta+\delta\eta\) gives
\begin{equation}
\theta_{k+1}(\eta+\delta\eta)-\theta_{k+1}(\eta)=-\delta\eta\,d_k,
\qquad
\left.\frac{\partial\theta_{k+1}}{\partial\eta}\right|_{(\theta_k,m_k,v_k,\Xi_k)}=-d_k,
\qquad
\|\delta\theta_{k+1}\|_2\le |\delta\eta|\,\|d_k\|_2.
\label{eq:rel2app_dtheta_deta}
\end{equation}
A second-order Taylor expansion of \(\mathcal J_{\rm rel}\) along the actual update direction yields
\begin{equation}
\mathcal J_{\rm rel}(\theta_{k+1};\Xi_k)=\mathcal J_{\rm rel}(\theta_k;\Xi_k)-\eta\,g_k^\top d_k+\frac{\eta^2}{2}d_k^\top H_k(\theta_k-\tau_k\eta d_k)d_k,
\label{eq:rel2app_taylor}
\end{equation}
where \(H_k=\nabla_\theta^2\mathcal J_{\rm rel}(\theta_k;\Xi_k)\) and \(\tau_k\in(0,1)\). If
\begin{equation}
g_k^\top d_k\ge c_k\|g_k\|_2\|d_k\|_2,
\qquad
\|H_k(\theta_k-\tau_k\eta d_k)\|_2\le M_k,
\label{eq:rel2app_ckMk}
\end{equation}
then a sufficient descent range is
\begin{equation}
0<\eta<\frac{2c_k\|g_k\|_2}{M_k\|d_k\|_2}.
\label{eq:rel2app_eta_range}
\end{equation}
Eq.~\eqref{eq:rel2app_eta_range} gives the precise mathematical meaning of a reasonable learning-rate interval.

The training-horizon sensitivity follows by summing the updates:
\begin{equation}
\theta_K=\theta_0-\eta\sum_{k=0}^{K-1}d_k,
\qquad
\theta_{K+\Delta K}-\theta_K=-\eta\sum_{k=K}^{K+\Delta K-1}d_k,
\qquad
\|\theta_{K+\Delta K}-\theta_K\|_2\le \eta\sum_{k=K}^{K+\Delta K-1}\|d_k\|_2.
\label{eq:rel2app_epoch}
\end{equation}
Therefore \(N_{\rm train}\) acts through the accumulated tail of the Adam directions, not through a purely formal iteration count.

The moment-parameter sensitivities are exact and nontrivial. Unrolling Eq.~\eqref{eq:rel2app_mkvk} gives
\begin{equation}
m_k=(1-\beta_1)\sum_{j=1}^k\beta_1^{k-j}\widetilde g_j,
\qquad
v_k=(1-\beta_2)\sum_{j=1}^k\beta_2^{k-j}(\widetilde g_j\odot\widetilde g_j).
\label{eq:rel2app_unroll}
\end{equation}
Differentiating Eq.~\eqref{eq:rel2app_unroll} gives
\begin{equation}
\frac{\partial m_k}{\partial\beta_1}=-\sum_{j=1}^k\beta_1^{k-j}\widetilde g_j+(1-\beta_1)\sum_{j=1}^k(k-j)\beta_1^{k-j-1}\widetilde g_j,
\label{eq:rel2app_dmdb1}
\end{equation}
\begin{equation}
\frac{\partial \widehat m_k}{\partial\beta_1}=\frac{(1-\beta_1^k)\frac{\partial m_k}{\partial\beta_1}+k\beta_1^{k-1}m_k}{(1-\beta_1^k)^2},
\label{eq:rel2app_dhatmdb1}
\end{equation}
\begin{equation}
\frac{\partial v_k}{\partial\beta_2}=-\sum_{j=1}^k\beta_2^{k-j}(\widetilde g_j\odot\widetilde g_j)+(1-\beta_2)\sum_{j=1}^k(k-j)\beta_2^{k-j-1}(\widetilde g_j\odot\widetilde g_j),
\label{eq:rel2app_dvdb2}
\end{equation}
\begin{equation}
\frac{\partial \widehat v_k}{\partial\beta_2}=\frac{(1-\beta_2^k)\frac{\partial v_k}{\partial\beta_2}+k\beta_2^{k-1}v_k}{(1-\beta_2^k)^2}.
\label{eq:rel2app_dhatvdb2}
\end{equation}
Thus \(\beta_1\) and \(\beta_2\) alter the optimizer through the full gradient-memory kernel and cannot be interpreted as harmless scalar prefactors.

The denominator shift and weight-decay sensitivities follow directly from the generalized Adam map. Since
\begin{equation}
\frac{\partial d_k}{\partial\varepsilon_{\rm A}}=-\frac{\widehat m_k}{(\sqrt{\overline v_k}+\varepsilon_{\rm A})^2},
\label{eq:rel2app_deps}
\end{equation}
one obtains
\begin{equation}
\frac{\partial\theta_{k+1}}{\partial\varepsilon_{\rm A}}=\eta\frac{\widehat m_k}{(\sqrt{\overline v_k}+\varepsilon_{\rm A})^2}.
\label{eq:rel2app_dtheta_deps}
\end{equation}
Likewise, because \(\widetilde g_k=g_k+\lambda_{\rm wd}\theta_k\), differentiation of Eq.~\eqref{eq:rel2app_mkvk} with respect to \(\lambda_{\rm wd}\) yields
\begin{equation}
\frac{\partial m_k}{\partial\lambda_{\rm wd}}=\beta_1\frac{\partial m_{k-1}}{\partial\lambda_{\rm wd}}+(1-\beta_1)\left(\theta_k+\lambda_{\rm wd}\frac{\partial\theta_k}{\partial\lambda_{\rm wd}}\right),
\label{eq:rel2app_dmwd}
\end{equation}
\begin{equation}
\frac{\partial v_k}{\partial\lambda_{\rm wd}}=\beta_2\frac{\partial v_{k-1}}{\partial\lambda_{\rm wd}}+2(1-\beta_2)\widetilde g_k\odot\left(\theta_k+\lambda_{\rm wd}\frac{\partial\theta_k}{\partial\lambda_{\rm wd}}\right).
\label{eq:rel2app_dvwd}
\end{equation}
At the uploaded value \(\lambda_{\rm wd}=0\), these formulas reduce to
\begin{equation}
\left.\frac{\partial m_k}{\partial\lambda_{\rm wd}}\right|_{\lambda_{\rm wd}=0}=\beta_1\left.\frac{\partial m_{k-1}}{\partial\lambda_{\rm wd}}\right|_{\lambda_{\rm wd}=0}+(1-\beta_1)\theta_k,
\label{eq:rel2app_dmwd0}
\end{equation}
\begin{equation}
\left.\frac{\partial v_k}{\partial\lambda_{\rm wd}}\right|_{\lambda_{\rm wd}=0}=\beta_2\left.\frac{\partial v_{k-1}}{\partial\lambda_{\rm wd}}\right|_{\lambda_{\rm wd}=0}+2(1-\beta_2)g_k\odot\theta_k.
\label{eq:rel2app_dvwd0}
\end{equation}
The AMSGrad switch is discrete, so its correct sensitivity notion is the jump functional
\begin{equation}
\Delta_{\rm AMS}\theta_{k+1}=\theta_{k+1}\big|_{\chi_{\rm AMS}=1}-\theta_{k+1}\big|_{\chi_{\rm AMS}=0}=-\eta\widehat m_k\left[\frac{1}{\sqrt{\max\{\overline v_{k-1},\widehat v_k\}}+\varepsilon_{\rm A}}-\frac{1}{\sqrt{\widehat v_k}+\varepsilon_{\rm A}}\right].
\label{eq:rel2app_amsjump}
\end{equation}

The collocation-count sensitivities act through stochastic variance. Denote the squared interior and boundary residual random variables by \(X_\Omega\), \(X_\downarrow\), \(X_\uparrow\), \(X_{\rm L}\), and \(X_{\rm R}\), where each variable already includes the full eight-field sum appearing in Eqs.~\eqref{eq:rel2app_Lint}--\eqref{eq:rel2app_Lright}. Writing their variances as \(\sigma_\Omega^2\), \(\sigma_\downarrow^2\), \(\sigma_\uparrow^2\), \(\sigma_{\rm L}^2\), and \(\sigma_{\rm R}^2\), one has
\begin{equation}
\operatorname{Var}(L_\Omega)=\frac{\sigma_\Omega^2}{N},
\qquad
\operatorname{Var}(L_\downarrow)=\frac{\sigma_\downarrow^2}{N_1},
\qquad
\operatorname{Var}(L_\uparrow)=\frac{\sigma_\uparrow^2}{N_1},
\qquad
\operatorname{Var}(L_{\rm L})=\frac{\sigma_{\rm L}^2}{N_1},
\qquad
\operatorname{Var}(L_{\rm R})=\frac{\sigma_{\rm R}^2}{N_1}.
\label{eq:rel2app_varchannels}
\end{equation}
Therefore
\begin{equation}
\operatorname{Var}(\mathcal J_{\rm rel})=\frac{\sigma_\Omega^2}{N}+\frac{\sigma_\downarrow^2+\sigma_\uparrow^2+\sigma_{\rm L}^2+\sigma_{\rm R}^2}{N_1},
\label{eq:rel2app_varJ}
\end{equation}
so that
\begin{equation}
\frac{\partial}{\partial N}\operatorname{Var}(\mathcal J_{\rm rel})=-\frac{\sigma_\Omega^2}{N^2},
\qquad
\frac{\partial}{\partial N_1}\operatorname{Var}(\mathcal J_{\rm rel})=-\frac{\sigma_\downarrow^2+\sigma_\uparrow^2+\sigma_{\rm L}^2+\sigma_{\rm R}^2}{N_1^2}.
\label{eq:rel2app_dvardN}
\end{equation}
If one prescribes variance budgets \(\delta_\Omega\) and \(\delta_\partial\), then sufficient admissible sampling ranges are
\begin{equation}
N\ge \frac{\sigma_\Omega^2}{\delta_\Omega^2},
\qquad
N_1\ge \frac{\sigma_\downarrow^2+\sigma_\uparrow^2+\sigma_{\rm L}^2+\sigma_{\rm R}^2}{\delta_\partial^2}.
\label{eq:rel2app_samplingrange}
\end{equation}
The seed \(s\) is discrete rather than differentiable, so the correct sensitivity notion is the realization gap
\begin{equation}
\Delta_{s_1,s_2}\mathcal J_{\rm rel}(\theta_k)=\mathcal J_{\rm rel}(\theta_k;\Xi_k^{(s_1)})-\mathcal J_{\rm rel}(\theta_k;\Xi_k^{(s_2)}).
\label{eq:rel2app_seedgap}
\end{equation}
Because all six uploaded runs keep \(s=888888\) fixed, Table~\ref{Tapp:rel2_lnn_lr} isolates learning-rate effects instead of mixing them with seed variability.

The transform parameters \(V_{\max},\epsilon_{\min},n_{\min},\delta_{\rm stab}\) enter the solver map explicitly. From Eqs.~\eqref{eq:rel2app_vtransform}--\eqref{eq:rel2app_endens},
\begin{equation}
\frac{\partial v_x}{\partial V_{\max}}=\tanh(r_1),
\qquad
\frac{\partial v_y}{\partial V_{\max}}=\tanh(r_2),
\label{eq:rel2app_dv_dVmax}
\end{equation}
\begin{equation}
\frac{\partial\epsilon}{\partial\epsilon_{\min}}=1,
\qquad
\frac{\partial n}{\partial n_{\min}}=1,
\label{eq:rel2app_dedn}
\end{equation}
while the raw-output sensitivities are
\begin{equation}
\frac{\partial v_x}{\partial r_1}=V_{\max}\operatorname{sech}^2(r_1),
\qquad
\frac{\partial v_y}{\partial r_2}=V_{\max}\operatorname{sech}^2(r_2),
\label{eq:rel2app_drv}
\end{equation}
\begin{equation}
\frac{\partial\epsilon}{\partial r_3}=\sigma(r_3),
\qquad
\frac{\partial n}{\partial r_4}=\sigma(r_4),
\label{eq:rel2app_dren}
\end{equation}
where \(\sigma\) denotes the logistic function. Because \(|\tanh(r)|<1\) and \(0<\sigma(r)<1\), the transform enforces the admissibility bounds
\begin{equation}
|v_x|<V_{\max},
\qquad
|v_y|<V_{\max},
\qquad
\epsilon>\epsilon_{\min},
\qquad
n>n_{\min}.
\label{eq:rel2app_transformbounds}
\end{equation}
The stabilizer \(\delta_{\rm stab}\) enters the constitutive map through Eq.~\eqref{eq:rel2app_constitutive}. Differentiating \(\gamma=(1-v_x^2-v_y^2+\delta_{\rm stab})^{-1/2}\) and \(T=p/(n+\delta_{\rm stab})\) gives
\begin{equation}
\frac{\partial\gamma}{\partial\delta_{\rm stab}}=-\frac{1}{2}(1-v_x^2-v_y^2+\delta_{\rm stab})^{-3/2},
\qquad
\frac{\partial T}{\partial\delta_{\rm stab}}=-\frac{p}{(n+\delta_{\rm stab})^2}.
\label{eq:rel2app_dstab}
\end{equation}
Thus \(\delta_{\rm stab}\) regularizes both the Lorentz factor and the temperature denominator.

The constitutive parameters \(\Gamma_{\rm ad},m,\eta_{\rm s},\zeta,\tau_\pi,\tau_\Pi\) enter the residuals analytically. From Eq.~\eqref{eq:rel2app_constitutive},
\begin{equation}
\frac{\partial p}{\partial\Gamma_{\rm ad}}=\epsilon-mn,
\qquad
\frac{\partial p}{\partial m}=-(\Gamma_{\rm ad}-1)n,
\label{eq:rel2app_dp}
\end{equation}
which shows how the equation-of-state pair \(\Gamma_{\rm ad},m\) propagates into every tensor channel containing the pressure. For the bulk relaxation residual
\begin{equation}
R_\Pi=\tau_\Pi(u\cdot\partial)\Pi+\Pi+\zeta\theta-S_\Pi,
\label{eq:rel2app_RPi}
\end{equation}
one has
\begin{equation}
\frac{\partial R_\Pi}{\partial\tau_\Pi}=(u\cdot\partial)\Pi,
\qquad
\frac{\partial R_\Pi}{\partial\zeta}=\theta.
\label{eq:rel2app_RPi_sens}
\end{equation}
For the three shear relaxation residuals
\begin{equation}
R_{ij}=\tau_\pi(u\cdot\partial)\pi_{ij}+\pi_{ij}-2\eta_{\rm s}\sigma^{ij}-S_{ij},
\qquad (ij)\in\{xx,xy,yy\},
\label{eq:rel2app_Rij}
\end{equation}
one has
\begin{equation}
\frac{\partial R_{ij}}{\partial\tau_\pi}=(u\cdot\partial)\pi_{ij},
\qquad
\frac{\partial R_{ij}}{\partial\eta_{\rm s}}=-2\sigma^{ij}.
\label{eq:rel2app_Rij_sens}
\end{equation}
Therefore \(\tau_\Pi\), \(\tau_\pi\), \(\zeta\), and \(\eta_{\rm s}\) enter the solver through exact first-order coefficients in the relaxation residuals rather than through vague indirect effects.

The manufactured-solution constants also admit exact sensitivities. Since the exact benchmark fields in Eqs.~\eqref{eq:rel2app_vx}--\eqref{eq:rel2app_piyy} depend linearly on their amplitudes, one has, for example,
\begin{equation}
\frac{\partial v_x^\ast}{\partial A_{v_x}}=\sin(\pi x)\sin(\pi y)(1+0.25\cos(\pi x)),
\qquad
\frac{\partial v_y^\ast}{\partial A_{v_y}}=\sin(\pi x)\sin(\pi y)(1-0.25\cos(\pi y)),
\label{eq:rel2app_dAvel}
\end{equation}
\begin{equation}
\frac{\partial\epsilon^\ast}{\partial E_0}=1,
\qquad
\frac{\partial\epsilon^\ast}{\partial A_{\epsilon 1}}=\cos(2\pi x)\sin(\pi y),
\qquad
\frac{\partial\epsilon^\ast}{\partial A_{\epsilon 2}}=\sin(\pi x)\cos(2\pi y),
\label{eq:rel2app_dAeps}
\end{equation}
\begin{equation}
\frac{\partial n^\ast}{\partial N_0}=1,
\qquad
\frac{\partial n^\ast}{\partial A_{n1}}=\sin(2\pi x)\sin(\pi y),
\qquad
\frac{\partial n^\ast}{\partial A_{n2}}=\cos(\pi x)\sin(2\pi y),
\label{eq:rel2app_dAn}
\end{equation}
with analogous formulas for \(\Pi^\ast\), \(\pi_{xx}^\ast\), \(\pi_{xy}^\ast\), and \(\pi_{yy}^\ast\). These exact derivatives immediately yield admissible parameter ranges that keep the manufactured benchmark physically regular. Because \(|\sin|\le1\) and \(|\cos|\le1\), one has the sufficient bounds
\begin{equation}
|v_x^\ast|\le 1.25A_{v_x},
\qquad
|v_y^\ast|\le 1.25A_{v_y},
\label{eq:rel2app_velbound}
\end{equation}
so a sufficient subluminal manufactured regime is
\begin{equation}
1.25^2(A_{v_x}^2+A_{v_y}^2)<1.
\label{eq:rel2app_subluminal}
\end{equation}
Likewise,
\begin{equation}
\epsilon^\ast\ge E_0-|A_{\epsilon 1}|-|A_{\epsilon 2}|,
\qquad
n^\ast\ge N_0-|A_{n1}|-|A_{n2}|,
\label{eq:rel2app_posbounds}
\end{equation}
so positivity of the manufactured energy and density is guaranteed by the sufficient conditions
\begin{equation}
E_0>|A_{\epsilon 1}|+|A_{\epsilon 2}|,
\qquad
N_0>|A_{n1}|+|A_{n2}|.
\label{eq:rel2app_posconds}
\end{equation}
These are the explicit mathematical bounds that define a reasonable amplitude regime for the manufactured benchmark itself.

The architecture parameters \(d,o,w,H,q,\alpha_0\) admit exact discrete or differential sensitivities. The total parameter count is already given in Eq.~\eqref{eq:rel2app_PLNN_general}. Therefore
\begin{equation}
P_{\rm LNN}(d+1,o,w,H,q)-P_{\rm LNN}(d,o,w,H,q)=w,
\label{eq:rel2app_dd}
\end{equation}
\begin{equation}
P_{\rm LNN}(d,o+1,w,H,q)-P_{\rm LNN}(d,o,w,H,q)=w+1,
\label{eq:rel2app_do}
\end{equation}
\begin{equation}
P_{\rm LNN}(d,o,w,H+1,q)-P_{\rm LNN}(d,o,w,H,q)=w^2+w,
\label{eq:rel2app_dH}
\end{equation}
\begin{equation}
P_{\rm LNN}(d,o,w,H,q+1)-P_{\rm LNN}(d,o,w,H,q)=w^2+2w.
\label{eq:rel2app_dq}
\end{equation}
Increasing the hidden width by one gives
\begin{align}
&P_{\rm LNN}(d,o,w+1,H,q)-P_{\rm LNN}(d,o,w,H,q)
\nonumber\
&=(H+q-1)\bigl((w+1)^2-w^2\bigr)+(d+H+o+2q),
\label{eq:rel2app_dw}
\end{align}
which at the uploaded values becomes
\begin{equation}
P_{\rm LNN}(2,8,65,4,4)-P_{\rm LNN}(2,8,64,4,4)=7\cdot129+22=925.
\label{eq:rel2app_dw_actual}
\end{equation}
The liquid-block Jacobian is
\begin{equation}
D\mathcal B_\ell(z)=I+\operatorname{Diag}(\alpha_\ell)\operatorname{Diag}\bigl(\operatorname{sech}^2(A_\ell z+c_\ell)\bigr)A_\ell,
\label{eq:rel2app_DB}
\end{equation}
so that
\begin{equation}
\|D\mathcal B_\ell(z)\|_2\le 1+\|\operatorname{Diag}(\alpha_\ell)\|_2\,\|A_\ell\|_2.
\label{eq:rel2app_DBbound}
\end{equation}
At initialization \(\|\operatorname{Diag}(\alpha_\ell^{(0)})\|_2=\alpha_0=0.5\), hence
\begin{equation}
\|D\mathcal B_\ell(z)\|_2\le 1+0.5\|A_\ell\|_2.
\label{eq:rel2app_DBinit}
\end{equation}
If a per-block amplification budget \(\kappa_\ell>1\) is imposed, then a sufficient admissible initialization range is
\begin{equation}
0\le \alpha_0\le \frac{\kappa_\ell-1}{\|A_\ell\|_2},
\qquad
\ell=1,2,3,4,
\label{eq:rel2app_alpha_range_layer}
\end{equation}
which becomes the global sufficient range
\begin{equation}
0\le \alpha_0\le \min_{\ell=1,2,3,4}\frac{\kappa-1}{\|A_\ell\|_2}
\label{eq:rel2app_alpha_range_global}
\end{equation}
under a common budget \(\kappa\). Furthermore,
\begin{equation}
\left.\frac{\partial \mathcal B_\ell(z)}{\partial\alpha_0}\right|_{\alpha_\ell=\alpha_0\mathbf 1}=\tanh(A_\ell z+c_\ell),
\qquad
\left\|\left.\frac{\partial \mathcal B_\ell(z)}{\partial\alpha_0}\right|_{\alpha_\ell=\alpha_0\mathbf 1}\right\|_2\le \sqrt{w},
\label{eq:rel2app_dBalpha}
\end{equation}
so the output sensitivity with respect to \(\alpha_0\) obeys the chain-rule bound
\begin{equation}
\left\|\frac{\partial u_\theta}{\partial\alpha_0}\right\|_2
\le
\|W_{\rm out}\|_2
\sum_{j=1}^4
\sqrt{w}
\prod_{\ell=j+1}^4
\bigl(1+\alpha_0\|A_\ell\|_2\bigr)\|W_{\ell-1}\|_2.
\label{eq:rel2app_dualpha}
\end{equation}

Finally, the evaluation-grid and metadata sensitivities close the analysis. On the uniform \(h\times h\) grid one has
\begin{equation}
\Delta x=\Delta y=\frac{1}{h-1},
\qquad
M_h=h^2,
\label{eq:rel2app_grid}
\end{equation}
so that at \(h=200\), \(\Delta x=\Delta y=1/199\) and \(M_h=40000\). If
\begin{equation}
\mathcal E_{\rm cont}(\theta)=\int_0^1\int_0^1\sum_{f\in\{v_x,v_y,\epsilon,n,\Pi,\pi_{xx},\pi_{xy},\pi_{yy}\}}|f_\theta-f^\ast|^2\,dx\,dy,
\label{eq:rel2app_Econt}
\end{equation}
then the discrete fieldwise MSE tables in Tables~\ref{Tapp:rel2_lnn_mse}--\ref{Tapp:rel2_lnn_max} are uniform-grid Riemann approximations of the corresponding continuum norms, and the usual smoothness estimate gives
\begin{equation}
\mathrm{MSE}_h(\theta)=\mathcal E_{\rm cont}(\theta)+\mathcal O(h^{-1}).
\label{eq:rel2app_riemann}
\end{equation}
Hence \(h\) is mathematically active for the reporting functionals but not for the training objective in Eq.~\eqref{eq:rel2app_J}. In contrast, \(N_{\rm print}\) and DPI do not enter either the optimization map or the error functionals, so
\begin{equation}
\frac{\partial\mathcal J_{\rm rel}}{\partial N_{\rm print}}=0,
\qquad
\frac{\partial\mathcal J_{\rm rel}}{\partial\mathrm{DPI}}=0,
\qquad
\frac{\partial\mathrm{MSE}_h}{\partial N_{\rm print}}=0,
\qquad
\frac{\partial\mathrm{MSE}_h}{\partial\mathrm{DPI}}=0.
\label{eq:rel2app_meta_zero}
\end{equation}
They affect only how often the run logs are saved and how densely the figures are rasterized.

The unique appendix summary figure for this benchmark appears in Fig.~\ref{Fapp:rel2_lnn_lr}. It shows the full six-rate loss histories together with the eight-field reconstructed solutions and the corresponding absolute-error maps. The figure therefore complements Tables~\ref{Tapp:rel2_lnn_lr}--\ref{Tapp:rel2_lnn_best_fields} by exposing the entire optimization-path variability and the full fieldwise reconstruction variability under the fixed LNN--PINN framework.

\begin{figure*}[t]
\centering
\includegraphics[width=\textwidth]{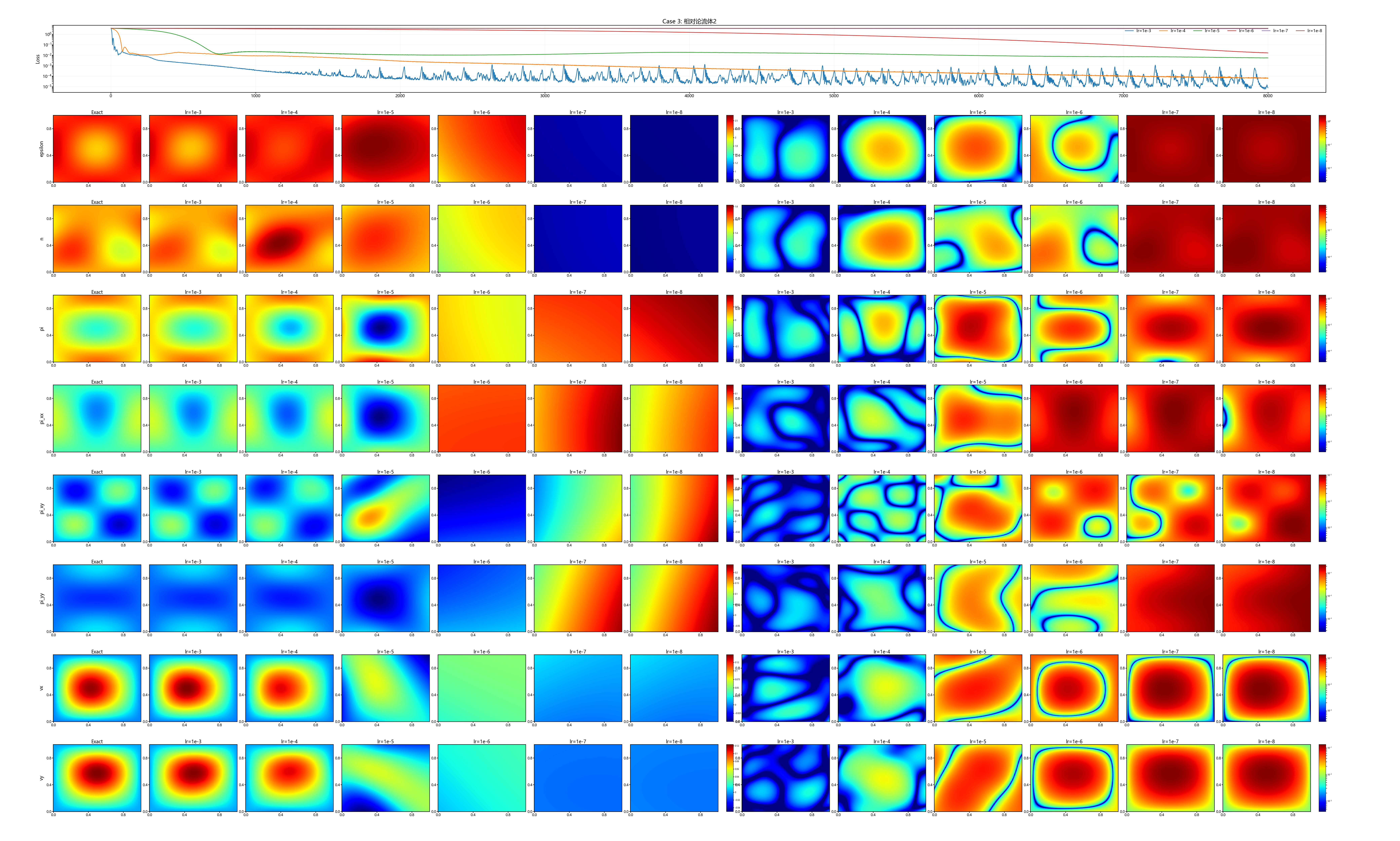}
\caption{{Learning-rate comparison for the uploaded LNN--PINN eight-field relativistic viscous-fluid benchmark. The upper panel reports the six loss histories associated with the learning-rate family in Eq.~\eqref{eq:rel2app_eta_train}. The lower panels report, for each of the eight principal fields, the reference solution, the six LNN--PINN reconstructions, and the associated absolute-error distributions. All runs keep the benchmark fields in Eqs.~\eqref{eq:rel2app_vx}--\eqref{eq:rel2app_piyy}, the objective in Eq.~\eqref{eq:rel2app_J}, the architecture in Eqs.~\eqref{eq:rel2app_h1}--\eqref{eq:rel2app_out}, the collocation counts in Eq.~\eqref{eq:rel2app_counts}, and the evaluation grid in Eq.~\eqref{eq:rel2app_grid} fixed.}}
\label{Fapp:rel2_lnn_lr}
\end{figure*}

Taken together, Eq.~\eqref{eq:rel2app_eta_range} gives the local descent-controlled admissible range for the learning rate, Eq.~\eqref{eq:rel2app_samplingrange} gives the variance-controlled admissible ranges for the interior and edge collocation counts, Eq.~\eqref{eq:rel2app_alpha_range_global} gives the block-amplification-controlled admissible range for the liquid-gate initialization, Eq.~\eqref{eq:rel2app_posconds} gives the positivity constraints for the manufactured energy and density baselines, Eq.~\eqref{eq:rel2app_subluminal} gives a sufficient subluminal regime for the manufactured velocities, and Eq.~\eqref{eq:rel2app_riemann} explains how the evaluation-grid side length controls reporting accuracy. Within the six uploaded runs summarized in Tables~\ref{Tapp:rel2_lnn_lr}--\ref{Tapp:rel2_lnn_best_fields}, $10^{-3}$) gives the strongest overall average field accuracy and the shortest runtime, while the larger-scale failures at \(10^{-7}\) and \(10^{-8}\) follow directly from the optimizer-sensitivity and transform-sensitivity mechanisms derived above.
\section{Learning-rate comparison, complete implementation parameters, and rigorous framework-based parameter-sensitivity analysis for the LNN--PINN relativistic viscous-fluid benchmark}
\label{f}

This appendix supplements Section~3.6 and focuses exclusively on the uploaded LNN--PINN implementation for the second steady relativistic viscous-fluid manufactured benchmark. The present appendix uses only valid external references, namely the generic physics-only composite objective in Eq.~\eqref{LOSS Total} and the framework figure in Fig.~\ref{LNN-PINN}. The case-specific eight-field relativistic system and its exact benchmark fields are not assigned separate local labels in the current main text, so we write the required case-level definitions explicitly here instead of invoking nonexistent references.

The unknown principal fields are
\begin{equation}
(v_x,v_y,\epsilon,n,\Pi,\pi_{xx},\pi_{xy},\pi_{yy}).
\label{eq:rel3app_fields}
\end{equation}
The uploaded scripts define the exact benchmark fields by
\begin{equation}
v_x^\ast(x,y)=B_{v_x1}\sin(2\pi x)\sin(\pi y)+B_{v_x2}\sin(\pi x)\sin(2\pi y),
\label{eq:rel3app_vx}
\end{equation}
\begin{equation}
v_y^\ast(x,y)=B_{v_y1}\sin(\pi x)\sin(2\pi y)-B_{v_y2}\sin(2\pi x)\sin(\pi y),
\label{eq:rel3app_vy}
\end{equation}
\begin{equation}
\epsilon^\ast(x,y)=E_1+B_{\epsilon1}\cos(2\pi x)\cos(\pi y)+B_{\epsilon2}\sin(\pi x)\sin(2\pi y),
\label{eq:rel3app_eps}
\end{equation}
\begin{equation}
n^\ast(x,y)=N_1+B_{n1}\sin(\pi x)\sin(\pi y)+B_{n2}\cos(2\pi x)\sin(2\pi y),
\label{eq:rel3app_n}
\end{equation}
\begin{equation}
\Pi^\ast(x,y)=B_{\Pi1}\cos(\pi x)\sin(2\pi y)+B_{\Pi2}\sin(2\pi x)\cos(\pi y),
\label{eq:rel3app_Pi}
\end{equation}
\begin{equation}
\pi_{xx}^\ast(x,y)=B_{xx1}\sin(2\pi x)\cos(\pi y)+B_{xx2}\cos(\pi x)\sin(2\pi y),
\label{eq:rel3app_pixx}
\end{equation}
\begin{equation}
\pi_{xy}^\ast(x,y)=B_{xy1}\sin(\pi x)\sin(\pi y)+B_{xy2}\sin(2\pi x)\sin(2\pi y),
\label{eq:rel3app_pixy}
\end{equation}
\begin{equation}
\pi_{yy}^\ast(x,y)=B_{yy1}\cos(2\pi x)\sin(\pi y)-B_{yy2}\sin(\pi x)\cos(2\pi y).
\label{eq:rel3app_piyy}
\end{equation}
The uploaded numerical amplitudes are
\begin{equation}
B_{v_x1}=0.085,\quad
B_{v_x2}=0.045,\quad
B_{v_y1}=0.075,\quad
B_{v_y2}=0.040,
\label{eq:rel3app_amp1}
\end{equation}
\begin{equation}
E_1=2.42,\quad
B_{\epsilon1}=0.17,\quad
B_{\epsilon2}=0.11,\quad
N_1=1.58,\quad
B_{n1}=0.095,\quad
B_{n2}=0.070,
\label{eq:rel3app_amp2}
\end{equation}
\begin{equation}
B_{\Pi1}=0.032,\quad
B_{\Pi2}=0.018,\quad
B_{xx1}=0.030,\quad
B_{xx2}=0.014,\quad
B_{xy1}=0.026,\quad
B_{xy2}=0.012,\quad
B_{yy1}=0.028,\quad
B_{yy2}=0.013.
\label{eq:rel3app_amp3}
\end{equation}

The uploaded model samples one interior set and four edge sets,
\begin{equation}
\mathcal S_\Omega=\{(x_i,y_i)\}_{i=1}^N,
\qquad
\mathcal S_\downarrow=\{(x_j,0)\}_{j=1}^{N_1},
\qquad
\mathcal S_\uparrow=\{(x_j,1)\}_{j=1}^{N_1},
\qquad
\mathcal S_\mathrm{L}=\{(0,y_j)\}_{j=1}^{N_1},
\qquad
\mathcal S_\mathrm{R}=\{(1,y_j)\}_{j=1}^{N_1},
\label{eq:rel3app_sets}
\end{equation}
with
\begin{equation}
N=1000,
\qquad
N_1=1000.
\label{eq:rel3app_counts}
\end{equation}
At every training step the code redraws these collocation sets independently from the corresponding uniform laws, so the loss is a stochastic empirical functional rather than a deterministic continuum functional.

The uploaded code builds the physical state through a transformed neural output. If the raw network output is denoted by
\begin{equation}
r_\theta(x,y)=\bigl(r_1,r_2,r_3,r_4,r_5,r_6,r_7,r_8\bigr),
\label{eq:rel3app_raw}
\end{equation}
then the actual field transform in the uploaded LNN--PINN scripts is
\begin{equation}
v_x=V_{\max}\tanh(r_1),
\qquad
v_y=V_{\max}\tanh(r_2),
\label{eq:rel3app_vtransform}
\end{equation}
\begin{equation}
\epsilon=\epsilon_{\min}+\operatorname{softplus}(r_3),
\qquad
n=n_{\min}+\operatorname{softplus}(r_4),
\label{eq:rel3app_endens}
\end{equation}
\begin{equation}
\Pi=r_5,
\qquad
\pi_{xx}=r_6,
\qquad
\pi_{xy}=r_7,
\qquad
\pi_{yy}=r_8,
\label{eq:rel3app_shearid}
\end{equation}
with the explicit transform constants
\begin{equation}
V_{\max}=0.45,
\qquad
\epsilon_{\min}=0.05,
\qquad
n_{\min}=0.05,
\qquad
\delta_{\mathrm{stab}}=10^{-12}.
\label{eq:rel3app_transform_constants}
\end{equation}

The uploaded architecture keeps the linear topology \(2\to64\to64\to64\to64\to8\) and inserts one width-preserving liquid residual block after each hidden activation. Writing \(\xi=[x\;y]^{\top}\in\mathbb{R}^2\), one has
\begin{equation}
z_1=\tanh(W_0\xi+b_0),
\qquad
h_1=z_1+\operatorname{Diag}(\alpha_1)\tanh(B_1z_1+c_1),
\label{eq:rel3app_h1}
\end{equation}
\begin{equation}
z_2=\tanh(W_1h_1+b_1),
\qquad
h_2=z_2+\operatorname{Diag}(\alpha_2)\tanh(B_2z_2+c_2),
\label{eq:rel3app_h2}
\end{equation}
\begin{equation}
z_3=\tanh(W_2h_2+b_2),
\qquad
h_3=z_3+\operatorname{Diag}(\alpha_3)\tanh(B_3z_3+c_3),
\label{eq:rel3app_h3}
\end{equation}
\begin{equation}
z_4=\tanh(W_3h_3+b_3),
\qquad
h_4=z_4+\operatorname{Diag}(\alpha_4)\tanh(B_4z_4+c_4),
\label{eq:rel3app_h4}
\end{equation}
\begin{equation}
r_\theta(x,y)=W_{\rm out}h_4+b_{\rm out}.
\label{eq:rel3app_out}
\end{equation}
The numerical architecture values are
\begin{equation}
d=2,
\qquad
o=8,
\qquad
w=64,
\qquad
H=4,
\qquad
q=4,
\qquad
\alpha_0=0.5,
\label{eq:rel3app_archvals}
\end{equation}
with \(\alpha_\ell^{(0)}=\alpha_0\mathbf{1}\) for \(\ell=1,2,3,4\).

At the constitutive level the uploaded scripts define
\begin{equation}
\gamma=\frac{1}{\sqrt{1-v_x^2-v_y^2+\delta_{\mathrm{stab}}}},
\qquad
p=(\Gamma_{\rm ad}-1)(\epsilon-mn),
\qquad
T=\frac{p}{n+\delta_{\mathrm{stab}}},
\label{eq:rel3app_constitutive}
\end{equation}
with the fixed constitutive parameters
\begin{equation}
\Gamma_{\rm ad}=1.4,
\qquad
m=0.8,
\qquad
\eta_{\rm shear}=0.06,
\qquad
\zeta=0.04,
\qquad
\tau_\pi=0.25,
\qquad
\tau_\Pi=0.22.
\label{eq:rel3app_constitutive_params}
\end{equation}
The eight interior PDE residual channels are denoted by
\begin{equation}
\mathcal R_\Omega=(R_E,R_{M_x},R_{M_y},R_n,R_\Pi,R_{xx},R_{xy},R_{yy}),
\label{eq:rel3app_resvec}
\end{equation}
where the first four channels come from the steady conservation laws and the last four channels come from the steady bulk and shear relaxation laws. The exact manufactured sources are generated by inserting the exact fields in Eqs.~\eqref{eq:rel3app_vx}--\eqref{eq:rel3app_piyy} into the same constitutive assembly and differentiating them analytically via automatic differentiation. The uploaded interior loss therefore takes the form
\begin{equation}
L_\Omega(\theta)=\frac{1}{N}\sum_{i=1}^N\sum_{a\in\{E,M_x,M_y,n,\Pi,xx,xy,yy\}}\bigl(R_a(x_i,y_i;\theta)-S_a(x_i,y_i)\bigr)^2,
\label{eq:rel3app_Lint}
\end{equation}
while each edge loss is the summed eight-field boundary mismatch,
\begin{equation}
L_\downarrow(\theta)=\frac{1}{N_1}\sum_{j=1}^{N_1}\sum_{f\in\{v_x,v_y,\epsilon,n,\Pi,\pi_{xx},\pi_{xy},\pi_{yy}\}}\bigl(f_\theta(x_j,0)-f^\ast(x_j,0)\bigr)^2,
\label{eq:rel3app_Lbot}
\end{equation}
\begin{equation}
L_\uparrow(\theta)=\frac{1}{N_1}\sum_{j=1}^{N_1}\sum_{f\in\{v_x,v_y,\epsilon,n,\Pi,\pi_{xx},\pi_{xy},\pi_{yy}\}}\bigl(f_\theta(x_j,1)-f^\ast(x_j,1)\bigr)^2,
\label{eq:rel3app_Ltop}
\end{equation}
\begin{equation}
L_{\rm L}(\theta)=\frac{1}{N_1}\sum_{j=1}^{N_1}\sum_{f\in\{v_x,v_y,\epsilon,n,\Pi,\pi_{xx},\pi_{xy},\pi_{yy}\}}\bigl(f_\theta(0,y_j)-f^\ast(0,y_j)\bigr)^2,
\label{eq:rel3app_Lleft}
\end{equation}
\begin{equation}
L_{\rm R}(\theta)=\frac{1}{N_1}\sum_{j=1}^{N_1}\sum_{f\in\{v_x,v_y,\epsilon,n,\Pi,\pi_{xx},\pi_{xy},\pi_{yy}\}}\bigl(f_\theta(1,y_j)-f^\ast(1,y_j)\bigr)^2,
\label{eq:rel3app_Lright}
\end{equation}
and the total uploaded objective is
\begin{equation}
\mathcal J_{\rm rel}(\theta)=L_\Omega(\theta)+L_\downarrow(\theta)+L_\uparrow(\theta)+L_{\rm L}(\theta)+L_{\rm R}(\theta).
\label{eq:rel3app_J}
\end{equation}
The learning-rate family is
\begin{equation}
\eta\in\left\{10^{-3},10^{-4},10^{-5},10^{-6},10^{-7},10^{-8}\right\},
\qquad
N_{\rm train}=8000.
\label{eq:rel3app_eta_train}
\end{equation}
The optimizer uses the Adam family with
\begin{equation}
\beta_1=0.9,
\qquad
\beta_2=0.999,
\qquad
\varepsilon_{\rm A}=10^{-8},
\qquad
\lambda_{\rm wd}=0,
\qquad
\chi_{\rm AMS}=0.
\label{eq:rel3app_adam_defaults}
\end{equation}
The evaluation grid and logging controls are
\begin{equation}
h=200,
\qquad
M_h=h^2=40000,
\qquad
N_{\rm print}=100,
\qquad
\mathrm{DPI}=300.
\label{eq:rel3app_eval_controls}
\end{equation}

\begin{table*}[t]
\centering
\scriptsize
\setlength{\tabcolsep}{4.1pt}
\renewcommand{\arraystretch}{1.10}
\caption{{Learning-rate comparison for the uploaded LNN--PINN eight-field relativistic viscous-fluid benchmark. The averaged global metrics are the arithmetic means of the eight fieldwise quantities listed in Tables~\ref{Tapp:rel3_lnn_mse}--\ref{Tapp:rel3_lnn_max}.}}
\label{Tapp:rel3_lnn_lr}
\resizebox{\textwidth}{!}{%
\begin{tabular}{cccccccc}
\toprule
Learning rate & $\mathcal J_{\mathrm{rel}}^{(8000)}$ & Avg. RMSE & Avg. MAE & Avg. rel$L_2$ & Avg. MaxErr & Train time (s) & Eval. time (s) \\
\midrule
$10^{-3}$ & 2.266074530e-03 & 6.688977610e-03 & 5.902523565e-03 & 1.007659500e-01 & 1.495373476e-02 & 1377.890340 & 17.063965 \\
$10^{-4}$ & 1.921621270e-03 & 9.796825176e-03 & 7.638779334e-03 & 1.417611254e-01 & 2.678152932e-02 & 2560.843930 & 11.053338 \\
$10^{-5}$ & 1.650525030e-01 & 5.931066380e-02 & 4.931004316e-02 & 7.842210056e-01 & 1.285905577e-01 & 2602.280200 & 8.729228 \\
$10^{-6}$ & 2.252492670e+00 & 7.398785350e-02 & 6.348093064e-02 & 1.390036696e+00 & 1.731745778e-01 & 2616.039760 & 7.231567 \\
$10^{-7}$ & 1.532460590e+01 & 3.576617888e-01 & 3.521822910e-01 & 2.898031204e+00 & 4.330625773e-01 & 1757.707690 & 7.458349 \\
$10^{-8}$ & 1.627523040e+01 & 3.722882903e-01 & 3.668734500e-01 & 3.130352353e+00 & 4.480976324e-01 & 1756.193080 & 8.716536 \\
\bottomrule
\end{tabular}}
\end{table*}

\begin{table*}[t]
\centering
\scriptsize
\setlength{\tabcolsep}{4.0pt}
\renewcommand{\arraystretch}{1.08}
\caption{{Fieldwise MSE values for all eight principal fields under the six uploaded LNN--PINN learning rates in the relativistic viscous-fluid benchmark. The smallest entry in each row is boldfaced.}}
\label{Tapp:rel3_lnn_mse}
\resizebox{\textwidth}{!}{%
\begin{tabular}{lcccccc}
\toprule
Field & $10^{-3}$ & $10^{-4}$ & $10^{-5}$ & $10^{-6}$ & $10^{-7}$ & $10^{-8}$ \\
\midrule
$v_x$ & \textbf{6.060608e-06} & 1.683742e-05 & 1.245980e-03 & 2.362096e-03 & 2.318204e-03 & 2.354087e-03 \\
$v_y$ & \textbf{5.050105e-06} & 6.295779e-06 & 9.435500e-04 & 1.806937e-03 & 1.901061e-03 & 1.873819e-03 \\
$\epsilon$ & \textbf{5.898105e-04} & 1.116533e-03 & 5.858615e-02 & 3.966799e-02 & 2.773762e+00 & 2.958396e+00 \\
$n$ & \textbf{1.523031e-04} & 4.234530e-04 & 7.276007e-03 & 1.876309e-02 & 5.843461e-01 & 6.407204e-01 \\
$\Pi$ & \textbf{2.547198e-05} & 4.533351e-05 & 5.541190e-04 & 5.756521e-04 & 3.355109e-03 & 6.750050e-03 \\
$\pi_{xx}$ & \textbf{2.271313e-06} & 2.107090e-05 & 3.159972e-04 & 8.233069e-03 & 1.035139e-02 & 5.311019e-03 \\
$\pi_{xy}$ & 1.858066e-05 & \textbf{4.530259e-06} & 2.530912e-04 & 9.321460e-04 & 4.364911e-04 & 2.181204e-03 \\
$\pi_{yy}$ & \textbf{1.718792e-06} & 1.864358e-05 & 5.711838e-04 & 3.758398e-04 & 2.531319e-02 & 2.699844e-02 \\
\bottomrule
\end{tabular}}
\end{table*}

\begin{table*}[t]
\centering
\scriptsize
\setlength{\tabcolsep}{4.0pt}
\renewcommand{\arraystretch}{1.08}
\caption{{Fieldwise RMSE values for all eight principal fields under the six uploaded LNN--PINN learning rates in the relativistic viscous-fluid benchmark. The smallest entry in each row is boldfaced.}}
\label{Tapp:rel3_lnn_rmse}
\resizebox{\textwidth}{!}{%
\begin{tabular}{lcccccc}
\toprule
Field & $10^{-3}$ & $10^{-4}$ & $10^{-5}$ & $10^{-6}$ & $10^{-7}$ & $10^{-8}$ \\
\midrule
$v_x$ & \textbf{2.461830e-03} & 4.103342e-03 & 3.529844e-02 & 4.860140e-02 & 4.814773e-02 & 4.851893e-02 \\
$v_y$ & \textbf{2.247244e-03} & 2.509139e-03 & 3.071726e-02 & 4.250809e-02 & 4.360116e-02 & 4.328763e-02 \\
$\epsilon$ & \textbf{2.428601e-02} & 3.341456e-02 & 2.420458e-01 & 1.991682e-01 & 1.665461e+00 & 1.719999e+00 \\
$n$ & \textbf{1.234111e-02} & 2.057797e-02 & 8.529951e-02 & 1.369784e-01 & 7.644254e-01 & 8.004501e-01 \\
$\Pi$ & \textbf{5.046977e-03} & 6.733016e-03 & 2.353973e-02 & 2.399275e-02 & 5.792330e-02 & 8.215869e-02 \\
$\pi_{xx}$ & \textbf{1.507087e-03} & 4.590305e-03 & 1.777631e-02 & 9.073626e-02 & 1.017418e-01 & 7.287674e-02 \\
$\pi_{xy}$ & 4.310529e-03 & \textbf{2.128440e-03} & 1.590884e-02 & 3.053107e-02 & 2.089237e-02 & 4.670336e-02 \\
$\pi_{yy}$ & \textbf{1.311027e-03} & 4.317822e-03 & 2.389945e-02 & 1.938659e-02 & 1.591012e-01 & 1.643120e-01 \\
\bottomrule
\end{tabular}}
\end{table*}

\begin{table*}[t]
\centering
\scriptsize
\setlength{\tabcolsep}{4.0pt}
\renewcommand{\arraystretch}{1.08}
\caption{{Fieldwise MAE values for all eight principal fields under the six uploaded LNN--PINN learning rates in the relativistic viscous-fluid benchmark. The smallest entry in each row is boldfaced.}}
\label{Tapp:rel3_lnn_mae}
\resizebox{\textwidth}{!}{%
\begin{tabular}{lcccccc}
\toprule
Field & $10^{-3}$ & $10^{-4}$ & $10^{-5}$ & $10^{-6}$ & $10^{-7}$ & $10^{-8}$ \\
\midrule
$v_x$ & \textbf{2.066875e-03} & 3.348049e-03 & 2.626318e-02 & 3.824281e-02 & 3.781525e-02 & 3.823877e-02 \\
$v_y$ & \textbf{1.920217e-03} & 1.983414e-03 & 2.548058e-02 & 3.366614e-02 & 3.454475e-02 & 3.417289e-02 \\
$\epsilon$ & \textbf{2.216381e-02} & 2.540382e-02 & 2.063906e-01 & 1.577169e-01 & 1.663212e+00 & 1.717882e+00 \\
$n$ & \textbf{9.695198e-03} & 1.636238e-02 & 6.904295e-02 & 1.256320e-01 & 7.630997e-01 & 7.991955e-01 \\
$\Pi$ & \textbf{4.804015e-03} & 5.477317e-03 & 1.879227e-02 & 2.053228e-02 & 5.236903e-02 & 7.803094e-02 \\
$\pi_{xx}$ & \textbf{1.229370e-03} & 3.303218e-03 & 1.459824e-02 & 8.838405e-02 & 9.777679e-02 & 6.730637e-02 \\
$\pi_{xy}$ & 4.235629e-03 & \textbf{1.754701e-03} & 1.368425e-02 & 2.796420e-02 & 1.530127e-02 & 4.203356e-02 \\
$\pi_{yy}$ & \textbf{1.105069e-03} & 3.477341e-03 & 2.022829e-02 & 1.570907e-02 & 1.533394e-01 & 1.581271e-01 \\
\bottomrule
\end{tabular}}
\end{table*}

\begin{table*}[t]
\centering
\scriptsize
\setlength{\tabcolsep}{4.0pt}
\renewcommand{\arraystretch}{1.08}
\caption{{Fieldwise relative $L_2$ errors for all eight principal fields under the six uploaded LNN--PINN learning rates in the relativistic viscous-fluid benchmark. The smallest entry in each row is boldfaced.}}
\label{Tapp:rel3_lnn_l2}
\resizebox{\textwidth}{!}{%
\begin{tabular}{lcccccc}
\toprule
Field & $10^{-3}$ & $10^{-4}$ & $10^{-5}$ & $10^{-6}$ & $10^{-7}$ & $10^{-8}$ \\
\midrule
$v_x$ & \textbf{5.145103e-02} & 8.575784e-02 & 7.377200e-01 & 1.015745e+00 & 1.006264e+00 & 1.014022e+00 \\
$v_y$ & \textbf{5.314204e-02} & 5.933524e-02 & 7.263910e-01 & 1.005216e+00 & 1.031065e+00 & 1.023651e+00 \\
$\epsilon$ & \textbf{1.002959e-02} & 1.379948e-02 & 9.995955e-02 & 8.225209e-02 & 6.877987e-01 & 7.103215e-01 \\
$n$ & \textbf{7.623918e-03} & 1.271237e-02 & 5.269513e-02 & 8.462060e-02 & 4.722359e-01 & 4.944908e-01 \\
$\Pi$ & \textbf{2.167059e-01} & 2.891006e-01 & 1.010743e+00 & 1.030195e+00 & 2.487097e+00 & 3.527710e+00 \\
$\pi_{xx}$ & \textbf{7.321036e-02} & 2.229850e-01 & 8.635266e-01 & 4.407731e+00 & 4.942349e+00 & 3.540161e+00 \\
$\pi_{xy}$ & 3.025733e-01 & \textbf{1.494037e-01} & 1.116705e+00 & 2.143097e+00 & 1.466519e+00 & 3.278295e+00 \\
$\pi_{yy}$ & \textbf{9.139151e-02} & 3.009947e-01 & 1.666027e+00 & 1.351436e+00 & 1.109092e+01 & 1.145417e+01 \\
\bottomrule
\end{tabular}}
\end{table*}

\begin{table*}[t]
\centering
\scriptsize
\setlength{\tabcolsep}{4.0pt}
\renewcommand{\arraystretch}{1.08}
\caption{{Fieldwise maximum absolute errors for all eight principal fields under the six uploaded LNN--PINN learning rates in the relativistic viscous-fluid benchmark. The smallest entry in each row is boldfaced.}}
\label{Tapp:rel3_lnn_max}
\resizebox{\textwidth}{!}{%
\begin{tabular}{lcccccc}
\toprule
Field & $10^{-3}$ & $10^{-4}$ & $10^{-5}$ & $10^{-6}$ & $10^{-7}$ & $10^{-8}$ \\
\midrule
$v_x$ & \textbf{5.922412e-03} & 1.024671e-02 & 1.423253e-01 & 1.090850e-01 & 1.058880e-01 & 1.081458e-01 \\
$v_y$ & \textbf{7.875663e-03} & 1.285107e-02 & 6.848134e-02 & 9.271904e-02 & 1.007788e-01 & 9.883013e-02 \\
$\epsilon$ & \textbf{5.065465e-02} & 8.862209e-02 & 4.600115e-01 & 6.574900e-01 & 1.863048e+00 & 1.903877e+00 \\
$n$ & \textbf{3.355682e-02} & 5.157423e-02 & 1.770303e-01 & 2.561685e-01 & 8.659976e-01 & 9.031647e-01 \\
$\Pi$ & \textbf{8.002438e-03} & 1.722534e-02 & 5.264746e-02 & 4.910175e-02 & 9.329003e-02 & 1.212676e-01 \\
$\pi_{xx}$ & \textbf{4.230476e-03} & 1.321806e-02 & 4.553898e-02 & 1.266442e-01 & 1.538613e-01 & 1.243358e-01 \\
$\pi_{xy}$ & \textbf{6.248086e-03} & 8.118458e-03 & 3.522235e-02 & 4.827890e-02 & 5.375760e-02 & 8.806160e-02 \\
$\pi_{yy}$ & \textbf{3.139334e-03} & 1.239627e-02 & 4.746725e-02 & 4.590917e-02 & 2.278794e-01 & 2.370984e-01 \\
\bottomrule
\end{tabular}}
\end{table*}

\begin{table*}[t]
\centering
\scriptsize
\setlength{\tabcolsep}{4.0pt}
\renewcommand{\arraystretch}{1.08}
\caption{{Complete non-physical implementation parameters for the uploaded LNN--PINN eight-field relativistic viscous-fluid benchmark.}}
\label{Tapp:rel3_lnn_params}
\resizebox{\textwidth}{!}{%
\begin{tabular}{llll}
\toprule
Category & Symbol / item & Value & Mathematical status \\
\midrule
Learning-rate family & $\eta$ & $\{10^{-3},10^{-4},10^{-5},10^{-6},10^{-7},10^{-8}\}$ & active \\
Training horizon & $N_{\rm train}$ & $8000$ & active \\
Interior sample count & $N$ & $1000$ & active \\
Boundary sample count & $N_1$ & $1000$ & active \\
Seed & $s$ & $888888$ & active through realization choice \\
Optimizer & Adam family & yes & active \\
First moment parameter & $\beta_1$ & $0.9$ & active \\
Second moment parameter & $\beta_2$ & $0.999$ & active \\
Adam denominator shift & $\varepsilon_{\rm A}$ & $10^{-8}$ & active \\
Weight decay & $\lambda_{\rm wd}$ & $0$ & active \\
AMSGrad switch & $\chi_{\rm AMS}$ & $0$ & active as discrete flag \\
Input dimension & $d$ & $2$ & active \\
Output dimension & $o$ & $8$ & active \\
Hidden width & $w$ & $64$ & active \\
Backbone hidden-stage count & $H$ & $4$ & active \\
Liquid-block count & $q$ & $4$ & active \\
Gate initialization & $\alpha_0$ & $0.5$ & active \\
Velocity limiter & $V_{\max}$ & $0.45$ & active \\
Energy floor & $\epsilon_{\min}$ & $0.05$ & active \\
Density floor & $n_{\min}$ & $0.05$ & active \\
Stability shift & $\delta_{\mathrm{stab}}$ & $10^{-12}$ & active \\
Adiabatic index & $\Gamma_{\rm ad}$ & $1.4$ & active \\
Rest mass & $m$ & $0.8$ & active \\
Shear viscosity & $\eta_{\rm shear}$ & $0.06$ & active \\
Bulk viscosity & $\zeta$ & $0.04$ & active \\
Shear relaxation time & $\tau_\pi$ & $0.25$ & active \\
Bulk relaxation time & $\tau_\Pi$ & $0.22$ & active \\
Evaluation-grid side length & $h$ & $200$ & active for reported metrics \\
Evaluation-point count & $M_h$ & $40000$ & derived active quantity \\
Parameter count & $P_{\rm LNN}$ & $30088$ & derived active quantity \\
Logging interval & $N_{\rm print}$ & $100$ & mathematically inactive for training and evaluation \\
Raster density & DPI & $300$ & mathematically inactive for training and evaluation \\
Execution metadata & device selection, CUDA warmup, output naming & present & inactive metadata \\
\bottomrule
\end{tabular}}
\end{table*}

The exact trainable-parameter count follows from direct counting. The first affine map contributes
\begin{equation}
P_{\mathrm{in}}=dw+w,
\label{eq:rel3app_Pin}
\end{equation}
the remaining \(H-1\) backbone hidden affine maps contribute
\begin{equation}
P_{\mathrm{hid}}=(H-1)(w^2+w),
\label{eq:rel3app_Phid}
\end{equation}
the output layer contributes
\begin{equation}
P_{\mathrm{out}}=wo+o,
\label{eq:rel3app_Pout}
\end{equation}
and each liquid block contributes one square matrix, one bias vector, and one gate vector,
\begin{equation}
P_{\mathrm{liq,one}}=w^2+w+w=w^2+2w.
\label{eq:rel3app_Pliq_one}
\end{equation}
Therefore
\begin{equation}
P_{\mathrm{LNN}}(d,o,w,H,q)=dw+w+(H-1)(w^2+w)+wo+o+q(w^2+2w).
\label{eq:rel3app_PLNN_general}
\end{equation}
Substituting \(d=2\), \(o=8\), \(w=64\), \(H=4\), and \(q=4\) into Eq.~\eqref{eq:rel3app_PLNN_general} gives
\begin{align}
P_{\mathrm{LNN}}
=30088.
\label{eq:rel3app_PLNN}
\end{align}

We now carry out the rigorous parameter-sensitivity analysis from the actual LNN--PINN framework in Eqs.~\eqref{eq:rel3app_h1}--\eqref{eq:rel3app_out} and from the actual relativistic objective in Eq.~\eqref{eq:rel3app_J}. Let
\begin{equation}
g_k=\nabla_\theta \mathcal J_{\mathrm{rel}}(\theta_k;\Xi_k),
\label{eq:rel3app_gk}
\end{equation}
where \(\Xi_k\) denotes the entire random collocation realization at training step \(k\). The generalized Adam family with explicit weight decay and AMSGrad switch reads
\begin{equation}
\widetilde g_k=g_k+\lambda_{\mathrm{wd}}\theta_k,
\label{eq:rel3app_gtilde}
\end{equation}
\begin{equation}
m_k=\beta_1m_{k-1}+(1-\beta_1)\widetilde g_k,
\qquad
v_k=\beta_2v_{k-1}+(1-\beta_2)(\widetilde g_k\odot \widetilde g_k),
\label{eq:rel3app_mkvk}
\end{equation}
\begin{equation}
\widehat m_k=\frac{m_k}{1-\beta_1^k},
\qquad
\widehat v_k=\frac{v_k}{1-\beta_2^k},
\label{eq:rel3app_hatmom}
\end{equation}
\begin{equation}
\overline v_k(\chi_{\mathrm{AMS}})
=
(1-\chi_{\mathrm{AMS}})\widehat v_k+\chi_{\mathrm{AMS}}\max\{\overline v_{k-1},\widehat v_k\},
\label{eq:rel3app_vbar}
\end{equation}
\begin{equation}
\theta_{k+1}=\theta_k-\eta\frac{\widehat m_k}{\sqrt{\overline v_k(\chi_{\mathrm{AMS}})}+\varepsilon_{\mathrm A}}.
\label{eq:rel3app_adam}
\end{equation}
Define the optimizer direction
\begin{equation}
d_k(\eta,\beta_1,\beta_2,\varepsilon_{\mathrm A},\lambda_{\mathrm{wd}},\chi_{\mathrm{AMS}})
=
\frac{\widehat m_k}{\sqrt{\overline v_k(\chi_{\mathrm{AMS}})}+\varepsilon_{\mathrm A}}.
\label{eq:rel3app_dk}
\end{equation}
Then Eq.~\eqref{eq:rel3app_adam} becomes
\begin{equation}
\theta_{k+1}=\theta_k-\eta d_k.
\label{eq:rel3app_adam_compact}
\end{equation}

The sensitivity with respect to the learning rate follows immediately. Freezing \((\theta_k,m_k,v_k,\Xi_k)\) and perturbing \(\eta\mapsto \eta+\delta\eta\), Eq.~\eqref{eq:rel3app_adam_compact} yields
\begin{equation}
\theta_{k+1}(\eta+\delta\eta)-\theta_{k+1}(\eta)=-\delta\eta\, d_k,
\label{eq:rel3app_eta_increment}
\end{equation}
hence
\begin{equation}
\left.\frac{\partial \theta_{k+1}}{\partial\eta}\right|_{(\theta_k,m_k,v_k,\Xi_k)}=-d_k,
\qquad
\|\delta\theta_{k+1}\|_2\le |\delta\eta|\,\|d_k\|_2.
\label{eq:rel3app_eta_derivative}
\end{equation}
A second-order expansion of \(\mathcal J_{\mathrm{rel}}\) along the actual update direction gives
\begin{equation}
\mathcal J_{\mathrm{rel}}(\theta_{k+1};\Xi_k)
=
\mathcal J_{\mathrm{rel}}(\theta_k;\Xi_k)
-
\eta\, g_k^\top d_k
+
\frac{\eta^2}{2}
d_k^\top
H_k(\theta_k-\tau_k\eta d_k)
d_k,
\label{eq:rel3app_taylor_eta}
\end{equation}
where
\begin{equation}
H_k=\nabla_\theta^2\mathcal J_{\mathrm{rel}}(\theta_k;\Xi_k),
\qquad
\tau_k\in(0,1).
\label{eq:rel3app_Hk}
\end{equation}
If
\begin{equation}
g_k^\top d_k\ge c_k\|g_k\|_2\|d_k\|_2,
\qquad
\|H_k(\theta_k-\tau_k\eta d_k)\|_2\le M_k,
\label{eq:rel3app_ckMk}
\end{equation}
then Eq.~\eqref{eq:rel3app_taylor_eta} yields the sufficient descent range
\begin{equation}
0<\eta<\frac{2c_k\|g_k\|_2}{M_k\|d_k\|_2}.
\label{eq:rel3app_eta_range}
\end{equation}

The training-horizon sensitivity follows by iterating Eq.~\eqref{eq:rel3app_adam_compact}:
\begin{equation}
\theta_K=\theta_0-\eta\sum_{k=0}^{K-1}d_k.
\label{eq:rel3app_thetaK}
\end{equation}
Hence enlarging the training horizon from \(K\) to \(K+\Delta K\) gives
\begin{equation}
\theta_{K+\Delta K}-\theta_K=-\eta\sum_{k=K}^{K+\Delta K-1}d_k,
\label{eq:rel3app_epoch_increment}
\end{equation}
and therefore
\begin{equation}
\|\theta_{K+\Delta K}-\theta_K\|_2
\le
\eta\sum_{k=K}^{K+\Delta K-1}\|d_k\|_2.
\label{eq:rel3app_epoch_bound}
\end{equation}

The moment parameters \(\beta_1\) and \(\beta_2\) also admit exact derivatives. Unrolling Eq.~\eqref{eq:rel3app_mkvk} gives
\begin{equation}
m_k=(1-\beta_1)\sum_{j=1}^{k}\beta_1^{k-j}\widetilde g_j,
\qquad
v_k=(1-\beta_2)\sum_{j=1}^{k}\beta_2^{k-j}(\widetilde g_j\odot \widetilde g_j).
\label{eq:rel3app_unroll}
\end{equation}
Differentiating Eq.~\eqref{eq:rel3app_unroll} with respect to \(\beta_1\) yields
\begin{equation}
\frac{\partial m_k}{\partial\beta_1}
=
-\sum_{j=1}^{k}\beta_1^{k-j}\widetilde g_j
+
(1-\beta_1)\sum_{j=1}^{k}(k-j)\beta_1^{k-j-1}\widetilde g_j,
\label{eq:rel3app_dm_db1}
\end{equation}
and therefore
\begin{equation}
\frac{\partial \widehat m_k}{\partial\beta_1}
=
\frac{(1-\beta_1^k)\frac{\partial m_k}{\partial\beta_1}+k\beta_1^{k-1}m_k}{(1-\beta_1^k)^2}.
\label{eq:rel3app_dhatm_db1}
\end{equation}
Likewise,
\begin{equation}
\frac{\partial v_k}{\partial\beta_2}
=
-\sum_{j=1}^{k}\beta_2^{k-j}(\widetilde g_j\odot \widetilde g_j)
+
(1-\beta_2)\sum_{j=1}^{k}(k-j)\beta_2^{k-j-1}(\widetilde g_j\odot \widetilde g_j),
\label{eq:rel3app_dv_db2}
\end{equation}
hence
\begin{equation}
\frac{\partial \widehat v_k}{\partial\beta_2}
=
\frac{(1-\beta_2^k)\frac{\partial v_k}{\partial\beta_2}+k\beta_2^{k-1}v_k}{(1-\beta_2^k)^2}.
\label{eq:rel3app_dhatv_db2}
\end{equation}

The denominator-shift sensitivity follows directly from Eq.~\eqref{eq:rel3app_dk}:
\begin{equation}
\frac{\partial d_k}{\partial \varepsilon_{\mathrm A}}
=
-\frac{\widehat m_k}{\left(\sqrt{\overline v_k(\chi_{\mathrm{AMS}})}+\varepsilon_{\mathrm A}\right)^2},
\label{eq:rel3app_deps}
\end{equation}
hence
\begin{equation}
\frac{\partial \theta_{k+1}}{\partial \varepsilon_{\mathrm A}}
=
\eta
\frac{\widehat m_k}{\left(\sqrt{\overline v_k(\chi_{\mathrm{AMS}})}+\varepsilon_{\mathrm A}\right)^2}.
\label{eq:rel3app_theta_eps}
\end{equation}

The weight-decay sensitivity enters through \(\widetilde g_k=g_k+\lambda_{\mathrm{wd}}\theta_k\). Differentiating Eq.~\eqref{eq:rel3app_mkvk} with respect to \(\lambda_{\mathrm{wd}}\) gives
\begin{equation}
\frac{\partial m_k}{\partial\lambda_{\mathrm{wd}}}
=
\beta_1\frac{\partial m_{k-1}}{\partial\lambda_{\mathrm{wd}}}
+
(1-\beta_1)\left(\theta_k+\lambda_{\mathrm{wd}}\frac{\partial\theta_k}{\partial\lambda_{\mathrm{wd}}}\right),
\label{eq:rel3app_dm_dwd}
\end{equation}
\begin{equation}
\frac{\partial v_k}{\partial\lambda_{\mathrm{wd}}}
=
\beta_2\frac{\partial v_{k-1}}{\partial\lambda_{\mathrm{wd}}}
+
2(1-\beta_2)\widetilde g_k\odot\left(\theta_k+\lambda_{\mathrm{wd}}\frac{\partial\theta_k}{\partial\lambda_{\mathrm{wd}}}\right).
\label{eq:rel3app_dv_dwd}
\end{equation}
At the uploaded value \(\lambda_{\mathrm{wd}}=0\), these become
\begin{equation}
\left.\frac{\partial m_k}{\partial\lambda_{\mathrm{wd}}}\right|_{\lambda_{\mathrm{wd}}=0}
=
\beta_1\left.\frac{\partial m_{k-1}}{\partial\lambda_{\mathrm{wd}}}\right|_{\lambda_{\mathrm{wd}}=0}
+
(1-\beta_1)\theta_k,
\label{eq:rel3app_dm_dwd0}
\end{equation}
\begin{equation}
\left.\frac{\partial v_k}{\partial\lambda_{\mathrm{wd}}}\right|_{\lambda_{\mathrm{wd}}=0}
=
\beta_2\left.\frac{\partial v_{k-1}}{\partial\lambda_{\mathrm{wd}}}\right|_{\lambda_{\mathrm{wd}}=0}
+
2(1-\beta_2)g_k\odot\theta_k.
\label{eq:rel3app_dv_dwd0}
\end{equation}

The AMSGrad switch \(\chi_{\mathrm{AMS}}\) is discrete rather than continuous. Therefore the correct sensitivity notion is a jump functional, not an ordinary derivative:
\begin{equation}
\Delta_{\mathrm{AMS}}\theta_{k+1}
=
\theta_{k+1}\big|_{\chi_{\mathrm{AMS}}=1}
-
\theta_{k+1}\big|_{\chi_{\mathrm{AMS}}=0}
=
-\eta\,\widehat m_k
\left[
\frac{1}{\sqrt{\max\{\overline v_{k-1},\widehat v_k\}}+\varepsilon_{\mathrm A}}
-
\frac{1}{\sqrt{\widehat v_k}+\varepsilon_{\mathrm A}}
\right].
\label{eq:rel3app_ams_jump}
\end{equation}

The collocation-count sensitivities act through stochastic variance. Define the squared residual random variables
\begin{equation}
X_\Omega=\sum_{a\in\{E,M_x,M_y,n,\Pi,xx,xy,yy\}}\bigl(R_a(x,y;\theta)-S_a(x,y)\bigr)^2,
\qquad
(x,y)\sim \mathrm{Unif}([0,1]^2),
\label{eq:rel3app_Xomega}
\end{equation}
\begin{equation}
X_\downarrow=\sum_f(f_\theta(x,0)-f^\ast(x,0))^2,
\qquad
X_\uparrow=\sum_f(f_\theta(x,1)-f^\ast(x,1))^2,
\qquad
x\sim \mathrm{Unif}([0,1]),
\label{eq:rel3app_Xud}
\end{equation}
\begin{equation}
X_{\mathrm L}=\sum_f(f_\theta(0,y)-f^\ast(0,y))^2,
\qquad
X_{\mathrm R}=\sum_f(f_\theta(1,y)-f^\ast(1,y))^2,
\qquad
y\sim \mathrm{Unif}([0,1]),
\label{eq:rel3app_Xlr}
\end{equation}
where \(f\) runs over the eight principal fields. Writing the channel variances as \(\sigma_\Omega^2,\sigma_\downarrow^2,\sigma_\uparrow^2,\sigma_{\mathrm L}^2,\sigma_{\mathrm R}^2\), one has
\begin{equation}
\operatorname{Var}(L_\Omega)=\frac{\sigma_\Omega^2}{N},
\qquad
\operatorname{Var}(L_\downarrow)=\frac{\sigma_\downarrow^2}{N_1},
\qquad
\operatorname{Var}(L_\uparrow)=\frac{\sigma_\uparrow^2}{N_1},
\qquad
\operatorname{Var}(L_{\rm L})=\frac{\sigma_{\mathrm L}^2}{N_1},
\qquad
\operatorname{Var}(L_{\rm R})=\frac{\sigma_{\mathrm R}^2}{N_1},
\label{eq:rel3app_vars}
\end{equation}
and therefore
\begin{equation}
\operatorname{Var}(\mathcal J_{\mathrm{rel}})
=
\frac{\sigma_\Omega^2}{N}
+
\frac{\sigma_\downarrow^2+\sigma_\uparrow^2+\sigma_{\mathrm L}^2+\sigma_{\mathrm R}^2}{N_1}.
\label{eq:rel3app_var_total}
\end{equation}
Differentiating Eq.~\eqref{eq:rel3app_var_total} gives
\begin{equation}
\frac{\partial}{\partial N}\operatorname{Var}(\mathcal J_{\mathrm{rel}})=-\frac{\sigma_\Omega^2}{N^2},
\qquad
\frac{\partial}{\partial N_1}\operatorname{Var}(\mathcal J_{\mathrm{rel}})=-\frac{\sigma_\downarrow^2+\sigma_\uparrow^2+\sigma_{\mathrm L}^2+\sigma_{\mathrm R}^2}{N_1^2}.
\label{eq:rel3app_dvars}
\end{equation}
If one prescribes variance tolerances \(\delta_\Omega\) and \(\delta_{\partial\Omega}\), then sufficient admissible sampling ranges are
\begin{equation}
N\ge \frac{\sigma_\Omega^2}{\delta_\Omega^2},
\qquad
N_1\ge \frac{\sigma_\downarrow^2+\sigma_\uparrow^2+\sigma_{\mathrm L}^2+\sigma_{\mathrm R}^2}{\delta_{\partial\Omega}^2}.
\label{eq:rel3app_sampling_ranges}
\end{equation}

The seed \(s\) is discrete rather than differentiable. Therefore the correct sensitivity notion is the realization gap
\begin{equation}
\Delta_{s_1,s_2}\mathcal J_{\mathrm{rel}}(\theta_k)
=
\mathcal J_{\mathrm{rel}}(\theta_k;\Xi_k^{(s_1)})-\mathcal J_{\mathrm{rel}}(\theta_k;\Xi_k^{(s_2)}).
\label{eq:rel3app_seed_gap}
\end{equation}
Since all six uploaded runs keep \(s=888888\) fixed, the learning-rate comparison isolates step-size effects rather than mixing them with seed variability.

The framework-based architecture sensitivities follow from the actual liquid residual block
\begin{equation}
\mathcal B_\ell(z)=z+\operatorname{Diag}(\alpha_\ell)\tanh(A_\ell z+c_\ell).
\label{eq:rel3app_block}
\end{equation}
Its Jacobian with respect to the input is
\begin{equation}
D\mathcal B_\ell(z)=I+\operatorname{Diag}(\alpha_\ell)\operatorname{Diag}\!\bigl(\operatorname{sech}^2(A_\ell z+c_\ell)\bigr)A_\ell.
\label{eq:rel3app_DB}
\end{equation}
Since
\begin{equation}
0<\operatorname{sech}^2(s)\le 1
\qquad
\text{for all }s\in\mathbb{R},
\label{eq:rel3app_sech}
\end{equation}
Eq.~\eqref{eq:rel3app_DB} yields
\begin{equation}
\|D\mathcal B_\ell(z)\|_2
\le
1+\|\operatorname{Diag}(\alpha_\ell)\|_2\,\|A_\ell\|_2.
\label{eq:rel3app_DB_bound}
\end{equation}
At initialization, Eq.~\eqref{eq:rel3app_archvals} gives
\begin{equation}
\|\operatorname{Diag}(\alpha_\ell^{(0)})\|_2=\alpha_0=0.5,
\label{eq:rel3app_alpha_norm}
\end{equation}
hence
\begin{equation}
\|D\mathcal B_\ell(z)\|_2\le 1+0.5\|A_\ell\|_2.
\label{eq:rel3app_DB_init}
\end{equation}
If one prescribes a per-block amplification budget \(\kappa_\ell>1\), then a sufficient admissible initialization range is
\begin{equation}
0\le \alpha_0\le \frac{\kappa_\ell-1}{\|A_\ell\|_2}.
\label{eq:rel3app_alpha_range_layer}
\end{equation}
Imposing a common budget \(\kappa>1\) on all four liquid blocks gives
\begin{equation}
0\le \alpha_0\le \min_{\ell=1,2,3,4}\frac{\kappa-1}{\|A_\ell\|_2}.
\label{eq:rel3app_alpha_range_global}
\end{equation}

The output sensitivity with respect to \(\alpha_0\) follows from the chain rule. First,
\begin{equation}
\left.\frac{\partial \mathcal B_\ell(z)}{\partial\alpha_0}\right|_{\alpha_\ell=\alpha_0\mathbf 1}=\tanh(A_\ell z+c_\ell),
\label{eq:rel3app_dB_dalpha}
\end{equation}
hence
\begin{equation}
\left\|\left.\frac{\partial \mathcal B_\ell(z)}{\partial\alpha_0}\right|_{\alpha_\ell=\alpha_0\mathbf 1}\right\|_2\le \sqrt{w}.
\label{eq:rel3app_dB_dalpha_bound}
\end{equation}
If \(\mathcal F_1,\ldots,\mathcal F_4\) denote the four stage maps, then
\begin{equation}
r_\theta(x,y)=W_{\rm out}\mathcal F_4\circ \mathcal F_3\circ \mathcal F_2\circ \mathcal F_1(\xi)+b_{\rm out},
\label{eq:rel3app_comp_output}
\end{equation}
and therefore
\begin{align}
\frac{\partial r_\theta}{\partial\alpha_0}
=
W_{\rm out}
\sum_{j=1}^4
\left(
\prod_{\ell=j+1}^4 D\mathcal F_\ell
\right)
\left.
\frac{\partial \mathcal B_j}{\partial\alpha_0}
\right|_{\alpha_j=\alpha_0\mathbf 1}.
\label{eq:rel3app_dr_dalpha}
\end{align}
Using Eq.~\eqref{eq:rel3app_dB_dalpha_bound} and
\begin{equation}
\|D\mathcal F_\ell\|_2\le (1+\alpha_0\|A_\ell\|_2)\|W_{\ell-1}\|_2,
\label{eq:rel3app_DF_bound}
\end{equation}
we obtain
\begin{equation}
\left\|
\frac{\partial r_\theta}{\partial\alpha_0}
\right\|_2
\le
\|W_{\rm out}\|_2
\sum_{j=1}^4
\sqrt{w}
\prod_{\ell=j+1}^4
(1+\alpha_0\|A_\ell\|_2)\|W_{\ell-1}\|_2.
\label{eq:rel3app_dr_dalpha_bound}
\end{equation}

The discrete sensitivities with respect to \(d,o,w,H,q\) follow exactly from Eq.~\eqref{eq:rel3app_PLNN_general}. Increasing the input dimension by one while fixing \((o,w,H,q)\) gives
\begin{equation}
P_{\mathrm{LNN}}(d+1,o,w,H,q)-P_{\mathrm{LNN}}(d,o,w,H,q)=w,
\label{eq:rel3app_ddim}
\end{equation}
which equals \(64\) at the uploaded width. Increasing the output dimension by one gives
\begin{equation}
P_{\mathrm{LNN}}(d,o+1,w,H,q)-P_{\mathrm{LNN}}(d,o,w,H,q)=w+1,
\label{eq:rel3app_do}
\end{equation}
which equals \(65\). Increasing the hidden-stage number by one gives
\begin{equation}
P_{\mathrm{LNN}}(d,o,w,H+1,q)-P_{\mathrm{LNN}}(d,o,w,H,q)=w^2+w,
\label{eq:rel3app_dH}
\end{equation}
which equals \(4160\). Increasing the liquid-block count by one gives
\begin{equation}
P_{\mathrm{LNN}}(d,o,w,H,q+1)-P_{\mathrm{LNN}}(d,o,w,H,q)=w^2+2w,
\label{eq:rel3app_dq}
\end{equation}
which equals \(4224\). Increasing the hidden width by one while fixing \((d,o,H,q)\) gives
\begin{align}
P_{\mathrm{LNN}}(d,o,w+1,H,q)-P_{\mathrm{LNN}}(d,o,w,H,q)
&=
(H+q-1)\bigl((w+1)^2-w^2\bigr)+(d+H+o+2q)
\nonumber\\
&=
(H+q-1)(2w+1)+(d+H+o+2q).
\label{eq:rel3app_dw}
\end{align}
At \(d=2\), \(o=8\), \(w=64\), \(H=4\), \(q=4\), Eq.~\eqref{eq:rel3app_dw} gives
\begin{equation}
7\cdot129+22=925.
\label{eq:rel3app_dw_actual}
\end{equation}

The same parameters also control transport amplification. If
\begin{equation}
\|W_{\ell-1}\|_2\le \rho,
\qquad
\|A_\ell\|_2\le \gamma,
\qquad
\ell=1,2,3,4,
\label{eq:rel3app_rhogamma}
\end{equation}
then Eq.~\eqref{eq:rel3app_DF_bound} yields
\begin{equation}
\|D\mathcal F_\ell\|_2\le \rho(1+\alpha_0\gamma),
\qquad \ell=1,2,3,4,
\label{eq:rel3app_uniform_DF}
\end{equation}
and therefore
\begin{equation}
\operatorname{Lip}\!\left(\mathcal F_4\circ\mathcal F_3\circ\mathcal F_2\circ\mathcal F_1\right)
\le
\bigl(\rho(1+\alpha_0\gamma)\bigr)^4.
\label{eq:rel3app_transport4}
\end{equation}
For a matched family with \(H=q\), the same reasoning gives
\begin{equation}
\operatorname{Lip}_H\le \bigl(\rho(1+\alpha_0\gamma)\bigr)^H.
\label{eq:rel3app_transportH}
\end{equation}
If one imposes a transport budget \(\Lambda_\star>1\), then the admissible depth range satisfies
\begin{equation}
H\le \frac{\log \Lambda_\star}{\log(\rho(1+\alpha_0\gamma))}
\qquad
\text{whenever }\rho(1+\alpha_0\gamma)>1.
\label{eq:rel3app_H_range}
\end{equation}

The evaluation-grid side length \(h\) and the derived point count \(M_h=h^2\) influence only the reporting functional, not the training dynamics. On \([0,1]^2\), the grid spacings satisfy
\begin{equation}
\Delta x=\Delta y=\frac{1}{h-1}.
\label{eq:rel3app_dxdy}
\end{equation}
Define the continuum fieldwise squared-error functional for each field \(f\) by
\begin{equation}
\mathcal E_{f,\mathrm{cont}}(\theta)=\int_0^1\int_0^1 |f_\theta(x,y)-f^\ast(x,y)|^2\,dx\,dy.
\label{eq:rel3app_Econt}
\end{equation}
Then the discrete MSE in the reporting tables is a uniform-grid Riemann approximation, so for smooth \(f_\theta\),
\begin{equation}
\mathrm{MSE}_{f,h}(\theta)=\mathcal E_{f,\mathrm{cont}}(\theta)+\mathcal O(h^{-1}).
\label{eq:rel3app_riemann}
\end{equation}
Thus the rigorous \(h\)-sensitivity enters through quadrature bias. If one prescribes a reporting bias budget \(\varepsilon_h\), then a sufficient admissible grid range satisfies
\begin{equation}
h\ge 1+\frac{C_q}{\varepsilon_h}.
\label{eq:rel3app_h_range}
\end{equation}
Since \(M_h=h^2\), the exact discrete increment of the evaluation-point count is
\begin{equation}
M_{h+1}-M_h=(h+1)^2-h^2=2h+1,
\label{eq:rel3app_Mh_increment}
\end{equation}
which equals \(401\) at \(h=200\).

The remaining numeric parameters \(N_{\rm print}\) and \(\mathrm{DPI}\) do not enter either Eq.~\eqref{eq:rel3app_J} or the fieldwise reporting metrics. Therefore their mathematical sensitivities with respect to the training objective and to the reported global errors vanish:
\begin{equation}
\frac{\partial \mathcal J_{\mathrm{rel}}}{\partial N_{\rm print}}=0,
\qquad
\frac{\partial \mathrm{MSE}_{f,h}}{\partial N_{\rm print}}=0,
\label{eq:rel3app_print_zero}
\end{equation}
\begin{equation}
\frac{\partial \mathcal J_{\mathrm{rel}}}{\partial \mathrm{DPI}}=0,
\qquad
\frac{\partial \mathrm{MSE}_{f,h}}{\partial \mathrm{DPI}}=0.
\label{eq:rel3app_dpi_zero}
\end{equation}

The unique summary figure for this appendix appears in Fig.~\ref{Fapp:rel3_lnn_lr}. It visualizes the six loss histories together with the six reconstructed field sets and their absolute-error maps under the six learning rates in Eq.~\eqref{eq:rel3app_eta_train}.

\begin{figure*}[t]
\centering
\includegraphics[width=\textwidth]{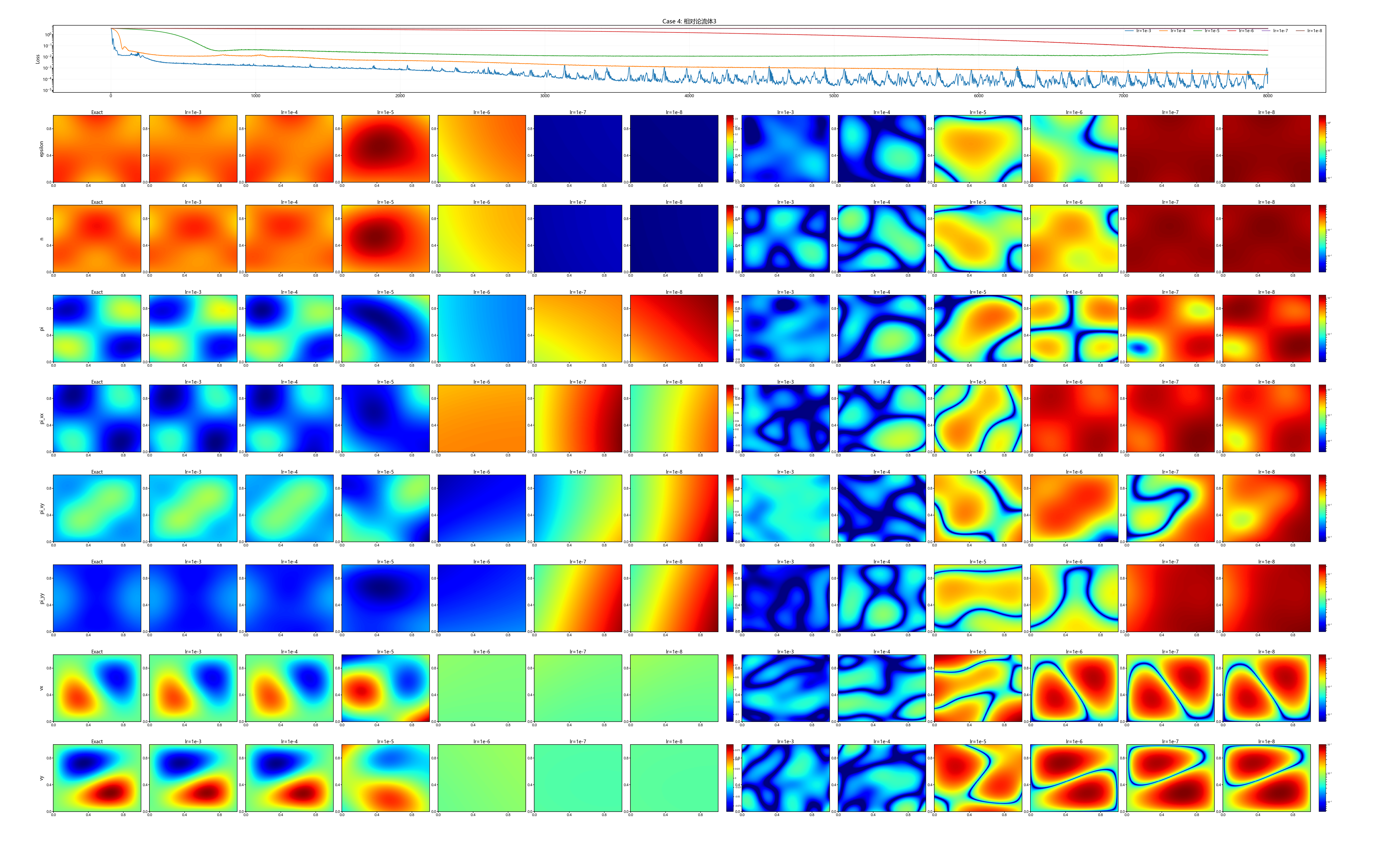}
\caption{Learning-rate comparison for the LNN--PINN eight-field relativistic viscous-fluid benchmark. The upper panel reports the full loss histories under the six learning rates in Eq.~\eqref{eq:rel3app_eta_train}. The lower panels report the reference fields, the corresponding six LNN--PINN reconstructions, and the associated absolute-error distributions for all eight principal fields. All six runs keep the stochastic relativistic operator, the constraint family, the objective in Eq.~\eqref{eq:rel3app_J}, the architecture in Eqs.~\eqref{eq:rel3app_h1}--\eqref{eq:rel3app_out}, the collocation counts in Eq.~\eqref{eq:rel3app_counts}, and the evaluation grid in Eq.~\eqref{eq:rel3app_eval_controls} fixed.}
\label{Fapp:rel3_lnn_lr}
\end{figure*}

Taken together, Eqs.~\eqref{eq:rel3app_eta_range}, \eqref{eq:rel3app_sampling_ranges}, \eqref{eq:rel3app_alpha_range_global}, \eqref{eq:rel3app_H_range}, and \eqref{eq:rel3app_h_range} give the rigorous framework-based meaning of a reasonable implementation regime for the uploaded LNN--PINN relativistic viscous-fluid scripts. The step size must lie inside the local Adam descent range controlled by curvature and gradient-direction alignment; the collocation counts must each exceed the threshold set by their own residual-variance budget; the gate initialization must remain below the block-amplification threshold imposed by the liquid matrices; the hidden depth must remain inside the transport-amplification budget induced by repeated liquid residual mappings; and the evaluation grid must be fine enough to keep quadrature bias below the reporting tolerance. Within the six uploaded runs, \(\eta=10^{-3}\) gives the best overall average fieldwise performance, the smallest average RMSE, the smallest average MAE, the smallest average relative \(L_2\) error, the smallest average maximum error, whereas \(\eta=10^{-4}\) gives the smallest terminal training loss. This appendix therefore shows not only which numerical choices work well for the relativistic eight-field benchmark, but also why each non-physical implementation parameter enters the LNN--PINN framework in the precise mathematical manner derived above.

}
\clearpage
\bibliographystyle{cas-model2-names}

\bibliography{cas-refs}



\end{document}